%% file: paper-2023-lmcs-macro-swarm.tex
\documentclass{lmcs}
\pdfoutput=1
\usepackage[utf8]{inputenc}

\usepackage{lastpage}
\lmcsdoi{21}{3}{13}
\lmcsheading{}{\pageref{LastPage}}{}{}%
{Jan.~23,~2024}{Aug.~05,~2025}{}

\usepackage{graphicx}
\usepackage{xcolor}
\usepackage{acronym}
\usepackage{hyperref}
\usepackage{comment}
\usepackage{tabularx}
\usepackage{makecell}
\usepackage{pifont}
\usepackage{wasysym}
\newcommand{\cmark}{\CIRCLE} 
\newcommand{\pmark}{\LEFTcircle} 
\newcommand{\xmark}{\Circle} 
\usepackage{subcaption}

\usepackage[capitalise]{cleveref}
\usepackage{listings}
\input{scala-def}
\newcommand{\neigh}{\rightsquigarrow}
\newcommand{\MacroSwarm}{{\sc{}MacroSwarm}}
\newcommand{\scafi}{{\sc{}ScaFi}}

\acrodef{api}[API]{Application Program Interface}
\acrodef{dsl}[DSL]{domain-specific language}
\acrodef{lidar}[LIDAR]{Laser Imaging Detection And Ranging}

\makeatletter
\AtBeginDocument
 {
   \def\ltx@label#1{\cref@label{#1}}
   \def\label@in@display@noarg#1{\cref@old@label@in@display{#1}}
\def\label@in@mmeasure@noarg#1{%
    \begingroup%
      \measuring@false%
      \cref@old@label@in@display{#1}
    \endgroup}%
 } %
\makeatother

\sloppypar

\let\oldsss\subsubsection
\renewcommand{\subsubsection}[1]{\oldsss{#1.}}

\begin{document}
\title[\MacroSwarm: A Framework for Swarm Programming]{MacroSwarm: A Field-based Compositional Framework for Swarm Programming}

\newcommand{\meta}[1]{{\color{blue}#1}}%
\newcommand{\revA}[1]{{#1}}%
\newcommand{\revB}[1]{{#1}}%

\author{Gianluca Aguzzi\lmcsorcid{0000-0002-1553-4561}}

\author{Roberto Casadei\lmcsorcid{0000-0001-9149-949X}}

\author{Mirko Viroli\lmcsorcid{0000-0003-2702-5702}}

\address{Alma Mater Studiorum---Università di Bologna, 
Cesena, Italy}	
\email{\{gianluca.aguzzi, roby.casadei, mirko.viroli\}@unibo.it}  
\begin{abstract}
\revA{
  Swarm behaviour engineering is an area of research that seeks to investigate methods and techniques
   for coordinating computation and action
   within groups of simple agents 
   to achieve complex global goals like \emph{pattern formation}, 
   \emph{collective movement},
   \emph{clustering}, and
   \emph{distributed sensing}.
  Despite recent progress in the analysis and engineering of swarms (of drones, robots, vehicles), 
   there is still a need for general design and implementation methods and tools
   that can be used to define complex swarm behaviour
   in a principled way.
  To contribute to this quest, 
   this article proposes a new field-based coordination approach, called \MacroSwarm{}, 
   to design and program swarm behaviour
   in terms of reusable and fully composable functional blocks embedding collective computation and coordination. 
  Based on the macroprogramming paradigm of aggregate computing, \MacroSwarm{} builds on the idea of expressing each swarm behaviour block 
  as a pure function, mapping sensing fields into actuation goal fields, e.g., including movement vectors.
In order to demonstrate the expressiveness, compositionality, and practicality of \MacroSwarm{} as a framework for swarm programming,
 we perform a variety of simulations 
 covering common patterns of  
 flocking, pattern formation, and collective decision-making.
\revB{The implications of the inherent self-stabilisation properties of field-based computations in \MacroSwarm{} are discussed, which formally guarantee some resilience properties and guided the design of the library.}
}
   
\end{abstract}

\maketitle          
\section{Introduction}

Recent technological advances 
 foster a vision of \emph{swarms} of mobile cyber-physical agents able to compute, coordinate with neighbours, and interact with the environment according to increasingly complex patterns, plans, and goals.
Notable examples include swarms of drones and robots~\cite{DBLP:journals/firai/SchranzUSE20}, 
 fleets of vehicles~\cite{DBLP:journals/jiii/TahirBHTP19},
 and crowds of wearable-augmented people~\cite{DBLP:journals/wc/GalininaMH0K18}.
In these domains,
 a prominent research problem
 is how to effectively engineer \emph{swarm behaviour}~\cite{DBLP:journals/swarm/BrambillaFBD13},
 i.e., how to promote 
 the emergence of desired global-level outcomes
 with inherent robustness and resiliency to changes and faults in the swarm or the environment.
Complex patterns can emerge through the interaction of simple agents~\cite{bonabeau1999swarmintelligence-book}, and centralised approaches 
 can suffer from scalability and dependability issues: as such,  
 we seek for an approach based on
 suitable distributed coordination models and languages to steer the micro-level activity of a possibly large set of agents.
This
 direction has been explored by various research threads related to coordination
 like  \emph{macroprogramming}~\cite{Casadei2023,regiment}, 
 \emph{spatial computing}~\cite{SpatialIGI2013},
 \emph{ensemble languages}
~\cite{scel2014taas,abd2020programming-cas-attribute-based}, 
 \emph{field-based coordination}~\cite{DBLP:journals/corr/Lluch-LafuenteL16,DBLP:journals/pervasive/MameiZL04}, and \emph{aggregate computing}~\cite{DBLP:journals/jlap/ViroliBDACP19}. 

Though a number of approaches and languages have been proposed 
 for specifying or programming swarm behaviour~\cite{Meld2007,%
DBLP:conf/icra/CarrollNS21,%
paros,%
DBLP:conf/isola/KosakHBWHR20,%
Koutsoubelias2016tecola,%
lima2018dolphin,%
Mottola2014voltron,%
DBLP:conf/iros/PinciroliB16,%
DBLP:conf/iros/YiDLD0WY20},
 a key feature that is generally missing or provided only to a limited extent
 is \emph{compositionality},
 namely the ability of combining blocks of simple swarm behaviour to construct swarm systems of increasing complexity in a controlled/engineered way.
Additionally, most of existing approaches tend to be pragmatic, not formally-founded and quite ad-hoc: they 
 enable construction of certain types of swarm applications
 but with limited support for analysis and principled design of complex applications (e.g.~\cite{lima2018dolphin,paros,DBLP:conf/iros/PinciroliB16,DBLP:conf/icra/CarrollNS21}).
Exceptions that provide a formal approach exist, but they are typically overly abstract, requiring additional effort to actually code and execute swarm control programs~\cite{DBLP:journals/csur/LuckcuckFDDF19}.

The goal of this work is 
 to introduce a formally-grounded \emph{\ac{api}},
 practical enough
 to concisely and elegantly encode a wide array of swarm behaviours.
This is based on the field-based coordination paradigm~\cite{DBLP:journals/jlap/ViroliBDACP19} and the field calculus~\cite{AVDPB-TOCL2019}:
each block of swarm behaviour is captured by a purely functional transformation of sensing fields into actuation fields including movement vectors,
and such a transformation declaratively captures the state/computation/interaction mechanisms necessary to achieve that behaviour.
Practically, such specifications can be programmed as Scala scripts in the \scafi{} framework \cite{DBLP:journals/softx/CasadeiVAP22,ACDV-LMCS2023}, a reference implementation for field-based coordination and aggregate computing.

Accordingly, we present \MacroSwarm{}, a \scafi{}-based framework to help programming with swarm behaviours by providing a set of blocks covering key swarming patterns as identified in literature~\cite{DBLP:journals/swarm/BrambillaFBD13}: flocking, pattern formation, consensus, and leader-follower behaviours.
\revB{This approach is backed by the formal framework of aggregate computing \cite{DBLP:journals/tomacs/ViroliABDP18,ADT-ISOLA2024}, for which self-stabilisation properties can be enacted: this is used to show that a significant portion of \MacroSwarm{} enjoys formally captured resiliency properties.}
\revB{To evaluate \MacroSwarm{}, we assess the efficacy of its constituent blocks both in isolation and in combination.  
We first conduct an in-depth analysis of the pattern formation block, 
a fundamental component underpinning numerous swarm behaviours. 
This analysis includes the impact of adversarial conditions, 
thereby demonstrating the robustness of the self-organising properties in this context.  
Subsequently, we present a simulated scenario involving a swarm of robots engaged in a collective decision-making task within a dynamic environment.  
This scenario illustrates how the individual blocks can be combined to achieve more complex emergent behaviour.
}

\revA{
Therefore,
 the main contribution of this work 
 is the design and implementation
 of \MacroSwarm{},
 \revB{the definition of a novel resiliency property for swarm 
 behaviours based on self-stabilisation of field-based computations,}
 and the development of simulations
 for assessing the correctness and effectiveness of the framework.
The \MacroSwarm{} library\footnote{The documentation is public available at \url{https://scafi.github.io/macro-swarm/}} is available as an open-source, permanently archived artefact~\cite{macroswarm-zenodo},
 and it is released on Maven Central for simple import with Maven-compatible dependency managers and build systems.
This article is a significant extension of conference paper~\cite{DBLP:conf/coordination/AguzziCV23}. 
Specifically, the extension consists of the following:
\revB{(i)} discussion of implication of self-stabilisation of algorithms to resiliency of swarming algorithms;
(ii) a largely extended experimental evaluation, covering a more comprehensive set of collective behaviour blocks; 
(iii) new functionality, i.e., based on a new block for achieving collective consensus towards a target value;
(iv) a broader discussion of related work;
and 
(v) additional clarifications and descriptions regarding important aspects of the framework like its overall architecture, execution model, and assumptions.
}

The remainder of this paper is organised as follows.
\Cref{sec:context} provides context and motivation.
\Cref{sec:background} reviews background on aggregate computing.
\Cref{sec:contrib} presents the main contribution of the paper, \MacroSwarm{}.
\revB{\Cref{sec:formal-selfstab} discusses resiliency properties as derived from self-stabilisation of field-based computations.}
\Cref{sec:eval} provides a simulation-based evaluation of the approach.
\Cref{sec:rw} reviews related work on swarm \revA{engineering}.
Finally, \Cref{sec:conc} provides a conclusion and future work.

\section{Context and Motivation}
\label{sec:context}

Engineering the collective behaviour of swarms is 
 a significant research challenge~\cite{DBLP:journals/swarm/BrambillaFBD13}.
Two main kinds of design methods can be identified~\cite{DBLP:journals/swarm/BrambillaFBD13}:
 \emph{automatic} design methods like evolutionary robotics~\cite{DBLP:series/sci/2008-108} or multi-agent reinforcement learning~\cite{DBLP:journals/tsmc/BusoniuBS08},
 also called \emph{behaviour-based} design, involving manually-implemented algorithms expressed via general-purpose or \acp{dsl}.
Our focus is on the latter category of methods and especially on \acp{dsl} for expressing swarm behaviour (which are reviewed in \Cref{sec:rw}).

Another main distinction is between \emph{centralised} (\emph{orchestration-based}) and \emph{decentralised} (\emph{choreographical}) approaches.
In the former category,
 programs generally specify tasks and relationships between tasks, 
 and these descriptions are used by a centralised entity
 to command the behaviour of the individual entities of the swarm.
By contrast,
 decentralised approaches
 do not rely on any centralised entity:
 each robot is driven by a control program 
 and the resulting execution is decentralised
 (e.g., based on interaction with neighbours, like in Meld~\cite{Meld2007}).
In this work, we focus on \emph{decentralised} solutions, for they support resilience and scalability by avoiding single-points-of-failure and bottlenecks.

In the general context of behaviour-based swarm design,
 researchers have pointed out various issues~\cite{DBLP:journals/swarm/BrambillaFBD13,DBLP:journals/scirobotics/TheraulazT20}
 like a general lack
 of \emph{top-down} design methods of collective behaviours
 (cf. the scientific issue of ``emergence programming''~\cite{varenne2015programming-emergence} and  ``self-organisation steering''~\cite{DBLP:journals/alife/GershensonTWS20}),
 the problem of formal verification and validation~\cite{DBLP:journals/csur/LuckcuckFDDF19},
 heterogeneity, 
 and operational/maintenance issues
 (e.g., scalability, adaptation, and security). 
\revA{
Specific challenges can also be found in the context of 
 specific kinds of swarm systems,
 such as (micro) aerial swarms~\cite{Abdelkader2021aerialswarms,%
Coppola2020microairswarms},
 specific domains,
 like agriculture~\cite{DBLP:journals/cea/AlbieroGUP22},
 or specific kinds of tasks,
 like simultaneous localisation and mapping (SLAM)~\cite{Kegeleirs2021slam}.
%
}

To address
 top-down swarm programming,
 an approach should provide
 the means to define and compose 
 blocks of high-level swarm behaviours. 
Regarding the kinds of blocks that can be provided,
 it is helpful to look at proposed taxonomies of collective/swarm behaviour.
In a prominent survey on swarm engineering~\cite{DBLP:journals/swarm/BrambillaFBD13},
 collective behaviours are classified into 
 (i) spatially-organising behaviours (e.g., pattern formation, morphogenesis),
 (ii) navigation behaviours (e.g., collective exploration, transport, and coordinated motion),
 (iii) collective decision-making (e.g., consensus achievement and task allocation),
 and
 (iv) others (e.g., human-swarm interaction and group size regulation).

\revA{Last but not least, the current literature 
 displays a quite sharp demarcation
 between techniques based on 
 formal specification methods for swarm behaviour~\cite{DBLP:journals/csur/LuckcuckFDDF19},
 which also promote verification,
 and more pragmatic approaches
 based on concrete and generally more usable \acp{dsl}.}
\revA{In a recent review on formal methods for swarm robotics engineering~\cite{DBLP:journals/csur/LuckcuckFDDF19}, 
 it is mentioned that two main shortcomings include (i) the toolchain and (i) the formalisation of the ``last step'' of turning the formal model into executable code.
Therefore, there is a need for approaches that suitably combine the value of formal methods and the practicality of programming approaches and \acp{dsl}.

In a nutshell,
 the possibility and opportunity of an 
 approach for \emph{formal-yet-practical 
 top-down behaviour-based design of decentralised swarm behaviour}
 provides the motivation for this work}.

\section{Background: Aggregate Computing}\label{sec:background} 

Aggregate computing~\cite{DBLP:journals/jlap/ViroliBDACP19}
 is a field-based coordination~\cite{DBLP:journals/pervasive/MameiZL04} and macroprogramming~\cite{Casadei2023} approach
 especially suitable to express
 the collective adaptive~\cite{DBLP:journals/sttt/NicolaJW20} and self-organising behaviour 
 of large groups of situated agents.
\revA{
To properly introduce the essentials of aggregate computing,
 we first present its system and execution models (\Cref{ac:model}),
 and then its programming model and constructs (\Cref{ac:programming}).
}
 
\subsection{System and execution model}\label{ac:model}
In aggregate computing,
 a system (also called an \emph{aggregate system}) 
 is modelled as a set of logical computing \emph{nodes} (also called \emph{devices}),
 where each node is 
 equipped with \emph{sensors} and \emph{actuators},
 and is  
 connected with other nodes
 according to some \emph{neighbouring relationships}.
This abstract logical model does not prescribe 
 particular technological solutions;
 instead,
 it uses minimal assumptions on the capabilities of devices (e.g., regarding synchrony, connectivity, and computing power).

The approach is generally used to program long-running control tasks that need several sensing, communication, computation, and actuation steps to be carried out.
Accordingly, the \emph{execution model} 
 is based on (or can be understood as) a repeated execution,
 by each device,
 of \emph{asynchronous sense--compute--interact rounds}---fundamentally mimicking self-organisation in biological systems~\cite{bonabeau1999swarmintelligence-book}.
For simplicity, we can consider each round to atomically consist of three steps:
\begin{itemize}
\item \textbf{Sense} -- the node's \emph{local context} is assessed, by sampling sensors and gathering the most recent (and not expired) message from any neighbour;

\item \textbf{Compute} -- the so-called \emph{aggregate program} is evaluated against the local context, producing an output (which can be used to describe actuations) and an internal output (invisible to programmers), called an \emph{export}, that contains the message to be sent to neighbours for coordination purposes;

\item \textbf{Interact} -- the export is sent to neighbours (logically, as a broadcast), and potential actuations can be performed.
\end{itemize}
\revA{Details such as scheduling policy, thresholds for message expiration, neighbouring relationship, and communication are not fixed by the model and may be tuned on a per-application basis.
This behaviour can be also perceived globally as s a single \emph{aggregate} machine 
  that executes the same program on all devices in parallel, 
  with the devices interacting with each other through the network.

\revB{
\paragraph{Augmented event structures as a model of system execution}
To capture the semantics of system execution, e.g., to reason about properties of computation such as self-stabilisation~\cite{DBLP:journals/tomacs/ViroliABDP18}, the framework of \emph{event structures}~\cite{DBLP:journals/tcs/NielsenPW81} can be used,
 following the interpretation approach of~\cite{DBLP:conf/coordination/AudritoBDV18}.
Accordingly, we consider a notion of an \emph{augmented event structure} (or event structure for short) $ \mathcal{E} = \langle E, \neigh d, p, t\rangle$ (see \Cref{fig:event-structure}), 
where 
\begin{itemize}
\item $E$ is a finite set of \emph{events} representing
 different rounds of execution, 
\item $\neigh \subseteq E \times E$ is the \emph{neighbouring relation} representing 
  message exchanges between events,
\end{itemize}
and the following augmentations provide additional information about the events:
\begin{itemize}
\item  $d : E \rightarrow \Sigma$ is a function that maps each event to a device identifier, 
\item $p : E \rightarrow \mathbb{R}^3$ is a function that maps each event to a position in a three-dimensional vector space, and
\item $t : E \rightarrow \mathbb{R}$ is a function that maps each event to time.
\end{itemize}
Intuitively, $p(e)$ and $t(e)$ represent altogether the spatio-temporal coordinates of the event.
Given a time interval $T\subseteq \mathbb{R}$ (e.g. $[t_1,t_2[$, or $[t_1,+\infty)$) we shall denote by $\mathcal{E}|_T$ the augmented event structure obtained by restricting $\mathcal{E}$ to all events whose time is within interval $T$.

%
%
In an event structure, we can define a \emph{field} $\phi: E \rightarrow V$ that maps each event to a value, which can be used to represent the evolution of a distributed property of the system---e.g., the field of temperatures perceived by mobile/stationary sensors spreads in the environment, or the field of movement directions (i.e., a field of vectors) feeding movement actuators.
Particularly, a \emph{field computation} over an augmented event structure with set of events $E$ is a function $f: \Phi_{in} \rightarrow \Phi_{out}$ where $\Phi_{in}$ and $\Phi_{out}$ are sets of fields sharing the same domain $E$. 

\begin{figure}
\includegraphics{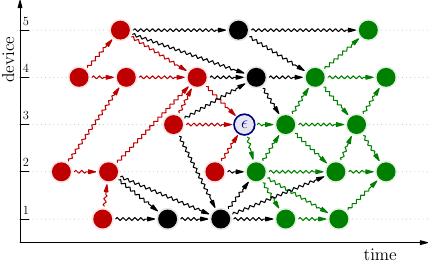}
\caption{\revB{An example of an event structure.
Each node and its incident edges represents a distinct sense--compute--interact round.
The ``sense'' step is given by the messages received by previous events, as well as any information the environment provides to it, e.g., according to its space-time location; notice that the state could be modelled as a self-message. The ``compute'' step is the data processing that it is supposed to happen at each event, mapping all context information to actuations and message data, which is delivered through the outgoing edges at the ``interact'' step.
Notice that the rounds of the same device are shown in the same row. Failure can be modelled through the loss of state, as shown for device $2$, where the second and third rounds are not connected.
Finally, notice that for any given event $\epsilon$, subsets of the event structure denoting its ``causal past'' (shown in red) and ``causal future'' (shown in green) can be determined through the transitive closure of the neighbouring relationship.
}}\label{fig:event-structure}
\end{figure}

Intuitively, in this context a program $P$ defines the sense-compute-interact behaviour at each device (namely, it is the single ``macroprogram'' executed in every device \cite{Casadei2023}): when applied to a certain event structure induces a certain field computation, and with a certain input field (inputs taken from sensors of various kinds) that computation produces an output field, formed by values at each event.
In general, such programs will be built so as to define the logic by which information gathered from the local context spreads from neighbourhood to neighbourhood, 
 and progressively gets integrated and transformed as it moves, 
 eventually converging towards proper local results that are globally coherent (e.g., of distributed sensing and actuation) 
 once environmental changes perturbing the system are incorporated.
}}
%
%
To understand how this aggregate program promotes collective adaptive behaviour, we now present the programming model.

\subsection{Programming model}\label{ac:programming}
%
%
%
%
On top of the field abstraction,
the \emph{field calculus}~\cite{AVDPB-TOCL2019}  (the base of aggregate computing)
 provides a minimal core language 
 which defines the primitives for 
 expressing ``space-time universal''~\cite{DBLP:conf/coordination/AudritoBDV18} distributed computations in terms of field manipulations.
Then, concrete languages like the Scala-internal \ac{dsl} \scafi{} ({\sc{}Sca}la {\sc{}Fi}elds)~\cite{DBLP:journals/softx/CasadeiVAP22,ACDV-LMCS2023} 
 can be used to actually develop aggregate programs.

The reader can refer to~\cite{ACDV-LMCS2023}
 for a full presentation of programming with \scafi{}.
Here, we briefly introduce its main language constructs and \revA{library building blocks: these will be the basis for encoding higher-level blocks swarm behaviour.
}

\subsubsection{Basic constructs} \revA{The minimal set of basic constructs covers state management, neighbourhood interaction, and functional application (also supporting behaviour branching).}

\revA{
\paragraph*{\textbf{Simple values and expressions.}}
Simple values and expressions can be interpreted both locally to one device, and globally as a field.
For instance:
}
\begin{lstlisting}
val threshold = 10
val deviceId = mid()
val random = scala.util.Random.nextInt(100)
\end{lstlisting}
\revA{Local value \lstinline|threshold| also denotes a static, uniform field  holding \lstinline|10| in every device.
Value \lstinline|deviceId| provides, locally, the identifier of the running device (as provided by built-in function \lstinline|mid|),
and, globally, the static field of device identifiers (since devices are assumed not to change their identifiers during execution).
Finally, value \lstinline|random| will locally change in each round, and so it denotes a generally non-static, non-uniform field of integers.
Notice that since \scafi{} is a Scala \ac{dsl},
 standard Scala constructs, values, and library functions can be used.
}

\paragraph*{\textbf{Construct \lstinline|rep|: stateful field evolution.}} Consider the following example.
\begin{lstlisting}
// def rep[T](init: T)(f: T => T): T
rep(0)(x => x+1) // type T=Int inferred
\end{lstlisting}
This purely local computation, when considered executed by all the devices in the system, yields a field of integers denoting 
 the number of rounds executed by each device.
This is obtained by applying function \lstinline|f| to the value computed the previous round (or \lstinline|init|, initially).

\paragraph*{\textbf{Construct \lstinline|foldhood|/\lstinline|nbr|: interaction with neighbours.}} Consider: 
\begin{lstlisting}
// def foldhood[A](init: => A)(acc: (A, A) => A)(expr: => A): A
// def nbr[A](expr: => A): A
foldhood[Set[ID]](Set.empty)(_++_){ Set(nbr(mid())) }
\end{lstlisting}
It yields, in each device, the set of identifiers of all its neighbours.
This is achieved by a purely functional fold over the collection of the singleton sets of neighbour identifiers, starting from the empty set, and aggregating using the set union operator (\lstinline|++|).
\revA{Notice that by-name type \lstinline|=> A|
 means the argument is passed unevaluated to the function. Therefore, the third argument to \lstinline|foldhood| (notice that, in Scala, singleton parameter lists can be denoted with braces as well as with parentheses) is not evaluated at the caller side but rather internally to the function.
}
Within the \lstinline|foldhood|, a \lstinline|nbr(e)| expression has the twofold role of sending and gathering the local value of \lstinline|e| to/from neighbours.
\revA{
Technically, argument \lstinline|expr| is evaluated fully for the running device, where the value of the expression within \lstinline|nbr| is kept for sharing, and once per each neighbour, where the \lstinline|nbr| expressions are substituted by the values shared by that neighbour.
}
Note that constructs \lstinline|rep| and \lstinline|foldhood|/\lstinline|nbr| can be combined to support the diffusion of information beyond direct neighbours.

\paragraph*{\textbf{Functional abstraction.}} New blocks can be defined with standard Scala functions:

\noindent\begin{minipage}{\textwidth}
\begin{lstlisting}[aboveskip=0.4cm]
def neighbouringField[T](f: => T): Set[T] = 
  foldhood[Set[T]](Set.empty)(_++_){ Set(nbr(f)) }

def neighbourIDs(): Set[ID] = 
  neighbouringField{ mid() }
  
def gradient(source: Boolean): Double = 
  rep(Double.PositiveInfinity){ distance =>
    mux(source) { 
      0.0 
    } {
      foldhoodPlus(Double.PositiveInfinity)(Math.min)(nbr{distance} + nbrRange)
    }
  }
\end{lstlisting}
\end{minipage}
\revA{
The latter block is called a \emph{gradient}~\cite{DBLP:journals/tomacs/ViroliABDP18} and implements the self-healing field of minimum distances from the source devices identified by Boolean field \lstinline|source|. It works as follows: \lstinline|rep| keeps track of the current gradient value \lstinline|distance| (initially, it is \lstinline|Double.PositiveInfinity|); with a \lstinline|mux(c)(t)(e)|, a purely functional selector that evaluates \lstinline|t| and \lstinline|e| and returns the former when \lstinline|c| is true and the latter otherwise,
 sources are given null distance (\lstinline|0.0|),
 and the other devices take as gradient value the minimum of the sum of a neighbour's gradient with the corresponding distance to the running device (as provided by neighbouring sensor \lstinline|nbrRange|).
The gradient is a basic pattern for implementing several self-organising behaviours~\cite{DBLP:journals/tomacs/ViroliABDP18}; for instance, information can spread or converge along the adaptive structure denoted by the gradient~\cite{DBLP:conf/saso/WolfH07}, or limited areas of influence may be obtained by truncating a gradient up to a certain distance threshold~\cite{DBLP:journals/fgcs/PianiniCVN21}.

One important thing to note is that each function can potentially encapsulate both the computation \emph{and} the communication of data (cf. \lstinline|nbr| expressions)
 needed to fully support a certain collective behaviour.
The other key aspect is the inherent adaptiveness that emerges by the combination of the execution model 
 (based on repeated sensing, computation, and interaction) and the program specification (which dictates how to algorithmically transform the input context into output decisions).
For instance, the gradient will progressively adapt in response to changes in the source set, topology (neighbourhoods and distance between neighbours), and current gradient values of neighbours, up to convergence once inputs cease to change (see \Cref{sec:acblocks} for more on this).
}

\paragraph*{\textbf{Construct \lstinline|branch|: splitting computation domains.}} Consider: 
\begin{lstlisting}
// def branch[A](cond: => Boolean)(th: => A)(el: => A)
branch(sense[Boolean]("hasTemperatureSensor")){
  val nearbyTemperatures: Set[Double] = 
    neighbouringField{ sense[Double]("temperature") }
  // ...
}{ noOp }
\end{lstlisting}
Here, computation is split into separate subsets of devices. Notice that neighbourhoods are restricted in each computation branch. So, in the first branch, it is ensured that only the neighbours with a temperature sensor are folded over.

\revB{
\subsubsection{Self-stabilising operators as main library constructs}\label{sec:acblocks}

A key feature of aggregate programming is the ability of identifying general-purpose constructs with desirable resiliency properties, such that their (functional) composition provably retains the same properties.
This paves the way for a methodology for the design of libraries such that all applications built on top guarantee such properties ``by construction''.

This approach is studied in \cite{DBLP:journals/tomacs/ViroliABDP18} for the notable case of so called ``self-stabilising'' computations, defined on top of the framework of augmented event structures as follows.
Inspired by the notation in \cite{ADT-ISOLA2024}, we denote an augmented event structure $\mathcal{E} 
$ 
as \emph{$(\Delta, \leadsto)$-adhering} (where $\Delta$ is a set of devices and $\leadsto$ a neighbouring relation for them) if:
\begin{itemize}
  \item (Topology) All events $e \in E$ follow the connection topology $\leadsto$.  That is, for all $e \in E$ with $d(e) = \delta\in\Delta$, the set of events influencing $e$, denoted $\{e': e' \neigh e\}$, is exactly the set $\{e_{\delta'} \mid \delta' \leadsto \delta\}$, where $e_{\delta'}$ is the most recent event on device $\delta'$ in $E$ happening before $e$. 
  \item (Fairness) For each device $\delta \in \Delta$, the set of events $\{e \in E \mid d(e) = \delta\}$ is infinite.
\end{itemize}
%
%
This notion captures system runs with a stable topology for device communications.
Conversely, we say that $\mathcal{E}$ is $(\Delta, \leadsto)$-fair if it is eventually $(\Delta, \leadsto)$-adhering, that is, if from some $t_0$ we have that projection $\mathcal{E}|_{[t_0,+\infty)}$ is $(\Delta, \leadsto)$-adhering; this variation captures the situation in which topology and communications may have transient changes, which however eventually disappear.
In a fair event structure, a field $\Phi$ is said to \emph{converge} to a limit $\Phi_{\text{lim}}: \Delta \rightarrow V$ if for all but finitely many events $e \in E$ with $d(e) = \delta$, $d(e) = d_{\text{lim}}(\delta)$, namely, for each device, $\Phi$ eventually associates to its events a constant value.
Finally, a field computation $f$ is \emph{self-stabilising} if, given any limit input data $\Phi^i_{\text{lim}}: \Delta \rightarrow V^i$ on a topology $\leadsto$, there exists a limit output data $\Phi^o_{\text{lim}}: \Delta \rightarrow V^o$ such that for every eventually $(\Delta, \leadsto)$-adhering augmented event structure $\mathcal{E}$ and input data $\Phi^i: E \rightarrow V^i$ with limit $\Phi^i_{\text{lim}}$, the output $\Phi^o: E \rightarrow V^o$ of the function $f$ on $\Phi^i$ has limit $\Phi^o_{\text{lim}}$. 
Informally, self-stabilisation entails that once topology and input data are fixed, computation eventually reaches a final stable configuration of output values, in spite of transient faults.
In Section~\ref{sec:formal-selfstab} we will show that this notion can entail resiliency in the behaviour of swarms as well.}

\revA{The \scafi{} library includes some general high-level operators} that are proved self-stabilising~\cite{DBLP:journals/tomacs/ViroliABDP18}. These will be leveraged in \Cref{sec:contrib} and hence are briefly described.
\begin{itemize}
\item \emph{\textbf{S}parse choice (leader election)~\cite{DBLP:conf/acsos/PianiniCV22}.} Block \lstinline|S(grain:Double):Boolean| can be used to yield a self-stabilising Boolean field which is \lstinline|true| in a sparse set of devices located at a mean distance \lstinline|grain|.

\item \emph{\textbf{G}radient-cast (distributed propagation)~\cite{DBLP:journals/tomacs/ViroliABDP18}.}
Block \lstinline[breaklines=true]|G[T](source:Boolean,value:T,acc:T=>T):T| is used to propagate \lstinline|value| from \lstinline|source| devices outwards along the gradient~\cite{DBLP:journals/tomacs/ViroliABDP18} of increasing distances from them, transforming the value through \lstinline|acc| along the way.

\item \emph{\textbf{C}ollect-cast (distributed collection)~\cite{DBLP:journals/cee/AudritoCDPV21}.} Block \lstinline[breaklines=true]|C[T](sink:Boolean,value:T,acc(T,T)=>T):T| is used to summarise distributed information into \lstinline|sink| devices, the \lstinline|value|s provided by devices around the system, while aggregating information through \lstinline|acc| along the gradient directed towards the sinks.
\end{itemize}
%
%

Examples further showing the compositionality of the approach are in \Cref{sec:contrib}.

\subsubsection{Aggregate computing for swarm programming}\label{ssec:ac-features-swarm}
\revA{
As covered in the following sections, we develop \MacroSwarm{} on top of aggregate computing.
This choice is motivated by peculiar features of aggregate computing (and its toolchain including \scafi{}) that make it particularly suitable for swarm programming.
We substantiate this statement 
 by briefly explaining, while synthesising from previous work, how such features help to address the challenges identified in \Cref{sec:context}.
}
\begin{itemize}
\item \emph{Top-down behaviour-based design.} It is promoted by the functional paradigm and the field abstraction~\cite{AVDPB-TOCL2019,DBLP:conf/ecoop/AudritoCDSV22},
 which together enable \emph{compositionality} and \emph{collective stance} in aggregate programming.
\item \emph{Scalability.} Since execution is fully decentralised and asynchronous, the approach is scalable to hundreds, thousands, and even more devices~\cite{DBLP:journals/eaai/CasadeiVAPD21}.
\item \emph{Formal approach.} Aggregate computing and \scafi{} are based on the field calculus~\cite{ACDV-LMCS2023,AVDPB-TOCL2019}, which enables formal analysis of programs and proofs of interesting properties like self-stabilisation~\cite{DBLP:journals/tomacs/ViroliABDP18} (more on this in \Cref{sec:formal-selfstab})
, universality~\cite{DBLP:conf/coordination/AudritoBDV18}, and others~\cite{DBLP:journals/jlap/ViroliBDACP19}.
\item \emph{Pragmatism.} Promoted by layers of abstractions, this is witnessed by open-source, maintained, concrete software artefacts like the \scafi{} \ac{dsl}~\cite{DBLP:journals/softx/CasadeiVAP22},  simulation platforms like Alchemist~\cite{DBLP:journals/jos/PianiniMV13} and {\sc{}ScaFi-Web}~\cite{DBLP:conf/coordination/AguzziCMPV21}\footnote{\url{https://scafi.github.io/web/}}, and the possibility to devise libraries of high-level functions~\cite{DBLP:journals/eaai/CasadeiVAPD21}.
\item \emph{Operational flexibility.} Concrete aggregate computing systems can be deployed and operated using different architectural styles~\cite{DBLP:journals/fi/CasadeiPPVW20} and execution policies~\cite{DBLP:journals/lmcs/PianiniCVMZ21}, supporting different technological and resource requirements.
\end{itemize}

\section{\MacroSwarm{}}
\label{sec:contrib}

\begin{figure}[t]
  \centering
  \begin{subfigure}[c]{0.4\textwidth}
    \includegraphics[width=\textwidth]{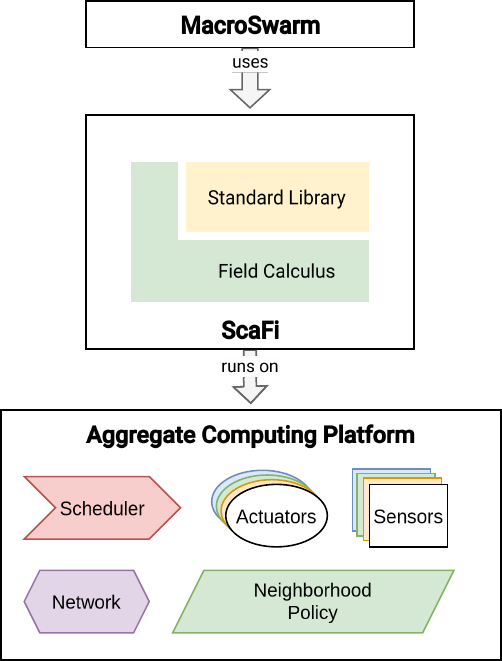}
    \caption{\MacroSwarm{} external architecture.}
    \label{fig:subfigure1}
  \end{subfigure}
  \hfill
  \begin{subfigure}[b]{0.8\textwidth}
    \includegraphics[width=\textwidth]{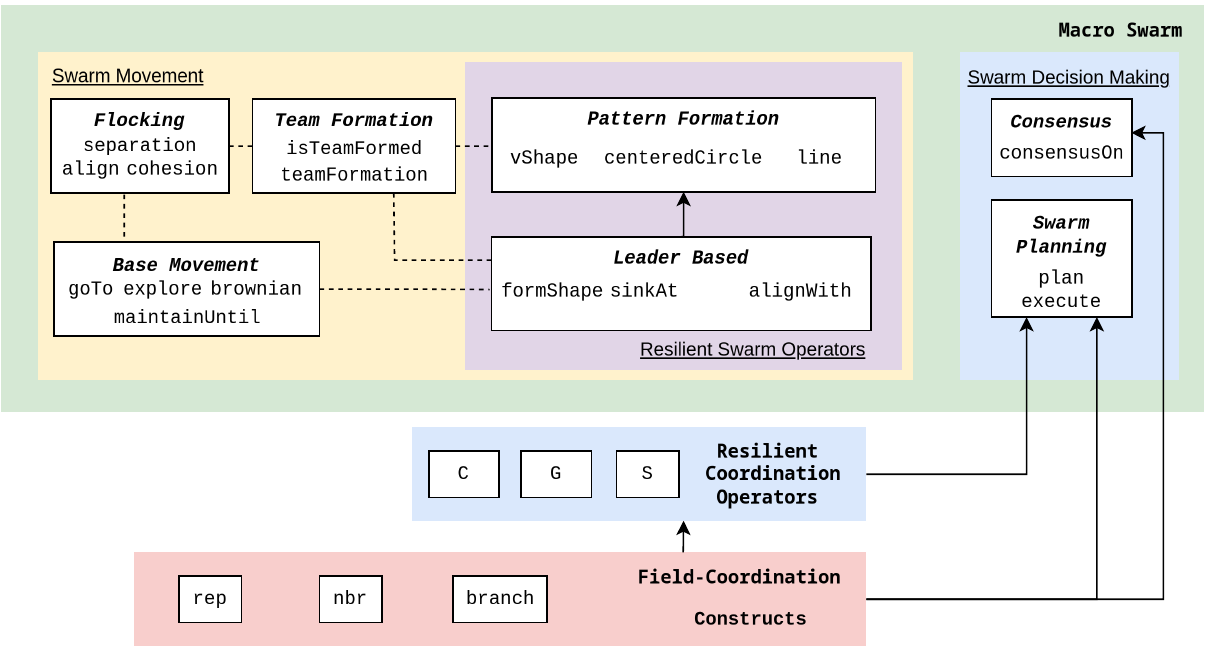}
    \caption{\ac{api} structure of \MacroSwarm{}.
    \revB{Solid lines represent strong dependencies (i.e., blocks that are used by other blocks), while dashed lines represent weak dependencies (i.e., blocks might be combined with other blocks to achieve a certain behaviour). 
    The \MacroSwarm{} (green rectangle) is based on both field-coordination constructs (red) and resilient operators (blue). The \ac{api} 
    is organised into modules, each one encapsulating a logically related set of behavioural blocks.}}
    \label{fig:subfigure2}
  \end{subfigure}
  
  \caption{Overall architecture of \MacroSwarm{}}
  \label{fig:architecture}
\end{figure}

This section presents the \MacroSwarm{} approach, architecture, and \ac{api}.

\revA{
\subsection{Architecture}\label{sec:macroswarm-arch}
The overall architecture of \MacroSwarm{} is shown in \Cref{fig:architecture}.
In particular, \Cref{fig:subfigure1} shows the ``external architecture'', namely how \MacroSwarm{} integrates and depends on other tools of the aggregate computing ecosystem.
Specifically, 
 \MacroSwarm{} has been implemented 
 as an extension of the \scafi{} aggregate programming language~\cite{DBLP:journals/softx/CasadeiVAP22,ACDV-LMCS2023} (cf. \Cref{ac:programming}),
 based on a library or \ac{api} of building blocks of swarm behaviour.
So, a \MacroSwarm{} program
 is essentially a \scafi{} program
 which in turn runs on an aggregate computing platform or middleware (i.e., an implementation of the execution model discussed in \Cref{ac:model}).

In \Cref{fig:subfigure2}, it is shown the organisation of the \MacroSwarm{} \ac{api}, namely its ``internal architecture''.
The \ac{api} is organised
 into multiple \emph{modules},
 each one encapsulating a logically related set of behavioural blocks.
The \ac{api} comprises 
 both general and highly reusable sets of coordination blocks (e.g., supporting information streams and leader election)
 and more swarm-specific sets of blocks (e.g., covering pattern formation or mobility patterns).
}

The key idea in the design of \MacroSwarm{}  
 lies in the representation of a swarm behavioural unit 
 as a function mapping sensing and parameter fields 
 to actuation fields (often, velocity vectors).

\revA{
\subsection{Actuation model}\label{subsec:act-model}

A main aspect that has been addressed revolves around the definition of an \emph{actuation model} for an aggregate computing system.
As explained in \Cref{ac:model}, 
 the devices of an aggregate computing system 
 undergo execution in rounds of sense--compute--act steps.
Therefore, we use the output of the \MacroSwarm{} program 
 to \emph{denote} what actuation has to be performed.
Then, the underlying platform is responsible for mapping actuation commands to actions for the raw (virtual or physical) actuators.
Moreover, to properly support actuation,
 coherently with modern robotic programming practice~\cite{robotics-for-programmers},
 in the \MacroSwarm{} \ac{api}, we separate the \emph{actuation intention} (i.e., the acting)
 from the \emph{actual actuation} (i.e., the control of raw actuators).

Therefore,
 the idea is that the output of each round
 sets, unsets, or revises (i.e. changes one or more parameters of) the actuation goal(s).
There are two \emph{modalities} of actuation that an actuation goal has to choose from:
\begin{itemize}
\item \emph{round-based}: the actuation goal is valid only until the end of the next round (at which point it has to be explicitly provided again to keep it valid);
\item \emph{long-standing}: the actuation goal is valid until revised or retracted.
\end{itemize}
According to the provided indication, the actual actuations 
 are performed as needed under-the-hood,
 possibly concurrently to the computation.
Unless explicitly indicated, the actuation goals use the round-based modality by default.

\revB{
Notice that this actuation model is purely functional (namely it produces an actuation goal as output but does not perform the actuation itself in the program evaluation) 
and is based on the assumption that the actuation is non-blocking and short in time.}
For certain applications, proper control of actuators may require multiple commands to be sent with low delay; in those cases, the application designer may interface directly with the platform and leave to the \MacroSwarm{} program only delay-tolerant actuation planning.
Indeed, being \MacroSwarm{} an extension of \scafi{}, which is a Scala-internal \ac{dsl}, then normal Scala code can be used for ad-hoc and integration logic.
}

\subsection{Movement blocks}\label{subsec:base}
These blocks control the movement of individual agents within the swarm. 
The simplest movement expressible 
 with \MacroSwarm{} is a collective constant movement (\Cref{fig:constant}), 
 described through a tuple 
 like \lstinline|Vector(x,y,z)|
 that devises the velocity vector of the swarm.
For instance, 
\begin{lstlisting}
Vector(2.5, 0, 0) // a constant field which is the same for all the agents
\end{lstlisting}
\revA{is a vector interpreted relatively to each device; the result is that all the devices will move to their right at a speed of 2.5 m/s.}
This vector must then be appropriately mapped 
 the right electrical stimulus for the underlying engine platform
 of the mobile robot of interest.
On top of this, 
 this module exposes several blocks to explore an environment. 
In particular, the \lstinline|brownian| block produces \revB{a velocity vector that simulates a Brownian motion.  
 In this model, nodes move in random directions with step sizes distributed around a scale $s$~\cite{DBLP:conf/antsw/DimidovOT16}.}
In addition to that simple logic, 
 there are movements based on \revA{absolute positioning systems like GPS. For instance, \lstinline|goTo| 
 yields a velocity vector that moves the system to eventually converge to a single location,
 and \lstinline|explore| 
 returns } a velocity vector that let the system explore a rectangle area defined through \lstinline|minBound| and \lstinline|maxBound|.
The last one is based on temporal blocks, 
  like \lstinline|maintainTrajectory| and \lstinline|maintainUntil|.
The former allows the systems to maintain a certain velocity for the time specified. 
 At that moment, a new velocity is generated according to the given strategy. 
The latter, instead, is used to maintain a certain velocity until a condition is met 
 (e.g., a target position is reached).
This module also exposes an \lstinline|obstacleAvoidance| block (\Cref{fig:obstacle-avoidance}), which creates a vector pointing away from obstacles.

Even if these blocks are quite simple, 
 it is still possible to combine them to create interesting behaviours. 
For instance, program 
\begin{lstlisting}
(maintainVelocity(browian()) + obstacleAvoidance(sense("obs"))).normalize
\end{lstlisting}
expresses a collective behaviour in which the nodes will explore the environment,
 while avoiding any obstacles perceived through a sensor. 
Notice how the composition is achieved by simply summing the computational fields produced by the sub-blocks.
Expression \lstinline|v.normalize| yields \lstinline|v| as a unit vector (of length 1), while keeping the same direction---useful when combining several vectors together.
A summary of the blocks exposed by this module is reported in the following listing:
\begin{lstlisting}
// Movement library
def brownian(scale: Double): Vector
// GPS Based
def goTo(target: Point3D): Vector
def explore(minBound: Point3D, maxBound: Point3D): Vector
// Temporal Based
def maintainTrajectory(trajectory: => Vector)(time: FiniteDuration):Vector
def maintainUntil(direction: Vector)(condition: Boolean): Vector
// Obstacle Avoidance
def obstacleAvoidance(obstacles: List[Vector]): Vector
\end{lstlisting}

\subsection{Flocking blocks}\label{subsec:flockig} 
In a swarm-like system, 
 it is often necessary to coordinate the movement of the entire swarm, 
 rather than just individual agents, to achieve emergent behaviours,
 and ensure that the nodes move cohesively, avoid collisions, 
 and strive to be aligned in a common direction. 
Therefore, in this module, 
 we have implemented the main blocks to support the \emph{flocking} of agents. 
Several models are available in the literature for this purpose.
 Particularly, \MacroSwarm{} exposes 
 the Vicsek~\cite{VicsekModeling1995}, 
 Cucker-Smale~\cite{CuckerSmaleModeling2007}, 
 and Reynolds (\Cref{fig:flock})~\cite{DBLP:conf/siggraph/Reynolds87} models. 
We have also exposed the individual blocks to implement Reynolds, 
 which are \lstinline|cohesion|, \lstinline|separation|, and \lstinline|alignment|. 
These blocks can be used individually by higher-level blocks to implement specific behaviours 
 (e.g., following a leader while avoiding collisions). 

Another essential aspect that emerges at this level is the concept of a \emph{variable neighbourhood}. 
Indeed, it may happen that the logical neighbourhood model used by aggregate computing 
 does not match the one used to coordinate the agents. 
 Thus, the node's visibility can be more \emph{restrictive} or \emph{extensive} 
 according to the neighbourhood model applied. 
In particular, in the case of Reynolds, 
 it is typical for the separation range to be different from that of alignment. 
Therefore, the flocking blocks accept a ``query'' strategy towards a variable neighbourhood.
 The main implementation of these queries are:
\begin{itemize}
  \item \lstinline|OneHopNeighborhood|: the same as the aggregate computing model;
  \item \lstinline|OneHopNeighborhoodWithinRange(radius: Double)|: it takes all the nodes in the neighbourhood within the given range.
\end{itemize}

The flocking models are typically described 
 by an iterated function in which the velocity at time $t+1$ depends on the velocity at time $t$.
Taking as an example the Vicsek rule, it is described as:
$ v_i(t + 1) = \frac{\sum_{j \in \mathcal{N}}v_j(t) }{|\mathcal{N}|} + \eta_i(t)$
where $\mathcal{N}$ is the neighbourhood of the node $i$ at time $t$, 
 $v_i(t)$ is the velocity of the node $i$ at time $t$, 
 and $\eta_i(t)$ is a random vector that models the noise of the model.
For this reason, 
 each block receives the previous velocity field as a parameter, 
 rather than encoding it internally within each block. 
This is because the previous velocities 
 may be influenced by other factors, 
 such as constant movements or a target position. 
Typical usage of this operator follows the following schema:
\begin{lstlisting}
rep(initialVelocity) { oldVelocity => flockingOperator(oldVelocity, ..) }  
\end{lstlisting}
For example, 
 the following program describes a collective movement 
 in which the nodes try to reach the position \texttt{(x,y)} while maintaining a distance of \texttt{k} meters from one another:
\begin{lstlisting}
rep(Point2D.Zero) {
  v => (goTo(Point2D(x, y)) + 
       separation(v, OneHopNeighbourhoodWithinRange(k))).normalize
}
\end{lstlisting}

\subsection{Leader-based blocks}\label{subsec:leader}
These blocks allow agents to follow a designated leader.
The idea behind leadership in swarm systems is that a leader 
 can act as a coordinator, influencing the followers that recognise it as such. 
In the context of aggregate computing, 
 leaders are typically defined as Boolean fields holding \lstinline|true| for leaders 
 and \lstinline|false| for non-leaders. 
Leaders can be predetermined (i.e., nodes with certain characteristics), 
 virtual (i.e., nodes that do not actually exist in the system but are simulated for collective movement steering),
 or chosen in space (e.g., using the \lstinline|S| block---see \Cref{sec:background}).
A leader can be thought of as creating an \emph{area of influence}, affecting the actions of its followers.

Currently, we have identified \lstinline|alignWithLeader|, \lstinline|sinkAt|, and \lstinline|formShape| (\Cref{fig:towards-leader}) 
 as essential blocks. 
The former propagates the leader's velocity throughout 
 its area of influence (e.g., via \lstinline|G|---see \Cref{sec:background}),
 with followers adjusting their velocity to it. 
However, sometimes it may also be desirable 
 to create a sort of attraction towards the leader, 
 so that the nodes remain cohesive with it. 
For this reason, the  \lstinline|sinkAt| block creates a computational field 
 in which nodes tend to move towards the leader. 
\revB{ 
Finally, the \lstinline|formShape| block allows the nodes to form a certain shape 
 around the leader.
In such case the formation is not necessarily a fixed shape, 
 but rather a shape that computed by the leader and propagated to the followers.
Particularly, this block collects the positions of the followers and computes the centroid of the shape. 
Then, it computes the relative position of each follower with respect to the centroid, 
 and finally, it computes the position of the follower with respect to the leader. 
}
These blocks are useful for higher-level blocks, 
 such as those associated with the creation of teams or spatial formations.
\revB{
  Furthermore, as they rely entirely on the resilient operators detailed in \Cref{sec:acblocks}, 
  they are inherently self-stabilising. 
  We highlight them as ``resilient swarm operators'', with a more thorough discussion of the implications of self-stabilising properties in this scenario provided in \Cref{sec:formal-selfstab}.
} 
\subsection{Team formation blocks}\label{subsec:team}
\begin{figure}[t]
\centering
\begin{subfigure}{0.32\textwidth}
  \centering
  {\includegraphics[width=\textwidth]{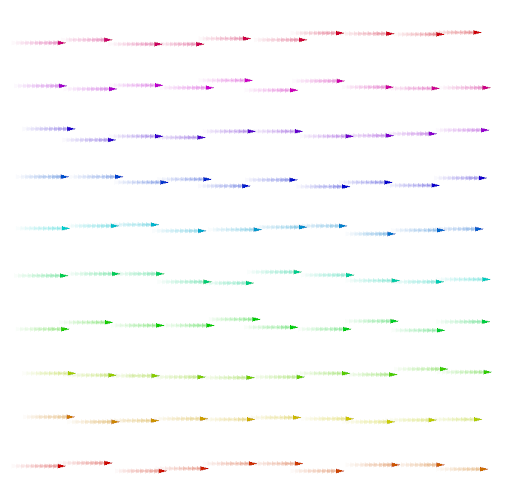}}
  \caption{}
  \label{fig:constant}
\end{subfigure}
\hfill
\begin{subfigure}{0.32\textwidth}
  \centering
  {\includegraphics[width=\textwidth]{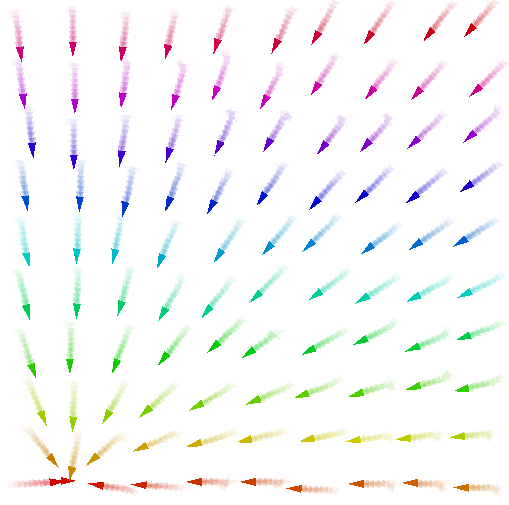}}
  \caption{}
  \label{fig:towards-leader}
\end{subfigure}
\hfill
\begin{subfigure}{0.32\textwidth}
  \centering
  {\includegraphics[width=\textwidth]{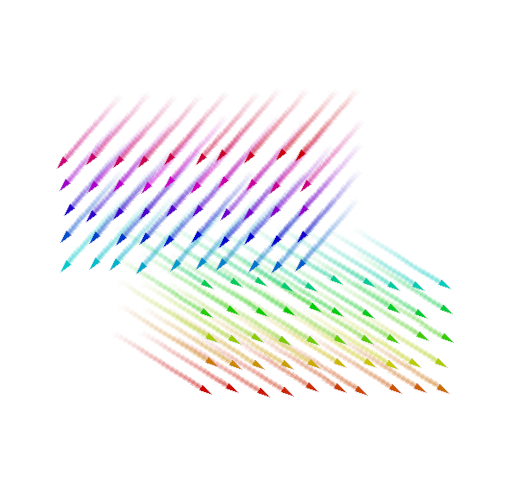}}
  \caption{}
  \label{fig:team}
\end{subfigure}
\hfill
\begin{subfigure}{0.32\textwidth}
  \centering
  {\includegraphics[width=\textwidth]{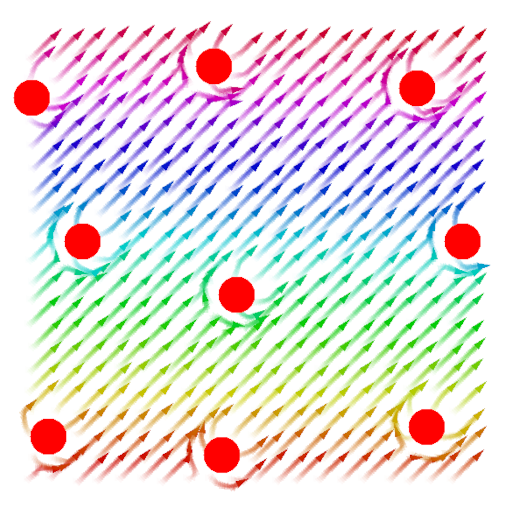}}
  \caption{}
  \label{fig:obstacle-avoidance}
\end{subfigure}
~
\begin{subfigure}{0.32\textwidth}
  \centering
  {\includegraphics[width=\textwidth]{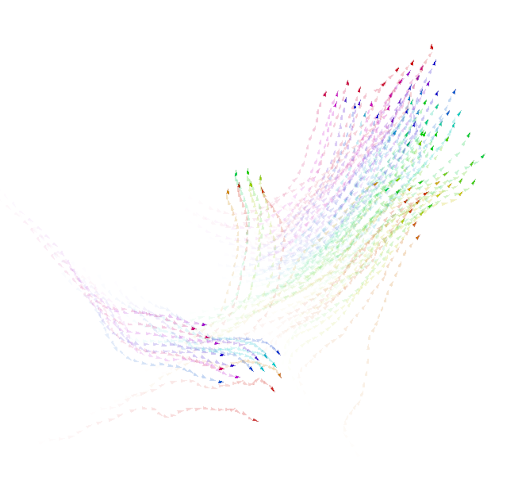}}
  \caption{}
  \label{fig:flock}
\end{subfigure}
\caption{Overview of swarm behaviours expressible with \MacroSwarm{}.}\label{fig:movement-overview}
\end{figure}

These blocks allow agents to form \emph{teams} or sub-groups within the swarm,
 useful e.g. for work division
 or situations requiring intervention by few agents.
In general, the formation of a team creates a ``split'' in the swarm logic, 
 conceptually creating multiple swarms with potentially different goals (cf. \Cref{fig:team}).
One way to create teams is by using the \lstinline|branch| construct (see \Cref{sec:background}). 
For example, the following program,
\begin{lstlisting}
def alignVelocity(id: Int) = 
  alighWithLeader(id == mid(), rep(browian())(x => x)
branch(mid() < 50) { alignVelocity(0) } { alignVelocity(50) }
\end{lstlisting}
creates two groups, each of which follows a certain velocity dictated by the leaders (0 and 50).

Other times, one needs to create teams based on the spatial structure of the network or when certain conditions are met. 
The \lstinline|teamFormation| block supports this scenario. 
By internally using \lstinline|S|, it allows for the creation of teams based on certain spatial constraints expressed through parameters \lstinline|intraDistance| (i.e., the distance between team members) and
 \lstinline|targetExtraDistance| (i.e., the size of the leader's area of influence). 
It is also possible to create teams based on predetermined leaders, denoted explicitly by Boolean fields.
Moreover, since team formation may take time to complete, or require conditions to be met (e.g., that at least $N$ members are present, or that the minimum distance between all nodes is less than a certain threshold),
 we also parameterise \lstinline|teamFormation| by a \lstinline|condition| predicate. 
An example of built-in predicate is \lstinline|isTeamFormed|, which verifies that each node under 
 the influence of the leader has the \lstinline|necessary| number of neighbours
 within a \lstinline|targetDistance| radius.
An example is as follows.
\begin{lstlisting}
teamFormation(targetIntraDistance = 30, // separation
  targetExtraDistance = 300, // influence of the leader
  condition = leader => isTeamFormed(leader, targetDistance = 40)
).velocity // use the velocity vector to create the Team
\end{lstlisting}
Each team must refer to a single leader, 
 who can coordinate the associated nodes 
 (using the APIs exposed by the \textbf{Leader-based block}). 
In particular, to execute a certain behaviour within a team, 
 the \lstinline|insideTeam| method must be used. 
 Given the ID of the leader to which a node belongs, 
 this method can define the movement logic relative to that leader.
For instance, this code aligns the followers with a velocity generated by a leader, 
\begin{lstlisting}[xrightmargin=-3.4pt]
team.insideTeam{leader => alignWithLeader(leader)(rep(brownian())(x => x))}
\end{lstlisting}

\subsection{Pattern formation blocks}\label{subsec:pattern}
Team formation blocks can be used to create groups of agents with certain characteristics.
However, sometimes we are also interested in the \emph{spatial structure} of the group. 
In swarm behaviours, the spatial structure of the swarm can be instrumental in performing certain tasks (e.g., coverage or transportation).
In \MacroSwarm{} some of the most idiomatic spatial structures are available. 

\revB{The implementation is based on \lstinline|formShape| and works as follow}: first of all, the formation of structures is based on the presence of a leader that
collects the hop-by-hop distances 
 of their followers (leveraging \lstinline|G| and \lstinline|C|) and
 sends them a direction in which they should go to form the required structure (using \lstinline|G|).

\begin{figure}[t]
  \centering
  \includegraphics[width=1.0\textwidth]{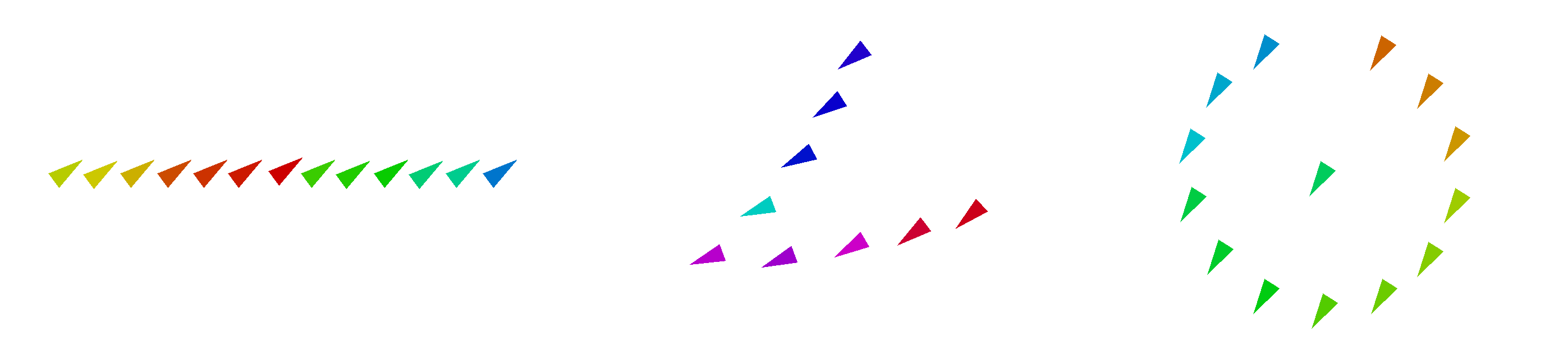}
  \caption{Examples of the supported patterns. From left to right: 
   line formation, v-like formation, and
   circular formation.
  }
   \label{fig:formations}
\end{figure}

The structures currently supported (\Cref{fig:formations}) are v-like shapes (\lstinline|vShape|), lines (\lstinline|line|), and circular formations (\lstinline|centeredCircle|). 
 These structures are both self-stabilising and \emph{self-healing}: if there is a disturbance 
 of the structure, 
 the group tends to reconstruct itself and return to a stable structure. 
Additionally, it is assumed that the leader has his own speed logic. 
 In this way,
 the group will follow the leader maintaining the chosen structure. 
  
\subsection{Swarm Planning blocks}\label{subsec:planner} 

With the previous blocks available, there is a need for a handy mechanism to express a series of \emph{plans} that 
 change over time and move the swarm towards different targets. 
For this reason, \MacroSwarm{} also exposes the concept of \emph{swarm planning}. 
The idea is to express a series of plans (or missions)
 defined by a \emph{behaviour} 
 (i.e., the logic of production of a velocity vector) 
 and a \emph{goal} (defined as a boolean predicate condition). 
At any given time, the swarm will be executing a certain sub-plan, 
 which will be considered complete only when the boolean condition is satisfied. 
At this point, the swarm will follow the next objective described by the overall plan.
The exposed API allows for the creation of these collective plans in the following way:
\begin{lstlisting}
execute.once {
  plan(goTo(goalOne).endWhen(isClose(goalOne)),
  plan(goTo(goalTwo).endWhen(isClose(goalTwo)),
}.run() // will trigger the execution of the plan
\end{lstlisting}
This snippet creates a plan 
 in which the nodes will first go to \lstinline|goalOne|, 
 and once reached (\lstinline|isClose| verifies that the node is close enough to the point passed), 
 it will move on to the next objective \lstinline|goalTwo|.
Since it is specified that the mission is executed \lstinline|once|, 
 after the completion of the last plan, the group will stop moving.
To make the group repeat the plan, 
 the \lstinline|repeat| method can be used instead of \lstinline|once|.
Note that there is no coordination between agents in the above code, 
 but you can enforce it using lower-level blocks (e.g., flocking or team-based behaviours).
For example, \MacroSwarm{} enables describing a swarm behaviour where:
(i) a group of nodes gathers around a leader,
(ii) the leader brings the entire group towards the \lstinline|goalOne|,
(iii) the leader brings the entire group towards the \lstinline|goalTwo|.
This can be described using the following code:
\begin{lstlisting}[xrightmargin=-5pt]
execute.once( // if it is repeated, you can use `repeat'
  plan{sinkAt(leaderX)}.endWhen{isTeamFormed(leaderX, targetDistance=100)},
  plan(goTo(goalOne)).endWhen{ G(leaderX, isClose(goalOne), x => x)},
  plan(goTo(goalTwo)).endWhen{ G(leaderX, isClose(goalTwo), x => x)},
).run()
\end{lstlisting}
The use of \lstinline|G| in this way is a recurrent pattern, 
 and in \scafi{} it is exposed through the \lstinline|broadcast[T](center: Boolean, value: T): T| block.
 
\subsection{Consensus}\label{subsec:consensus}
\revB{
Consensus, within the context of swarm robotics, 
refers to the ability of a group of robots to reach agreement on a common decision or choice through distributed interactions.
This collective behaviour enables a swarm of robots to converge on a single decision among several alternatives, 
typically optimising the performance of the system as a whole. 
Consensus achievement is crucial for cohesive action despite limitations in individual robot capabilities,   
particularly when the optimal choice is dynamic, 
not immediately obvious due to limited perception, 
or requires coordinated actions like movement, 
task allocation, 
or environmental monitoring.
Two main classes of swarm consensus exist, categorised by communication methods, direct and indirect communication.
In direct \emph{communication approaches}, robots explicitly exchange information about their preferred choices or relevant data (e.g., voting, sharing sensor readings, negotiating preferences). 
Examples include Gutiérrez et al.~\cite{DBLP:journals/nca/GutierrezCMMD10}, 
where robots share information about foraging areas, 
and Montes de Oca et al.~\cite{DBLP:journals/swarm/OcaFSPBD11}, 
where robots vote on a path.
Whereas in \emph{indirect communication} robots infer the swarm's preference through indirect cues like spatial distribution or density. 
This resembles stigmergy in natural systems. 
Examples include Garnier et al.~\cite{DBLP:conf/ecal/GarnierJJGACT05,DBLP:journals/adb/GarnierGAJT09},
 where robots achieve consensus on shelter selection by observing spatial distribution, 
Wessnitzer and Melhuish~\cite{DBLP:conf/ecal/WessnitzerM03}, 
where robots decide based on proximity to a target, 
and Parker and Zhang~\cite{DBLP:journals/ijrr/ParkerZ11} using a quorum-sensing inspired mechanism.
  
In \MacroSwarm{} applications, achieving consensus is crucial for coordinated actions across the potentially large and dispersed robot collective. 
\MacroSwarm{} addresses this need by providing the \lstinline|consensus| block, 
allowing nodes to converge on a common value based on local and neighbourhood preferences. 
Among the implemented consensus algorithms available, \MacroSwarm{} implements swarm consensus \cite{DBLP:journals/swarm/ZhengL23}, 
a simple yet effective direct communication approach where each node maintains a preference distribution over available options. 
This distribution is iteratively updated by incorporating the exhibited decisions 
(highest preference options) 
and their associated certainty levels 
(calculated based on the entropy of the preference distribution) 
received from neighbouring nodes, eventually leading to convergence towards a shared decision with high certainty across the swarm.
The algorithm API proceeds as follows:
}
\begin{lstlisting}
def consensus(preferences: List[Double], neighbourhoodWeight: ID => Double): Int
\end{lstlisting}
\revA{
Where \lstinline|preferences| is the list of preferences of the node, 
 and \lstinline|neighbourhoodWeight| is a function that returns the weight of the neighbour with the given ID.
It returns the index of the preference that the node has converged to.
}

\revB{
\section{On resiliency of pattern formation}\label{sec:formal-selfstab}

Formally assessing resiliency results of swarming algorithms for large and dynamic set of mobile drones as supported in \MacroSwarm{} is extremely difficult, due to the intrinsic intractability of the problem---large and unbounded number of states, noisy environment, loss of messages, and so on (see \Cref{rw:properties} for more details about this problem and related works).
However, aggregate computing provides some support for checking and enacting resiliency properties, e.g., by the notion of self-stabilisation as described in \Cref{sec:acblocks}.
Namely, the whole pattern formation component of the library is built on top of blocks \texttt{S}, \texttt{G} and \texttt{C} (cf. \Cref{sec:acblocks}), hence it is self-stabilising.
In this section we elaborate on this aspect, and discuss the implications of self-stabilisation in the context of mobile devices.
We shall see that a self-stabilising computation aimed at pattern formation actually implies convergence of a swarm of devices to that pattern, provided that the speed to computation and communications is high enough with respect to the rate of movement of devices.

\subsection{Field computations for movement and pattern formation}

We start formally framing \MacroSwarm{} algorithms (and executions) for pattern formation in the context of augmented event structures.

We call a \emph{formation pattern} (or pattern for short) $F$ a set of positions in space including the origin $0 \in F$. 
Let $\Delta$ be the set of devices and $P = \{p(\delta) : \delta\in\Delta\}$ be the set of their positions. 
We say that the devices $\Delta$ are in formation $F$ with respect to an anchor device $\delta_a \in \Delta$ 
and an error tolerance $\epsilon > 0$ if their positions match $F$ modulo error $\epsilon$ and a translation given by vector $p(\delta_a)$.
Namely, there should be a ``formation mapping'' $\mu:\Delta\mapsto F$ such that $\mu(\delta_a) = 0$ and for every $\delta \in \Delta$, 
$|p(\delta) - \mu(\delta) - p(\delta_a)| \leq \epsilon$. 

Formations, and any other swarming algorithm, are achieved by field computations producing at each device a movement vector (or versor), feeding an actuator that controls the actual movement of the device.
To capture this mechanism, we first denote as \emph{movement field} a field $\Phi_m:\Delta\mapsto \texttt{V}_0$ where $\texttt{V}_0$ is the set of versors in space, including vector $0$.
Then, a \emph{movement field computation} is a function $f: \Phi_{in} \rightarrow \Phi_{m}$ producing a movement field.
We say that a movement field computation ``resiliently computes a formation $F$ (with mapping $\mu$, anchor device $\delta_a\in\Delta$, and error approximation $\epsilon$)'' if it is self-stabilising and converges to a limit movement field such that the output at any $\delta$ is a versor directed towards a position that is distant less than $\epsilon$ than the ideal \emph{target position} $\mu(\delta) - p(\delta_a)$.

\subsection{A quasistatic approximation}

A technical problem in translating self-stabilisation results in the context of movement computations is that when movement of devices is continuous, topology is not guaranteed to eventually freeze as the definition of self-stabilisation actually requires: hence, in a continuously moving system, convergence may generally fail to be achieved even if the implementation of algorithms is self-stabilising.
%
%
At a closer look, however, we note that mobile systems like swarms always feature a trade-off between speed of computation/communication and rate of change of positions: on the one hand, if the topology of devices (namely their position and neighbourhood relation) changes too quickly (e.g., because of noise in movement and because of other transient effects) clearly the configuration of devices will hardly reach the desired formation; on the other hand, if we can assume computation/communication (i.e., field computations) be quick enough to reach convergence before significant changes in topology actually occurs, then we may expect to correctly steer the swarm to the sought configuration.
Formally identifying the threshold of convergence involves investigation of real-time aspects of the selected algorithms, but this is open research outside the scope of this paper---the recent work in \cite{ADT-ISOLA2024} draws a roadmap.
In this paper we can still state a result concerning the eventual reachability of correct formation in the case computation/communication is fast enough with respect to the rate of change.

To do so, we rely as baseline on the well-known \emph{quasistatic approximation} notion used in physics, which assumes that a big-step evolution is made by a sequence of small-step state changes, each state being in equilibrium.

Specifically, given a self-stabilising field computation $f$, we say that an augmented event structure $\mathcal{E}$ over devices $\Delta$ performs a \emph{quasistatic evolution with respect to $f$} if there exists an increasing sequence of points in time $t_0, t_1, t_2,\dots$ such that for each $i$: (i) the projection $\mathcal{E}|_{[t_i, t_{i+1}[}$ is $(\Delta, \leadsto)$-adhering for some $\leadsto$, and (ii) $f$ reaches convergence on $\Delta$ before $t_{i+1}$.
This notion resembles quasistatic approximation in that system topology changes slowly, namely only at times $t_0, t_1, t_2,\dots$, and after a change the system goes back to a stable limit behaviour (an equilibrium) before the next change.
Note that in one such quasistatic evolution, each device can still move infinite many-times, but does so only at time $t_i$ for some $i$---more specifically, at the first event happening at a time $t\geq t_i$.

\subsection{Convergence guarantee in the quasistatic approximation}

We can hence state the following result of ``convergence to a formation in quasistatic evolutions''.
Consider a movement field computation $f$ resiliently computing formation $F$ with mapping $\mu$ and error $\epsilon$, and run it over an augmented event structure $\mathcal{E}$ performing quasistatic evolution with respect to $f$.
In such a system, we noted that each device $\delta$ is allowed to move infinitely many times: we assume each time it moves, let it be at event $e$, it does so in the direction of the output versor computed by $f$ at $e$, and with a length of movement that reduces its distance from its target point of at least a fixed distance $d_{min}>0$, or directly going to the target point otherwise.
Under these hypotheses, the set of devices eventually reach formation $F$ (with error $\epsilon$).

Note that this statement can easily be shown to hold, since with quasistatic evolution a device is moved only when field computation $f$ reached its limit behaviour, and for a movement field computation this gives for that device a correct versor by definition: moving the device along that direction necessarily decreases the overall sum of distances of devices to their target position until reaching the target formation within error $\epsilon$ at each device.

Though the ``quasistatic approximation'' is an ideal that hardly fits several interesting swarming scenarios, it still gives an important baseline.
In fact, it is then easy to show that a much more realistic notion of ``near-equilibrium evolution'' would guarantee convergence as well: namely, it is not needed that movement field computation reaches convergence to the expected exact versors for devices be moved, instead, it is sufficient that a reasonable approximation of such versors is obtained just before moving, provided these movements reduce the distance from target in all cases but for a finite number of times, i.e., eventually.
Similarly, movement of devices may be subject to errors due to noise or external intervention, again provided these movements reduce the distance from target in all cases but for a finite number of times.
As a matter of fact, the proposed algorithms typically rather quickly provide proper versors in spite of change, hence leading to resiliency in pattern formation as evaluated by simulation in next section.
}

\section{Evaluation}
\label{sec:eval}

\revA{To validate the proposed approach and \ac{api} we extensively verify the main blocks proposed and then define a simulated \emph{find-and-rescue} case study,
to show 
the ability of \MacroSwarm{} to express complex swarm behaviours (\Cref{subsec:case-study}).} 
Then, we discuss the results of the case study and the applicability of the proposed approach in real-world scenarios (\Cref{subsec:discussion}).
The simulations are public available at \url{https://doi.org/10.5281/zenodo.10529068}.
\subsection{Block verification}
\label{subsec:block-verification}

\subsubsection{Evaluation Goals}
\revB{
This section validates the main components of the \MacroSwarm{}, focusing on \emph{pattern formations} and \emph{consensus} mechanisms. We address the following goals:

\begin{itemize} 
  \item \textbf{G1: Correctness.} Verify the correct implementation of both pattern formation components (i.e., by assessing the convergence of drones to the expected formation) and consensus mechanisms (i.e., by evaluating the convergence of drones to a common decision). 
  \item \textbf{G2: Resiliency.} Assess the resilience of these blocks by evaluating the drones' ability to maintain formation/decision in the presence of adversarial events. 
\end{itemize}

For \textbf{G1}, we conduct multiple simulations with varying initial drone positions (i.e., different dynamical histories deforming the lattice topology) 
to assess convergence to expected values. 
Due to the emergent nature of these computations, transient phases may exhibit invalid output configurations. 
Therefore, simulations are run sufficiently long to ensure eventual convergence to the expected configuration.
For \textbf{G2}, we evaluate resilience by introducing adversarial events, specifically \emph{message loss}, \emph{perception position errors}, and \emph{node failures}.
Message loss is simulated by randomly dropping messages between nodes with probability $D$. 
Perception position errors are simulated by adding Gaussian noise to node positions: 
$$ x_p = x + \mathcal{N}(0, P^2), \quad y_p = y + \mathcal{N}(0, P^2) $$ 
where $(x_p, y_p)$ are the perceived coordinates, $(x, y)$ the real coordinates, 
and $\mathcal{N}(0, P^2)$ represents Gaussian noise with mean 0 and variance $P^2$.
Node failures are simulated by randomly removing $K$\%
of the nodes at a given time $T_k$
}
\subsubsection{Simulation Setup and Scenario}
Our simulation toolchain is based on the \emph{Alchemist} simulator~\cite{DBLP:journals/jos/PianiniMV13} and its Alchemist-ScaFi integration~\cite{DBLP:journals/softx/CasadeiVAP22}.

Each simulation consist of a network of drones configured in a partially deformed 2D lattice composed of 7 rows and 7 columns (for a total of 49 drones), 
 covering an area of 1000x1000 meters. 
 Each drone within this network had a communication range of 200 meters. 
A central leader drone is designated within the lattice to test the drones' capacity for pattern formation around this leader 
 and to ensure the reproducibility of the resultant formations.
\revB{
Each simulation run lasts for 4000 seconds, 
 during which the drones operated at a maximum velocity of 20km/h. 
To ensure the statistical robustness of our findings,
 we replicated each experiment 16 times for each configuration (considering the free variable $D$, $P$, and $K$, see \Cref{tab:free_variables} for details)
 and computed the average results. 
 These repetitions were conducted with the drones starting from randomised positions relative to the lattice to eliminate any biases.
At $T_k=2000$ seconds we introduce the node failures event (in the case of pattern formation)
 removing K\% of the nodes from the lattice.
Finally, we have also tried different lattice configurations to ensure the scalability of the proposed approach (8x8 and 9x9 lattices).
} 
\begin{table}[h]
  \centering
  \caption{Free Variables in Resilience Evaluation (G2)}
  \label{tab:free_variables}
  \begin{tabular}{lll}
  \hline
  \textbf{Variable} & \textbf{Description} & \textbf{Values} \\ \hline
  $D$ & Message Loss Probability & $\{0, 0.1, 0.2, 0.4, 0.7 \}$ \\
  $P$ & Standard Deviation of Position Noise & $\{0, 1, 5, 10 \}$ (Units: meters) \\
  $K$ & Percentage of Node Failures & $\{0, 0.1, 0.2, 0.3 \}$ \\
  $T_k$ & Time of Node Failure Event & 2000 seconds \\
  \hline
  \end{tabular}
\end{table}

\revB{
\subsubsection{Metrics}
To evaluate the correctness of the pattern formation,
 we mainly use the error in the formation's shape and the time to reach the desired formation.
This error is computed as the number of drones that are not in the desired formation, for a given time step $t$.
To evaluate if a drone is in the desired formation, we compute the distance between the drone and the desired position, 
 and we consider the drone in the formation if this distance is less than a threshold $\epsilon$ (following the discussion about the quasistatic approximation in \Cref{sec:formal-selfstab}).
Formally, the error at time $t$ is computed as:
$$
\text{error}(t) = \sum_{i=1}^{N} \mathbb{I}(\|p_i(t) - p_i^* \| > \epsilon)
$$
where $N$ is the number of drones, 
$p_i(t)$ is the position of drone $i$ at time $t$, 
$p_i^*$ is the desired position of drone $i$ (computed with a perfect communication),
and $\mathbb{I}$ is the indicator function.
This metric, however, does not give any indication about how ``close'' the drones are to the desired formation.
For this reason, we also compute specific metrics for each pattern formation.
For the V-shape formation, we compute the average angular alignment of the lead drone in relation to the others over time. It gives an indication of how close the drones are to the desired angle.
Formally, let $D = {d_1, d_2, ..., d_n}$ be the set of $n$ drones in the V-shape formation, where $d_1$ is the lead drone.
Let $\phi_{t}$ be the angle between the position of the lead drone $d_1$ at time $t$ and the position of drone $d_i$ at time $t$, 
namely $\phi_{t} = \arctan\left(\frac{y_i(t) - y_1(t)}{x_i(t) - x_1(t)}\right)$. 
This is then averaged for the whole system, namely

$$\alpha(t) = \frac{1}{n-1}\sum_{i=2}^n \phi_{t}$$

For the separation formation, 
we compute the average distance of the nearest 4 drones to the other drones over time. 
It gives an indication of how close the drones are to the desired distance. 
For each drone $d_i$ at time $t$, 
we find the four nearest drones excluding itself, denoted by $N_i(t)$. The average distance of the nearest 4 drones to drone $d_i$ at time $t$ is: 
$\overline{\delta}^N_i(t) = \frac{1}{4} \sum_{d_j \in N_i(t)} \rho(p_i(t), p_j(t))$. 
The overall average distance for all drones at time $t$ is: $$\overline{\delta}(t) = \frac{1}{n} \sum_{i=1}^n \overline{\delta}^N_i(t)$$

For the line formation, 
we compute the average vertical variation of the lead drone in relation to the others over time. 
It is computed as the average of the vertical distance between the lead drone and the other drones. 
Let $y_i(t)$ be the vertical position of drone $d_i$ at time $t$, the average vertical variation at time $t$ is: 
$$\overline{v}(t) = \frac{1}{n-1}\sum_{i=2}^n |y_i(t) - y_1(t)|$$

Finally, for the circle formation,
we compute the average distance of the lead drone in relation to the others over time.
It gives an indication of how close the drones are to the desired radius.
For each drone $d_i$ at time $t$,
we compute the distance between the lead drone and the drone $d_i$, denoted by $\rho(p_1(t), p_i(t))$.
The overall average distance for all drones at time $t$ is:
$$\overline{\rho}(t) = \frac{1}{n-1} \sum_{i=2}^n \rho(p_1(t), p_i(t))$$
}

\subsubsection{Results}\label{subsec:results-shapes}
\begin{figure}
  \centering
  \begin{subfigure}[b]{0.3\textwidth}
    \includegraphics[width=\textwidth]{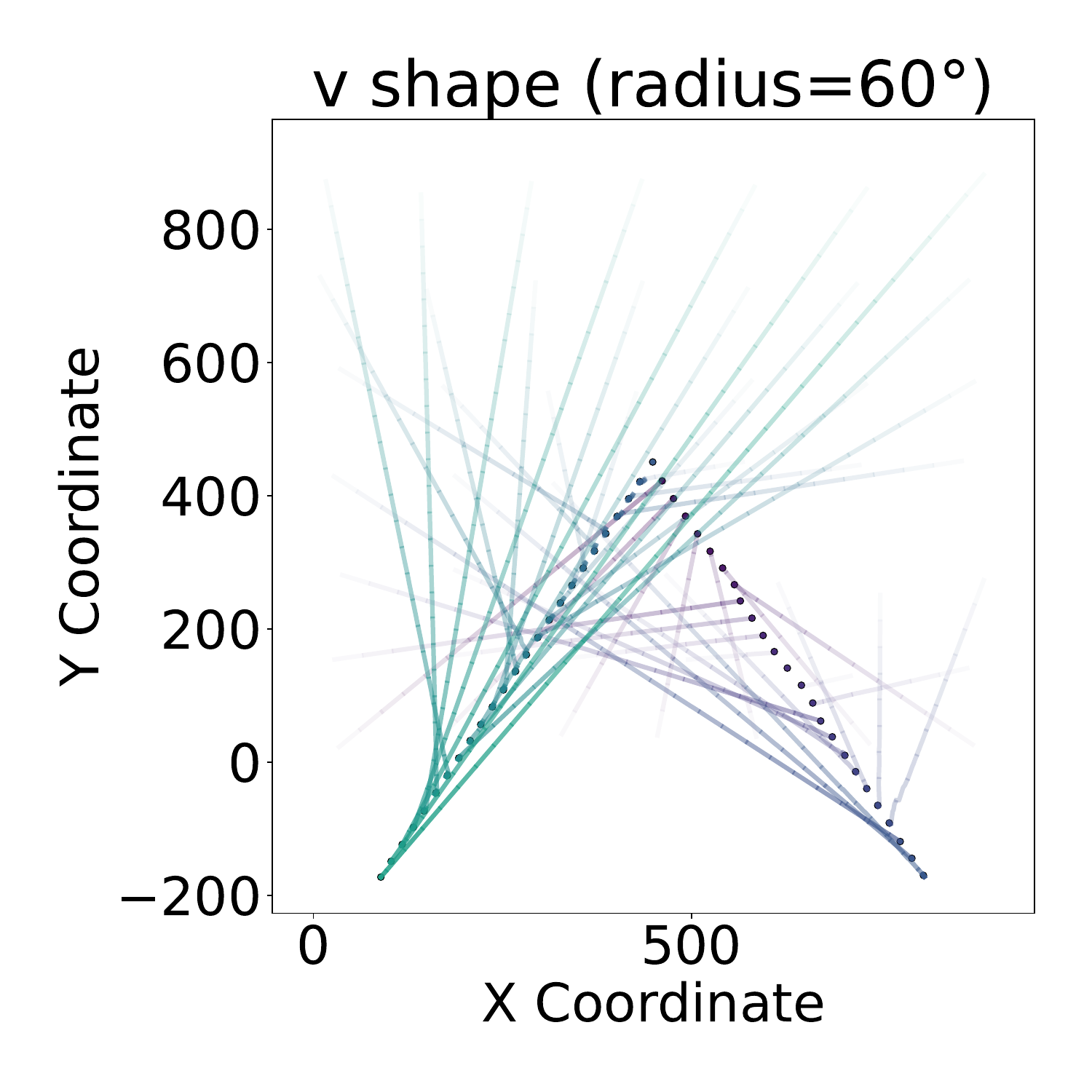}
  \end{subfigure}
  \hfill
  \begin{subfigure}[b]{0.3\textwidth}
    \includegraphics[width=\textwidth]{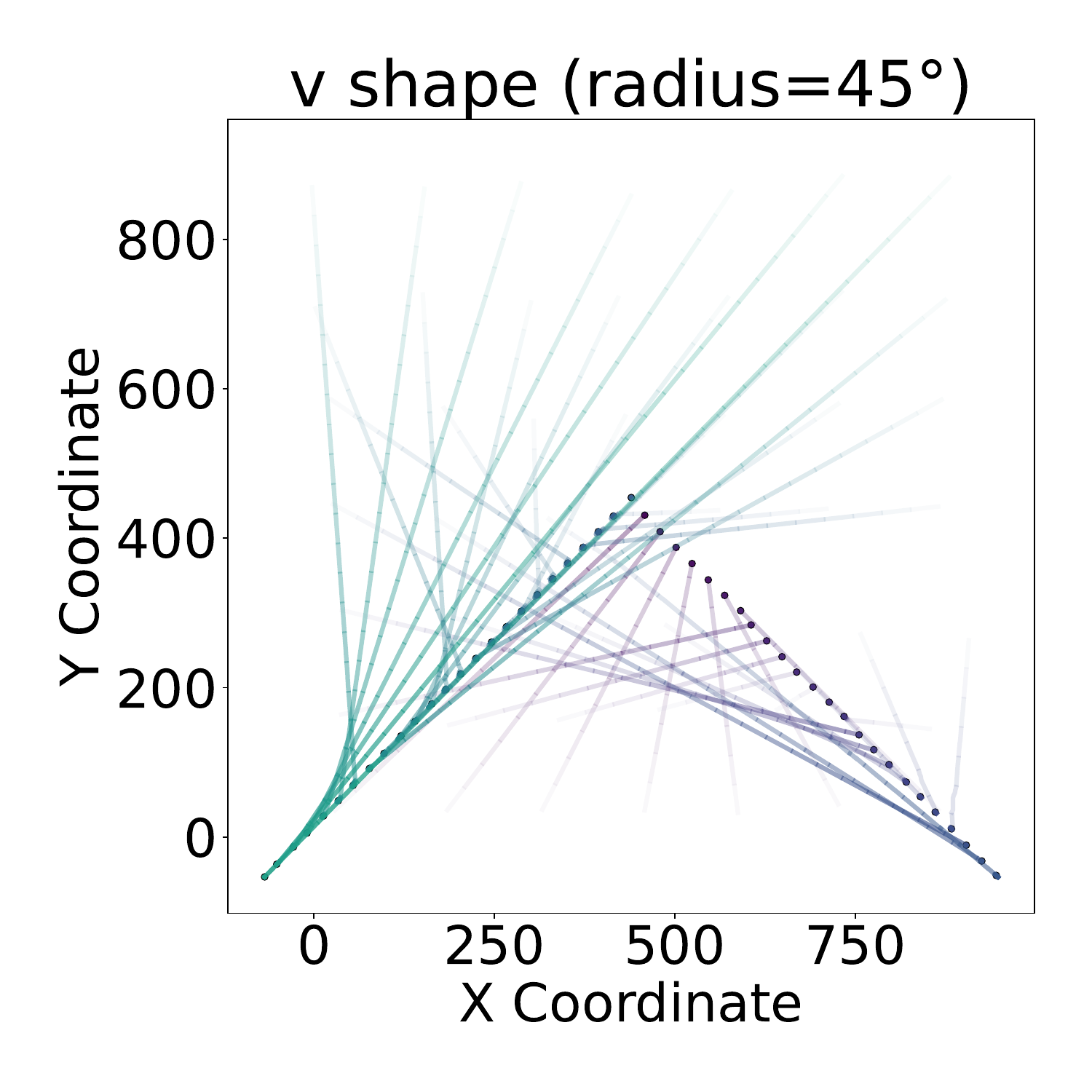}
  \end{subfigure}
  \hfill
  \begin{subfigure}[b]{0.3\textwidth}
    \includegraphics[width=\textwidth]{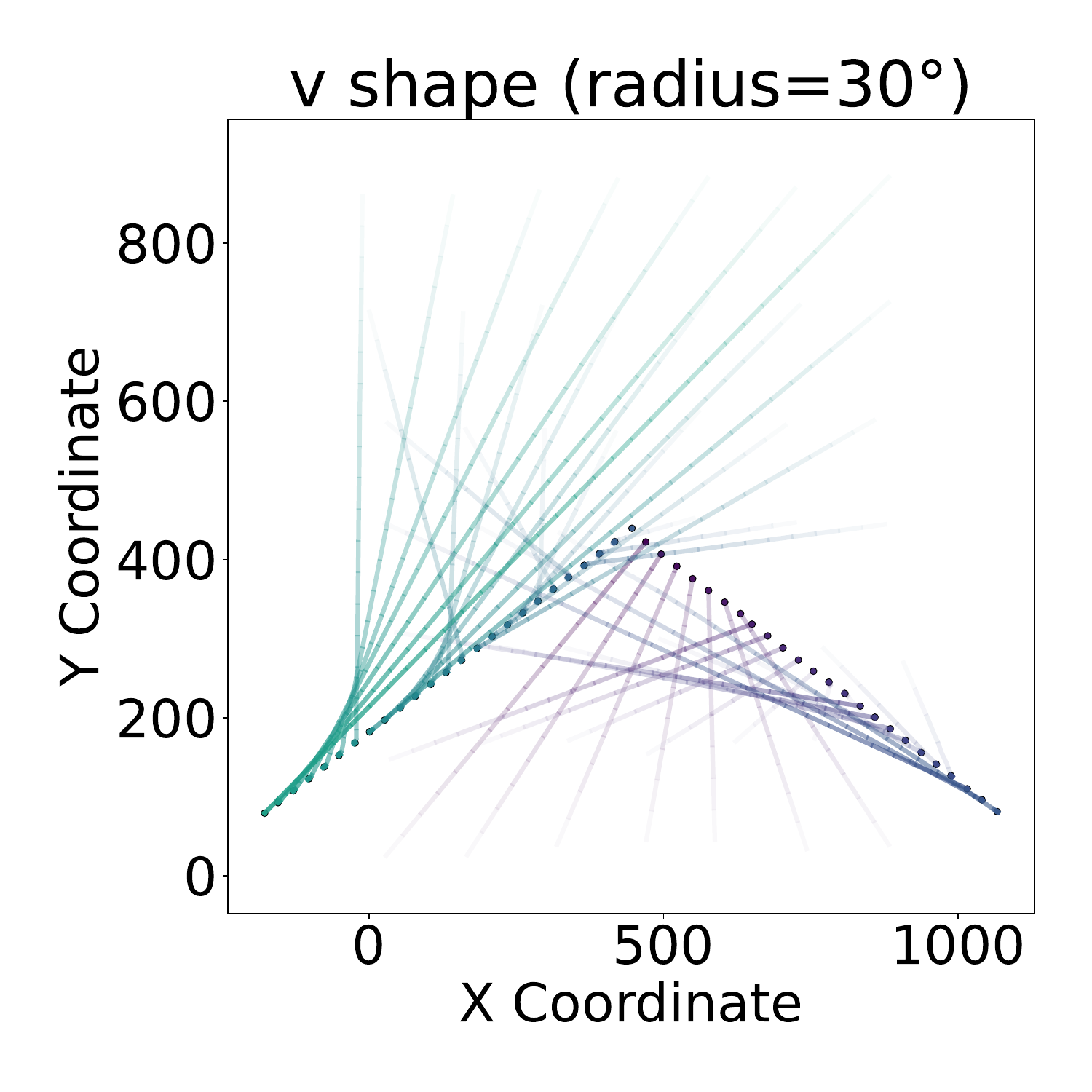}
  \end{subfigure}
  
  \begin{subfigure}[b]{0.3\textwidth}
    \includegraphics[width=\textwidth]{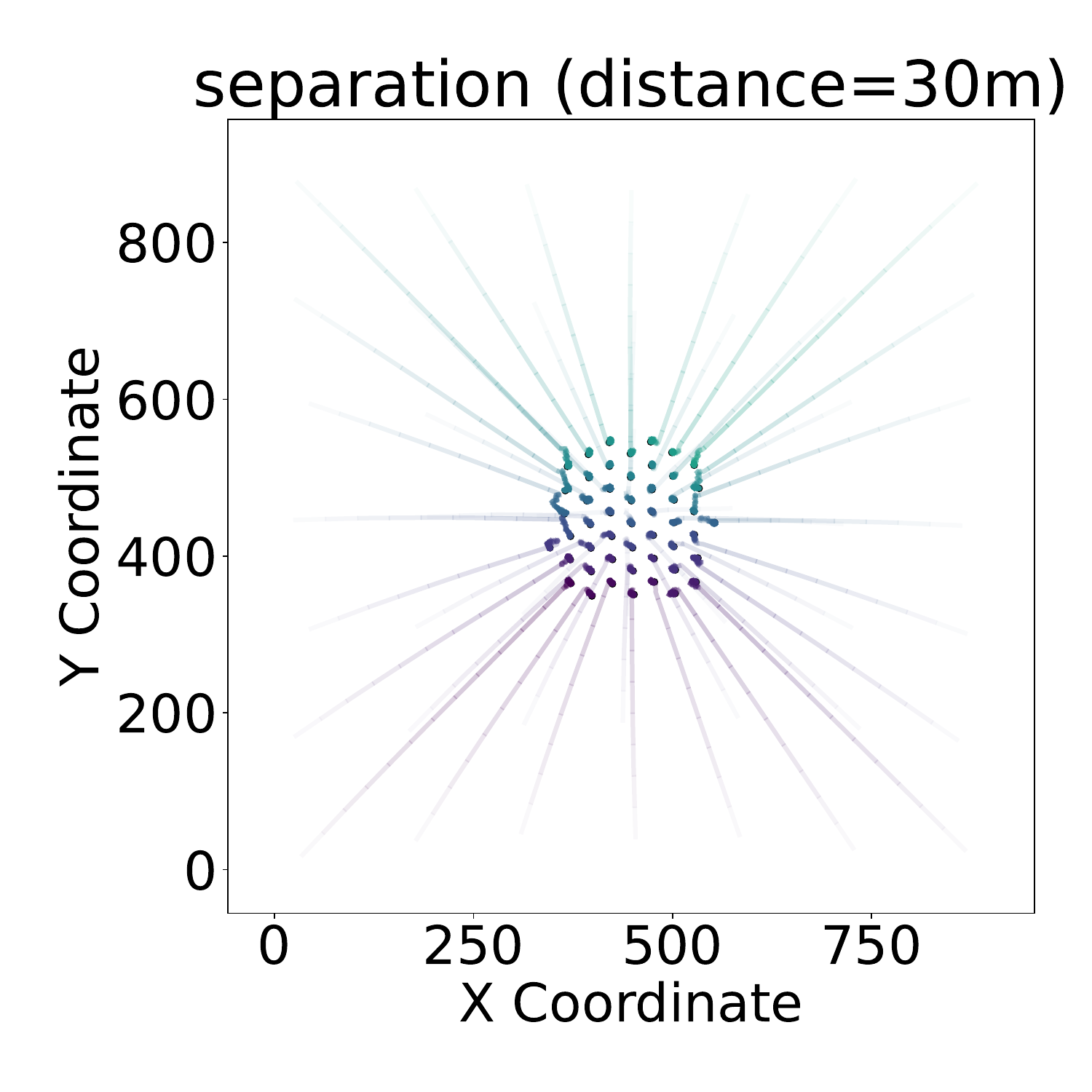}
  \end{subfigure}
  \hfill
  \begin{subfigure}[b]{0.3\textwidth}
    \includegraphics[width=\textwidth]{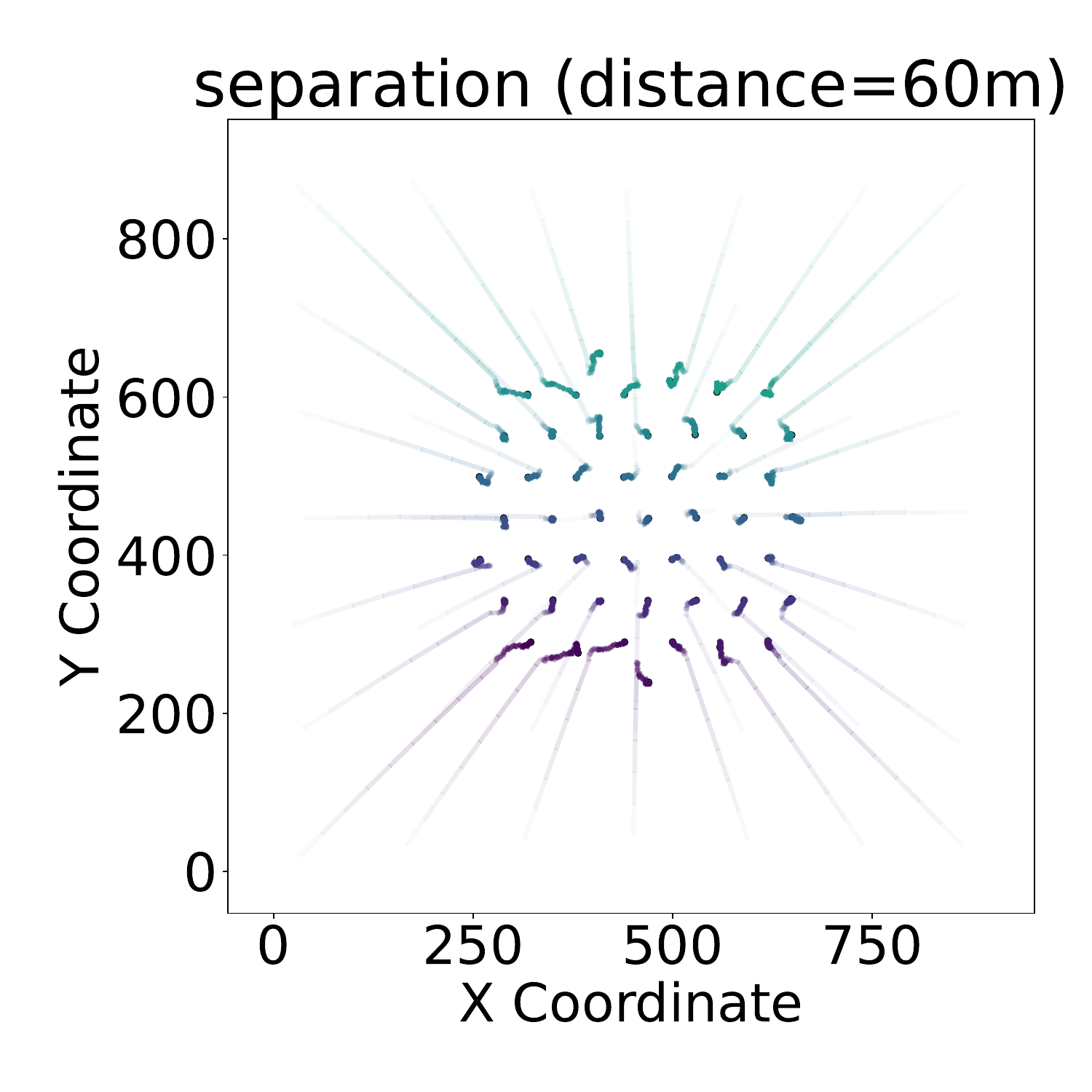}
  \end{subfigure}
  \hfill
  \begin{subfigure}[b]{0.3\textwidth}
    \includegraphics[width=\textwidth]{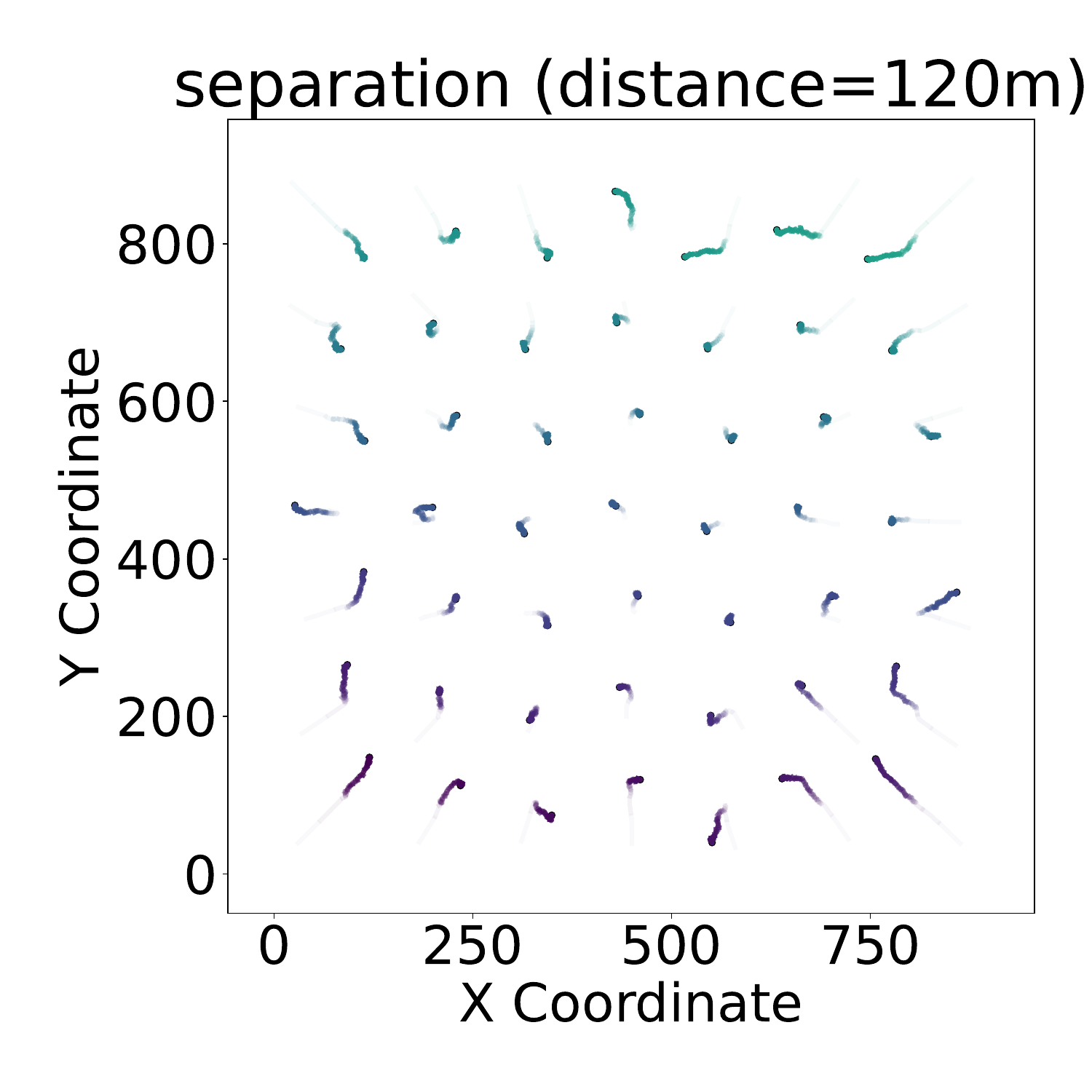}
  \end{subfigure}
  \centering
  \begin{subfigure}[b]{0.3\textwidth}
    \includegraphics[width=\textwidth]{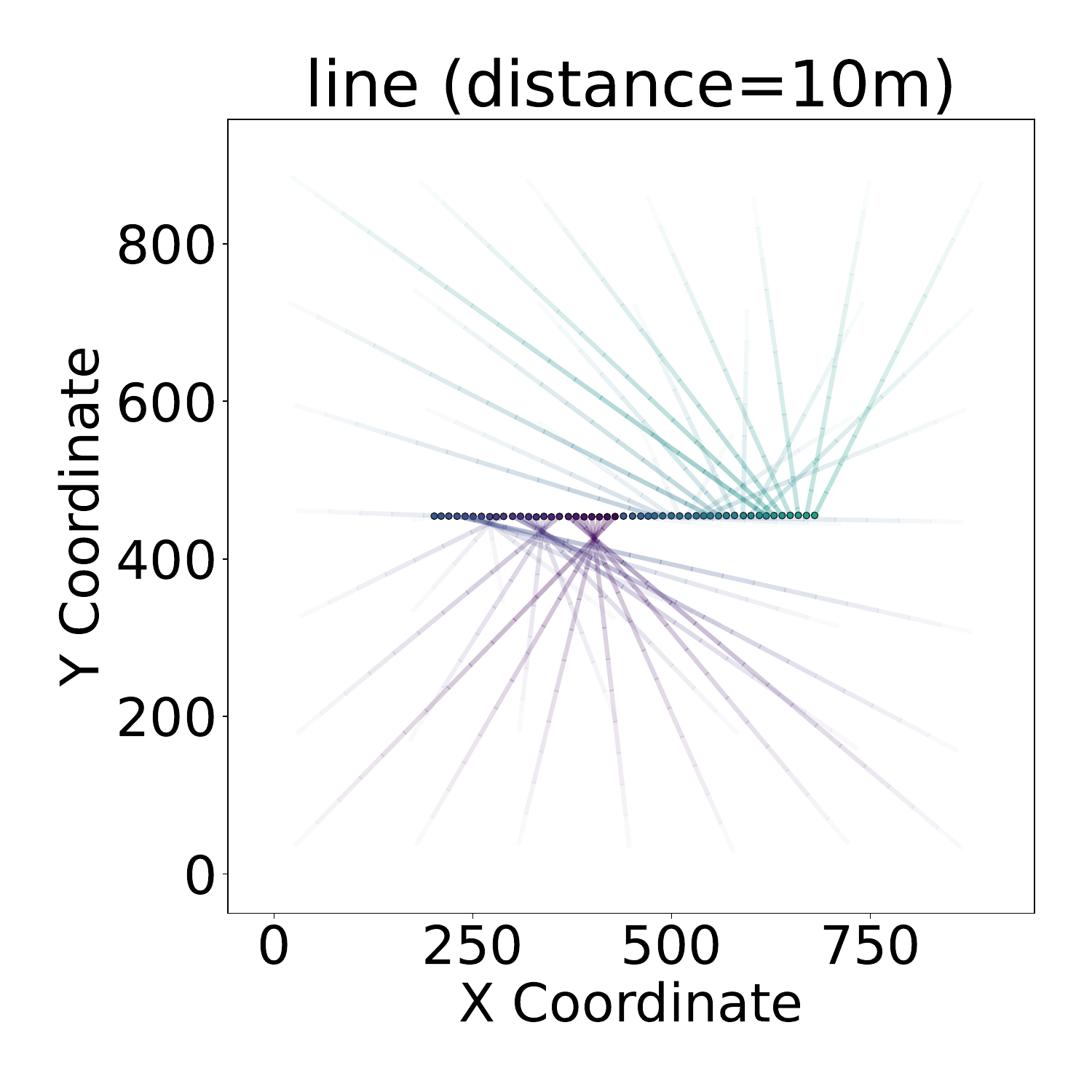}
  \end{subfigure}
  \hfill
  \begin{subfigure}[b]{0.3\textwidth}
    \includegraphics[width=\textwidth]{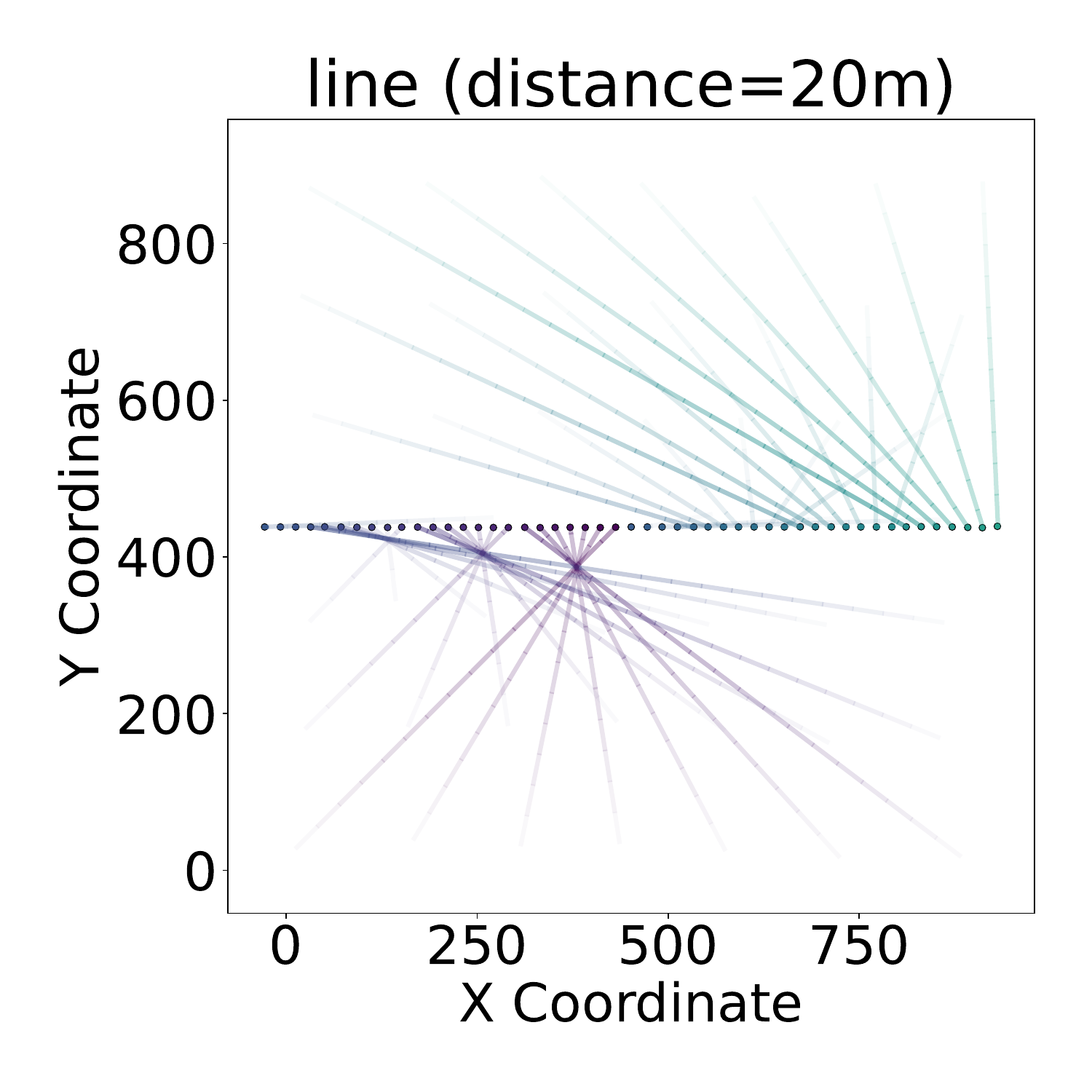}
  \end{subfigure}
  \hfill
  \begin{subfigure}[b]{0.3\textwidth}
    \includegraphics[width=\textwidth]{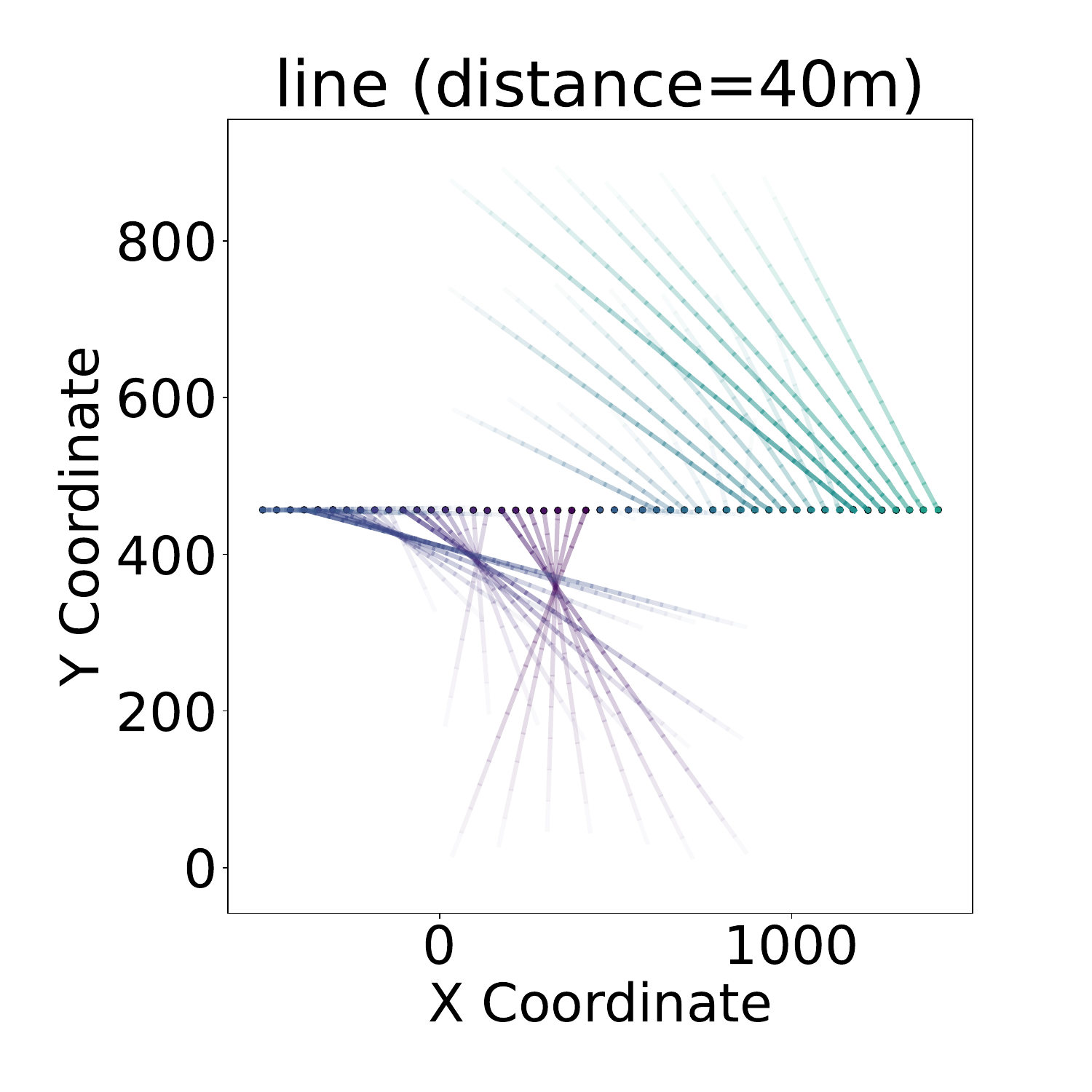}
  \end{subfigure}
  \centering
  \begin{subfigure}[b]{0.3\textwidth}
    \includegraphics[width=\textwidth]{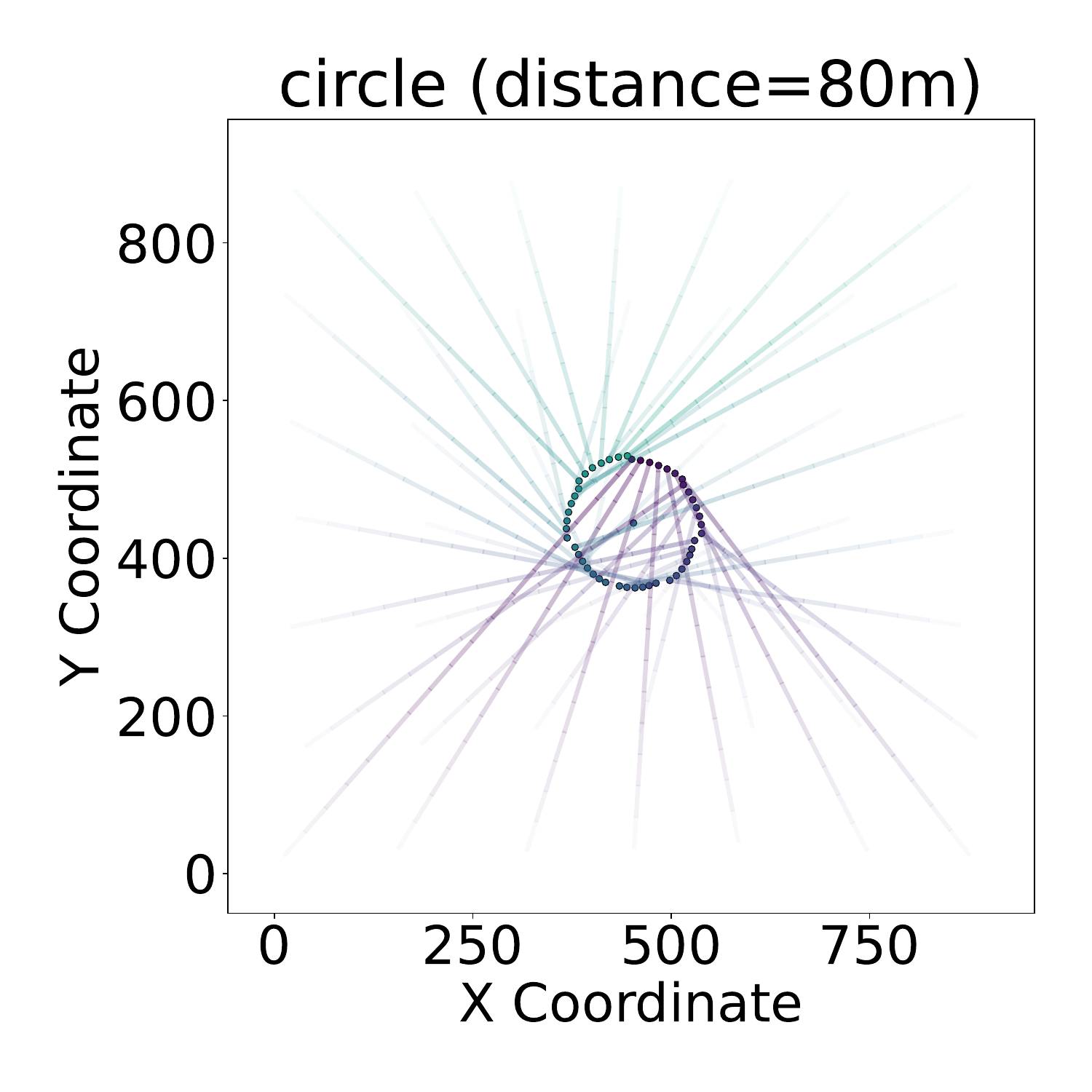}
  \end{subfigure}
  \hfill
  \begin{subfigure}[b]{0.3\textwidth}
    \includegraphics[width=\textwidth]{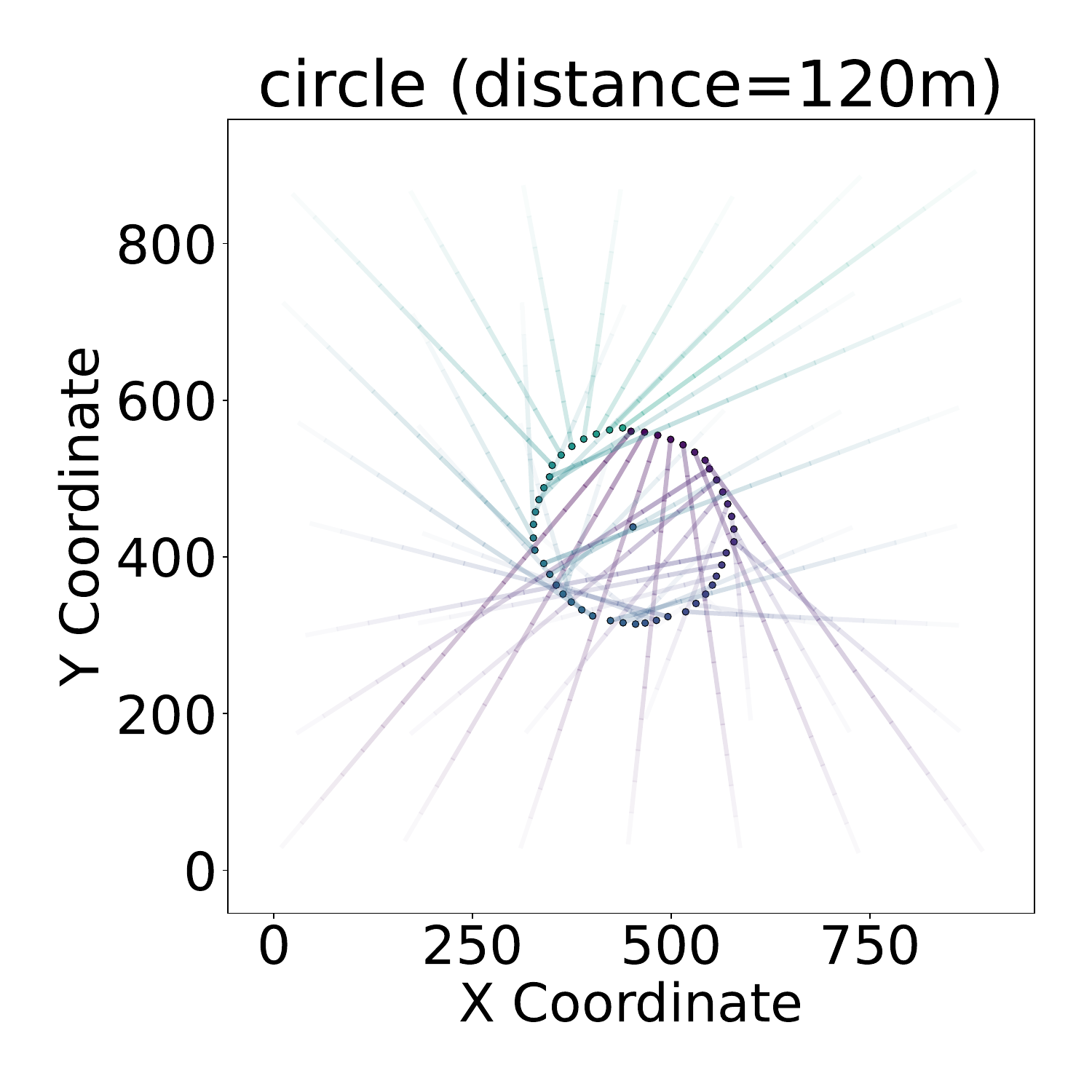}
  \end{subfigure}
  \hfill
  \begin{subfigure}[b]{0.3\textwidth}
    \includegraphics[width=\textwidth]{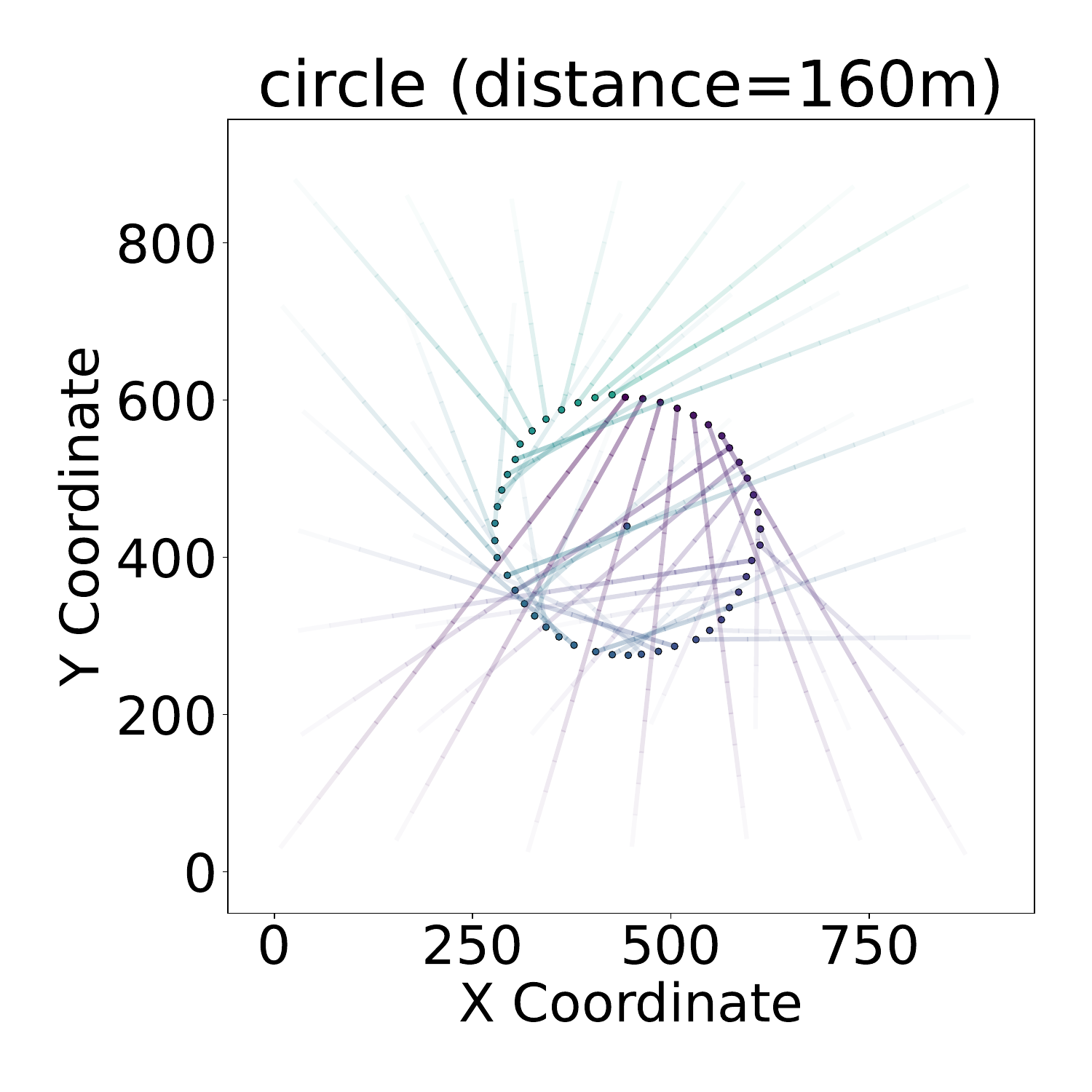}
  \end{subfigure}
  \caption{
    \revB{
    Pattern formation under ideal conditions (no message loss or perception noise).
    The node are depicted as circles coloured by their ID.
    The alpha of the colour is proportional to the time, the oldest nodes are more transparent. This is to show the evolution of the formation over time.}
  }
  \label{fig:pattern-eval}
\end{figure}

\begin{figure}
  \centering
  \begin{subfigure}[b]{0.3\textwidth}
    \includegraphics[width=\textwidth]{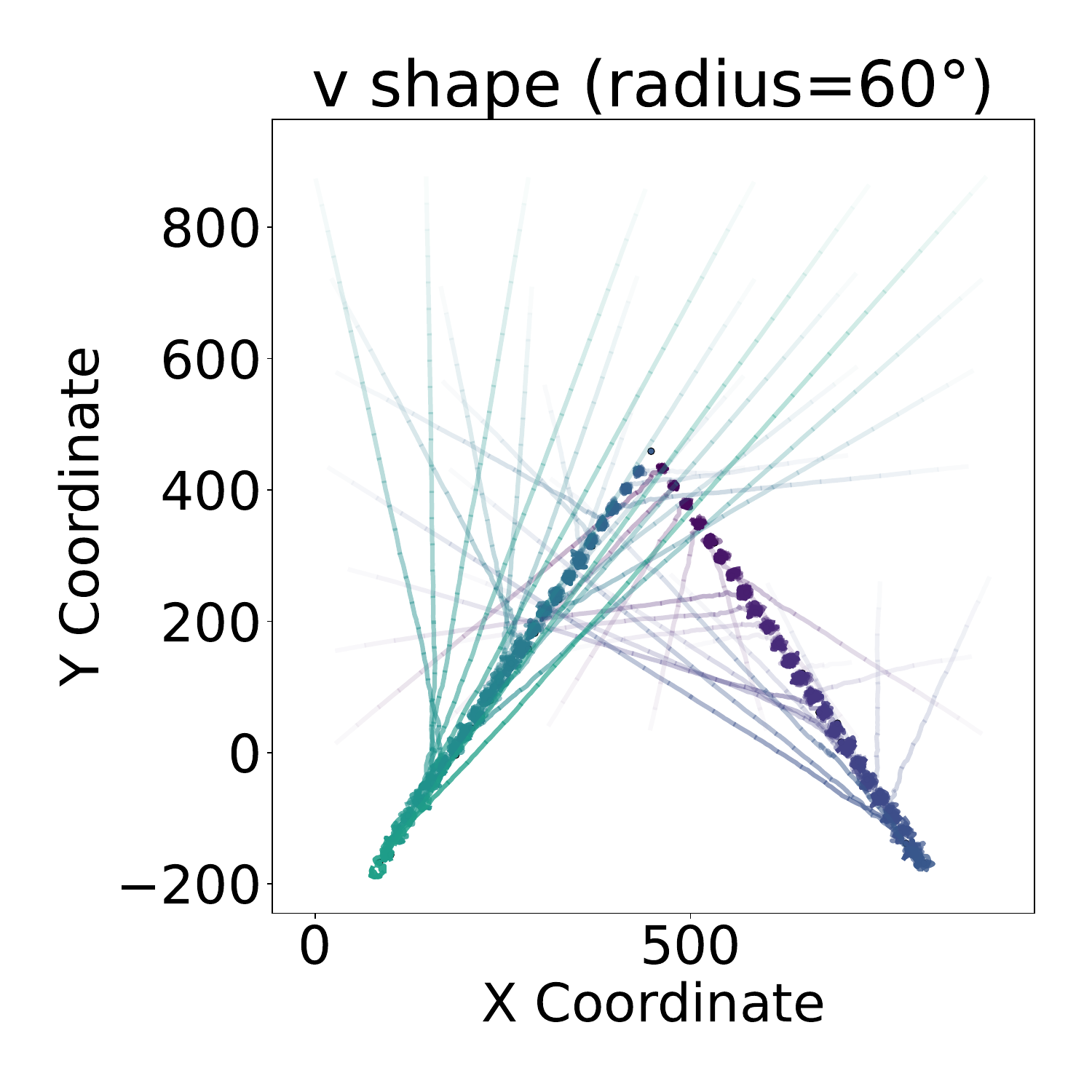}
  \end{subfigure}
  \hfill
  \begin{subfigure}[b]{0.3\textwidth}
    \includegraphics[width=\textwidth]{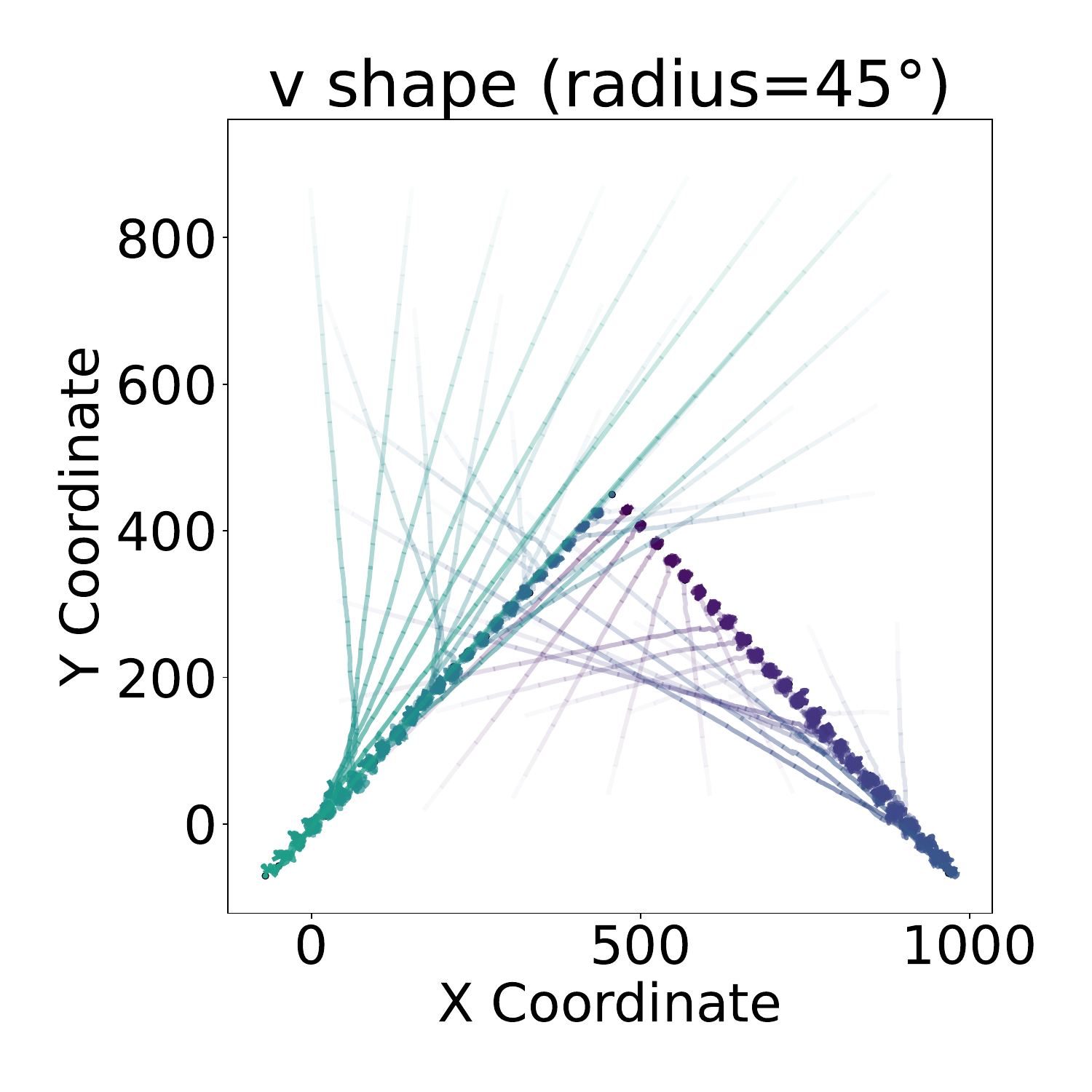}
  \end{subfigure}
  \hfill
  \begin{subfigure}[b]{0.3\textwidth}
    \includegraphics[width=\textwidth]{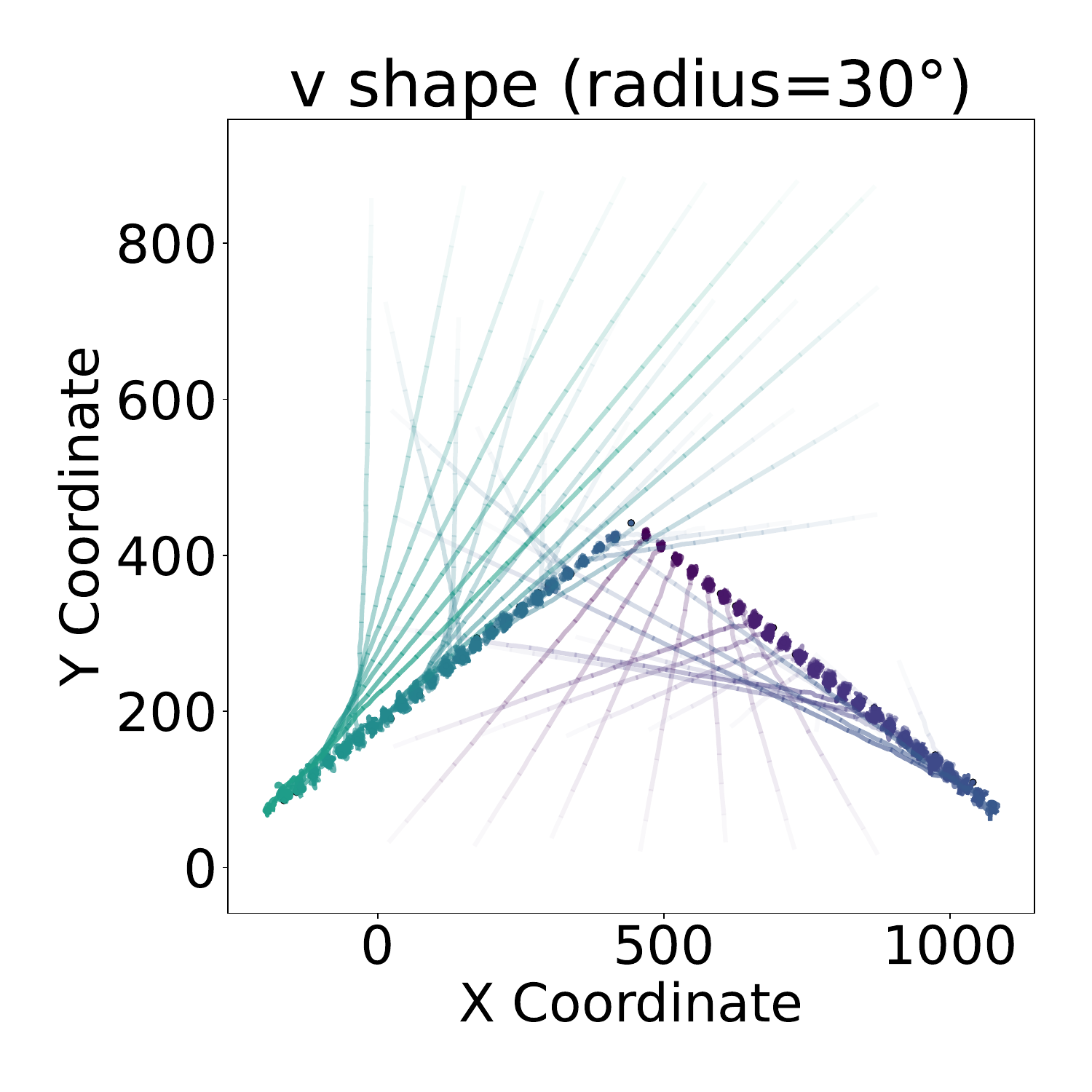}
  \end{subfigure}
  
  \begin{subfigure}[b]{0.3\textwidth}
    \includegraphics[width=\textwidth]{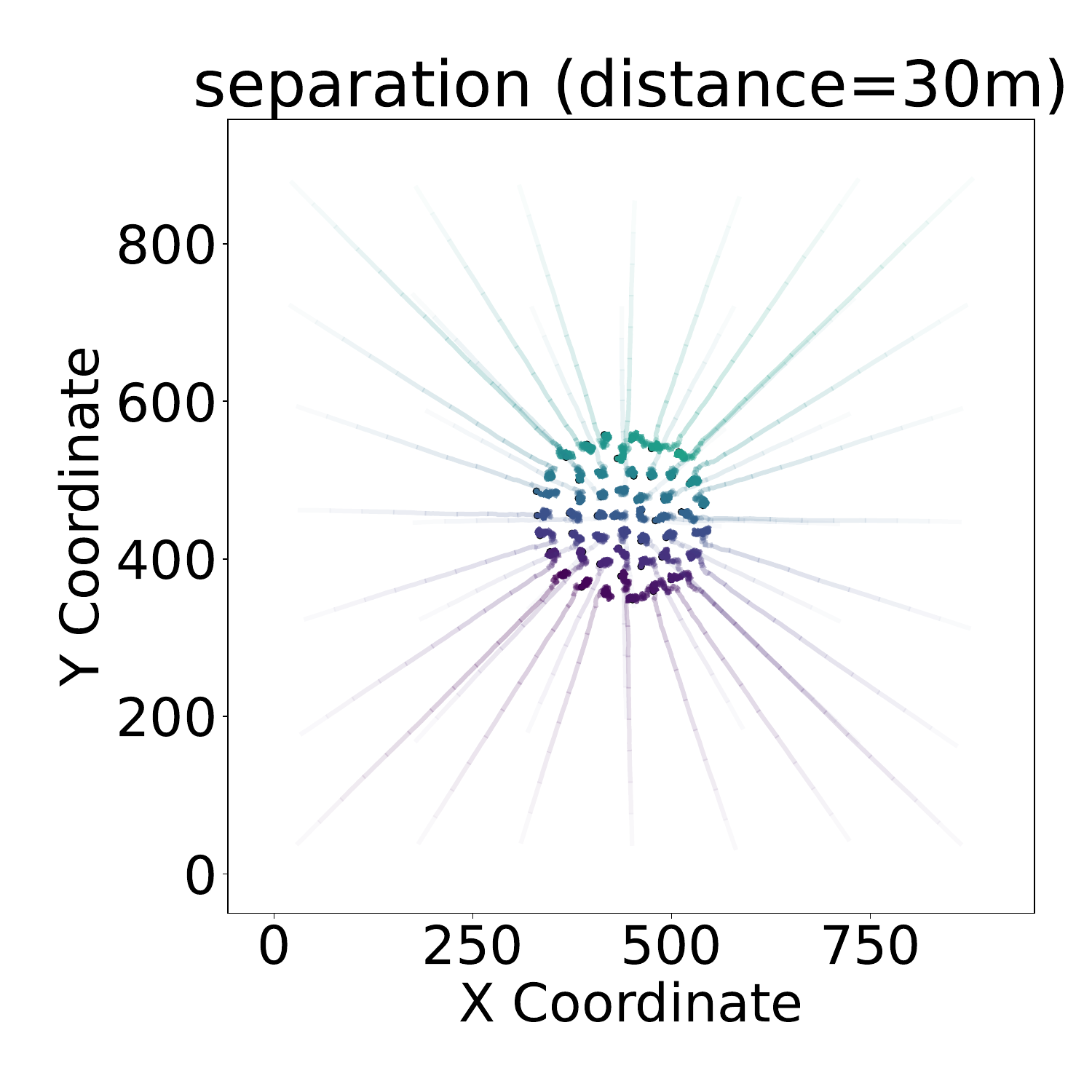}
  \end{subfigure}
  \hfill
  \begin{subfigure}[b]{0.3\textwidth}
    \includegraphics[width=\textwidth]{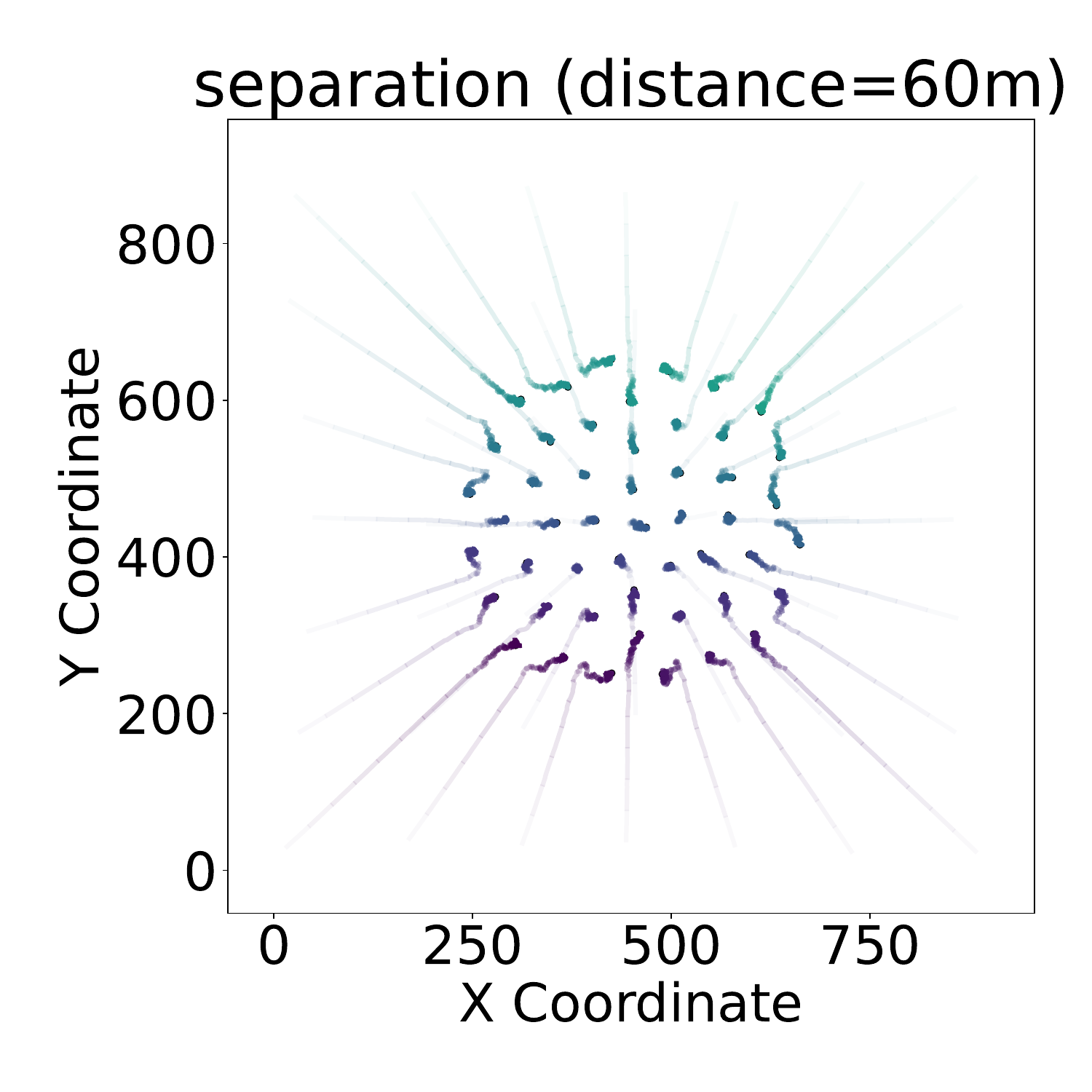}
  \end{subfigure}
  \hfill
  \begin{subfigure}[b]{0.3\textwidth}
    \includegraphics[width=\textwidth]{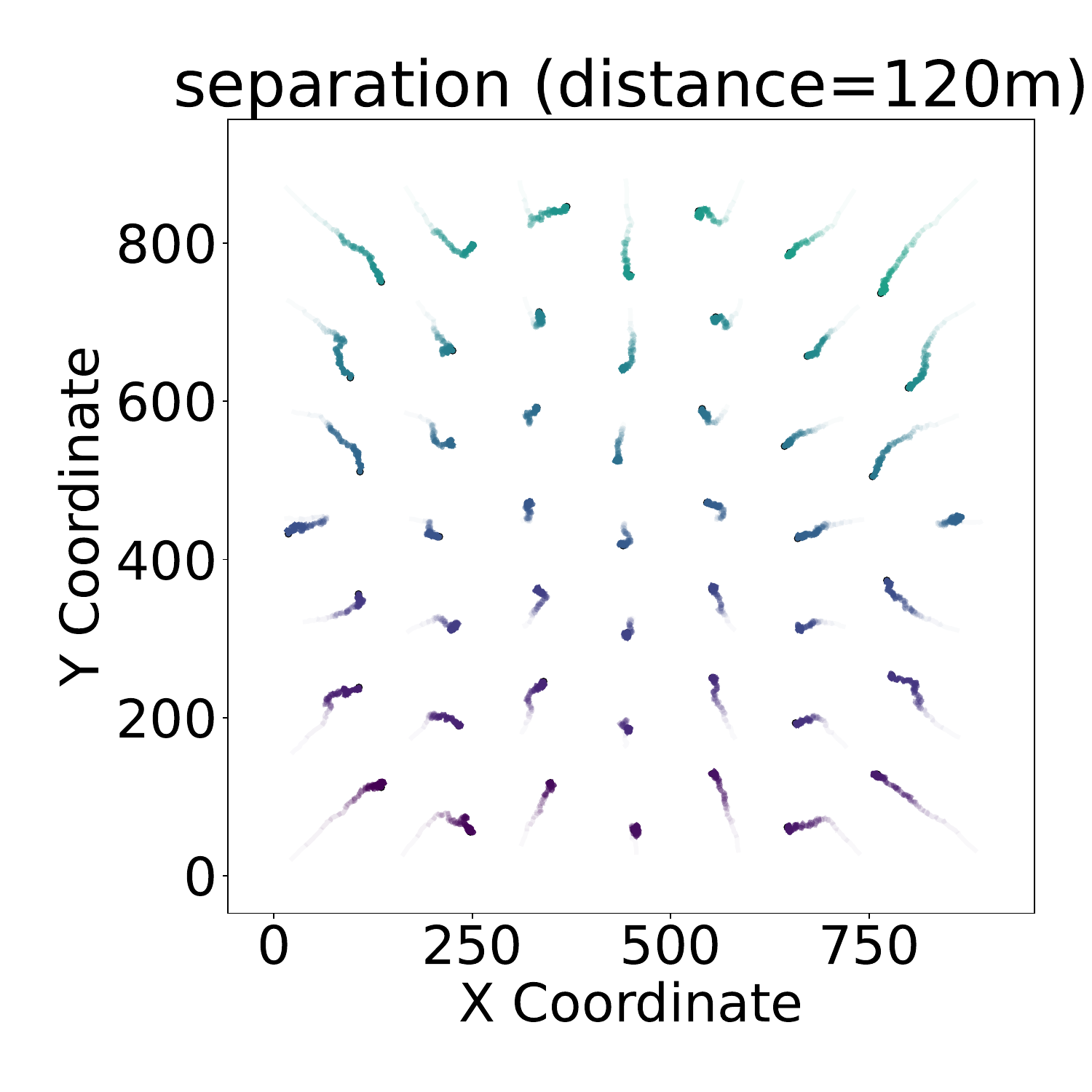}
  \end{subfigure}
  \centering
  \begin{subfigure}[b]{0.3\textwidth}
    \includegraphics[width=\textwidth]{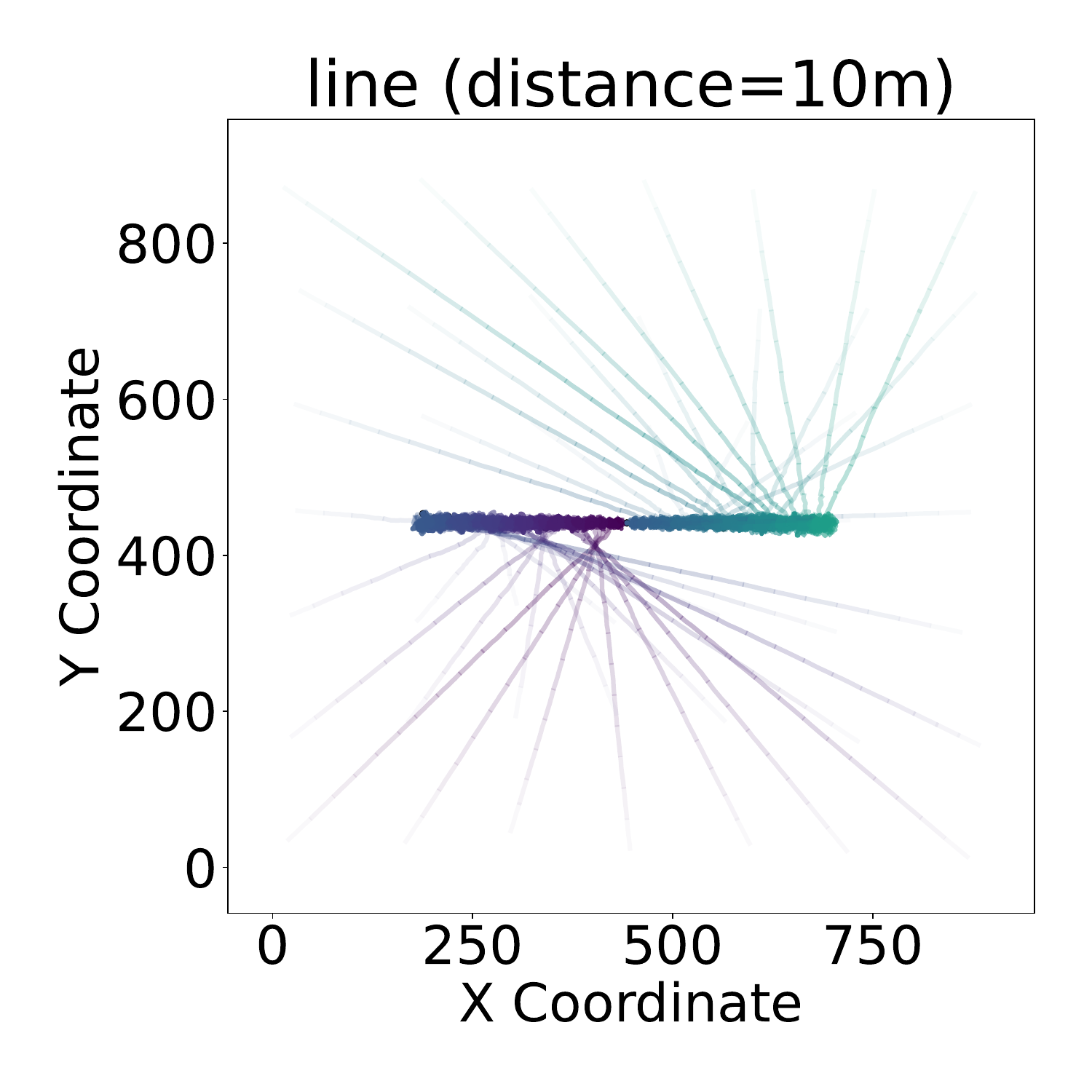}
  \end{subfigure}
  \hfill
  \begin{subfigure}[b]{0.3\textwidth}
    \includegraphics[width=\textwidth]{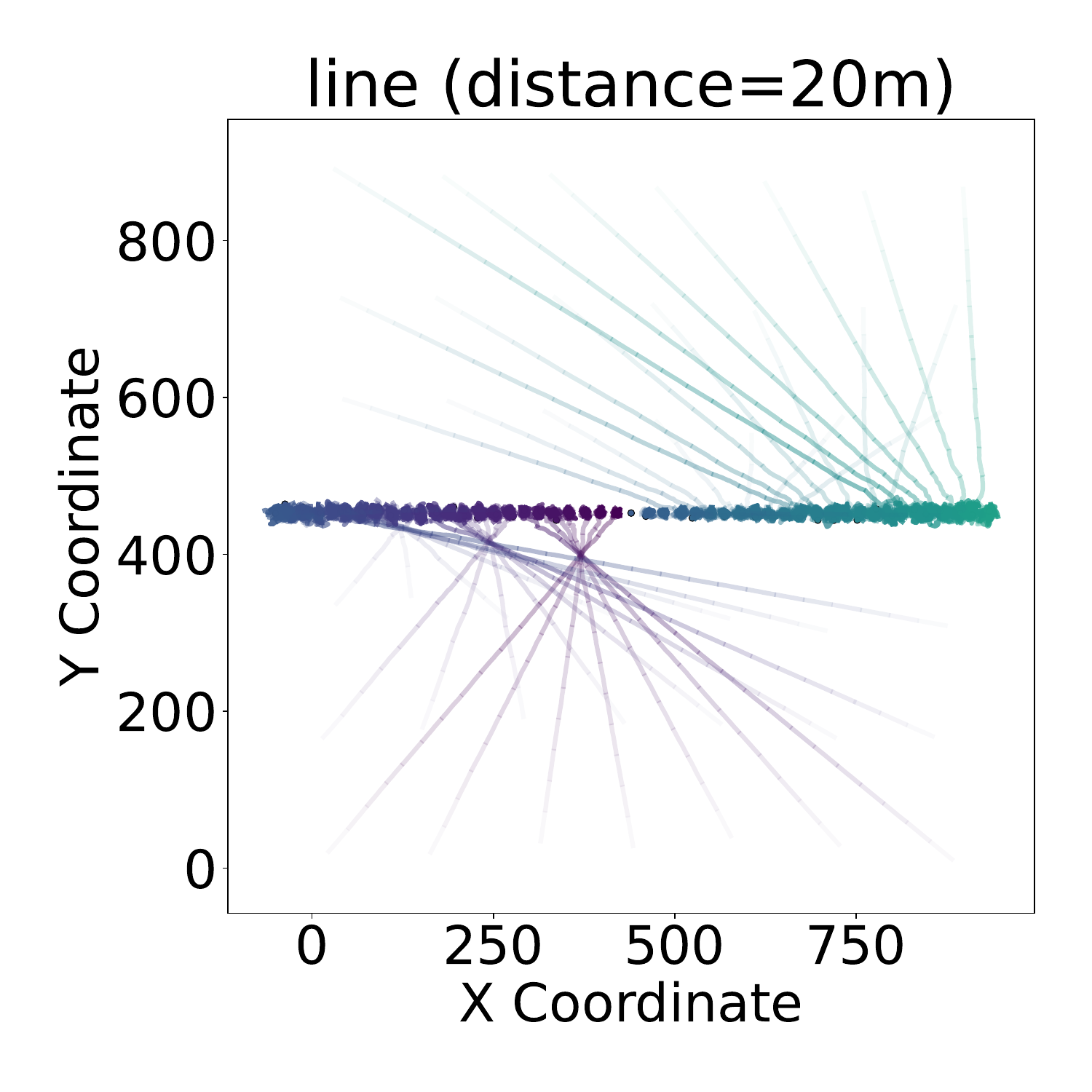}
  \end{subfigure}
  \hfill
  \begin{subfigure}[b]{0.3\textwidth}
    \includegraphics[width=\textwidth]{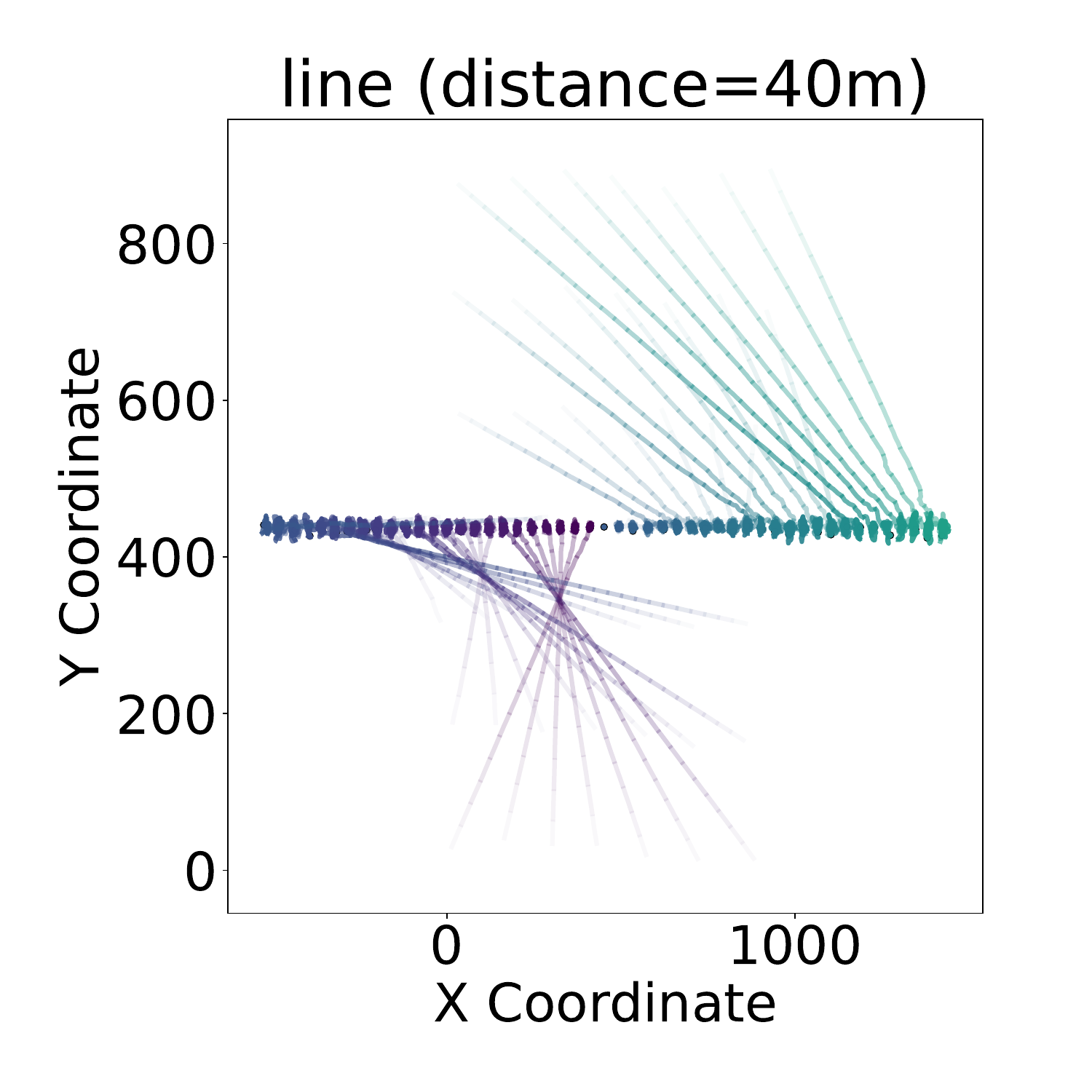}
  \end{subfigure}
  \centering
  \begin{subfigure}[b]{0.3\textwidth}
    \includegraphics[width=\textwidth]{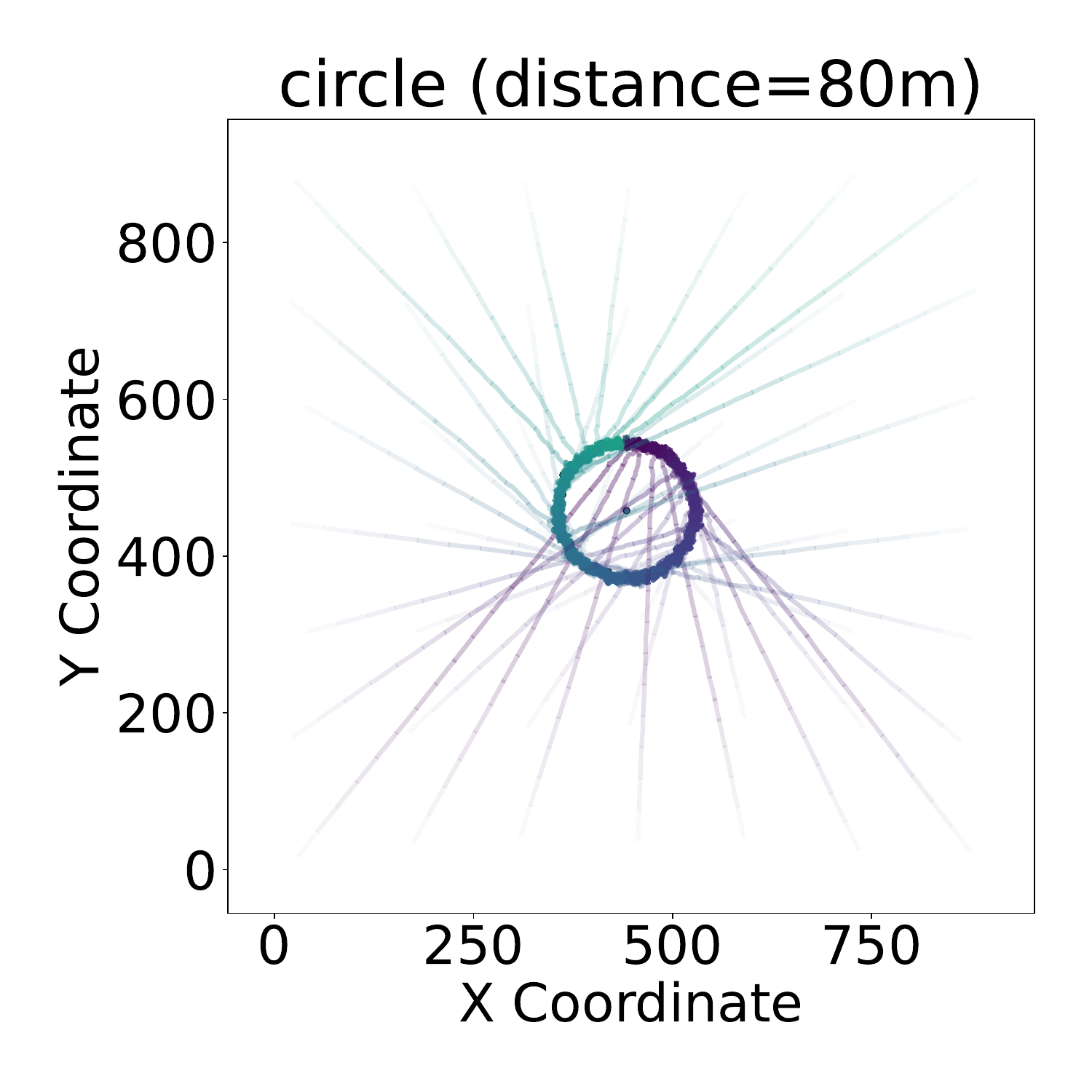}
  \end{subfigure}
  \hfill
  \begin{subfigure}[b]{0.3\textwidth}
    \includegraphics[width=\textwidth]{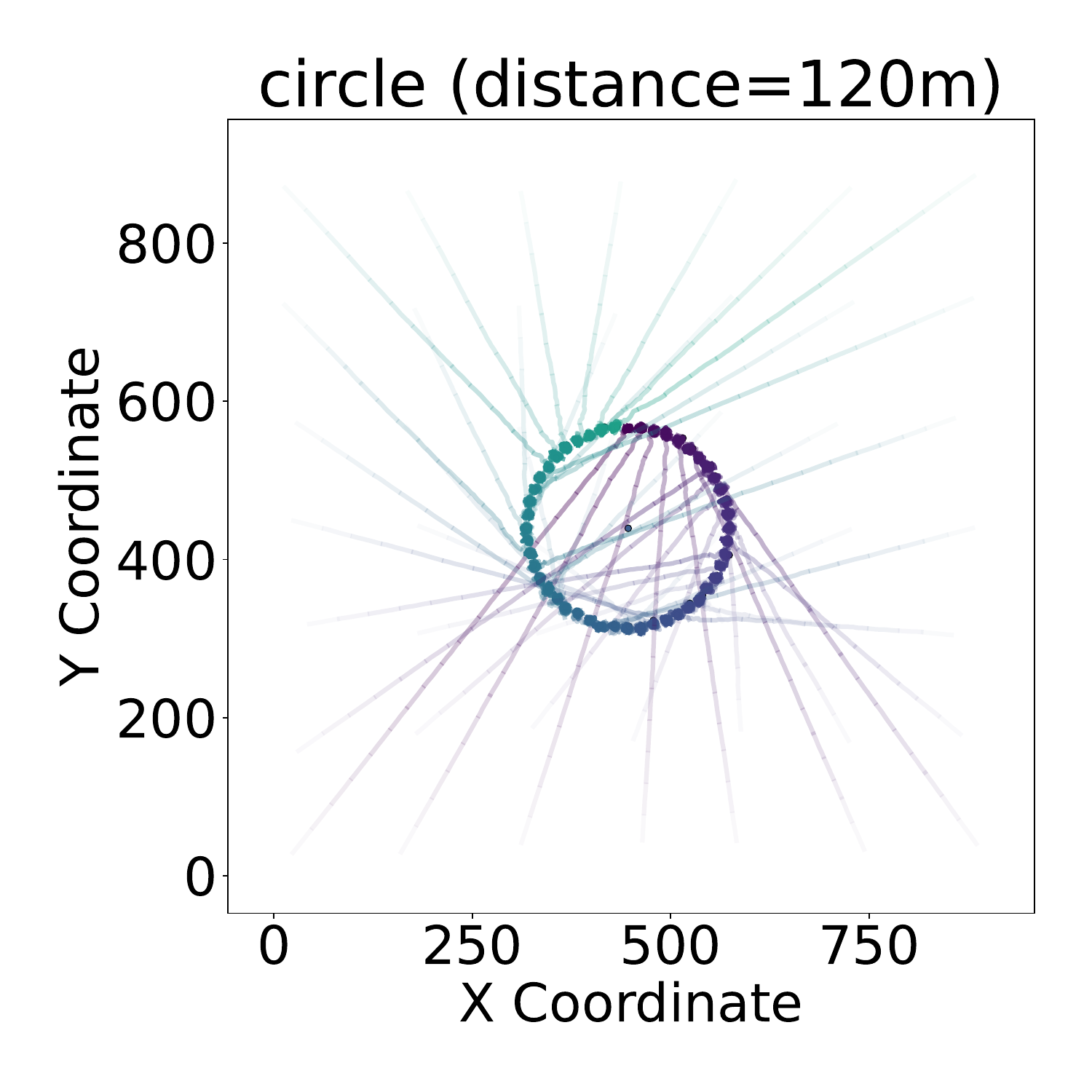}
  \end{subfigure}
  \hfill
  \begin{subfigure}[b]{0.3\textwidth}
    \includegraphics[width=\textwidth]{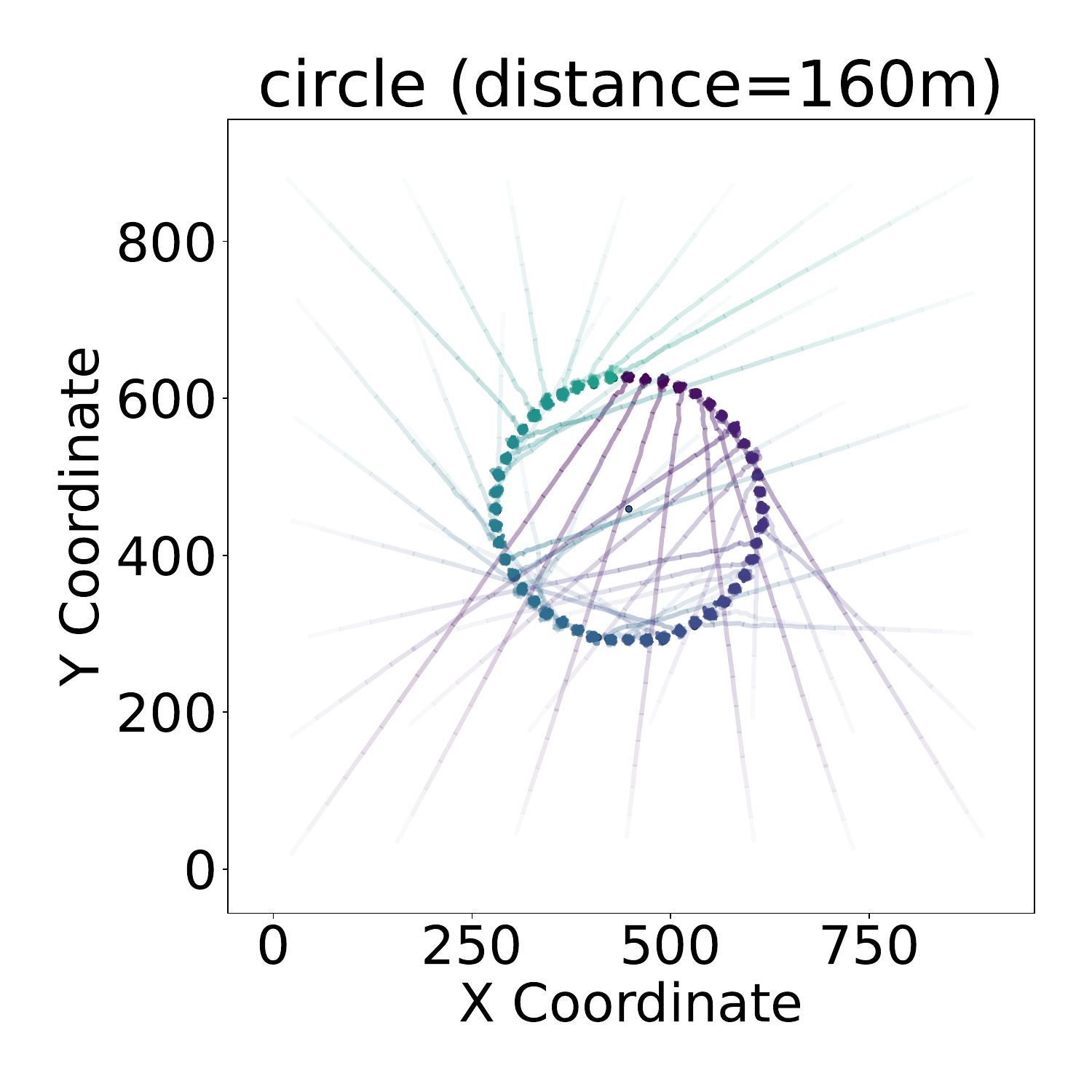}
  \end{subfigure}
  \caption{
    \revB{
    Shows the effect of 10.0 meters of noise in the position perception ($P=10$).
    Even if the structure is not perfect, the drones are able to form the desired shape.}
  }
  \label{fig:pattern-eval-error-perception}
\end{figure}

\begin{figure}
  \centering
  \begin{subfigure}[b]{0.3\textwidth}
    \includegraphics[width=\textwidth]{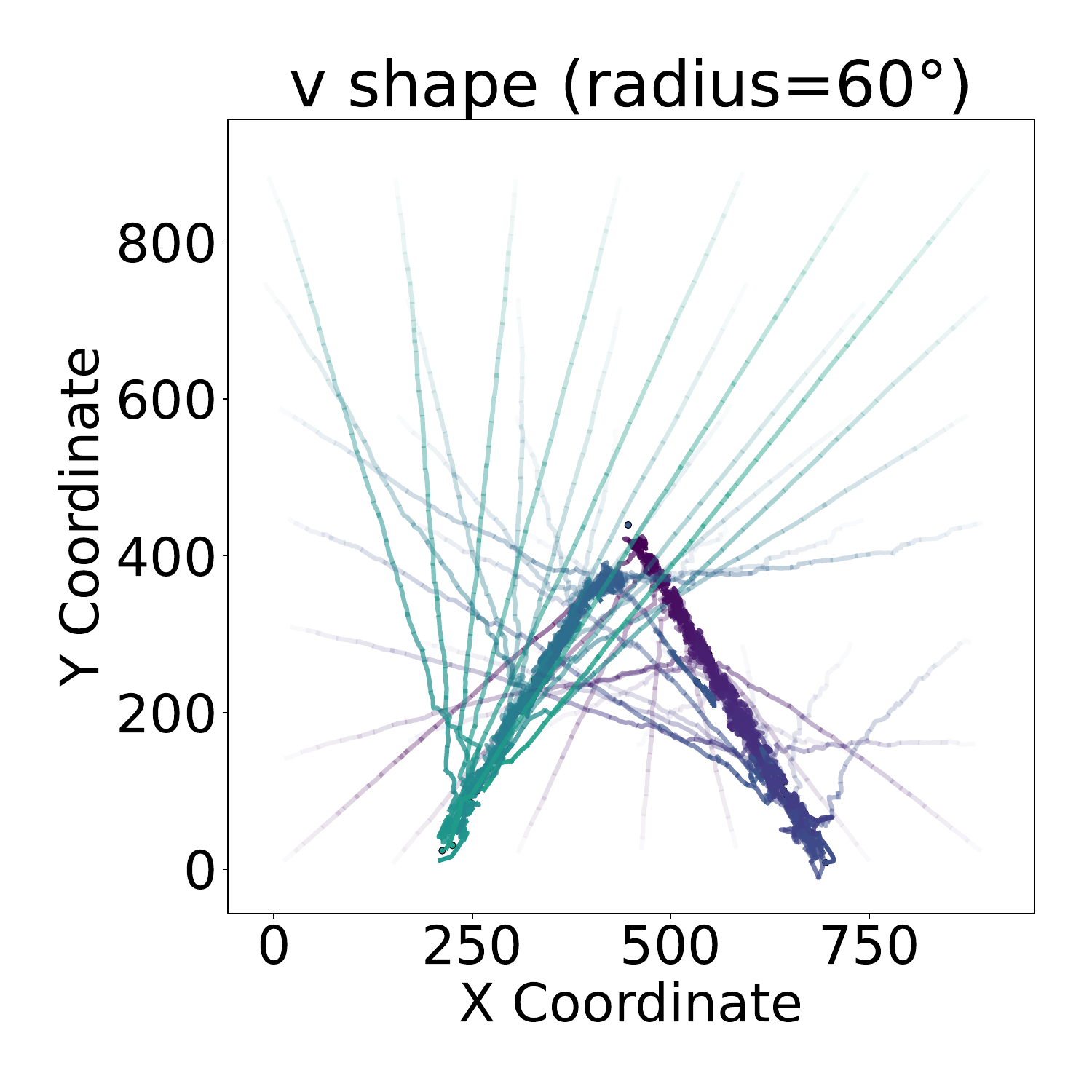}
  \end{subfigure}
  \hfill
  \begin{subfigure}[b]{0.3\textwidth}
    \includegraphics[width=\textwidth]{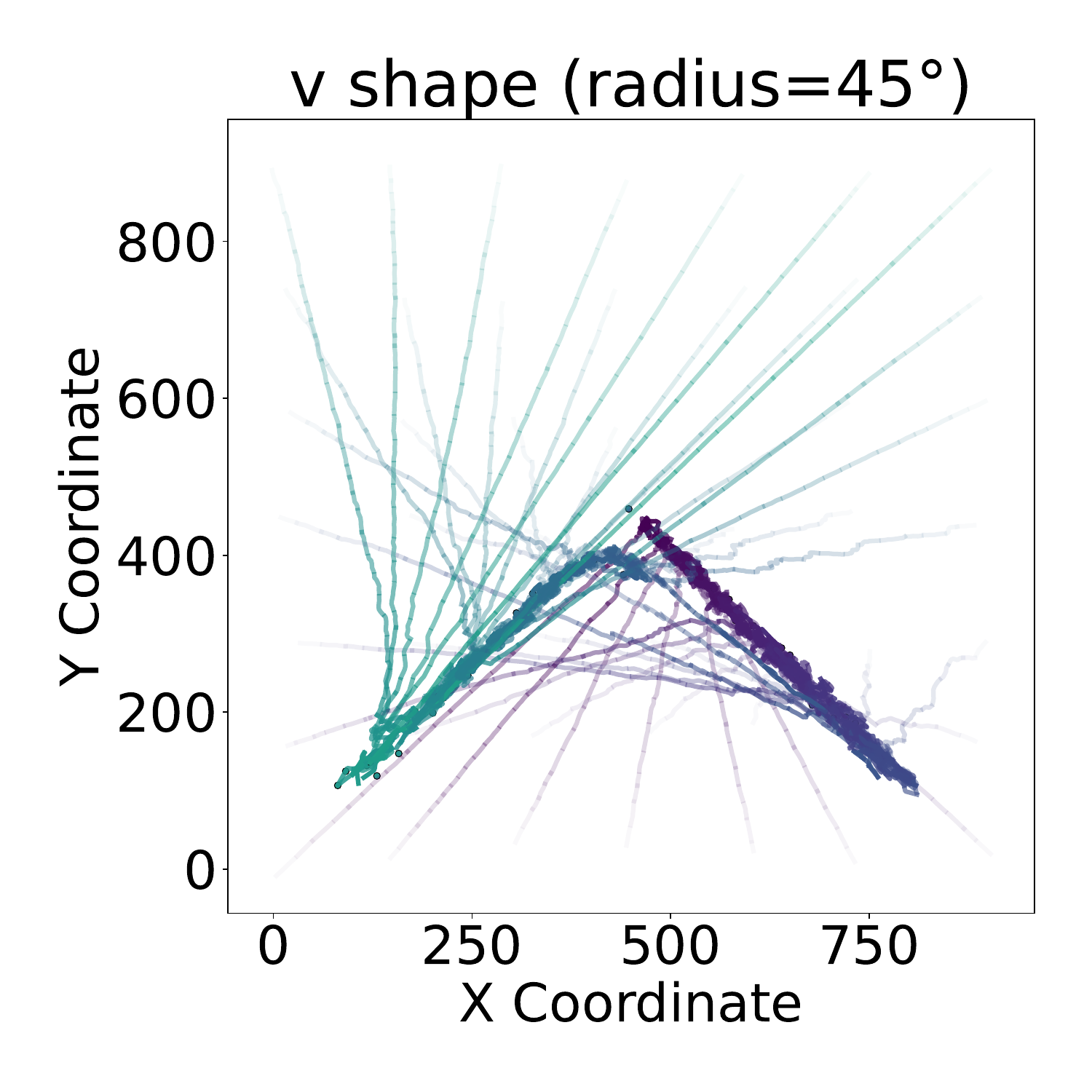}
  \end{subfigure}
  \hfill
  \begin{subfigure}[b]{0.3\textwidth}
    \includegraphics[width=\textwidth]{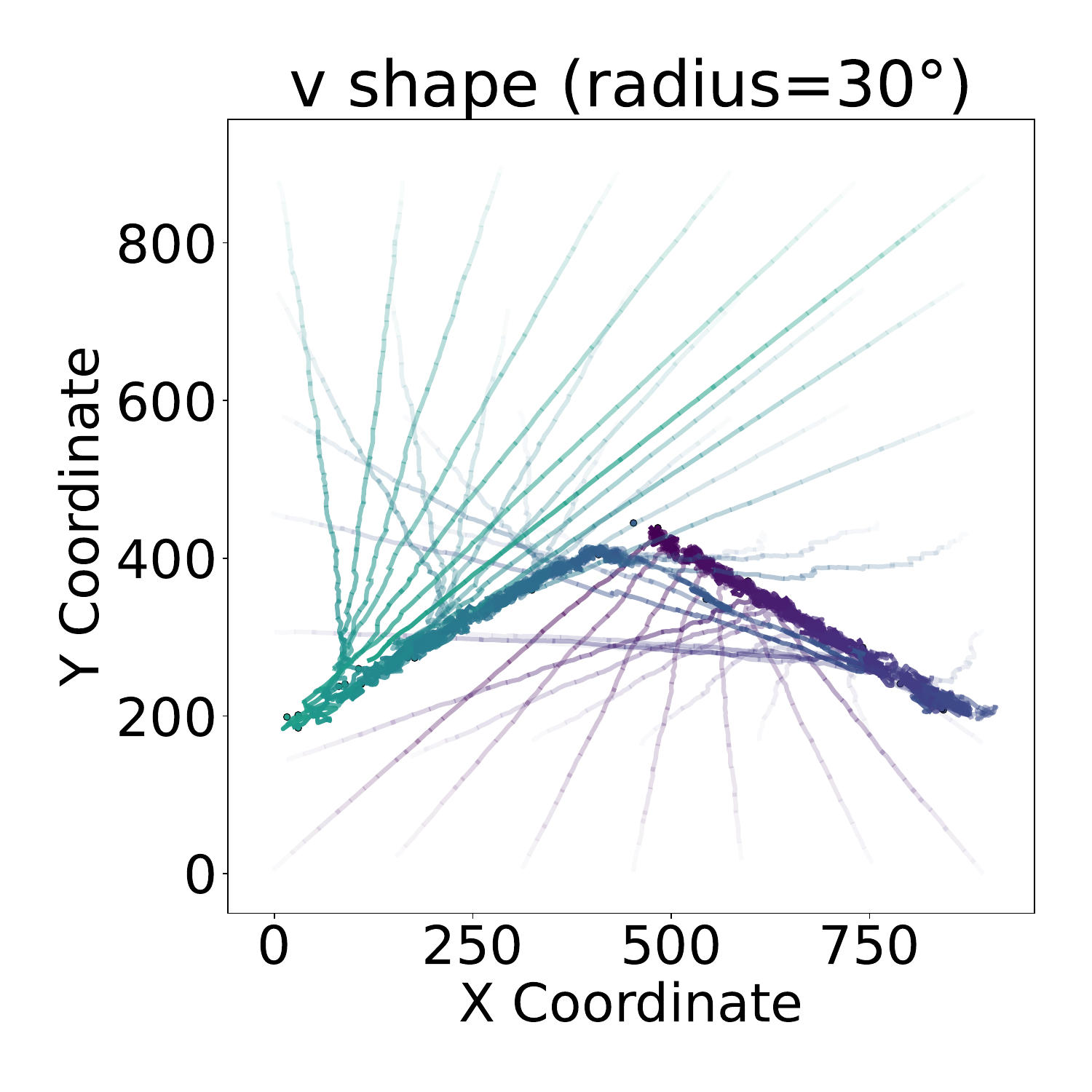}
  \end{subfigure}
  
  \begin{subfigure}[b]{0.3\textwidth}
    \includegraphics[width=\textwidth]{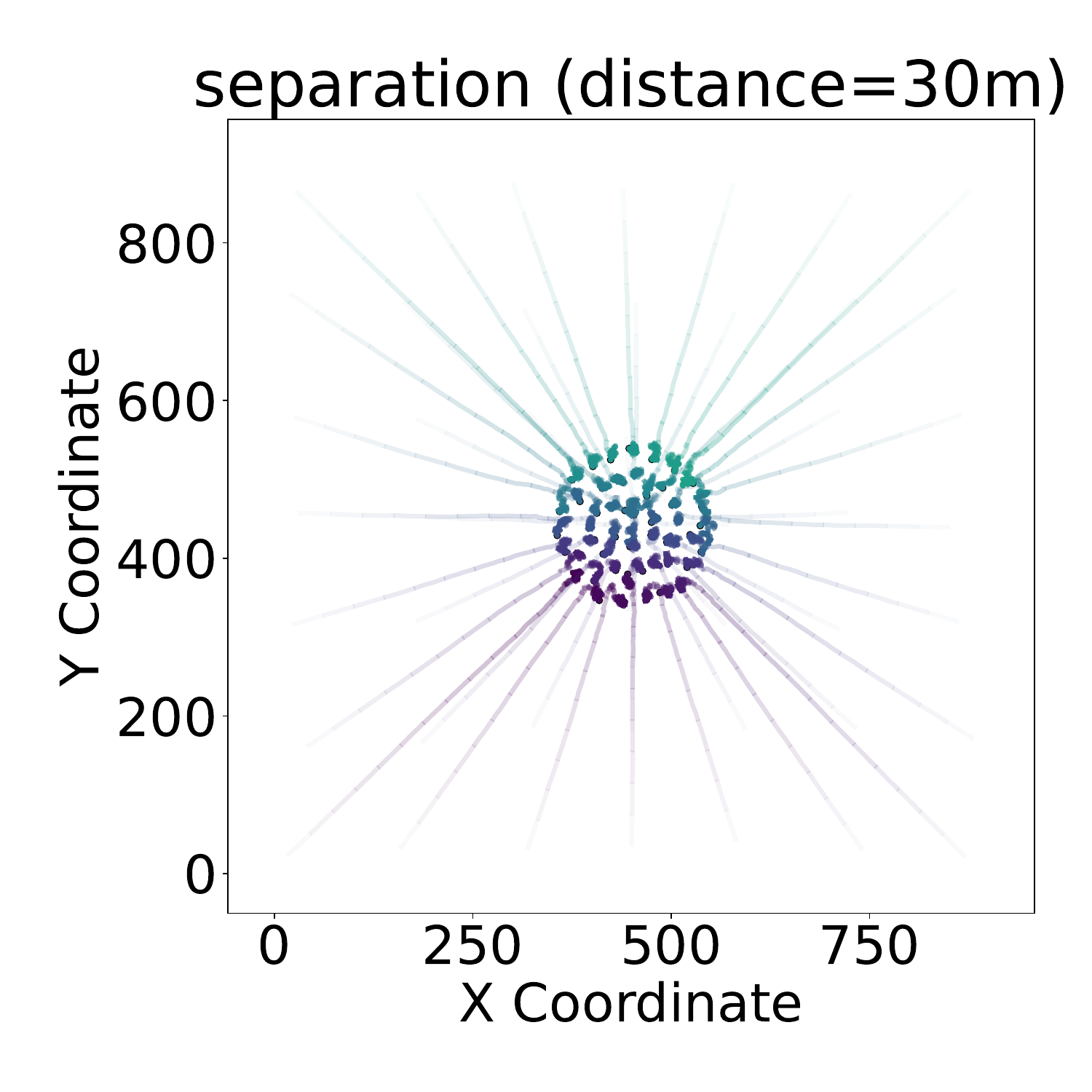}
  \end{subfigure}
  \hfill
  \begin{subfigure}[b]{0.3\textwidth}
    \includegraphics[width=\textwidth]{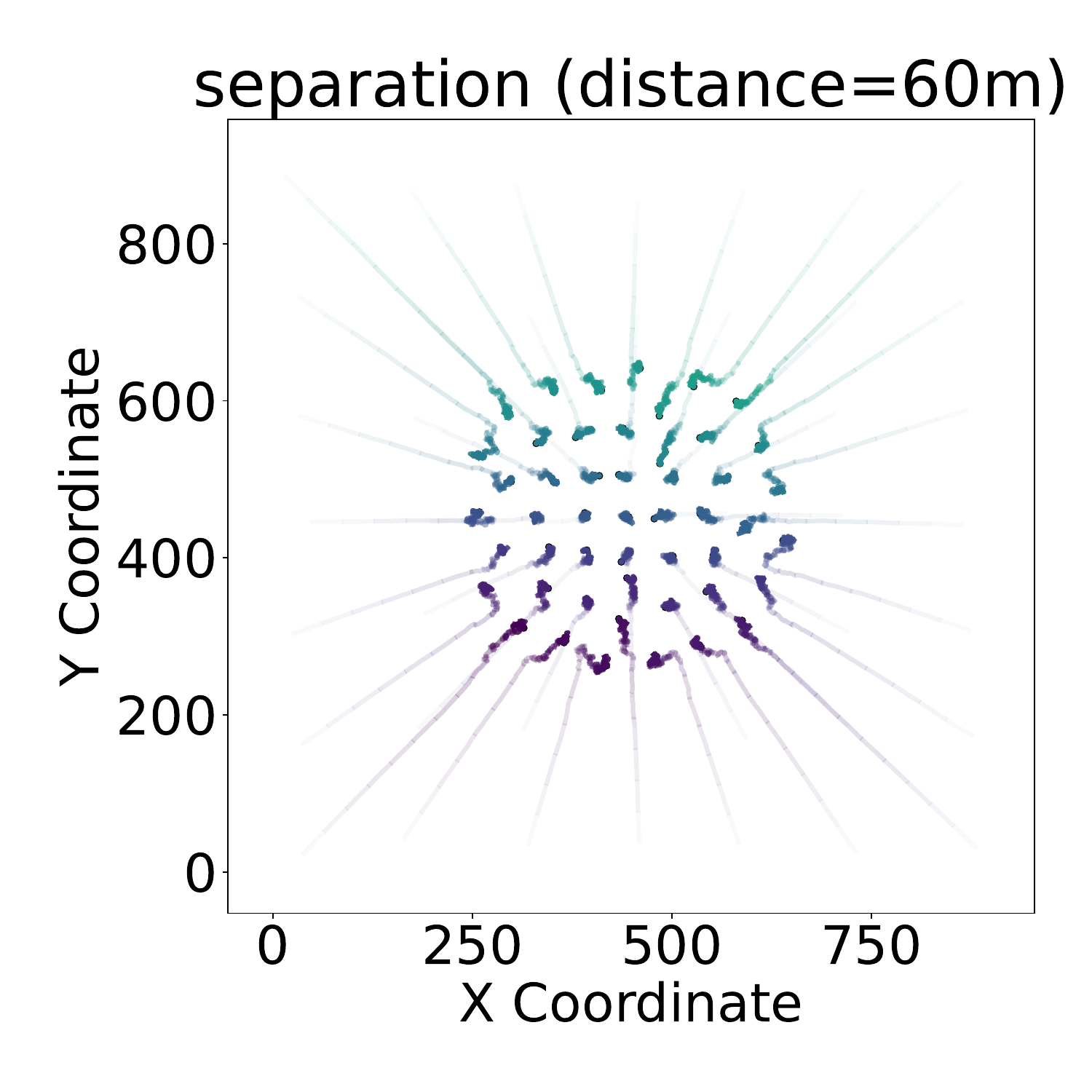}
  \end{subfigure}
  \hfill
  \begin{subfigure}[b]{0.3\textwidth}
    \includegraphics[width=\textwidth]{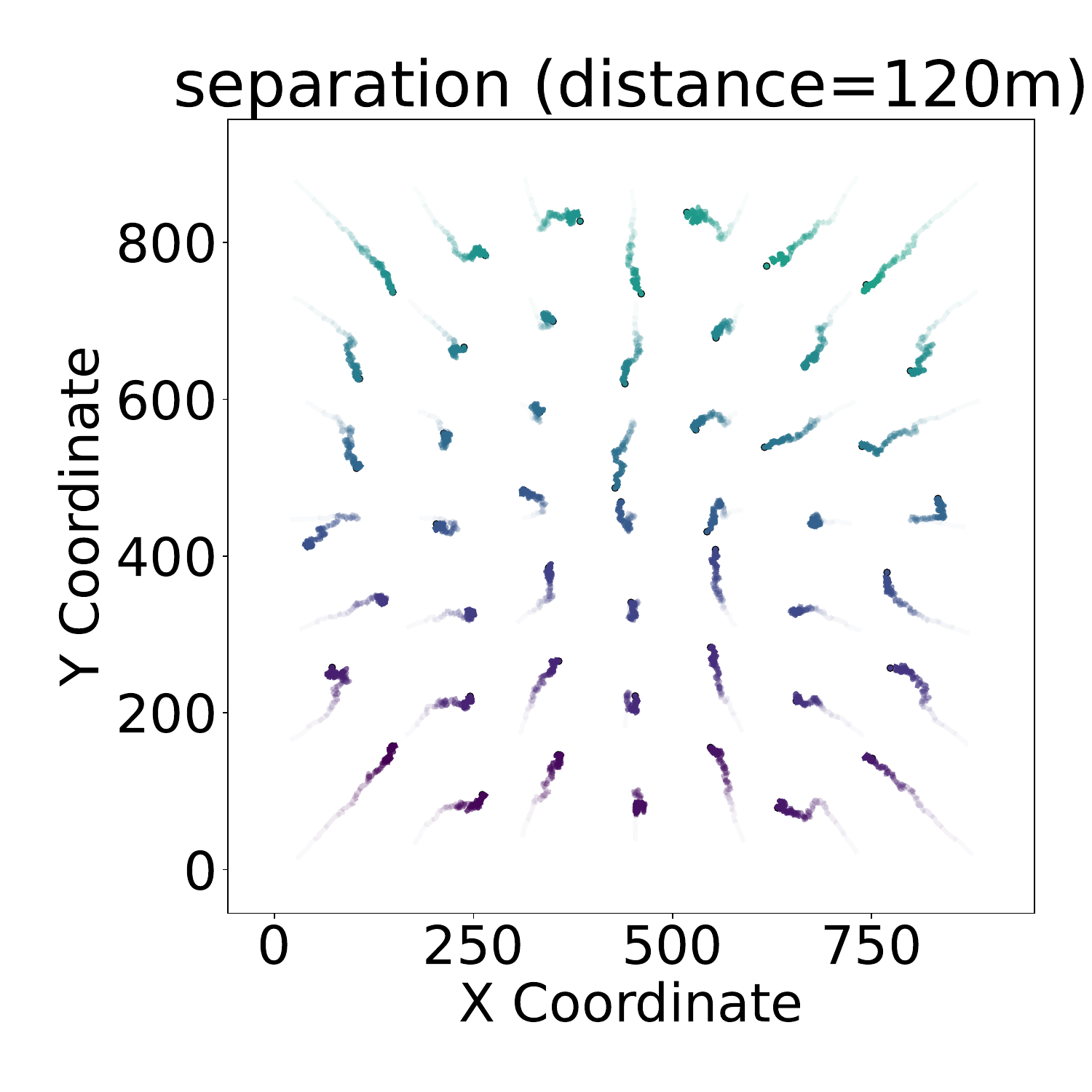}
  \end{subfigure}
  \centering
  \begin{subfigure}[b]{0.3\textwidth}
    \includegraphics[width=\textwidth]{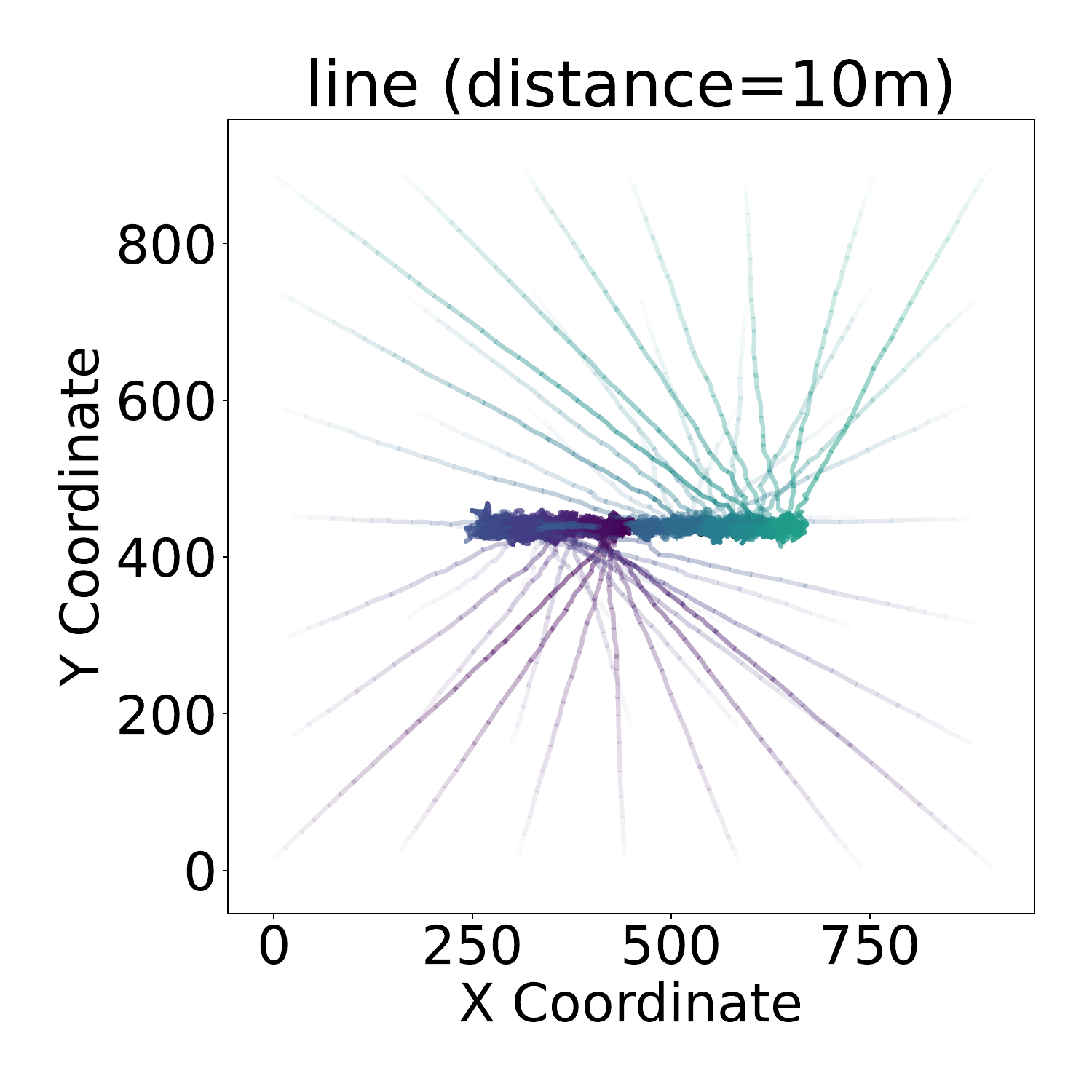}
  \end{subfigure}
  \hfill
  \begin{subfigure}[b]{0.3\textwidth}
    \includegraphics[width=\textwidth]{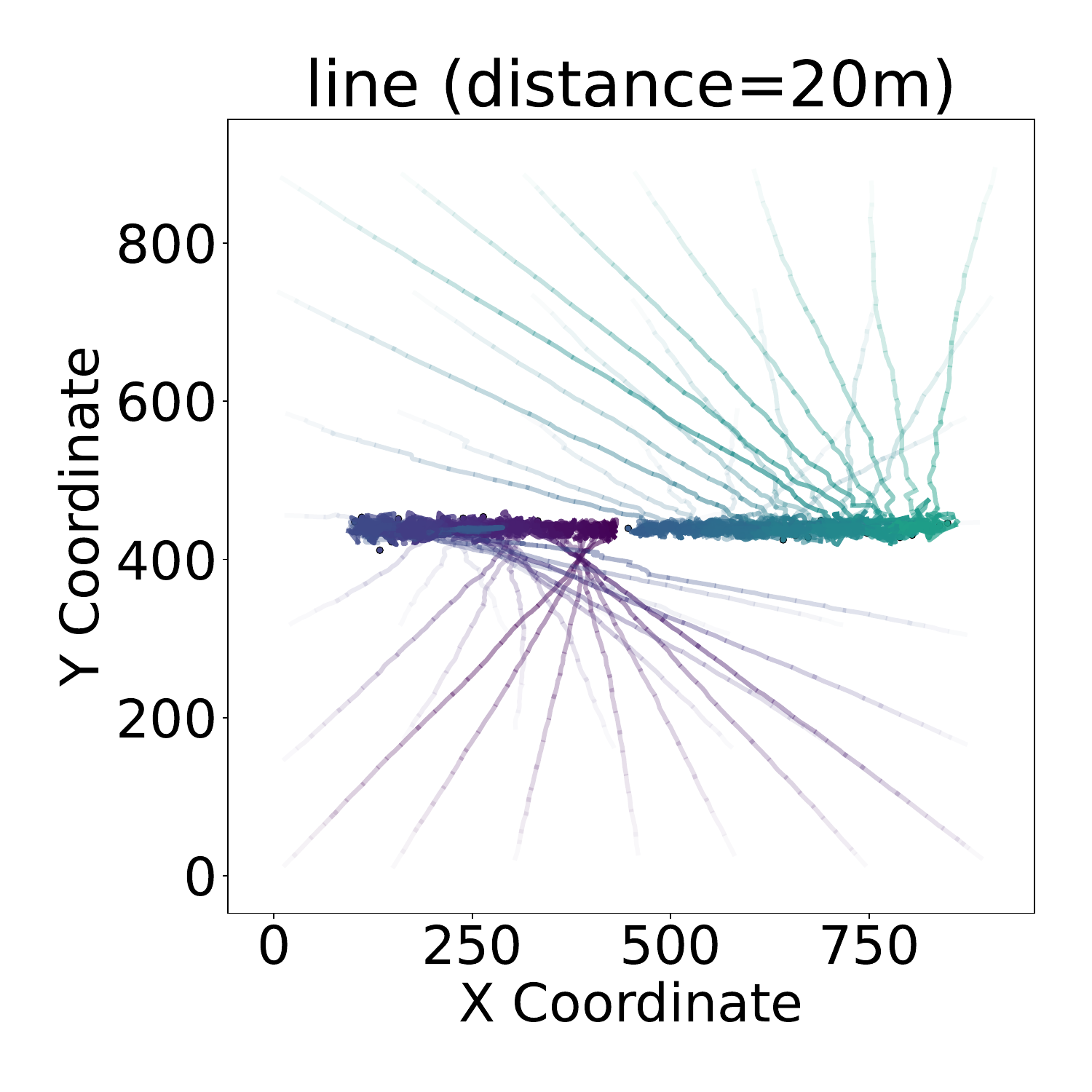}
  \end{subfigure}
  \hfill
  \begin{subfigure}[b]{0.3\textwidth}
    \includegraphics[width=\textwidth]{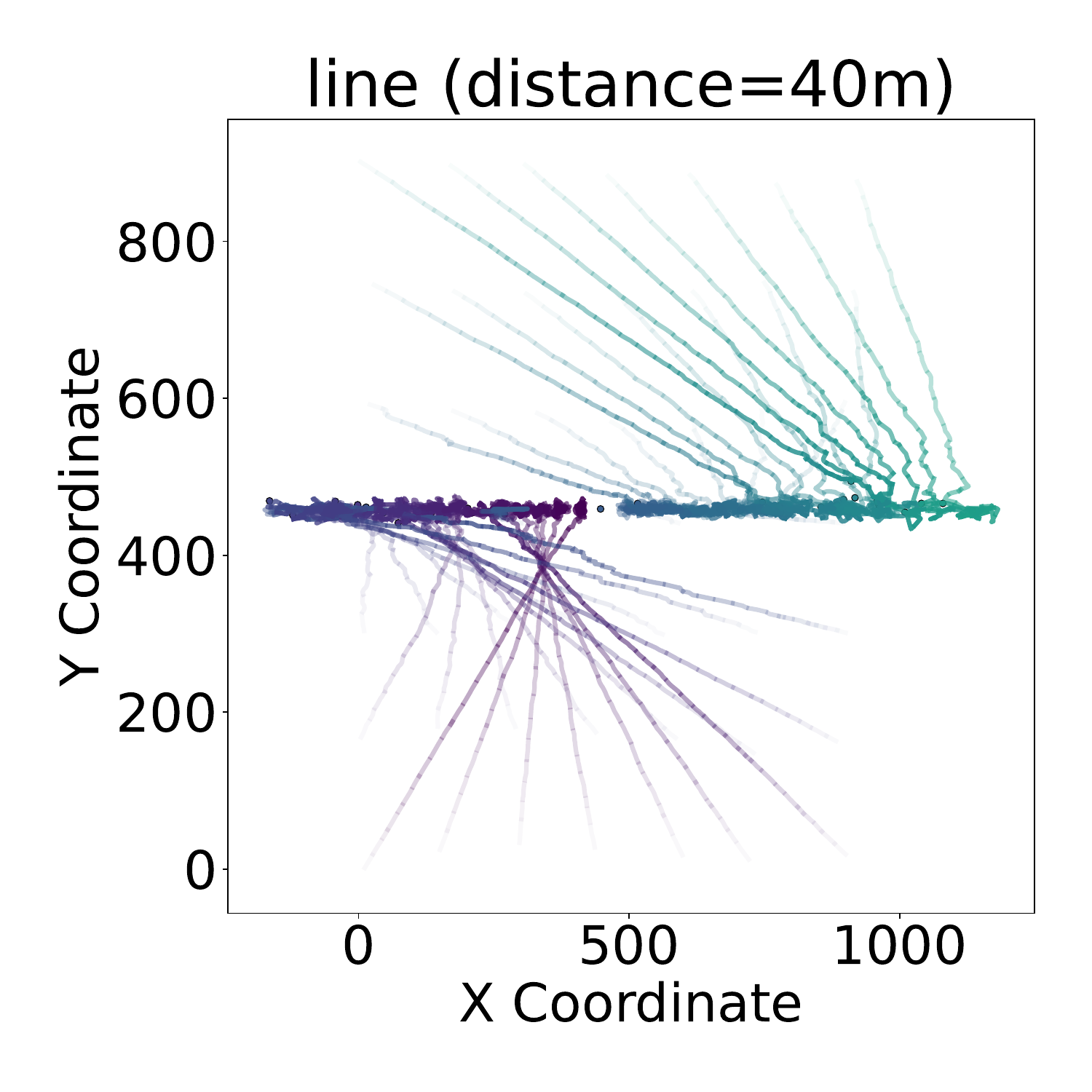}
  \end{subfigure}
  \centering
  \begin{subfigure}[b]{0.3\textwidth}
    \includegraphics[width=\textwidth]{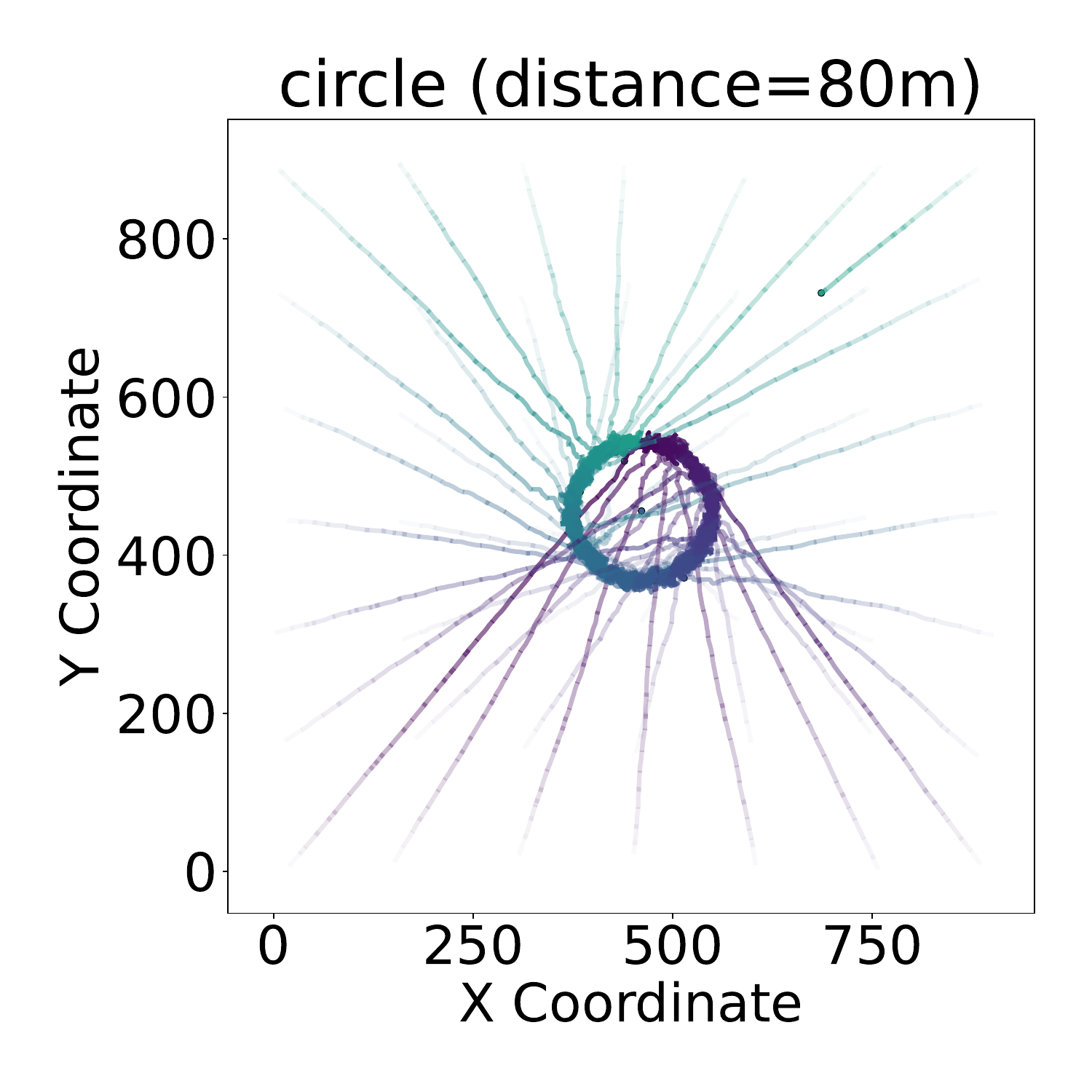}
  \end{subfigure}
  \hfill
  \begin{subfigure}[b]{0.3\textwidth}
    \includegraphics[width=\textwidth]{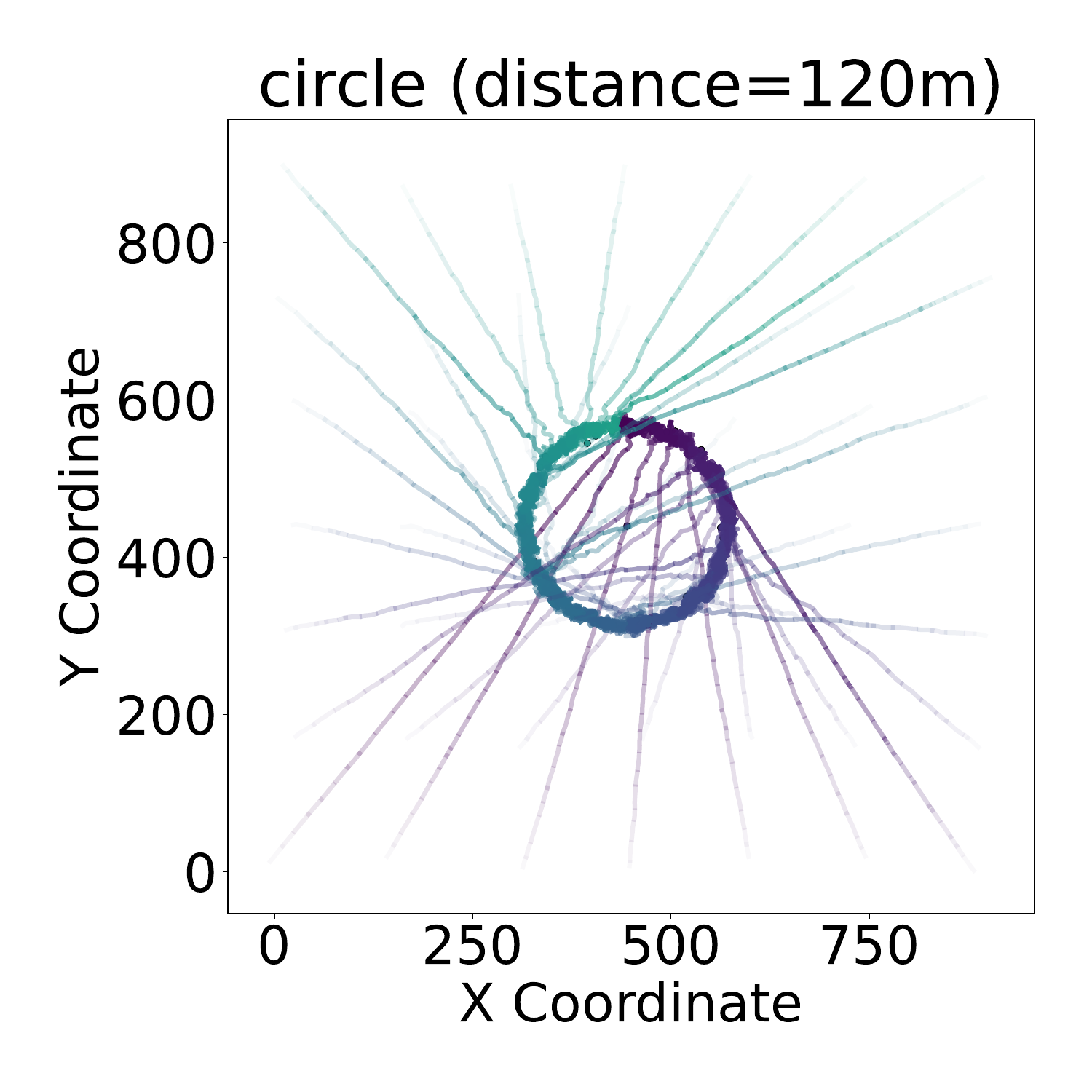}
  \end{subfigure}
  \hfill
  \begin{subfigure}[b]{0.3\textwidth}
    \includegraphics[width=\textwidth]{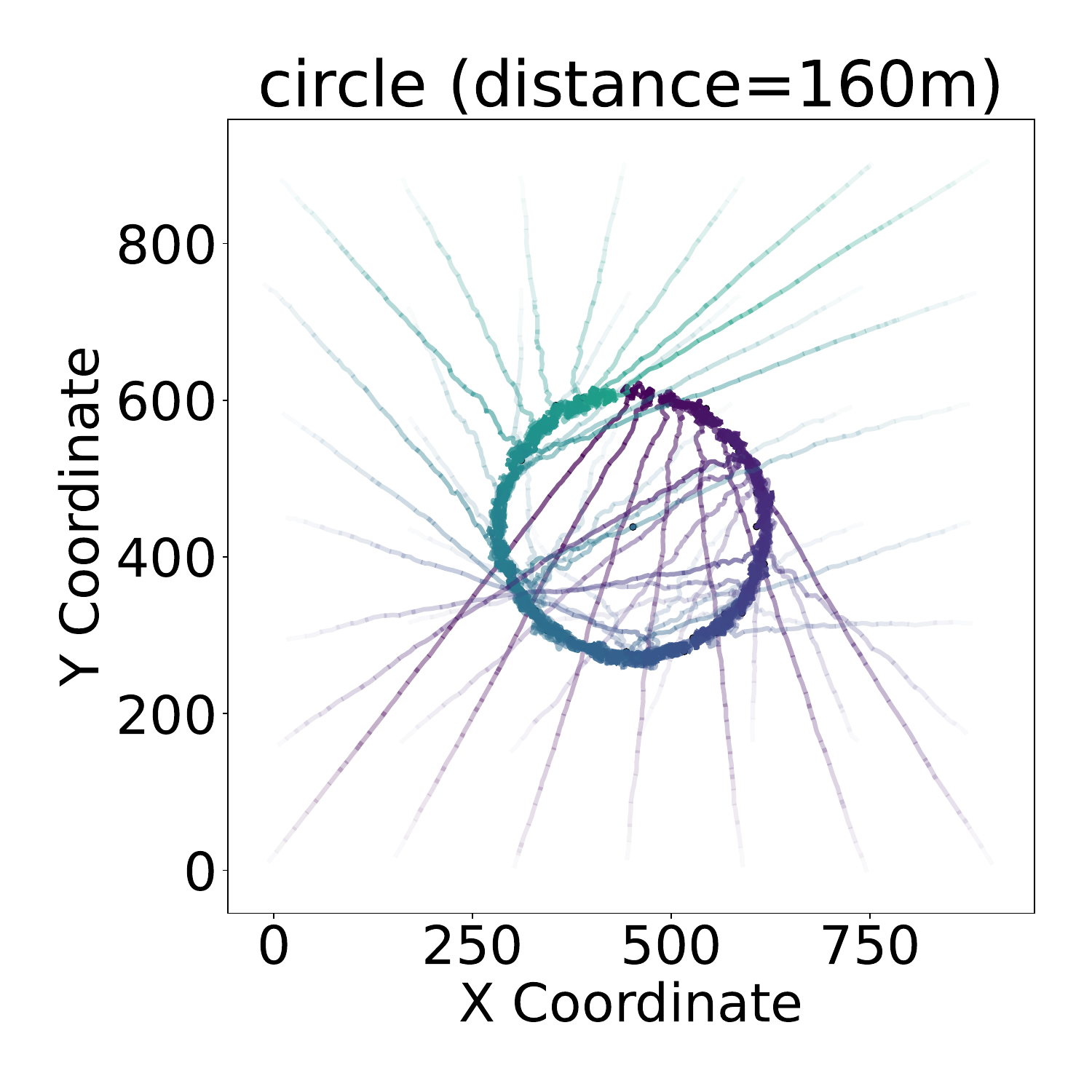}
  \end{subfigure}
  \caption{
    \revB{
    Shows the effect of a 70\% chance of message loss ($D=0.7$) and 10.0 meters of noise in the position perception ($P = 10$).
    V shape and line are strongly affected by the noise, while the circle is less affected but still with some very big errors (e.g., in distance=40m there is a node that is very far from the others).}
  }
  \label{fig:pattern-eval-error-both}
\end{figure}

\begin{figure}
  \centering
  \begin{subfigure}[b]{0.24\textwidth}
    \includegraphics[width=\textwidth]{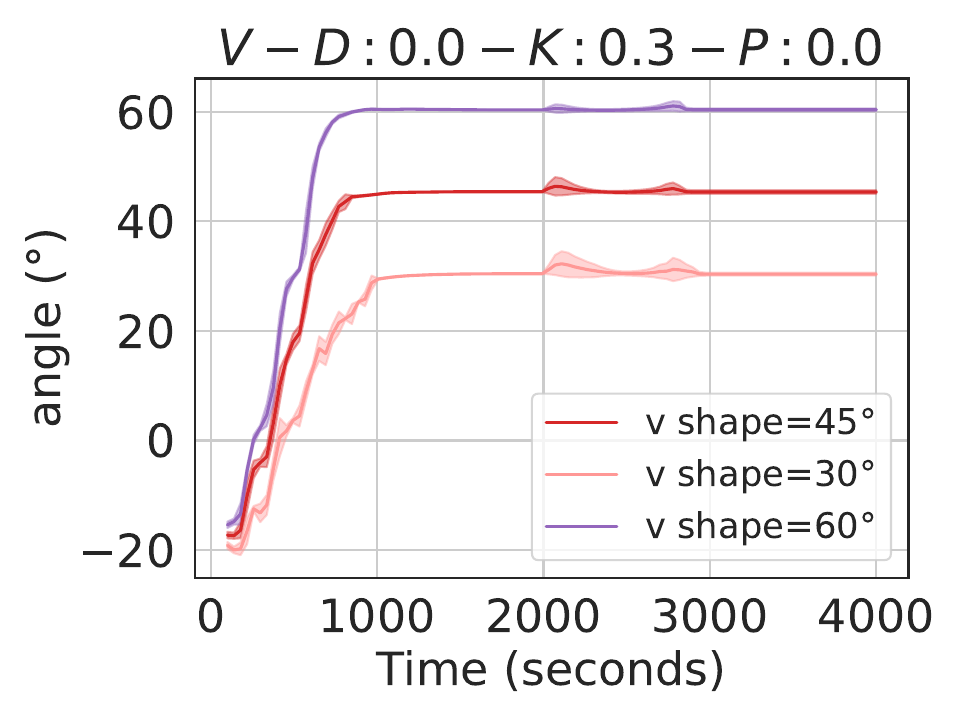}
    \caption{}\label{fig:detail-vshape-0}
  \end{subfigure}
  \begin{subfigure}[b]{0.24\textwidth}
    \includegraphics[width=\textwidth]{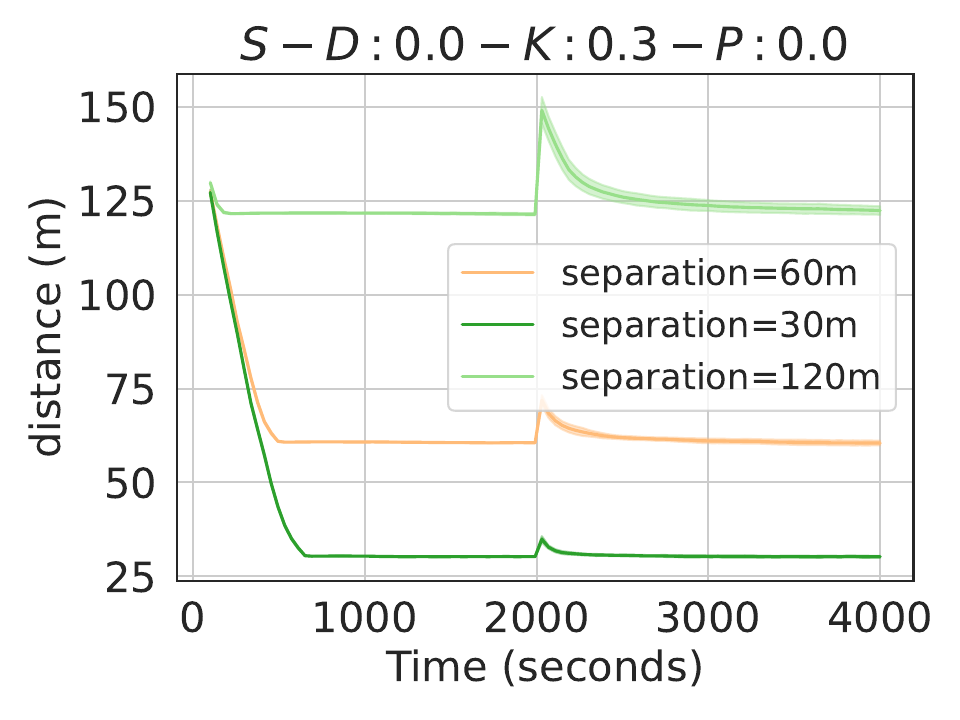}
    \caption{}\label{fig:detail-separation-0}
  \end{subfigure}
  \begin{subfigure}[b]{0.24\textwidth}
    \includegraphics[width=\textwidth]{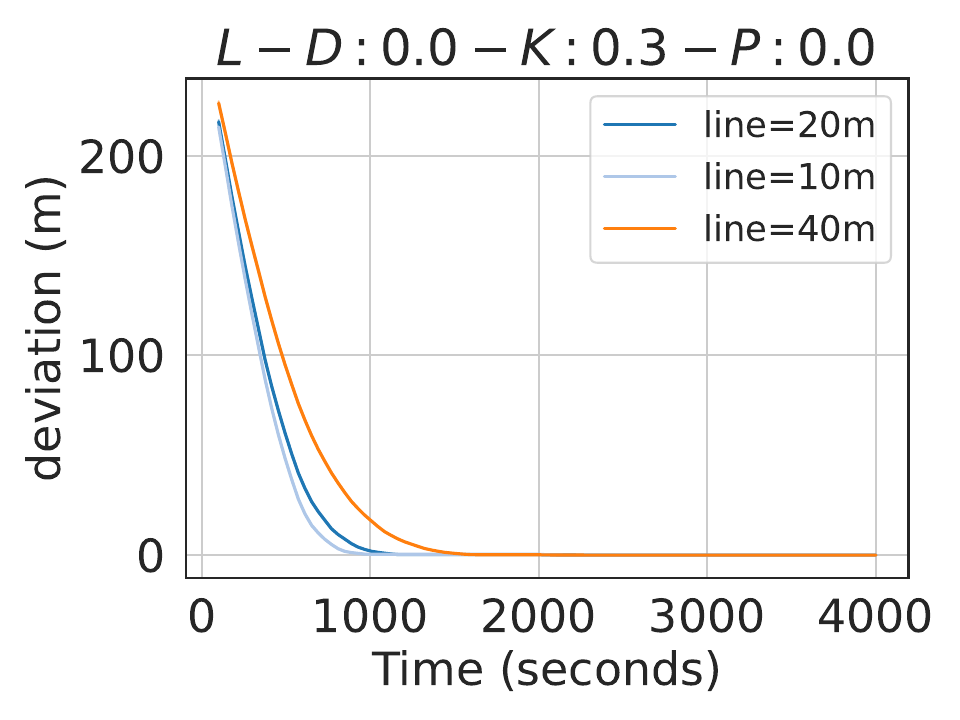}
    \caption{}\label{fig:detail-line-0}
  \end{subfigure}
  \begin{subfigure}[b]{0.24\textwidth}
    \includegraphics[width=\textwidth]{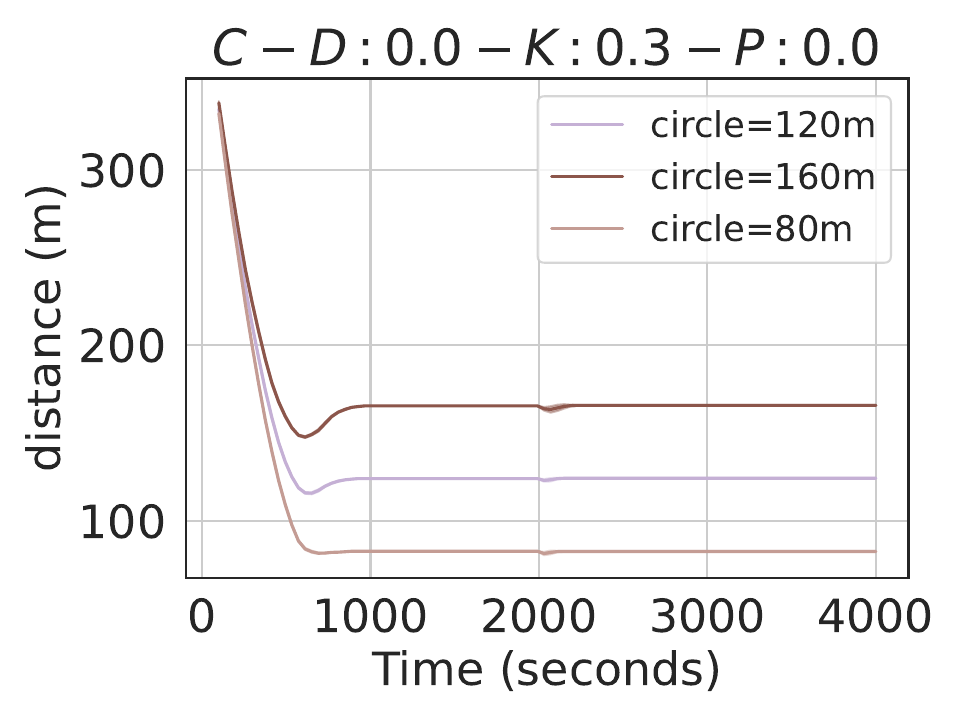}
    \caption{}\label{fig:detail-circle-0}
  \end{subfigure}
  \begin{subfigure}[b]{0.24\textwidth}
    \includegraphics[width=\textwidth]{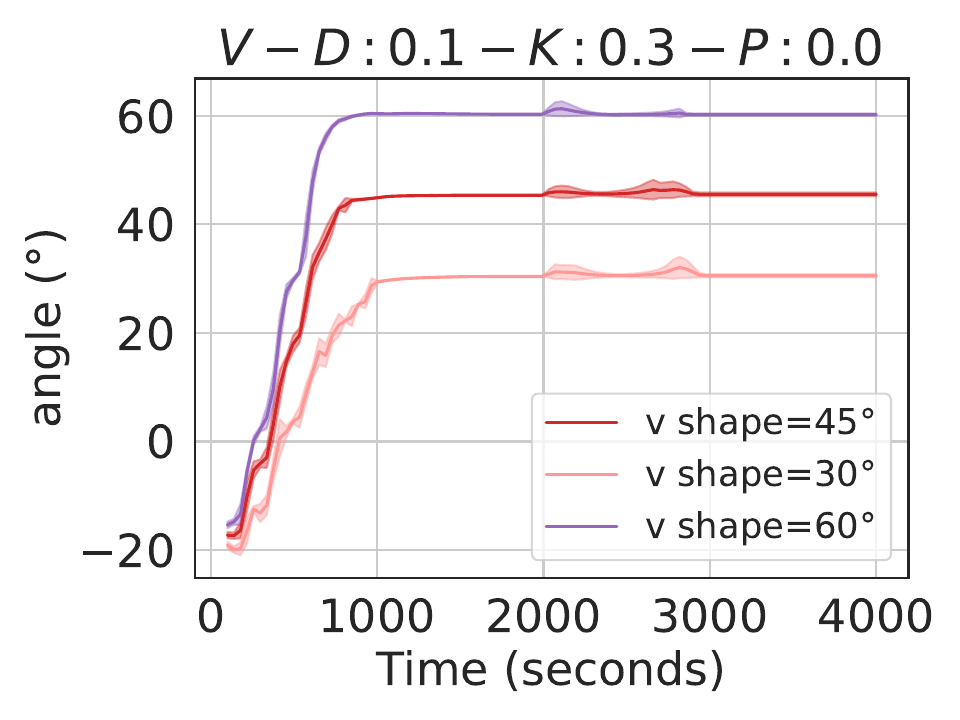}
    \caption{}\label{fig:detail-vshape-0.1}
  \end{subfigure}
  \begin{subfigure}[b]{0.24\textwidth}
    \includegraphics[width=\textwidth]{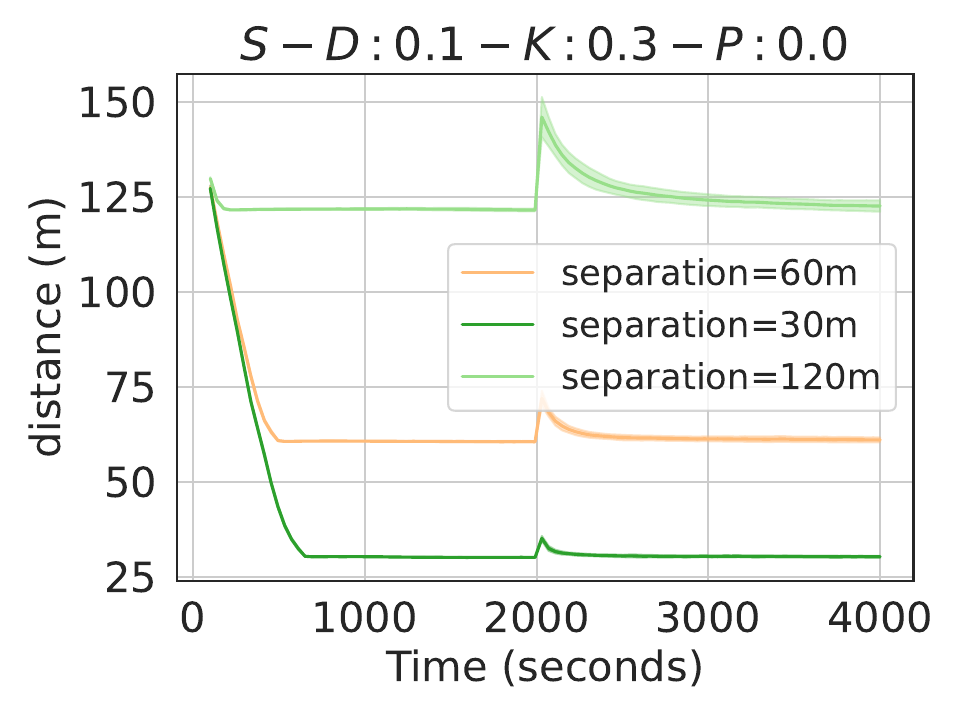}
    \caption{}\label{fig:detail-separation-0.1}
  \end{subfigure}
  \begin{subfigure}[b]{0.24\textwidth}
    \includegraphics[width=\textwidth]{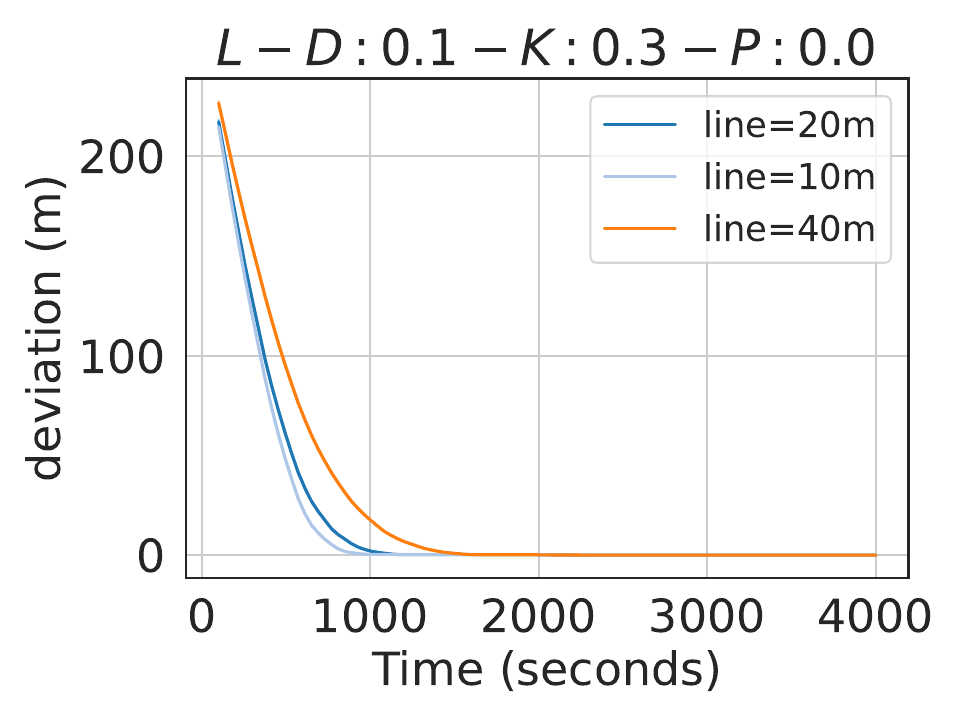}
    \caption{}\label{fig:detail-line-0.1}
  \end{subfigure}
  \begin{subfigure}[b]{0.24\textwidth}
    \includegraphics[width=\textwidth]{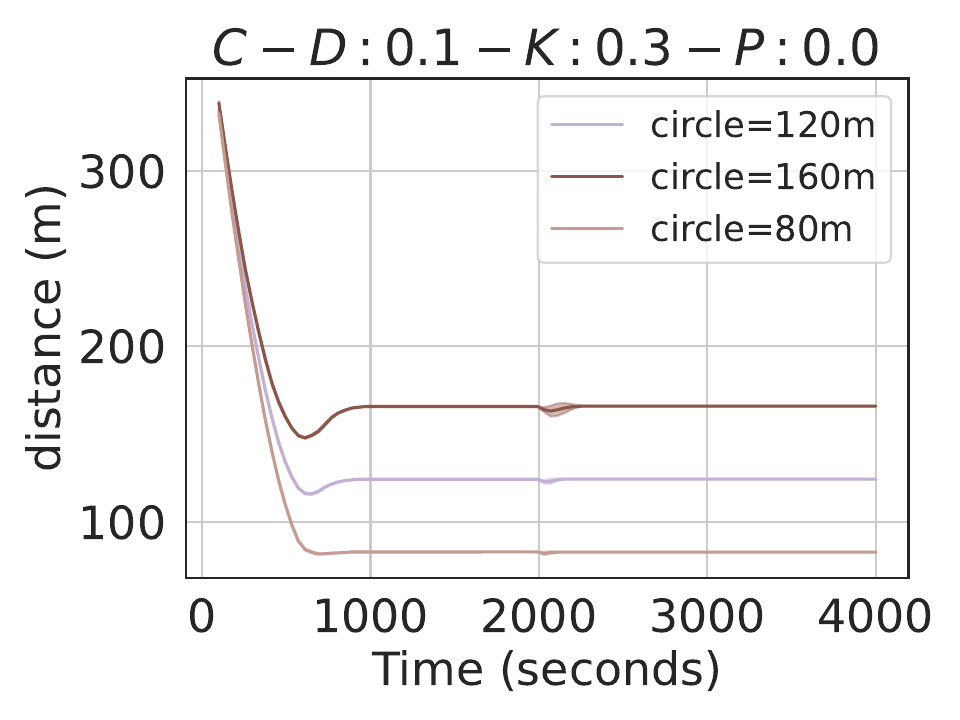}
    \caption{}\label{fig:detail-circle-0.1}
  \end{subfigure}
   \begin{subfigure}[b]{0.24\textwidth}
    \includegraphics[width=\textwidth]{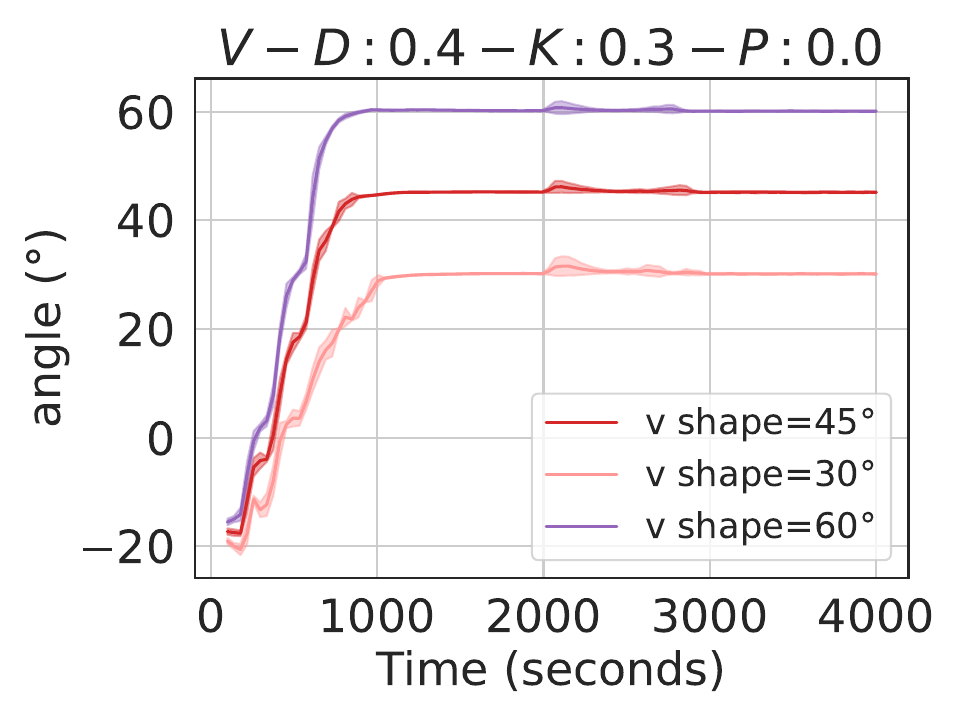}
    \caption{}\label{fig:detail-vshape-0.3}
  \end{subfigure}
  \begin{subfigure}[b]{0.24\textwidth}
    \includegraphics[width=\textwidth]{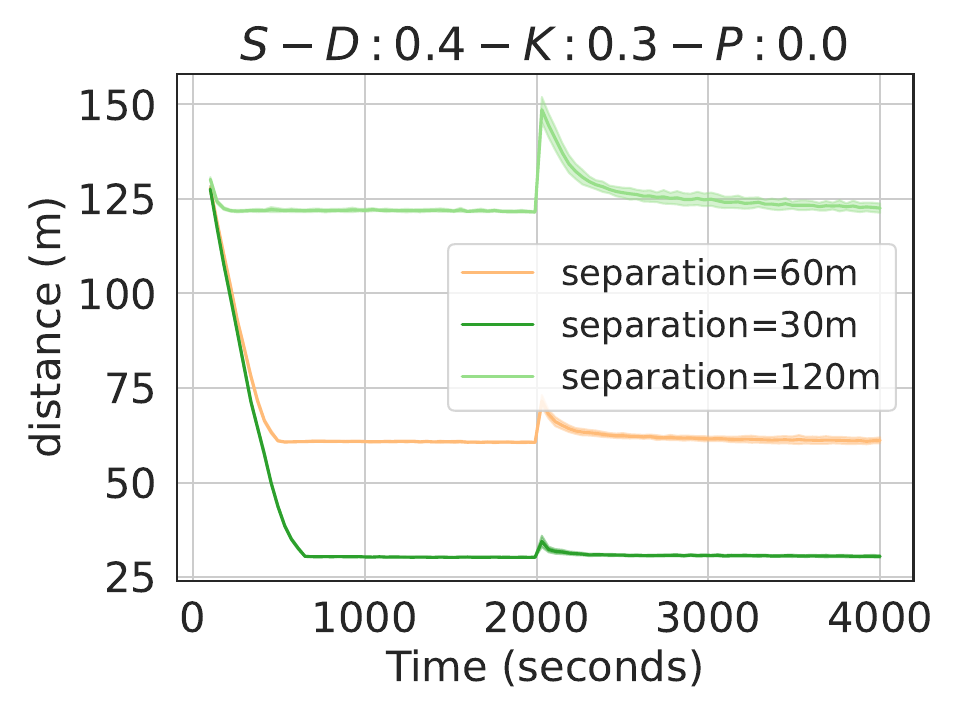}
    \caption{}\label{fig:detail-separation-0.3}
  \end{subfigure}
  \begin{subfigure}[b]{0.24\textwidth}
    \includegraphics[width=\textwidth]{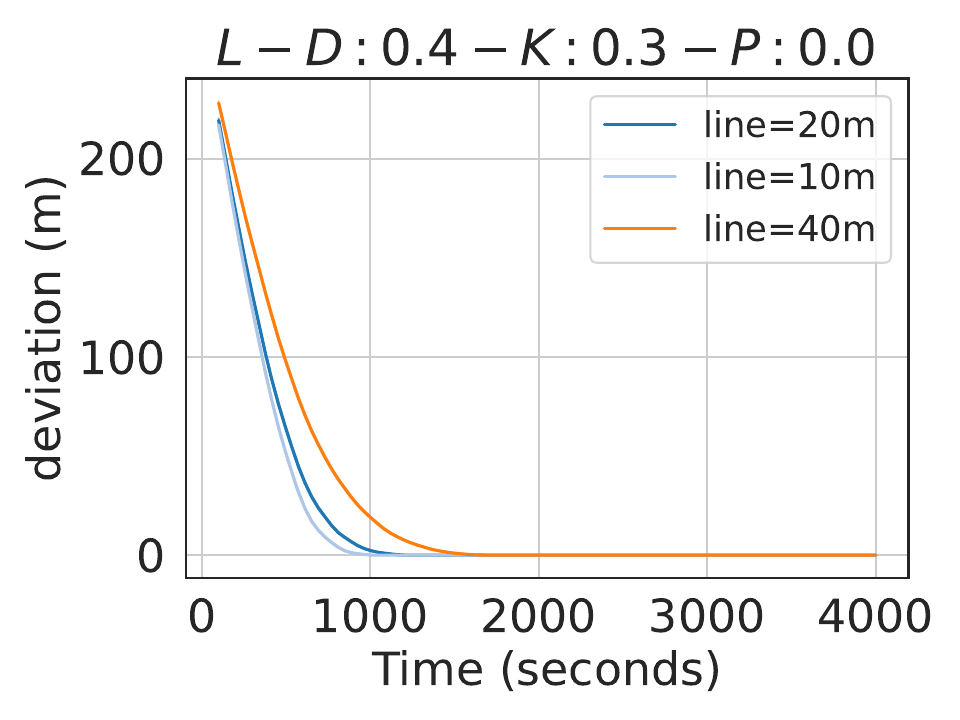}
    \caption{}\label{fig:detail-line-0.3}
  \end{subfigure}
  \begin{subfigure}[b]{0.24\textwidth}
    \includegraphics[width=\textwidth]{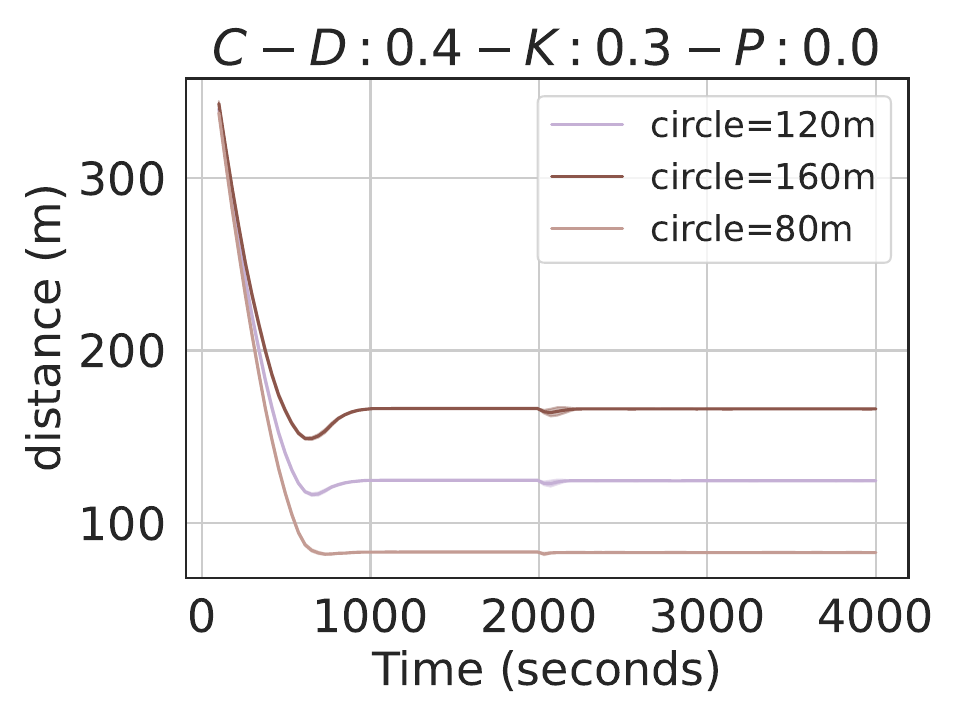}
    \caption{}\label{fig:detail-circle-0.3}
  \end{subfigure}
  \begin{subfigure}[b]{0.24\textwidth}
  \includegraphics[width=\textwidth]{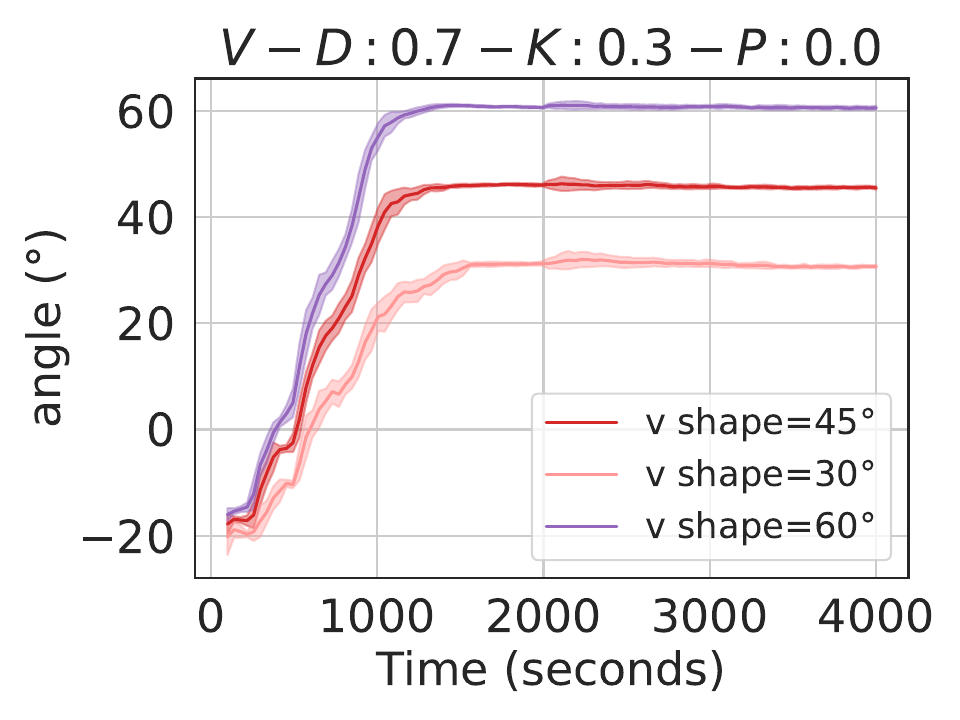}
  \caption{}\label{fig:detail-vshape-0.7}
  \end{subfigure}
  \begin{subfigure}[b]{0.24\textwidth}
    \includegraphics[width=\textwidth]{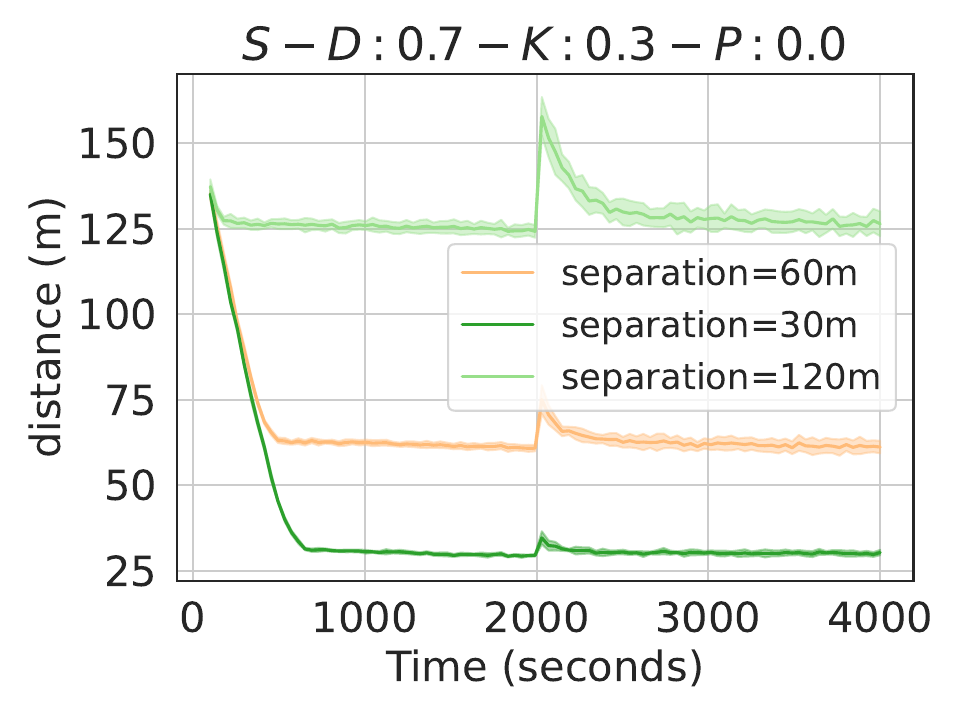}
    \caption{}\label{fig:detail-separation-0.7}
  \end{subfigure}
  \begin{subfigure}[b]{0.24\textwidth}
    \includegraphics[width=\textwidth]{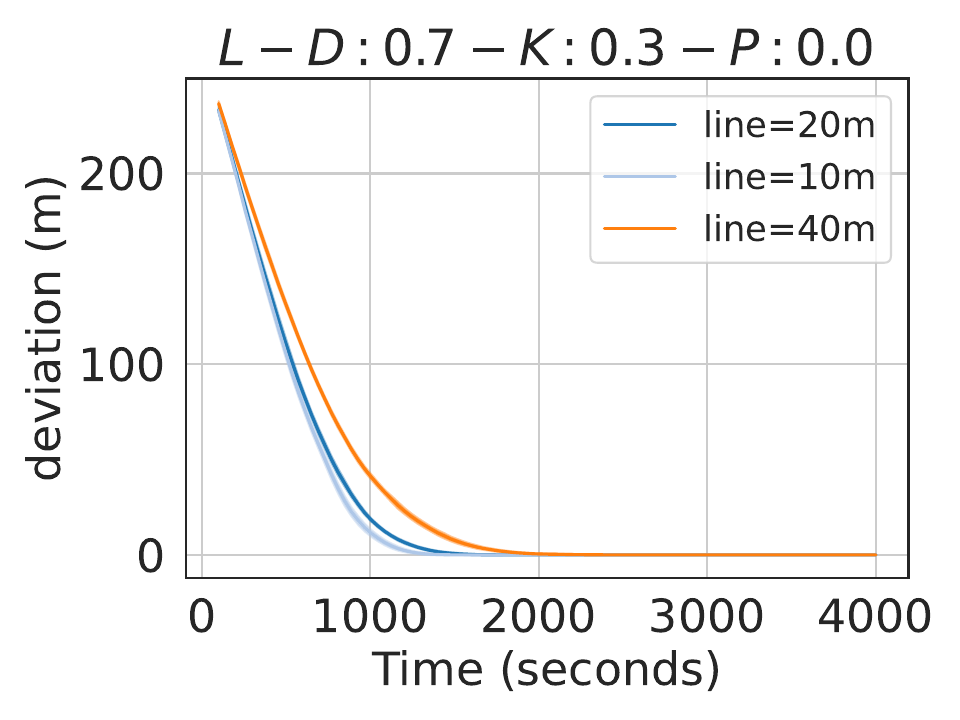}
    \caption{}\label{fig:detail-line-0.7}
  \end{subfigure}
  \begin{subfigure}[b]{0.24\textwidth}
    \includegraphics[width=\textwidth]{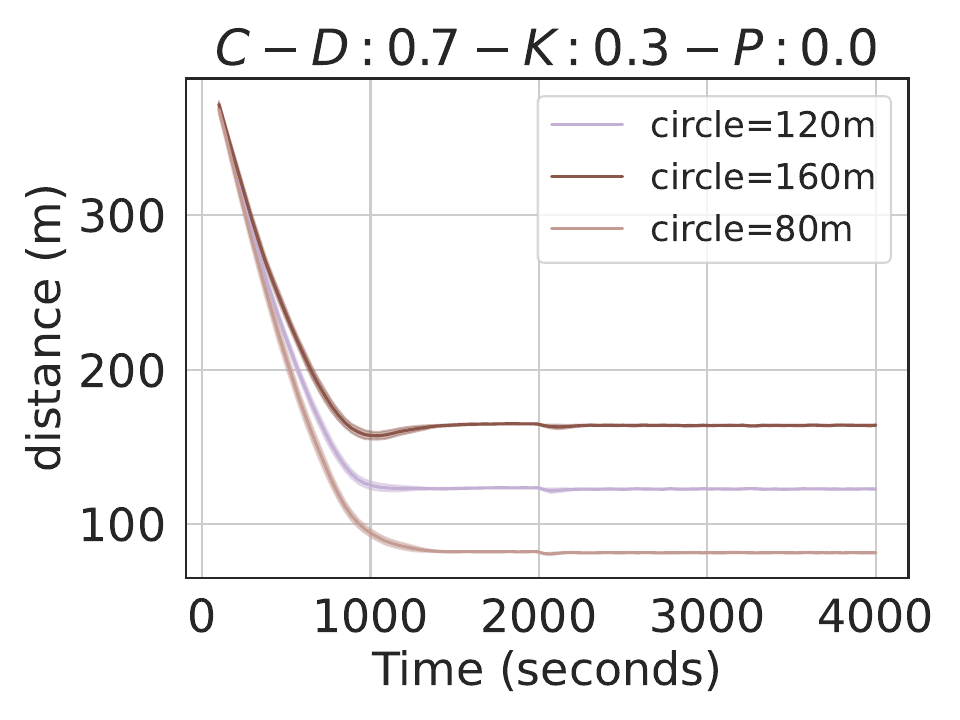}
    \caption{}\label{fig:detail-circle-0.7}
  \end{subfigure}
  \caption{
  \revB{  
  Convergence analysis of the pattern formation blocks. 
  Each of them, after an initial transient, converges to the desired value.
  Moreover, observing the effect of node faults (T $>$ 2000s) we can see that the system is able to recover the desired shape.
  The message loss seems to have a bigger impact on V shape and separation formation, while the line and circle are less affected. 
  In all structures, the message loss introduces a longer transient phase.}}
  
  \label{fig:pattern-eval-depth-message-loss}
\end{figure}

\begin{figure}
  \begin{subfigure}[b]{0.24\textwidth}
    \includegraphics[width=\textwidth]{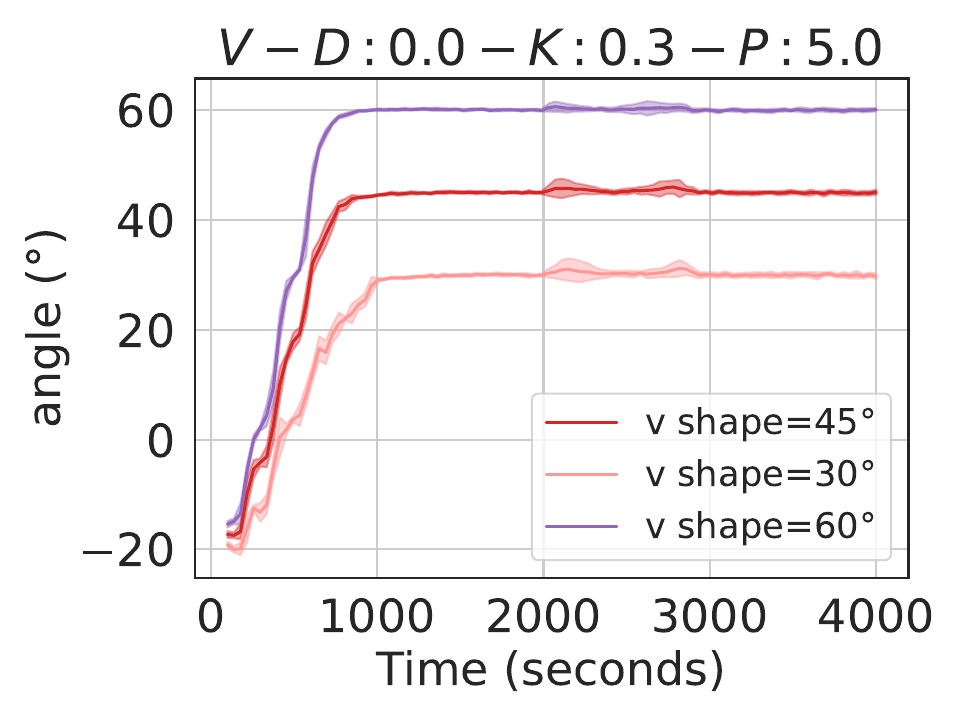}
    \caption{}\label{fig:detail-vshape-5.0}
  \end{subfigure}
  \begin{subfigure}[b]{0.24\textwidth}
    \includegraphics[width=\textwidth]{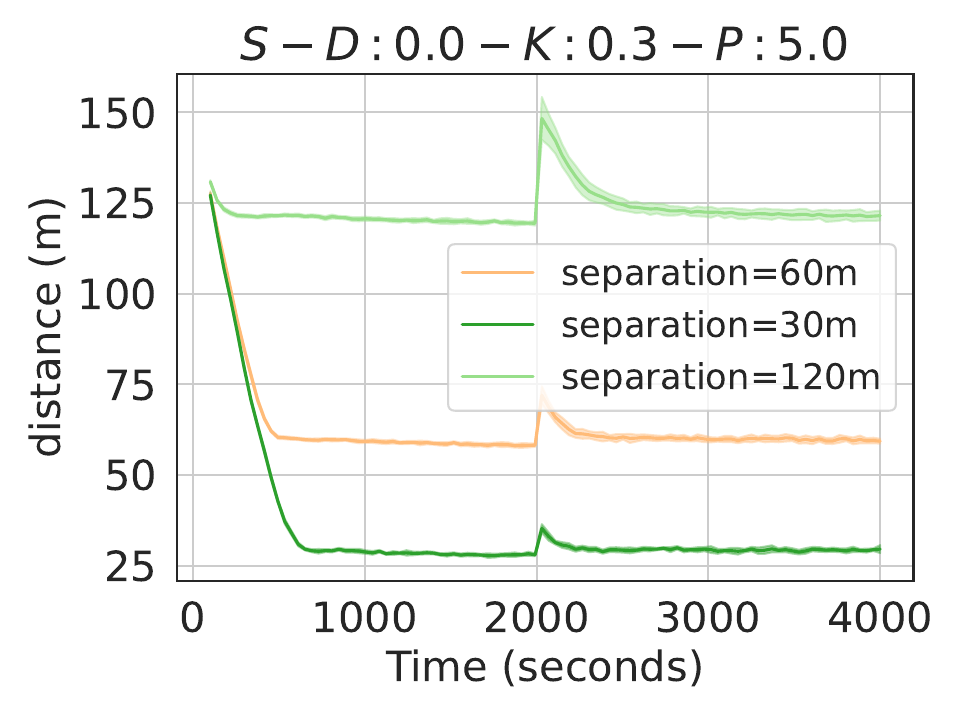}
    \caption{}\label{fig:detail-separation-5.0}
  \end{subfigure}
  \begin{subfigure}[b]{0.24\textwidth}
    \includegraphics[width=\textwidth]{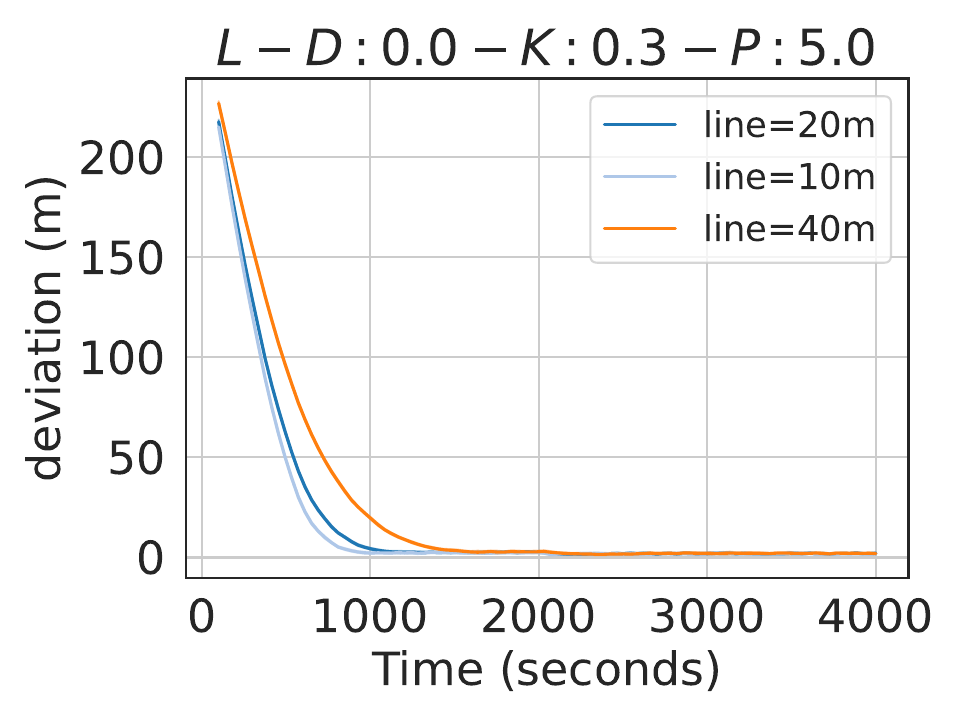}
    \caption{}\label{fig:detail-line-5.0}
  \end{subfigure}
  \begin{subfigure}[b]{0.24\textwidth}
    \includegraphics[width=\textwidth]{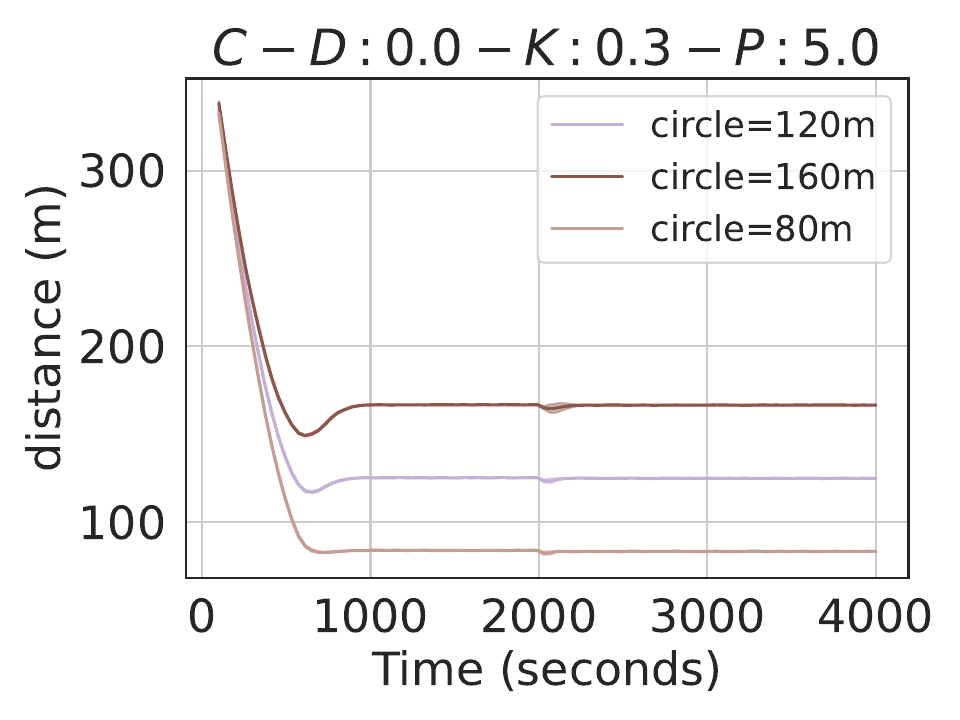}
    \caption{}\label{fig:detail-circle-5.0}
  \end{subfigure}
  \begin{subfigure}[b]{0.24\textwidth}
    \includegraphics[width=\textwidth]{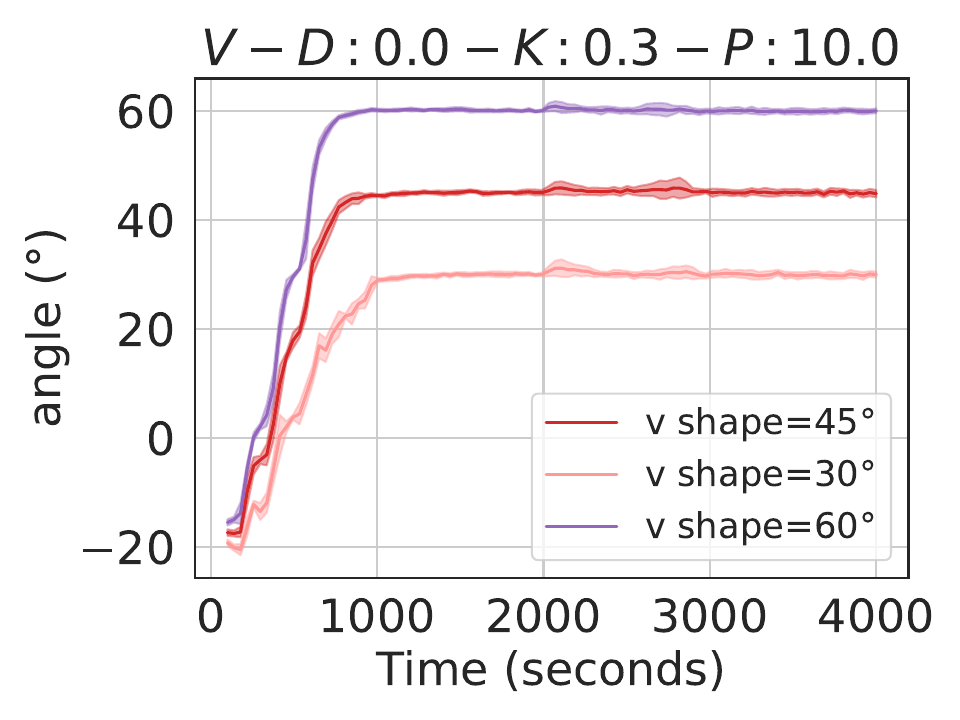}
    \caption{}\label{fig:detail-vshape-10.0}
  \end{subfigure}
  \begin{subfigure}[b]{0.24\textwidth}
    \includegraphics[width=\textwidth]{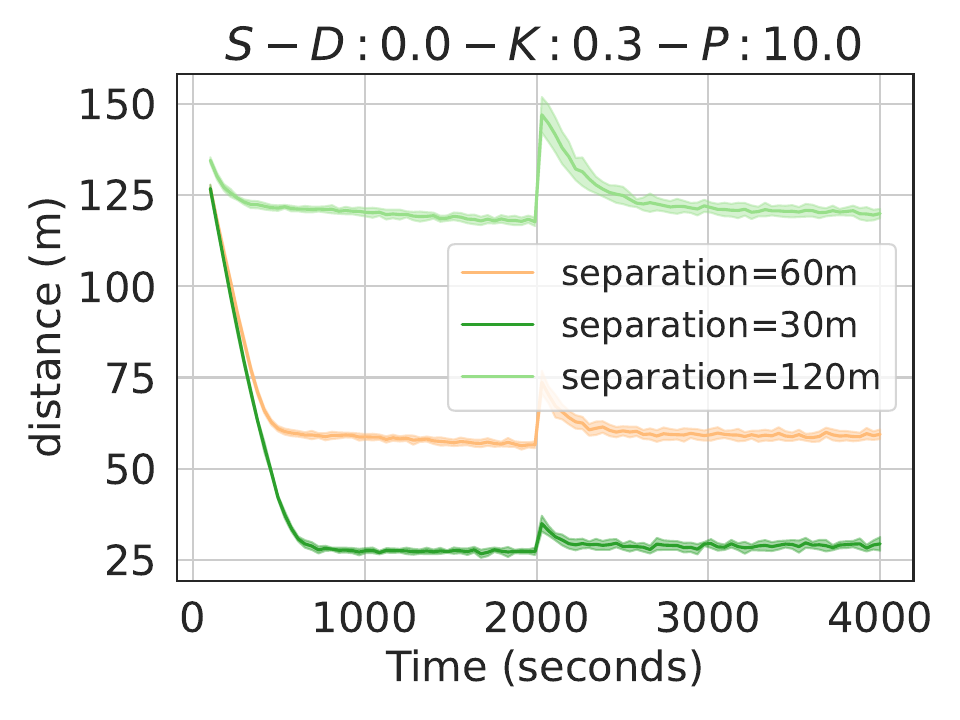}
    \caption{}\label{fig:detail-separation-10.0}
  \end{subfigure}
  \begin{subfigure}[b]{0.24\textwidth}
    \includegraphics[width=\textwidth]{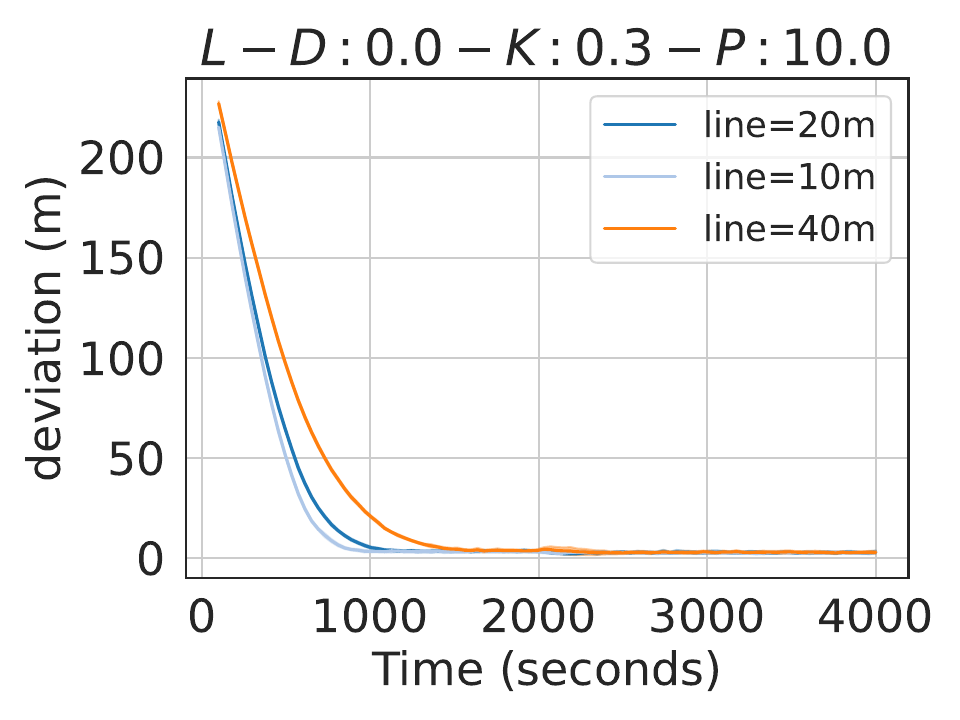}
    \caption{}\label{fig:detail-line-10.0}
  \end{subfigure}
  \begin{subfigure}[b]{0.24\textwidth}
    \includegraphics[width=\textwidth]{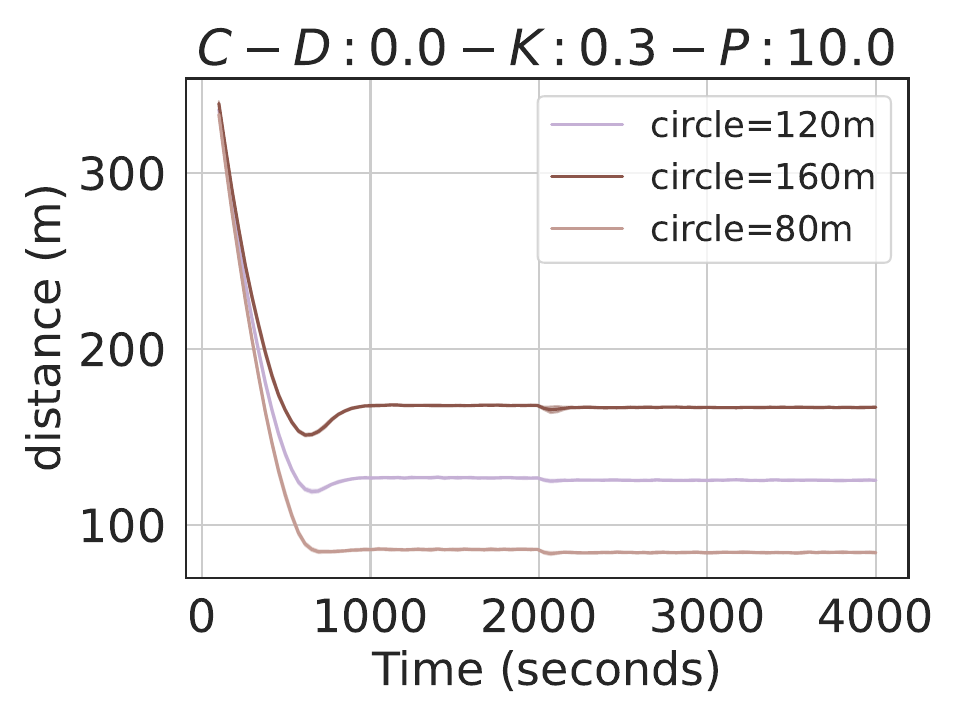}
    \caption{}\label{fig:detail-circle-10.0}
  \end{subfigure}
  \caption{\revB{Shows the effect of perception noise in the position of the drones.
  This introduces a small deformation in the structure, but the drones are still able to form the desired shape--more details in the errors analysis.}}
  \label{fig:pattern-eval-depth-perception}
\end{figure}

\begin{figure}
  \centering
  \begin{subfigure}[b]{0.24\textwidth}
    \includegraphics[width=\textwidth]{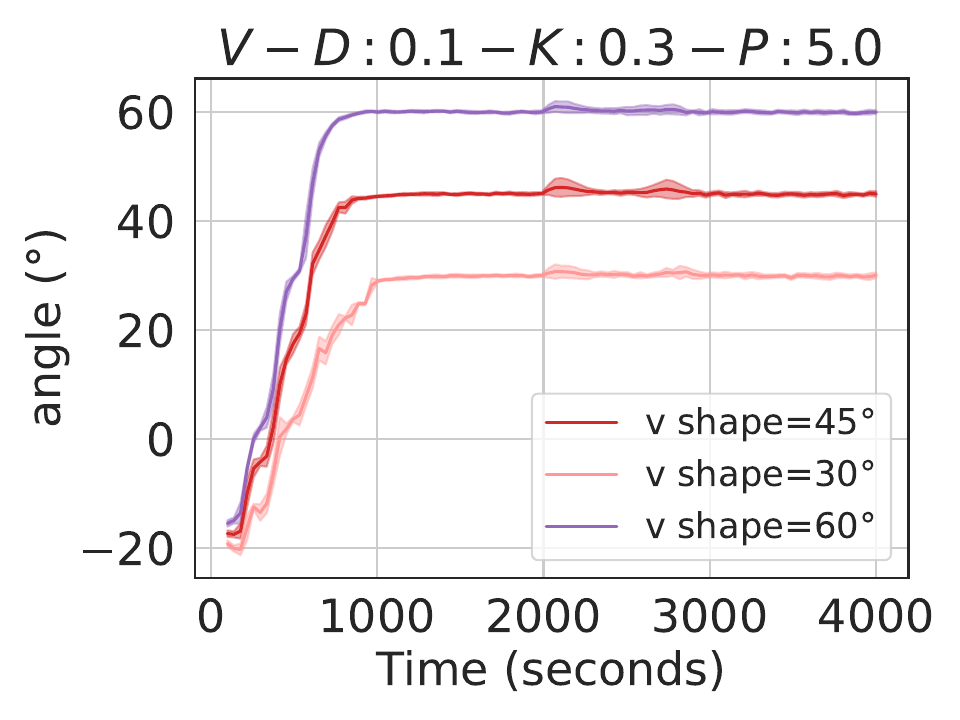}
    \caption{}\label{fig:detail-vshape-0.1-5.0}
  \end{subfigure}
  \begin{subfigure}[b]{0.24\textwidth}
    \includegraphics[width=\textwidth]{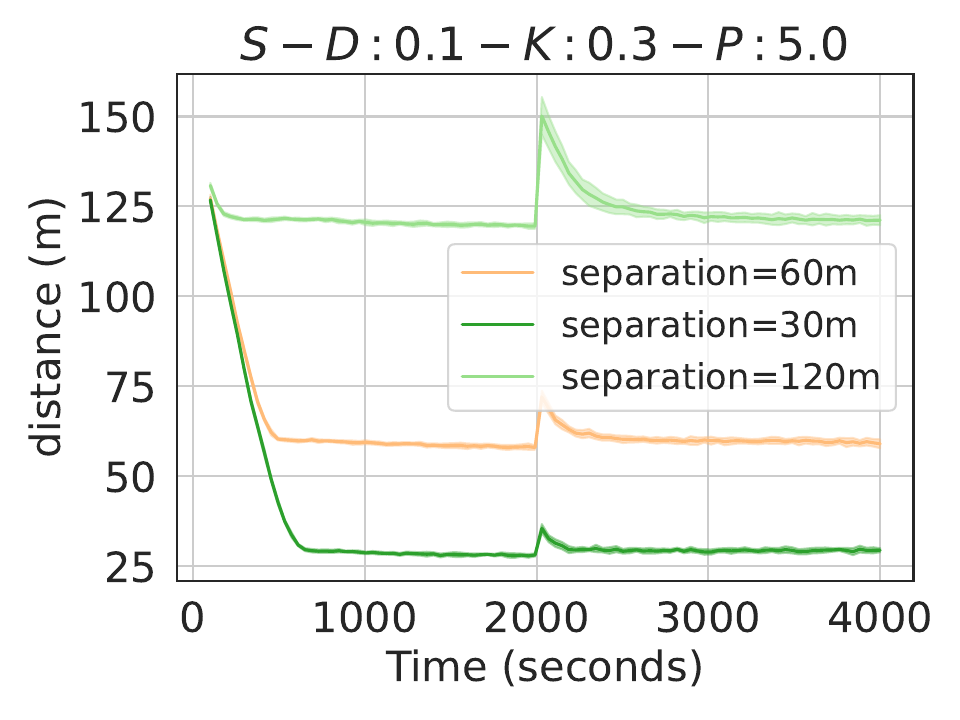}
    \caption{}\label{fig:detail-separation-0.1-5.0}
  \end{subfigure}
  \begin{subfigure}[b]{0.24\textwidth}
    \includegraphics[width=\textwidth]{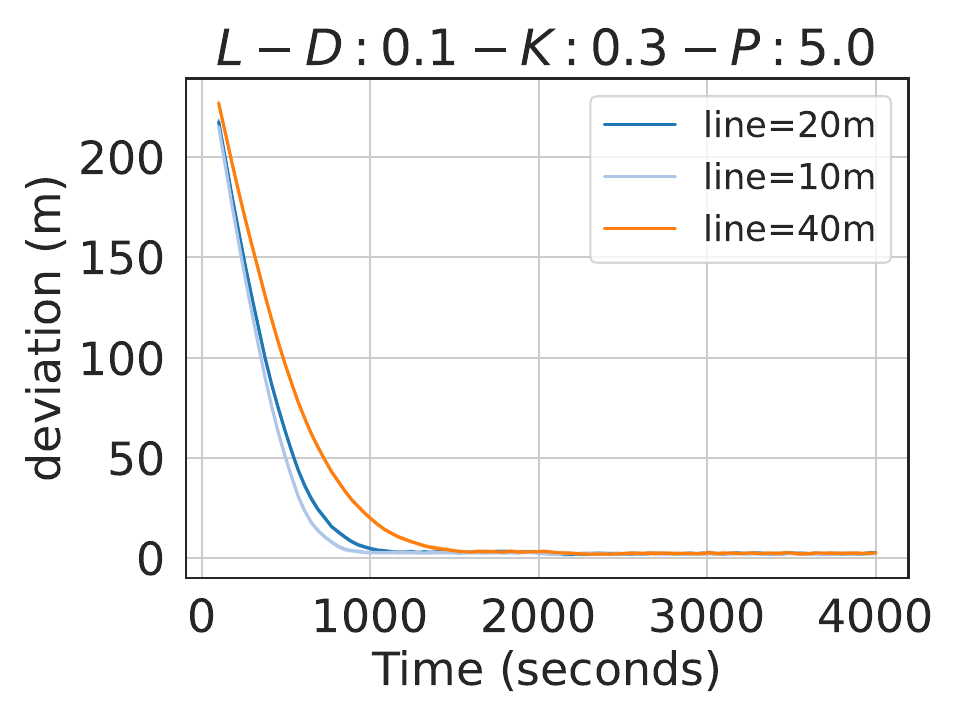}
    \caption{}\label{fig:detail-line-0.1-5.0}
  \end{subfigure}
  \begin{subfigure}[b]{0.24\textwidth}
    \includegraphics[width=\textwidth]{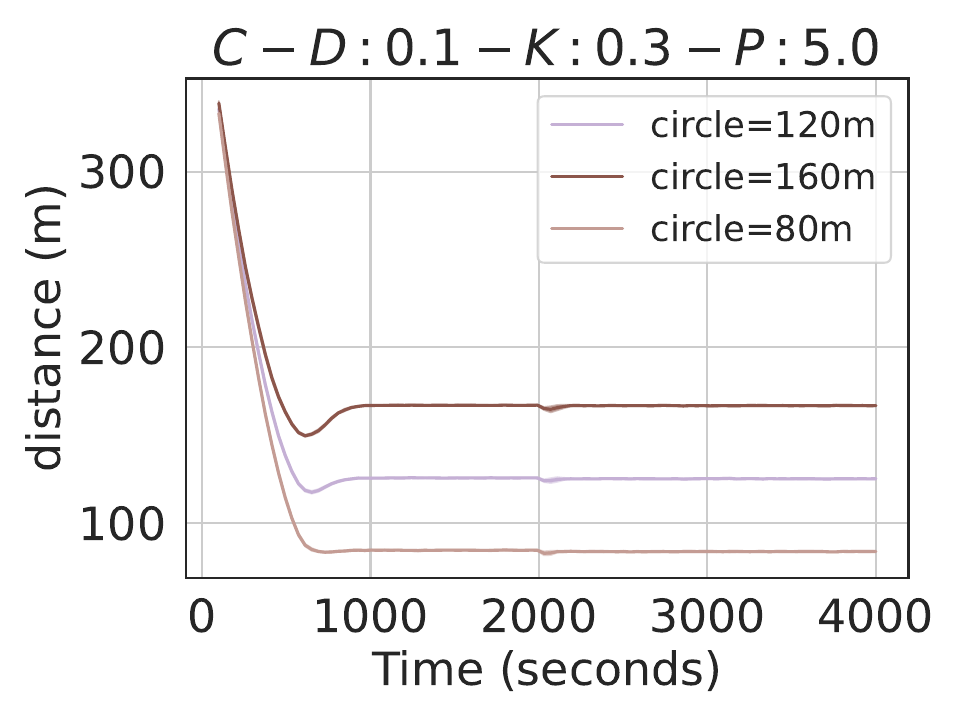}
    \caption{}\label{fig:detail-circle-0.1-5.0}
  \end{subfigure}\hfill

  \begin{subfigure}[b]{0.24\textwidth}
    \includegraphics[width=\textwidth]{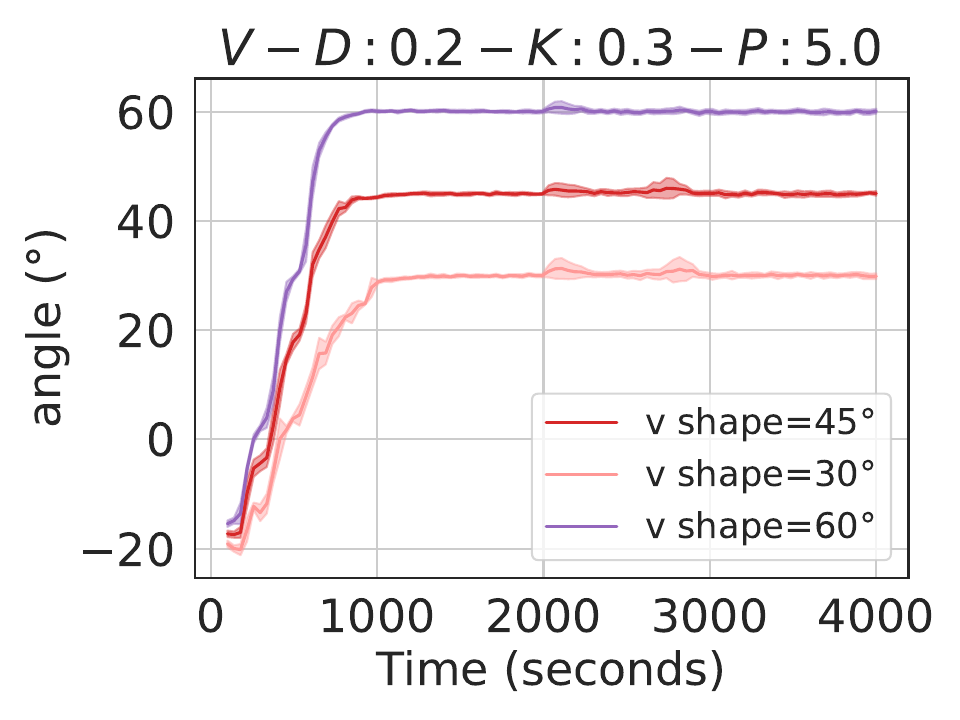}
    \caption{}\label{fig:detail-vshape-0.2-5.0}
  \end{subfigure}
  \begin{subfigure}[b]{0.24\textwidth}
    \includegraphics[width=\textwidth]{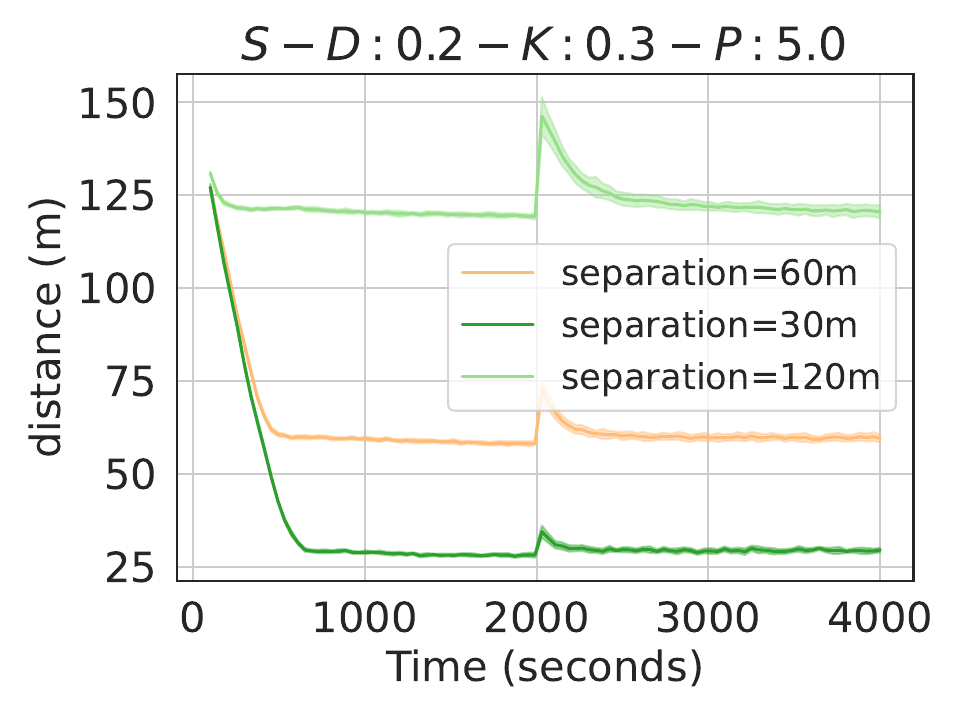}
    \caption{}\label{fig:detail-separation-0.2-5.0}
  \end{subfigure}
  \begin{subfigure}[b]{0.24\textwidth}
    \includegraphics[width=\textwidth]{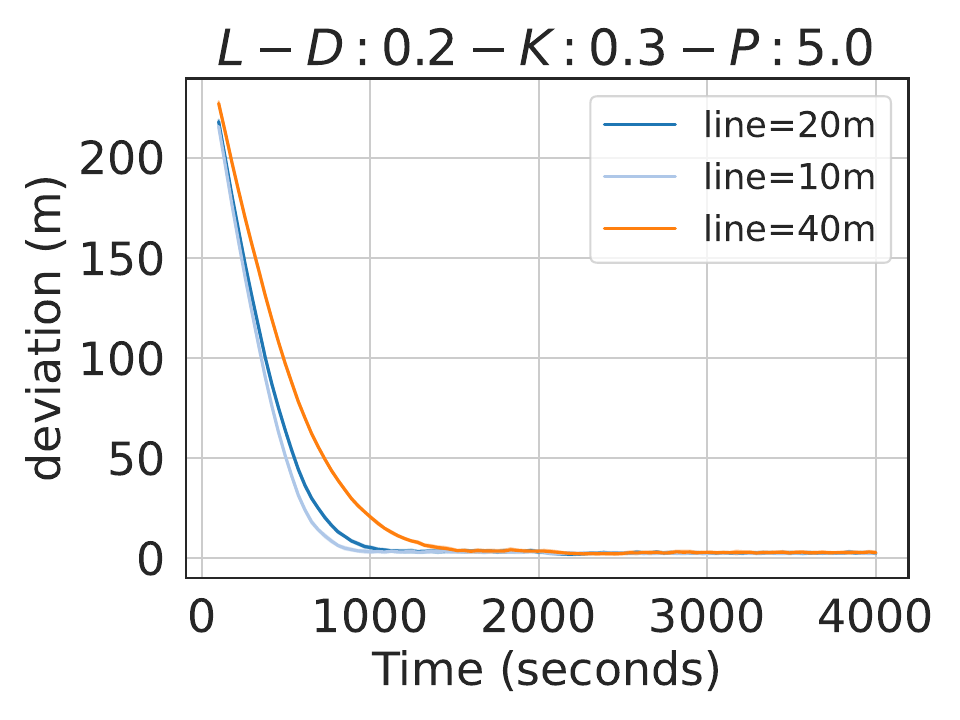}
    \caption{}\label{fig:detail-line-0.2-5.0}
  \end{subfigure}
  \begin{subfigure}[b]{0.24\textwidth}
    \includegraphics[width=\textwidth]{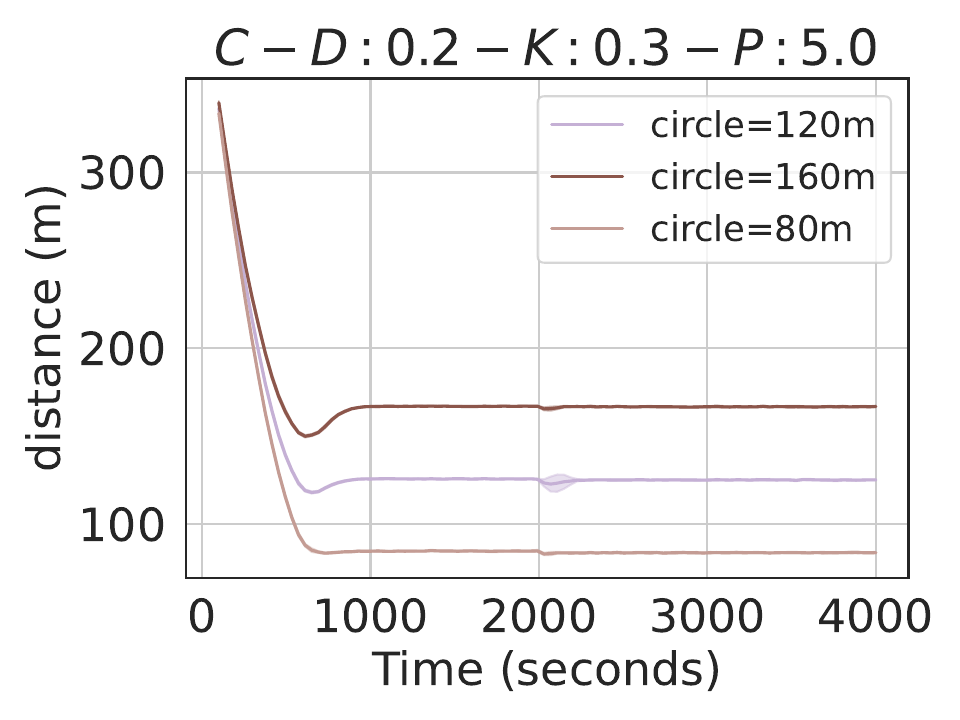}
    \caption{}\label{fig:detail-circle-0.2-5.0}
  \end{subfigure}\hfill

  \begin{subfigure}[b]{0.24\textwidth}
    \includegraphics[width=\textwidth]{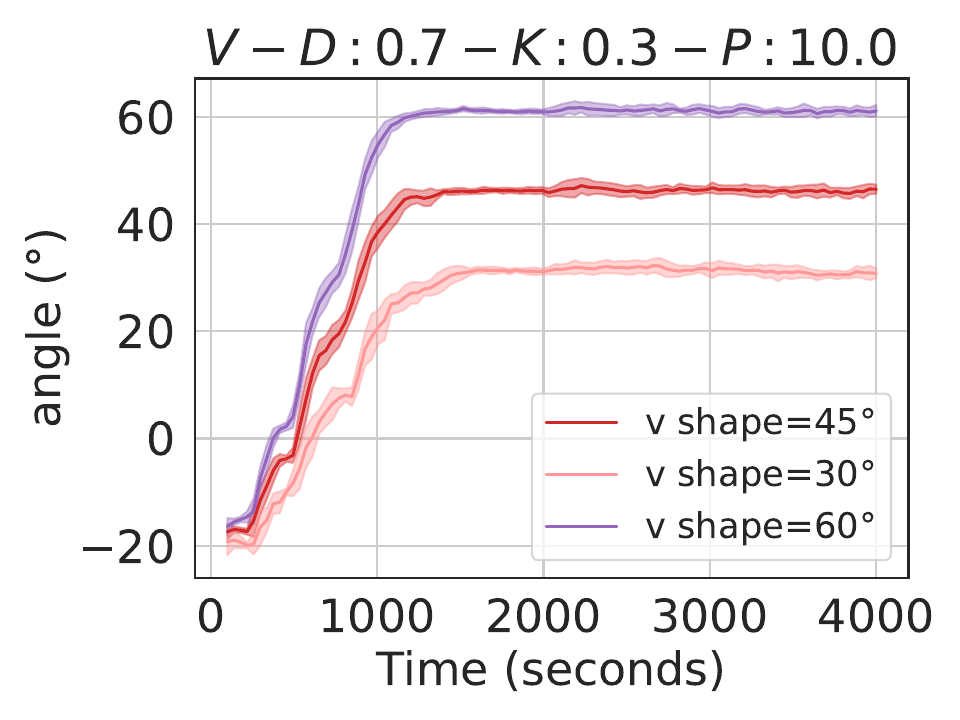}
    \caption{}\label{fig:detail-vshape-0.7-10.0}
  \end{subfigure}
  \begin{subfigure}[b]{0.24\textwidth}
    \includegraphics[width=\textwidth]{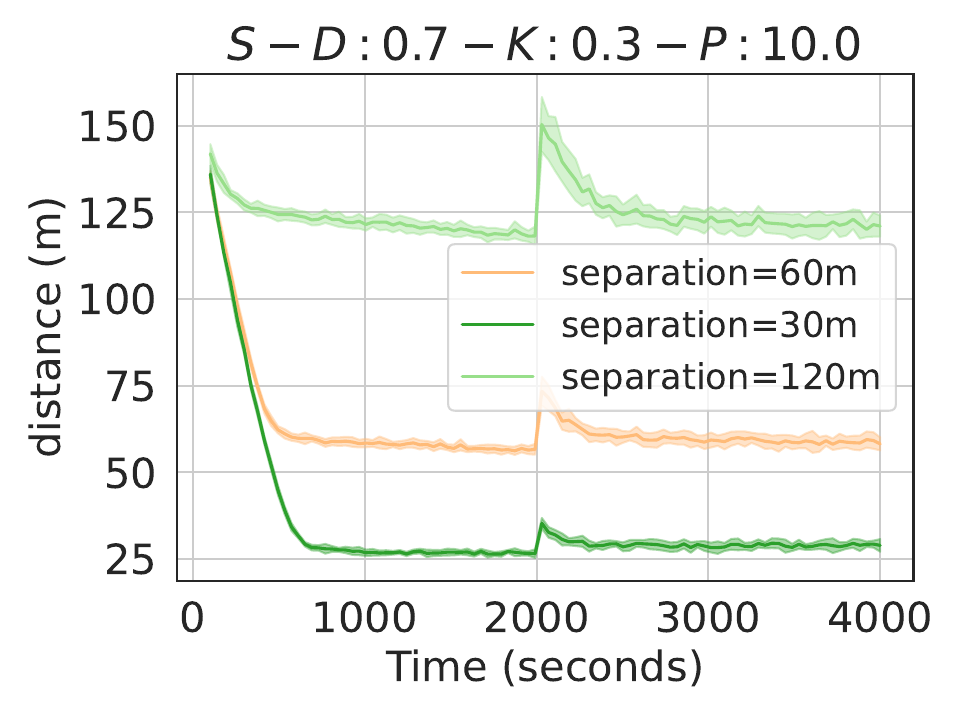}
    \caption{}\label{fig:detail-separation-0.7-10.0}
  \end{subfigure}
  \begin{subfigure}[b]{0.24\textwidth}
    \includegraphics[width=\textwidth]{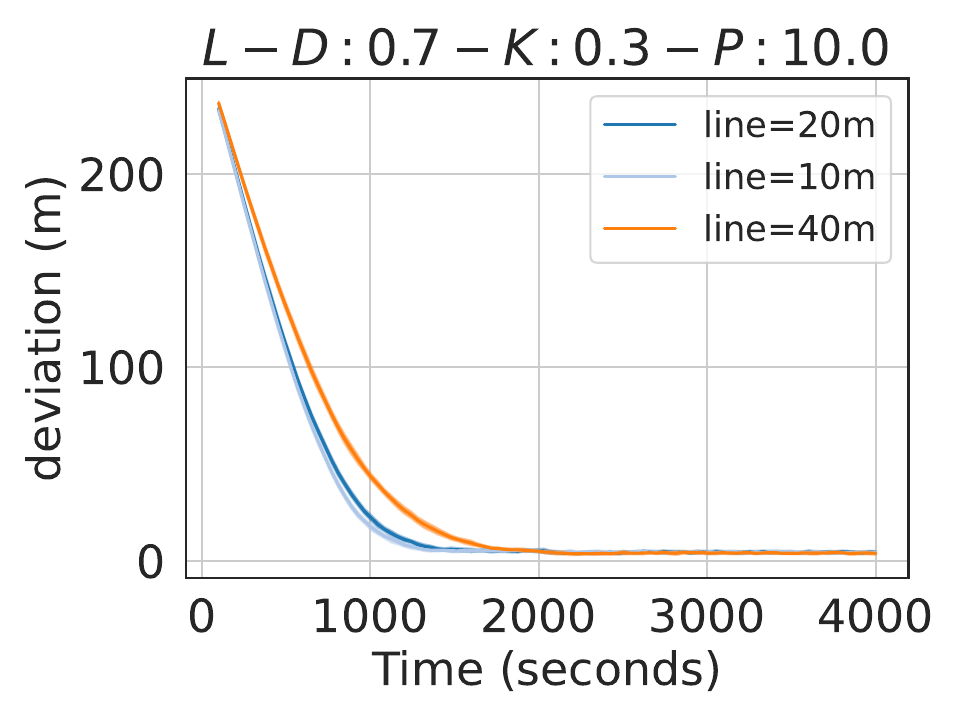}
    \caption{}\label{fig:detail-line-0.7-10.0}
  \end{subfigure}
  \begin{subfigure}[b]{0.24\textwidth}
    \includegraphics[width=\textwidth]{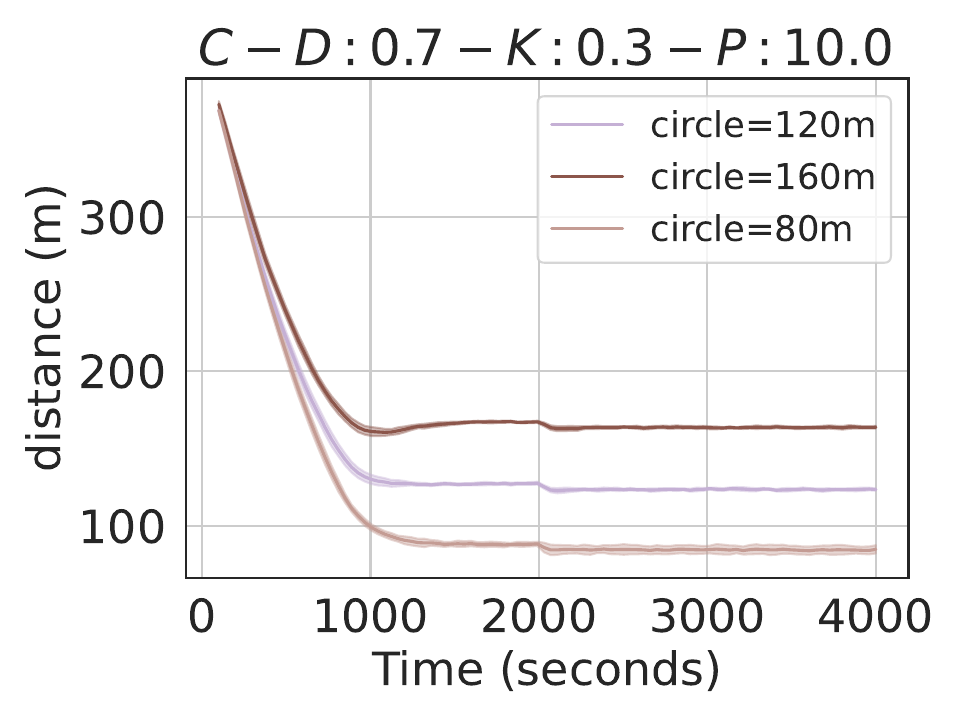}
    \caption{}\label{fig:detail-circle-0.7-10.0}
  \end{subfigure}\hfill

  \caption{\revB{The effect of perception noise in the position of the drones and message loss.
  With a reasonable amount of noise (P=5.0 and D=0.2) the drones are still able to form the desired shape, whereas with a higher amount of noise (P=10.0 and D=0.7) the structure is deformed.}}
  \label{fig:pattern-eval-depth-both}
\end{figure}

\begin{figure}
  \centering
  \begin{subfigure}[b]{0.3\textwidth}
    \includegraphics[width=\textwidth]{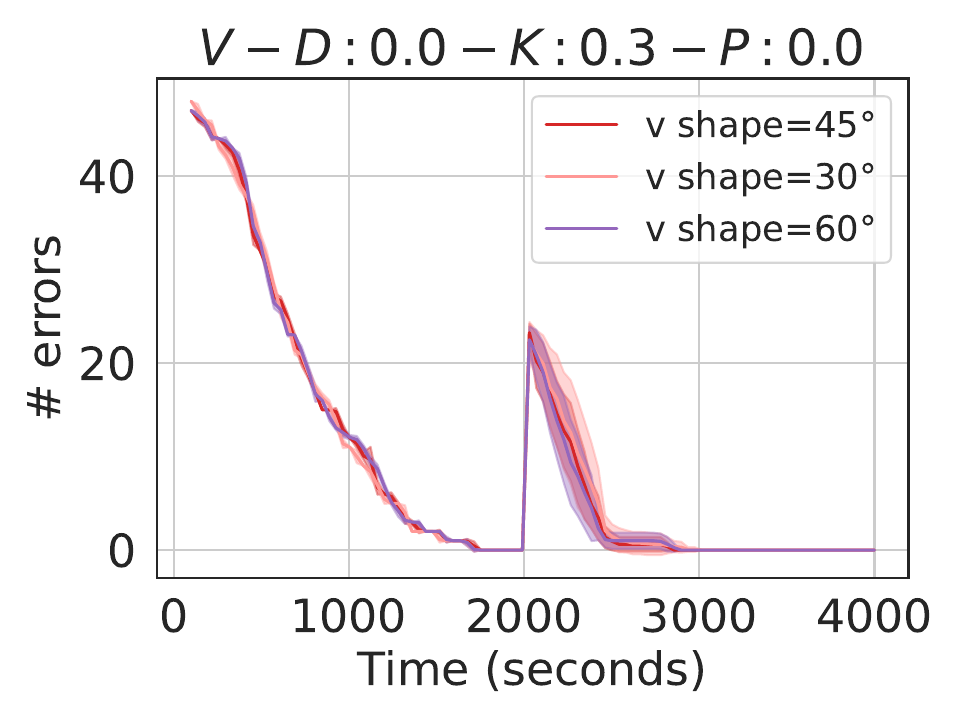}
    \caption{}\label{fig:detail-vshape-errors-0.0}
  \end{subfigure}
  \begin{subfigure}[b]{0.3\textwidth}
    \includegraphics[width=\textwidth]{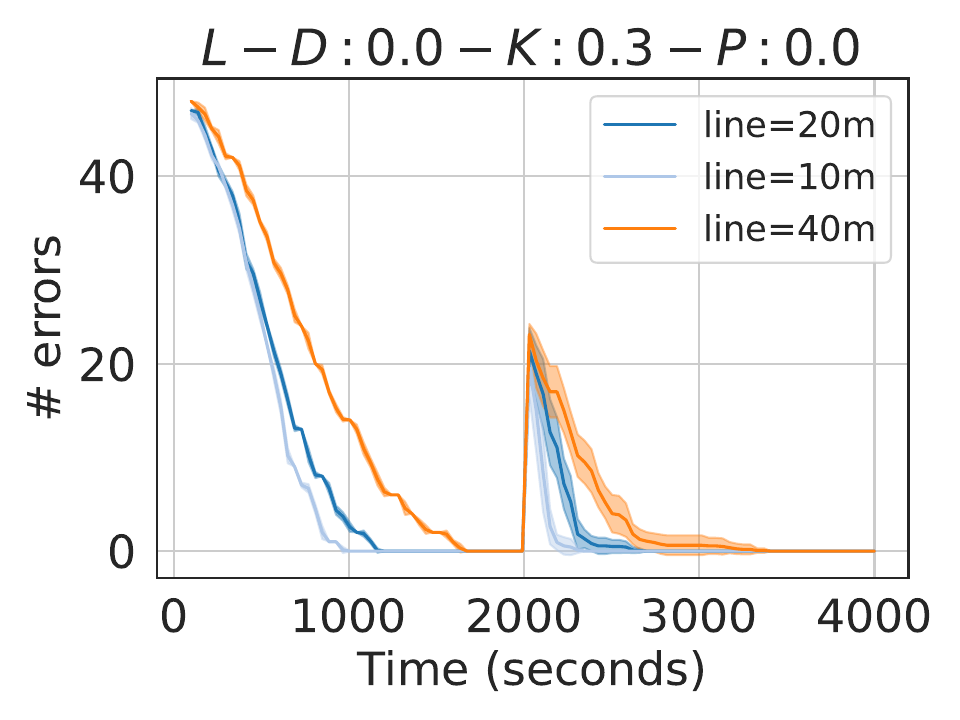}
    \caption{}\label{fig:detail-line-errors-0.0}
  \end{subfigure}
  \begin{subfigure}[b]{0.3\textwidth}
    \includegraphics[width=\textwidth]{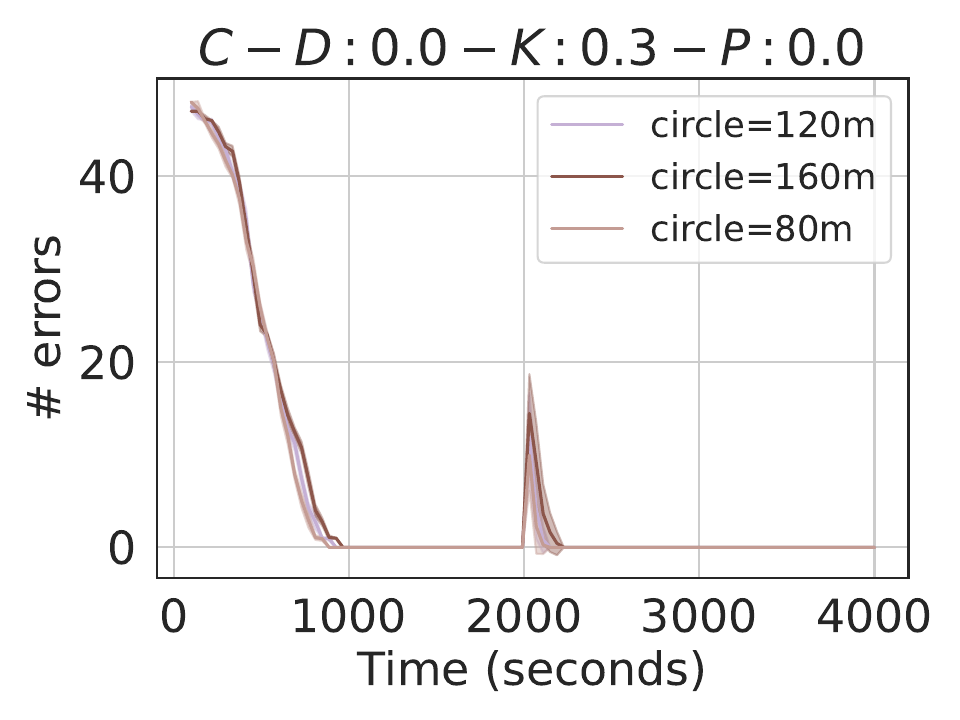}
    \caption{}\label{fig:detail-circle-errors-0.0}
  \end{subfigure}\hfill

  \begin{subfigure}[b]{0.3\textwidth}
    \includegraphics[width=\textwidth]{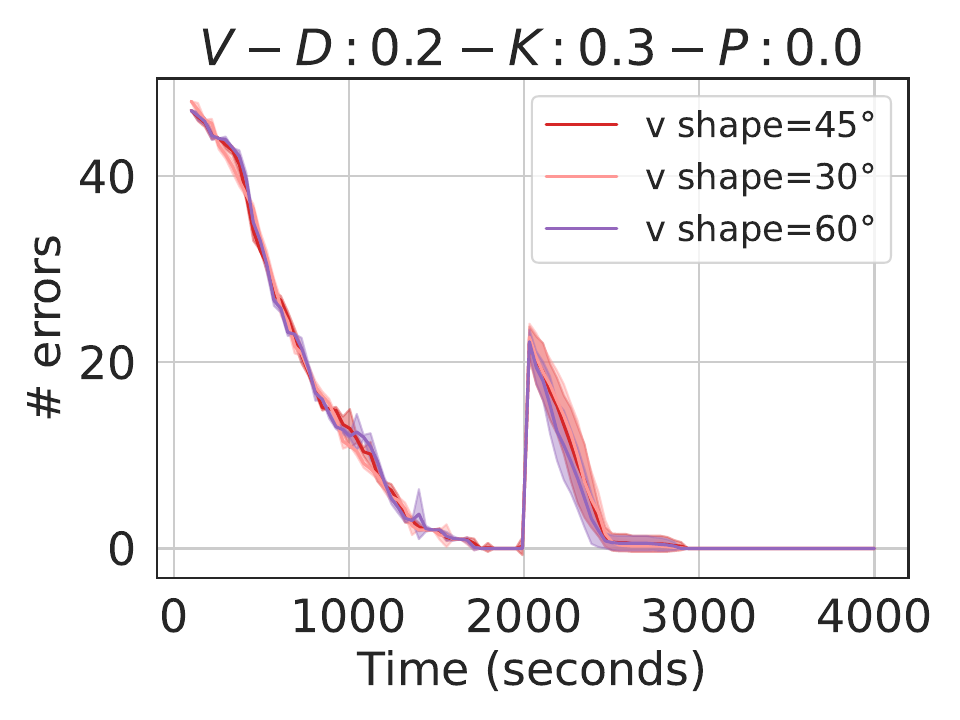}
    \caption{}\label{fig:detail-vshape-errors-0.2}
  \end{subfigure}
  \begin{subfigure}[b]{0.3\textwidth}
    \includegraphics[width=\textwidth]{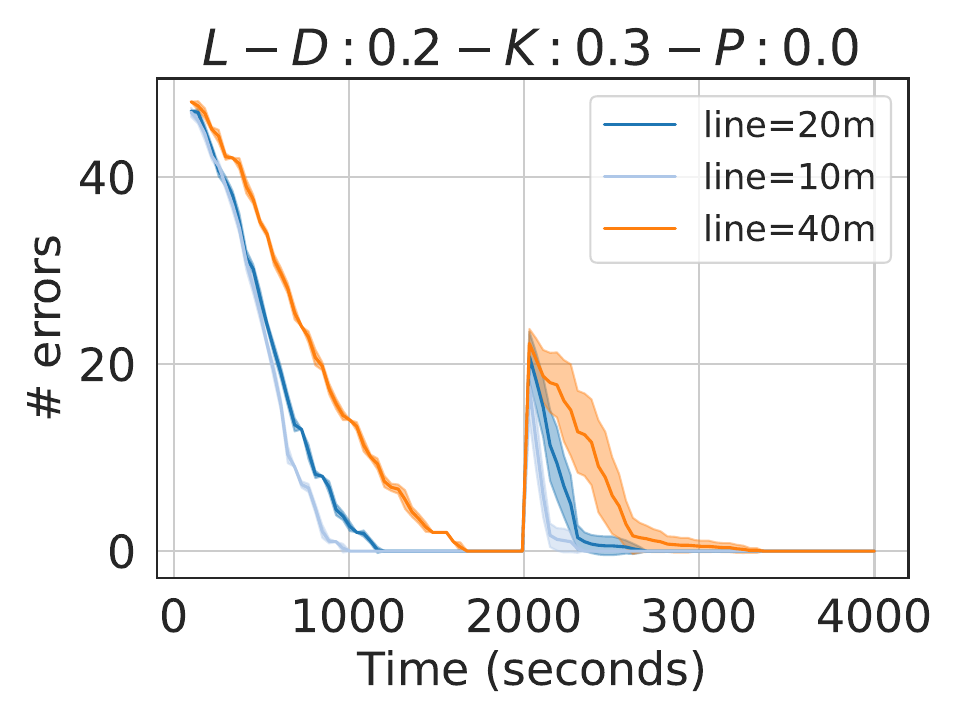}
    \caption{}\label{fig:detail-line-errors-0.2}
  \end{subfigure}
  \begin{subfigure}[b]{0.3\textwidth}
    \includegraphics[width=\textwidth]{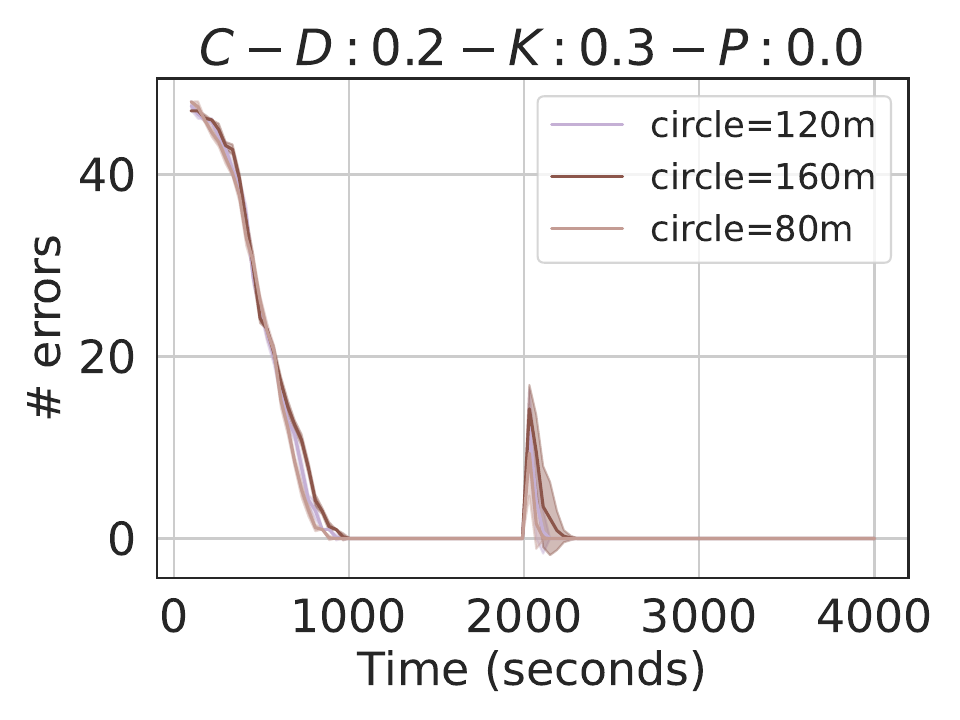}
    \caption{}\label{fig:detail-circle-errors-0.2}
  \end{subfigure}\hfill
  
  \begin{subfigure}[b]{0.3\textwidth}
    \includegraphics[width=\textwidth]{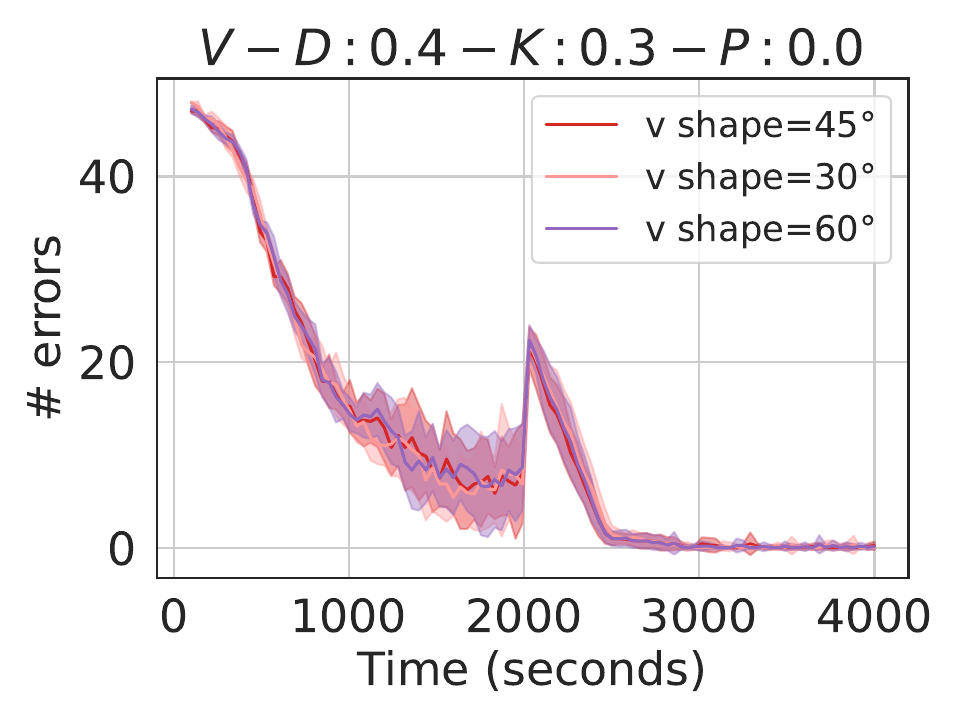}
    \caption{}\label{fig:detail-vshape-errors-0.4}
  \end{subfigure}
  \begin{subfigure}[b]{0.3\textwidth}
    \includegraphics[width=\textwidth]{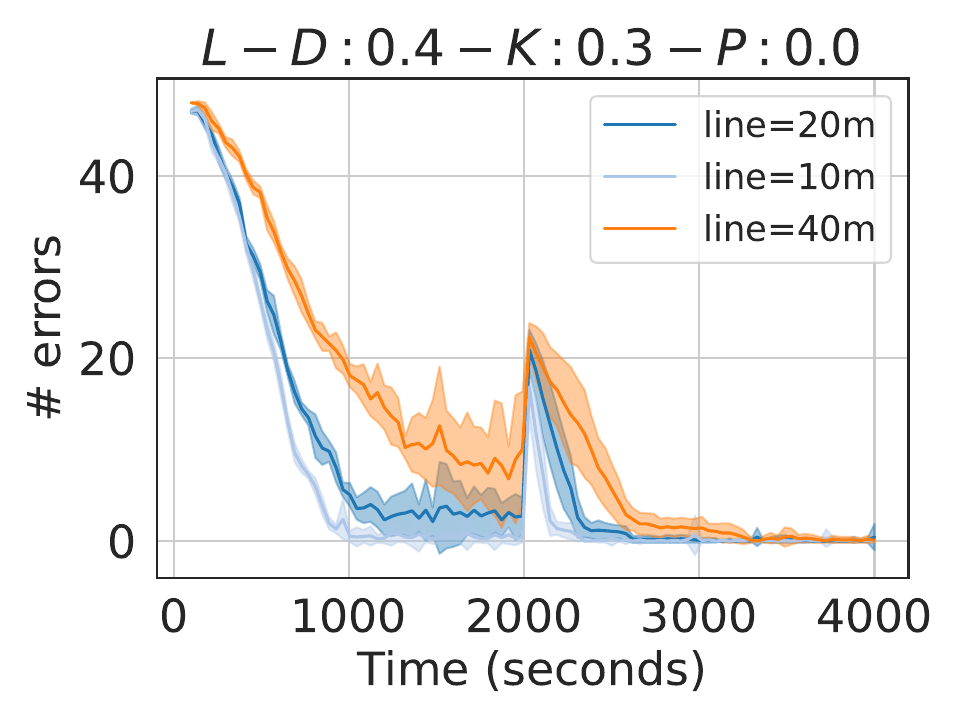}
    \caption{}\label{fig:detail-line-errors-0.4}
  \end{subfigure}
  \begin{subfigure}[b]{0.3\textwidth}
    \includegraphics[width=\textwidth]{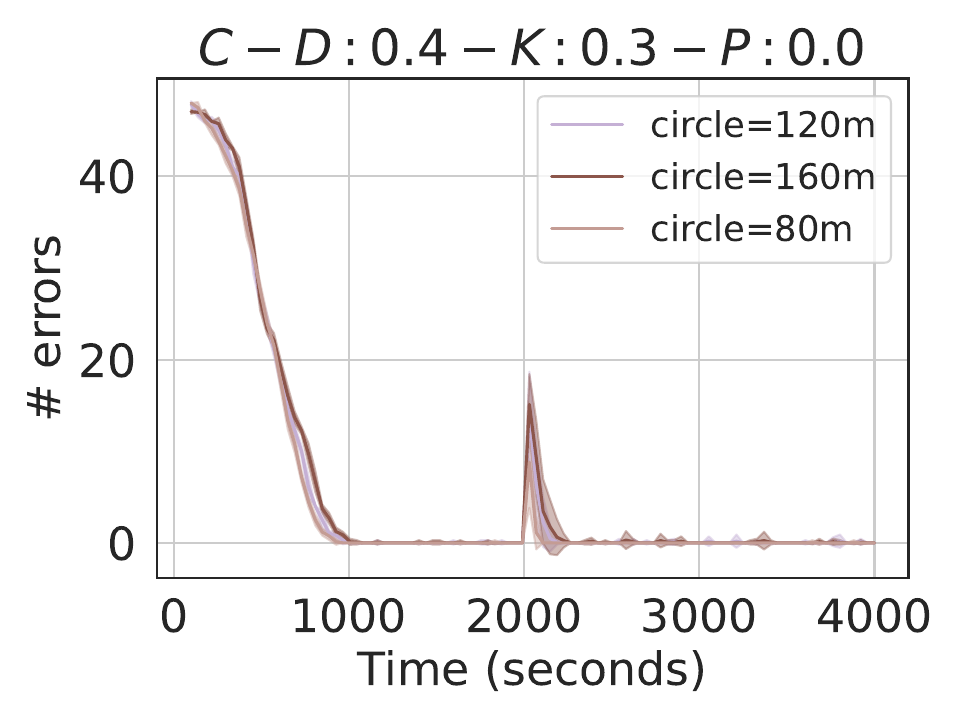}
    \caption{}\label{fig:detail-circle-errors-0.4}
  \end{subfigure}

  \begin{subfigure}[b]{0.3\textwidth}
    \includegraphics[width=\textwidth]{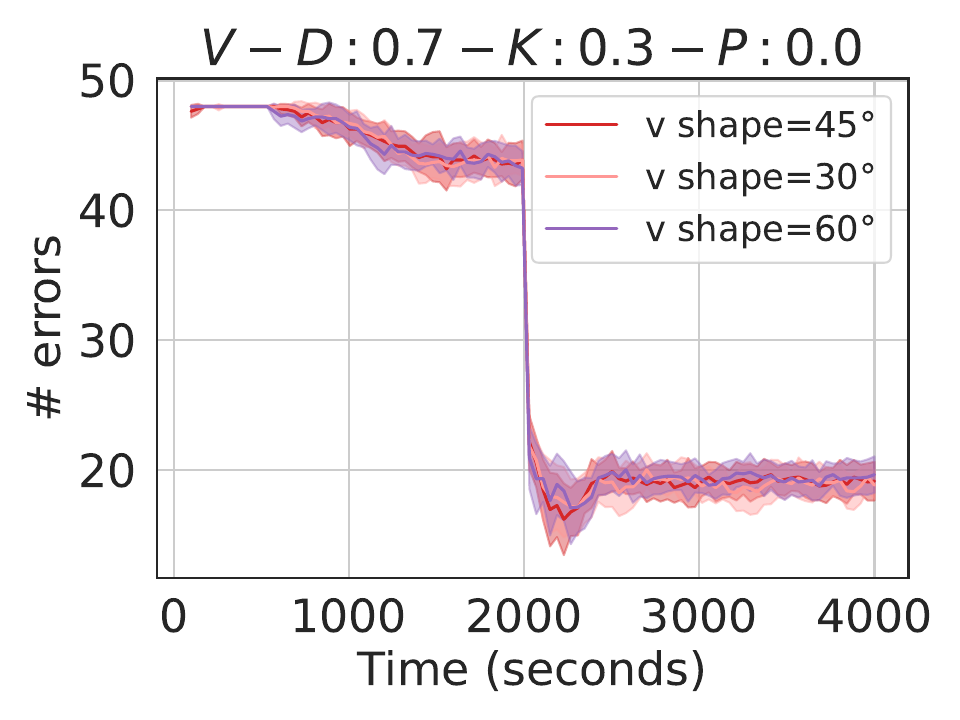}
    \caption{}\label{fig:detail-vshape-errors-0.7}
  \end{subfigure}
  \begin{subfigure}[b]{0.3\textwidth}
    \includegraphics[width=\textwidth]{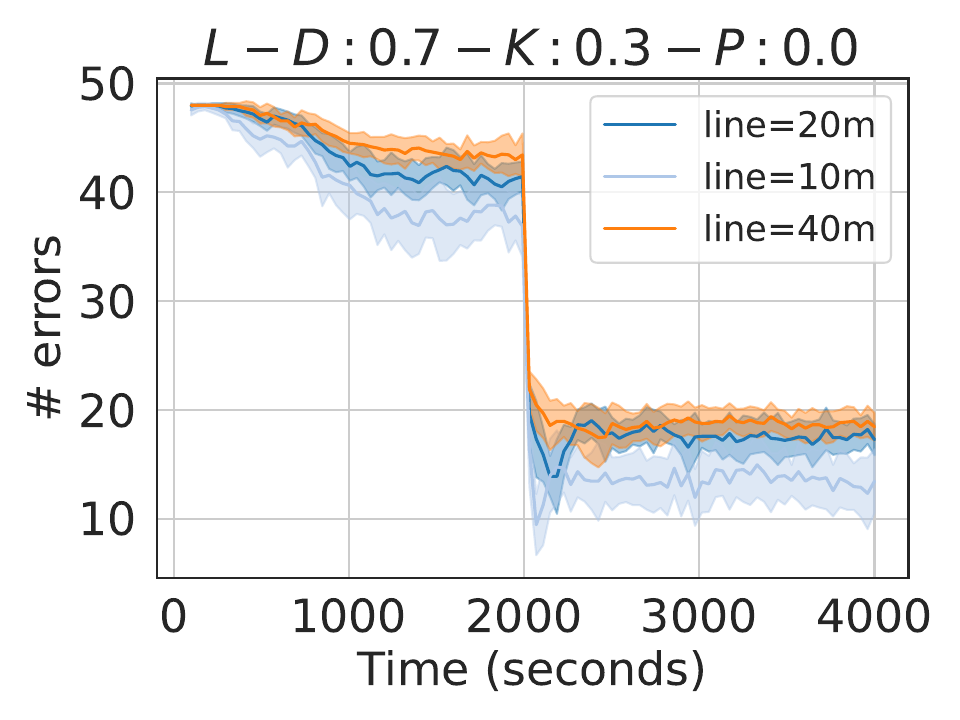}
    \caption{}\label{fig:detail-line-errors-0.7}
  \end{subfigure}
  \begin{subfigure}[b]{0.3\textwidth}
    \includegraphics[width=\textwidth]{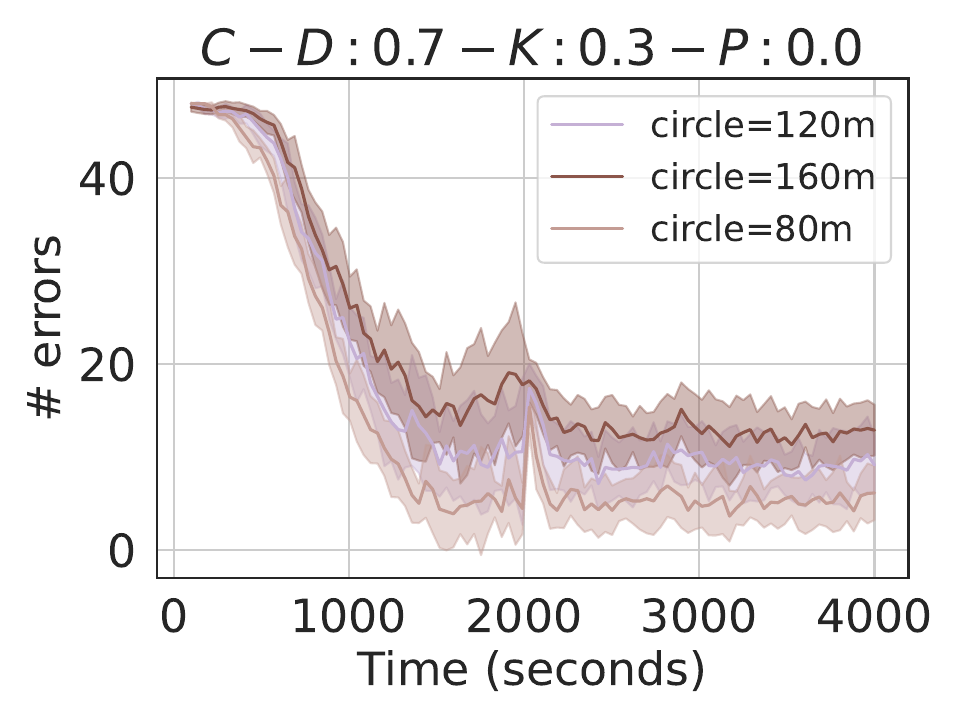}
    \caption{}\label{fig:detail-circle-errors-0.7}
  \end{subfigure}
  \caption{\revB{Analyses error propagation in pattern formation due to node failures and message loss.  Under ideal conditions, nodes converge to their desired positions.   
  The system demonstrates resilience to moderate message loss (e.g., D=0.2), exhibiting a longer convergence time but ultimately achieving the desired shape.  
  Higher message loss (D=0.4 and D=0.7) results in increased positional errors (V and line formations), 
  due to the increased number of hops required for message delivery. 
  Even at the highest loss rate (D=0.7), a pattern is formed, albeit with significant positional inaccuracies.}
  }\label{fig:pattern-eval-depth-errors-message-loss}
\end{figure}

\begin{figure}
  \centering
  \begin{subfigure}[b]{0.3\textwidth}
    \includegraphics[width=\textwidth]{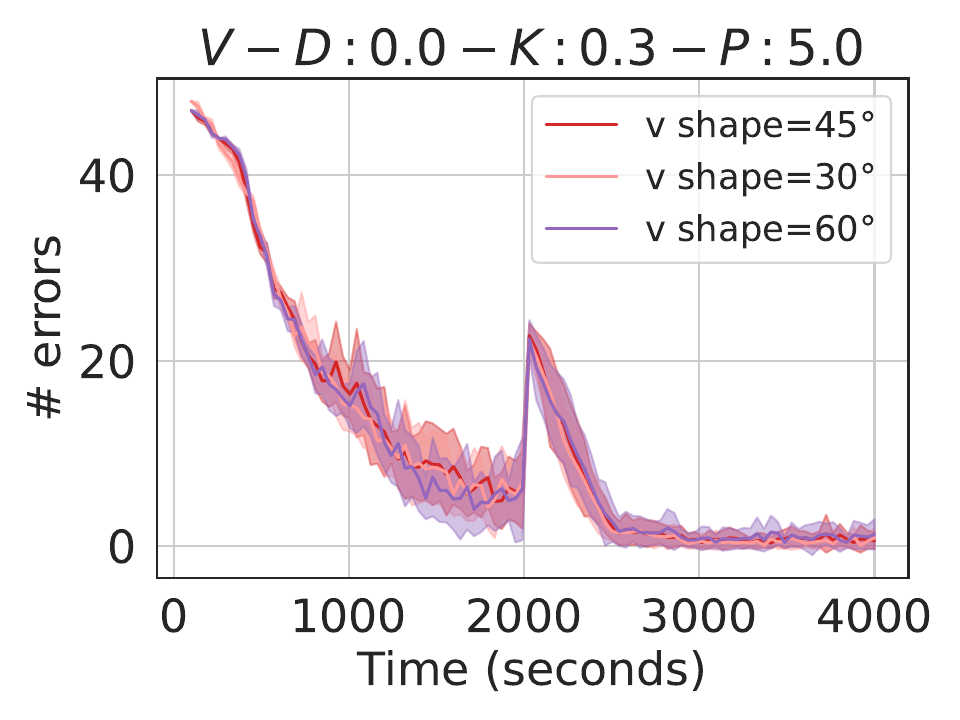}
    \caption{}\label{fig:detail-vshape-errors-5.0}
  \end{subfigure}
  \begin{subfigure}[b]{0.3\textwidth}
    \includegraphics[width=\textwidth]{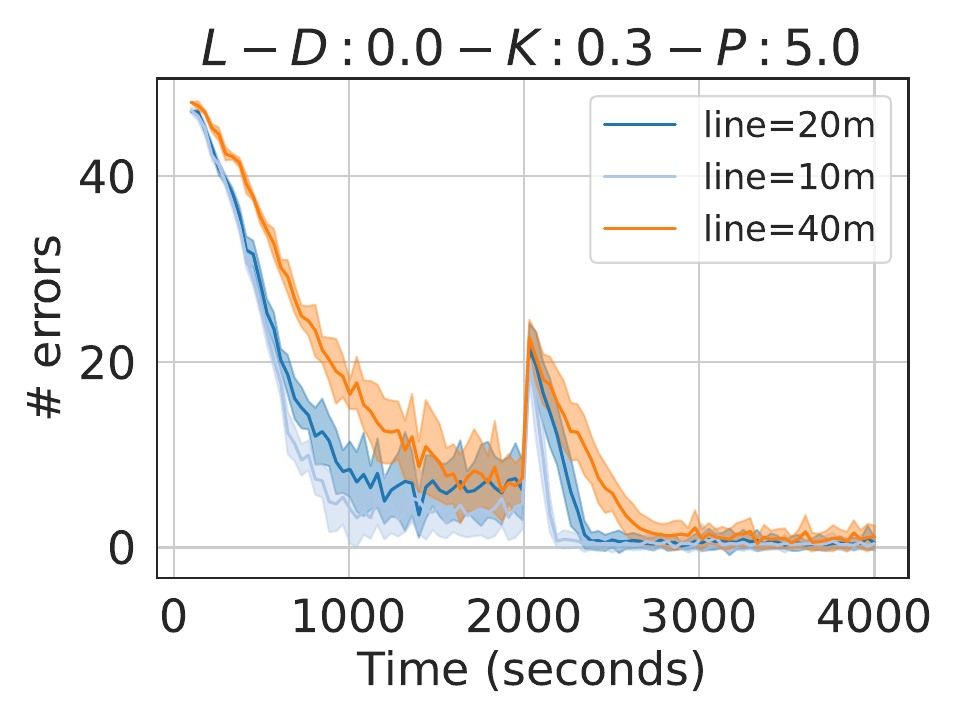}
    \caption{}\label{fig:detail-line-errors-5.0}
  \end{subfigure}
  \begin{subfigure}[b]{0.3\textwidth}
    \includegraphics[width=\textwidth]{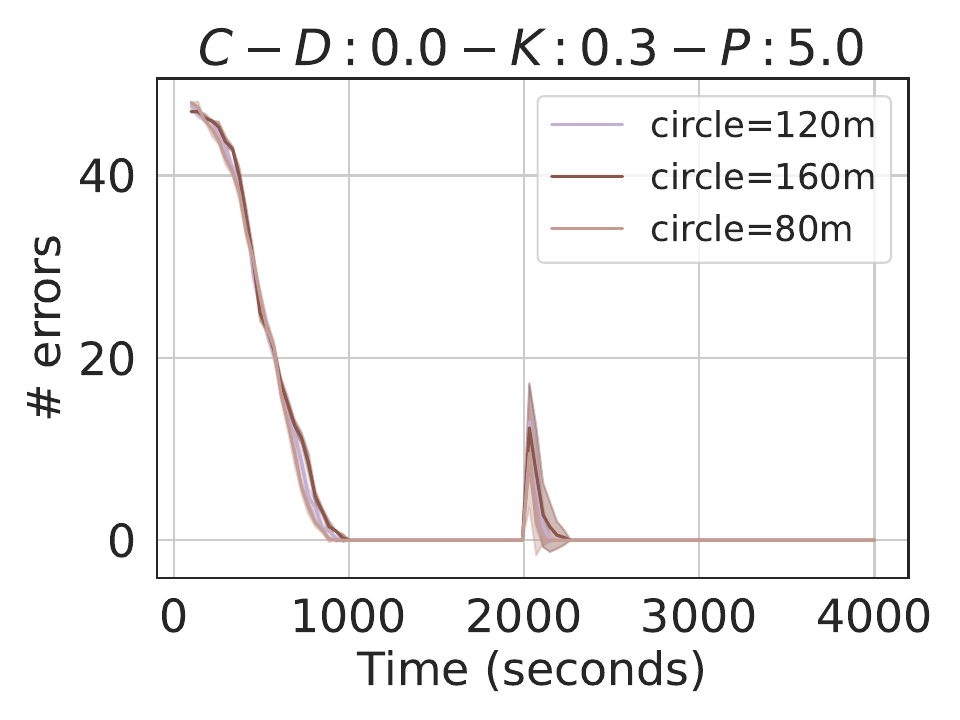}
    \caption{}\label{fig:detail-circle-errors-5.0}
  \end{subfigure}\hfill

  \begin{subfigure}[b]{0.3\textwidth}
    \includegraphics[width=\textwidth]{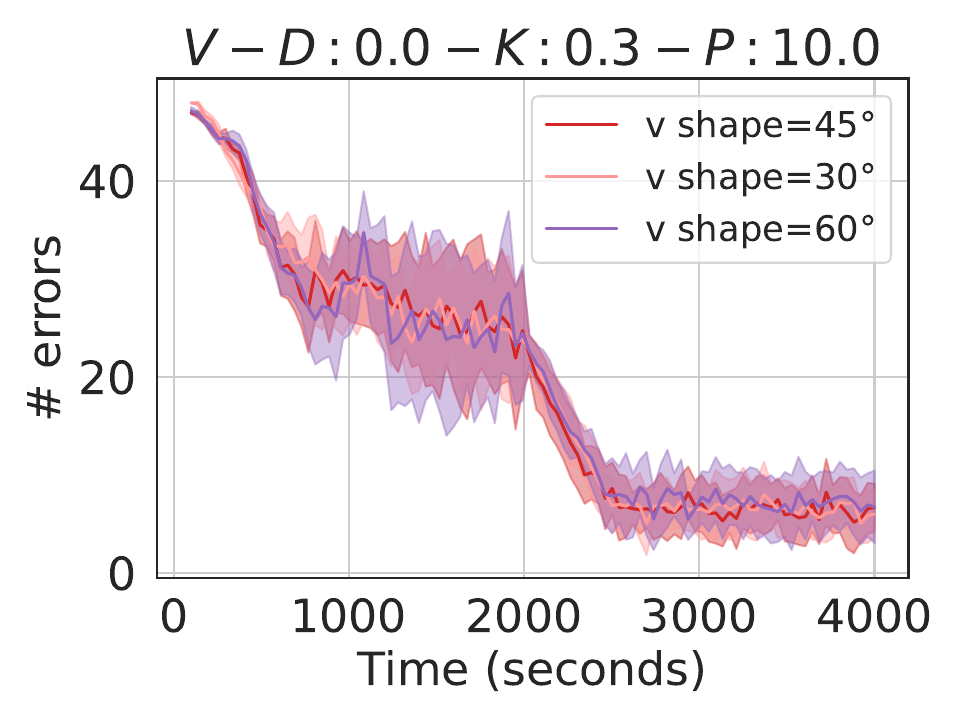}
    \caption{}\label{fig:detail-vshape-errors-10.0}
  \end{subfigure}
  \begin{subfigure}[b]{0.3\textwidth}
    \includegraphics[width=\textwidth]{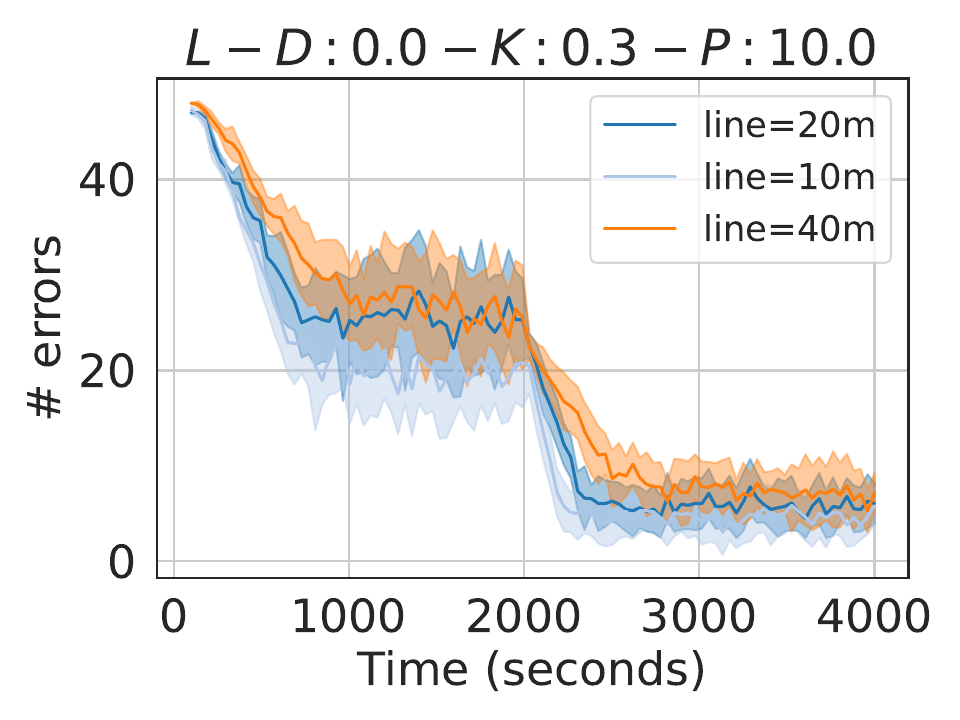}
    \caption{}\label{fig:detail-line-errors-10.0}
  \end{subfigure}
  \begin{subfigure}[b]{0.3\textwidth}
    \includegraphics[width=\textwidth]{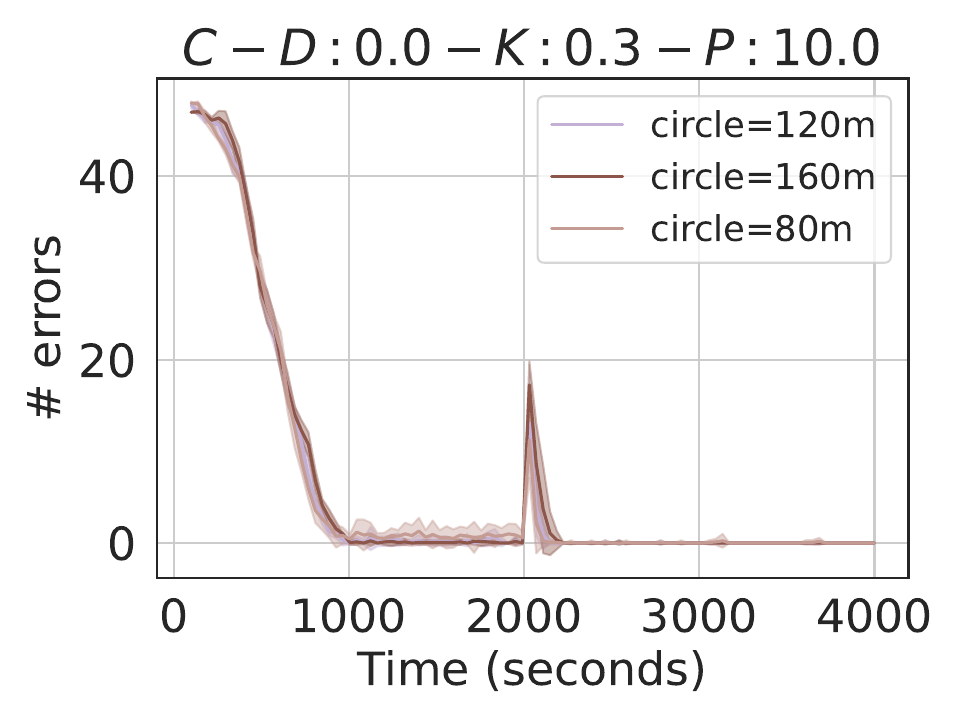}
    \caption{}\label{fig:detail-circle-errors-10.0}
  \end{subfigure}
  \caption{\revB{Effect of perception noise in the position of the drones.
    With a reasonable amount of noise (P=5.0) the drones are still able to form the desired shape, whereas with a higher amount of noise (P=10.0) the structure is strongly deformed
    particularly in V shape and line formation.}}
  \label{fig:pattern-eval-depth-errors-perception}
\end{figure}

\begin{figure}
  \centering
  \begin{subfigure}[b]{0.3\textwidth}
    \includegraphics[width=\textwidth]{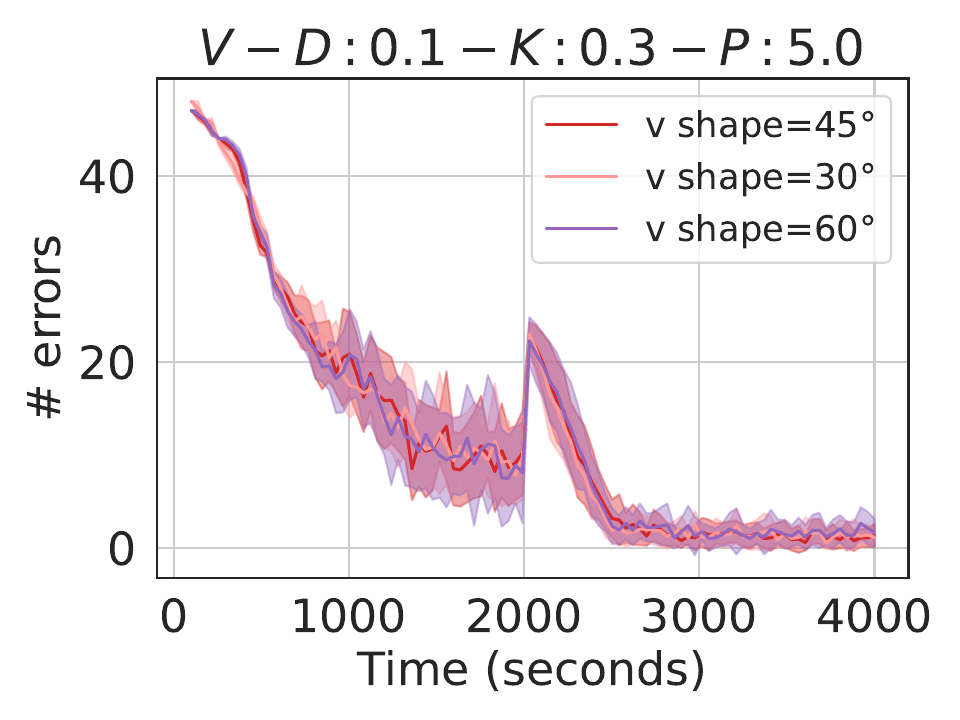}
    \caption{}\label{fig:detail-vshape-errors-0.1-5.0}
  \end{subfigure}
  \begin{subfigure}[b]{0.3\textwidth}
    \includegraphics[width=\textwidth]{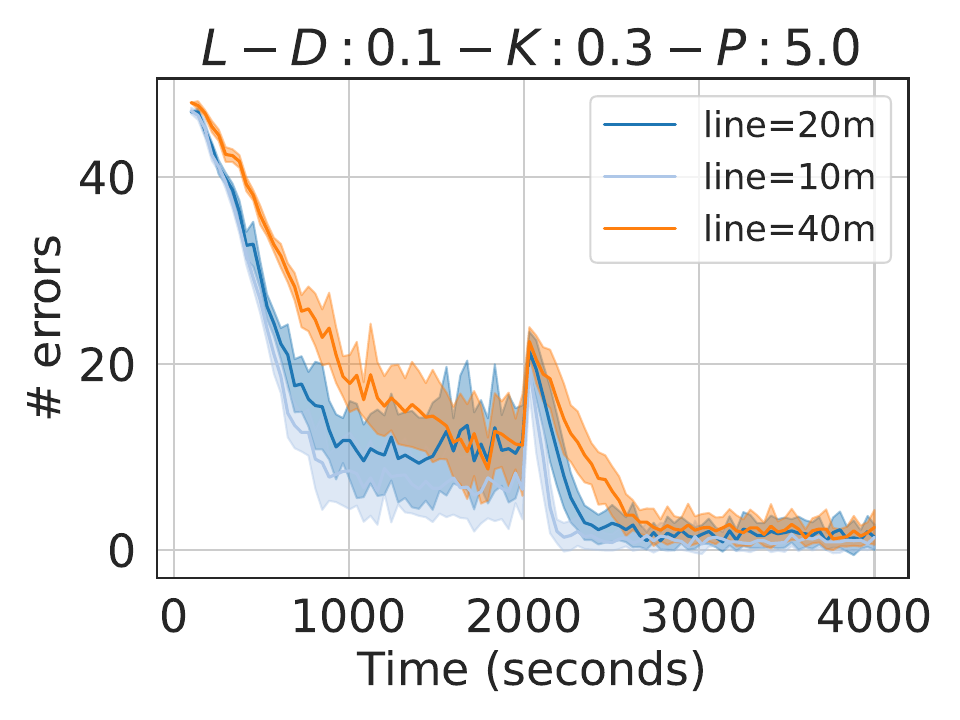}
    \caption{}\label{fig:detail-line-errors-0.1-5.0}
  \end{subfigure}
  \begin{subfigure}[b]{0.3\textwidth}
    \includegraphics[width=\textwidth]{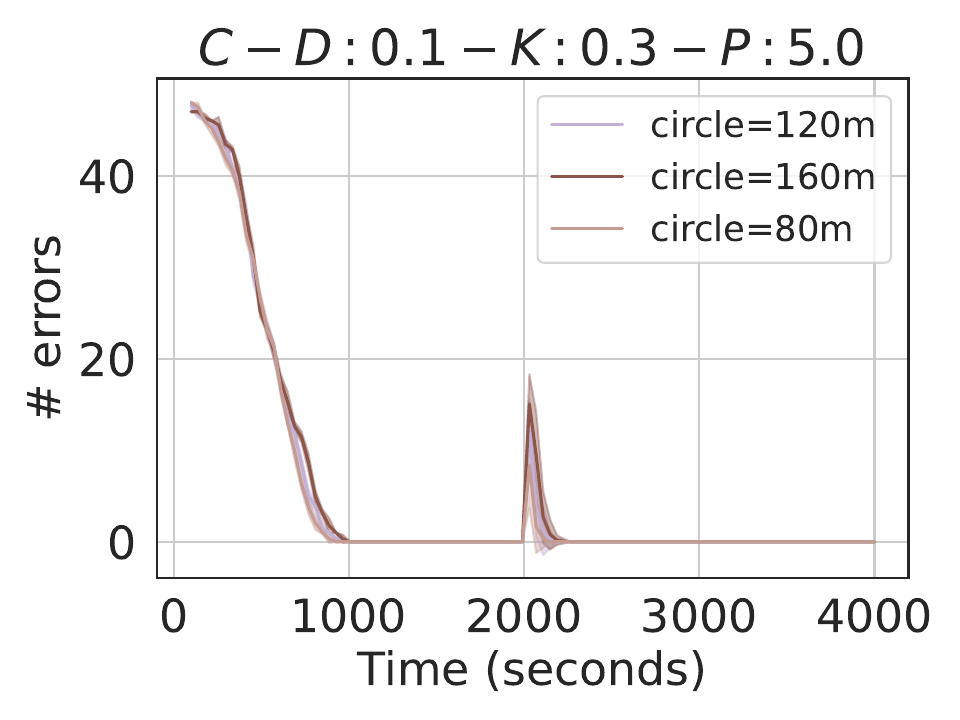}
    \caption{}\label{fig:detail-circle-errors-0.1-5.0}
  \end{subfigure}\hfill

  \begin{subfigure}[b]{0.3\textwidth}
    \includegraphics[width=\textwidth]{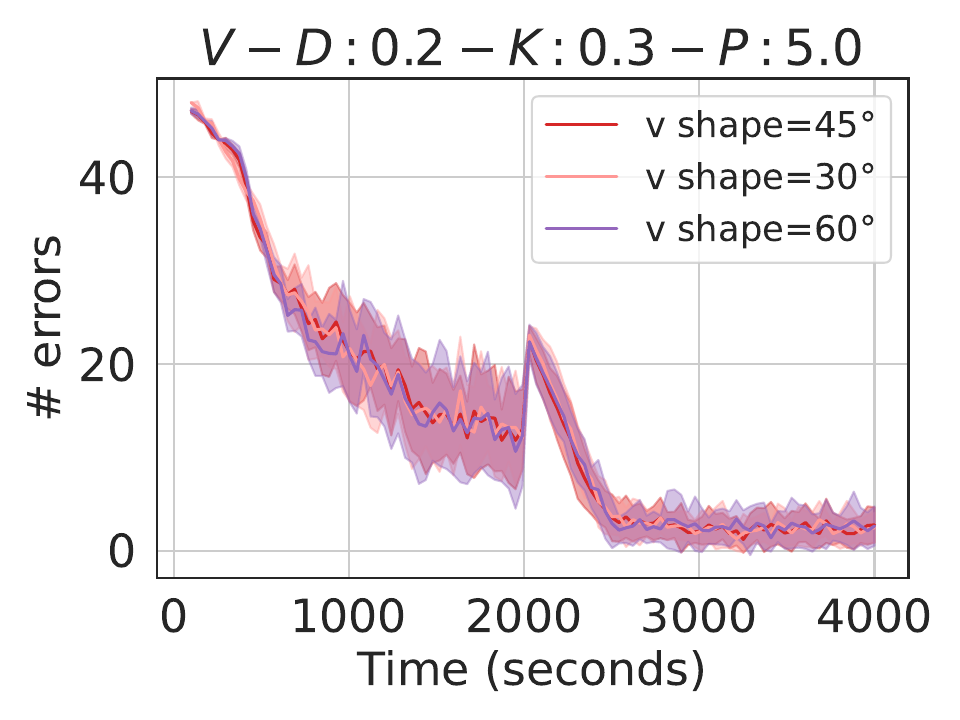}
    \caption{}\label{fig:detail-vshape-errors-0.2-5.0}
  \end{subfigure}
  \begin{subfigure}[b]{0.3\textwidth}
    \includegraphics[width=\textwidth]{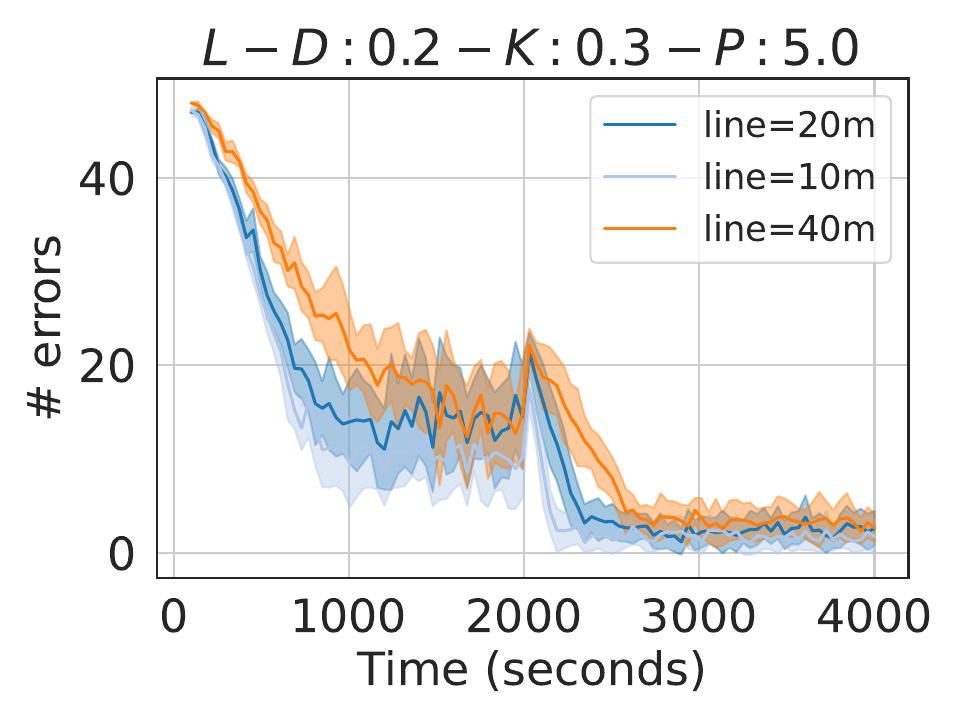}
    \caption{}\label{fig:detail-line-errors-0.2-5.0}
  \end{subfigure}
  \begin{subfigure}[b]{0.3\textwidth}
    \includegraphics[width=\textwidth]{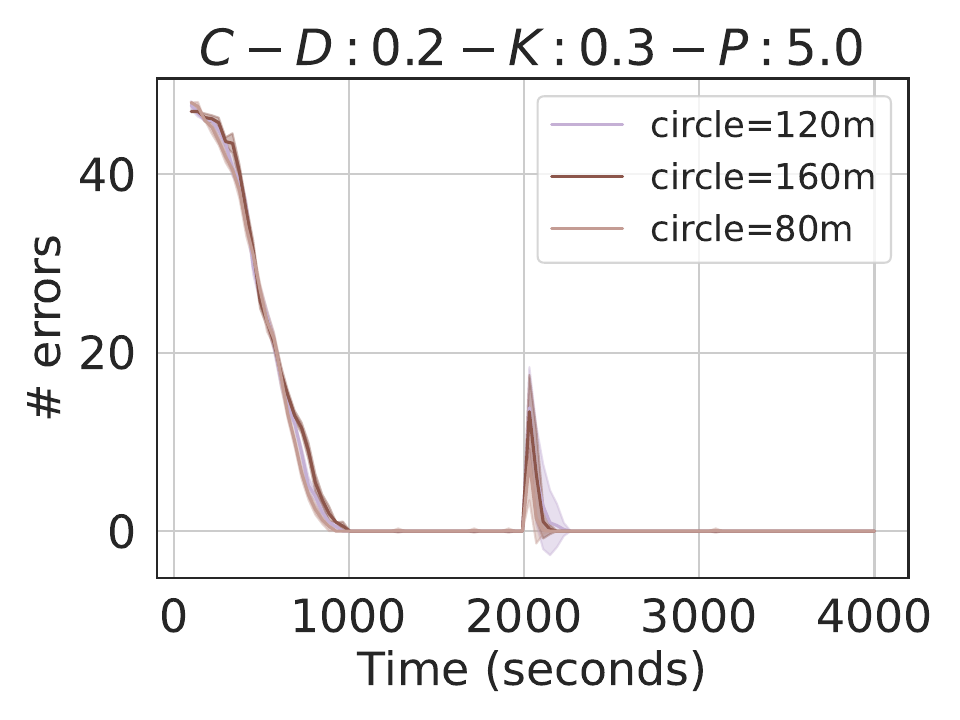}
    \caption{}\label{fig:detail-circle-errors-0.2-5.0}
  \end{subfigure}\hfill

  \begin{subfigure}[b]{0.3\textwidth}
    \includegraphics[width=\textwidth]{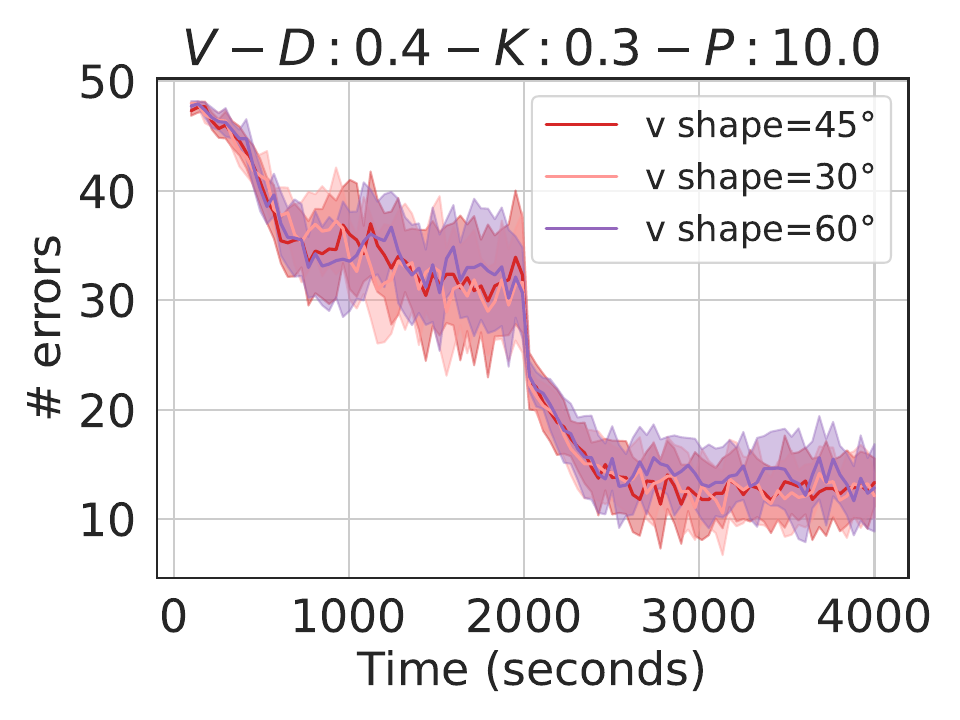}
    \caption{}\label{fig:detail-vshape-errors-0.4-10.0}
  \end{subfigure}
  \begin{subfigure}[b]{0.3\textwidth}
    \includegraphics[width=\textwidth]{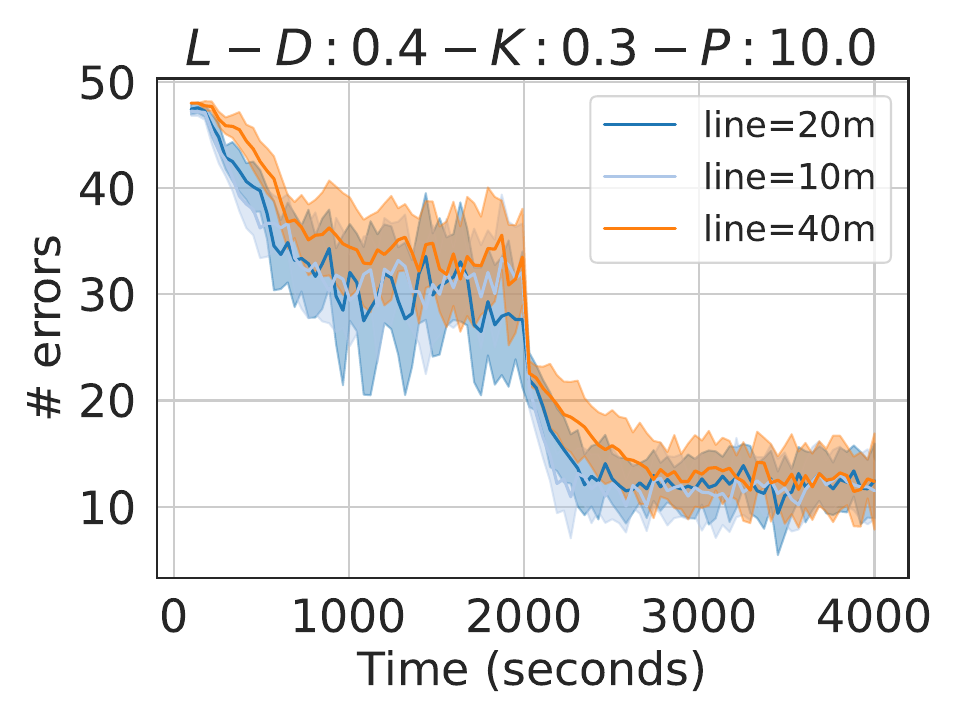}
    \caption{}\label{fig:detail-line-errors-0.4-10.0}
  \end{subfigure}
  \begin{subfigure}[b]{0.3\textwidth}
    \includegraphics[width=\textwidth]{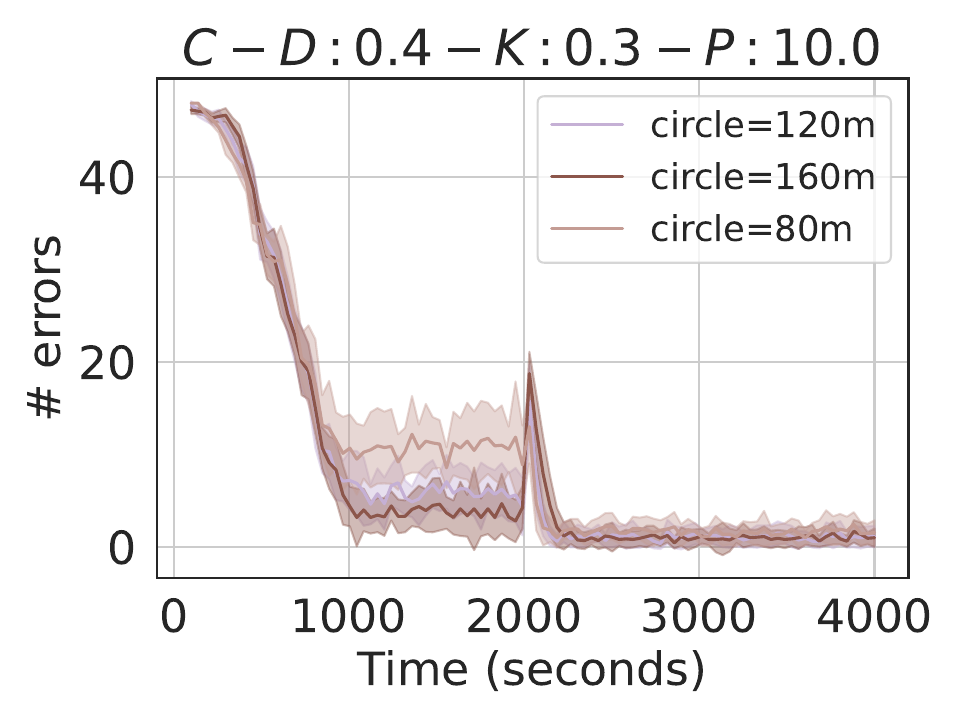}
    \caption{}\label{fig:detail-circle-errors-0.4-10.0}
  \end{subfigure}\hfill

  \caption{\revB{Effect of perception noise in the position of the drones and message loss.
    With a reasonable amount of noise (P=5.0 and D=0.2) the drones are still able to form the desired shape, whereas with a higher amount of noise (P=10.0 and D=0.4) the structure is strongly deformed. In this case, also the circle formation is strongly affected ($\sim$ 10 errors out of 50 drones).}}  
  
  \label{fig:pattern-eval-depth-errors-both}
\end{figure}

\begin{figure}
  \begin{subfigure}[b]{0.3\textwidth}
    \includegraphics[width=\textwidth]{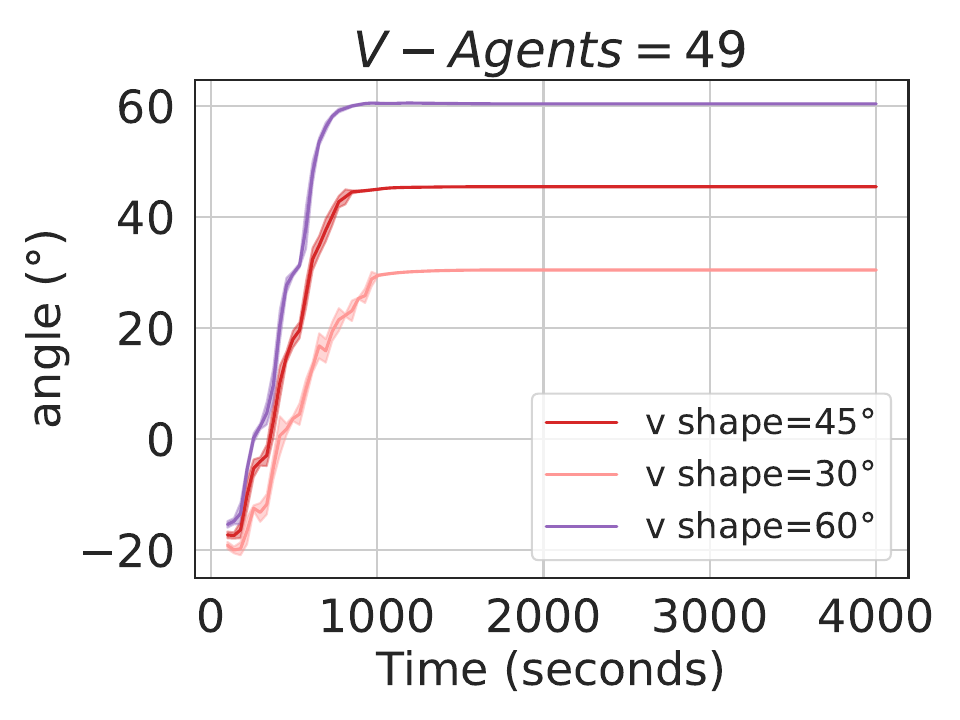}
    \caption{}\label{fig:detail-vshape-errors-0.4-10.0}
  \end{subfigure}
  \begin{subfigure}[b]{0.3\textwidth}
    \includegraphics[width=\textwidth]{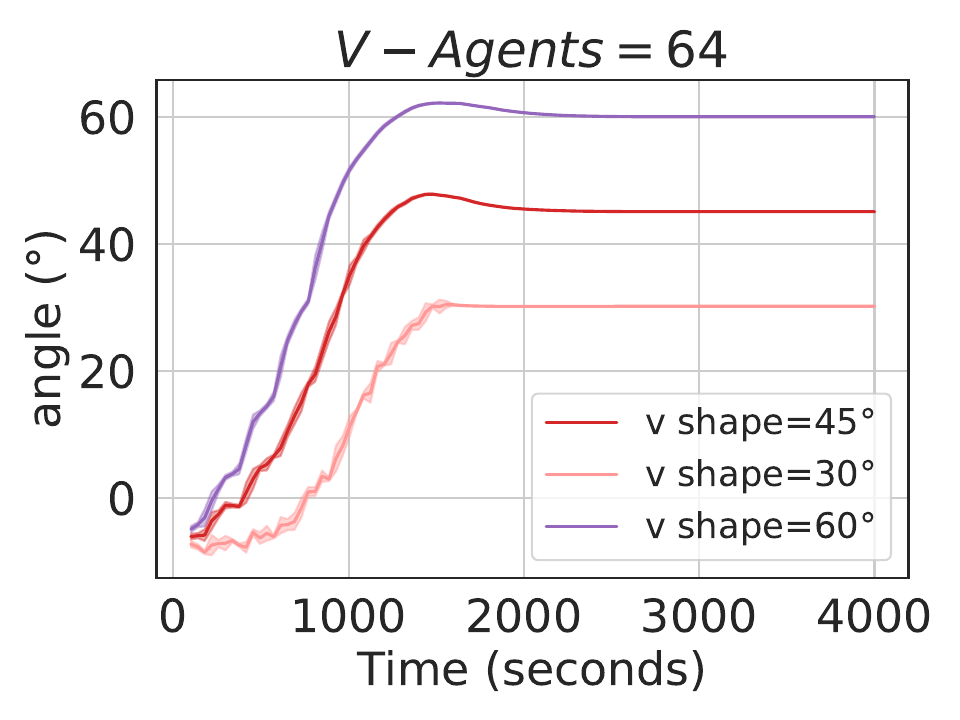}
    \caption{}\label{fig:detail-line-errors-0.4-10.0}
  \end{subfigure}
  \begin{subfigure}[b]{0.3\textwidth}
    \includegraphics[width=\textwidth]{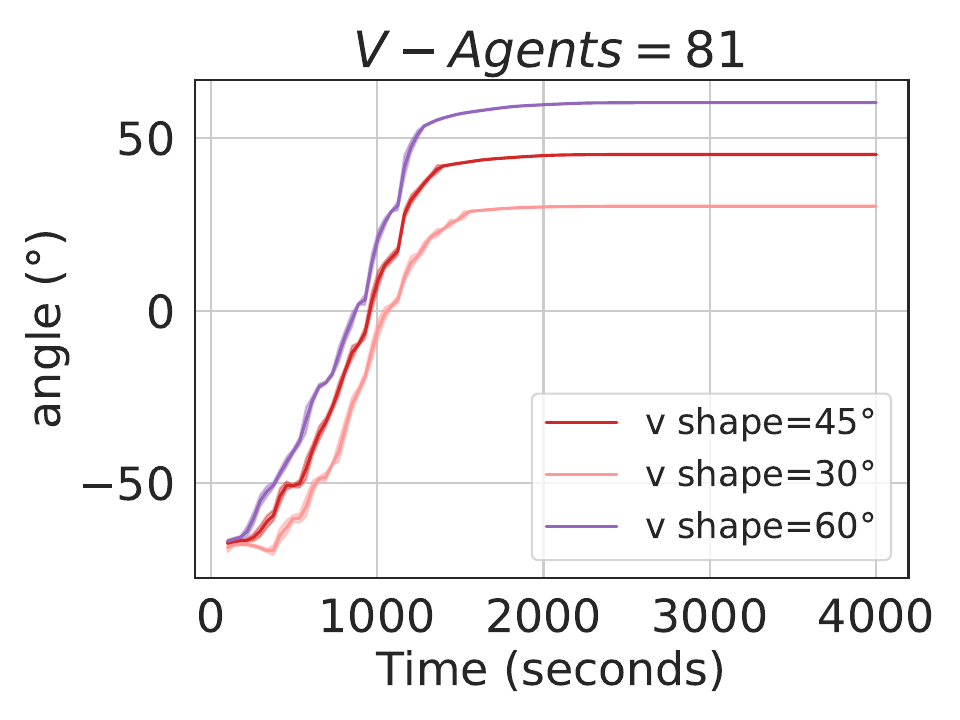}
    \caption{}\label{fig:detail-circle-errors-0.4-10.0}
  \end{subfigure}
  \begin{subfigure}[b]{0.3\textwidth}
    \includegraphics[width=\textwidth]{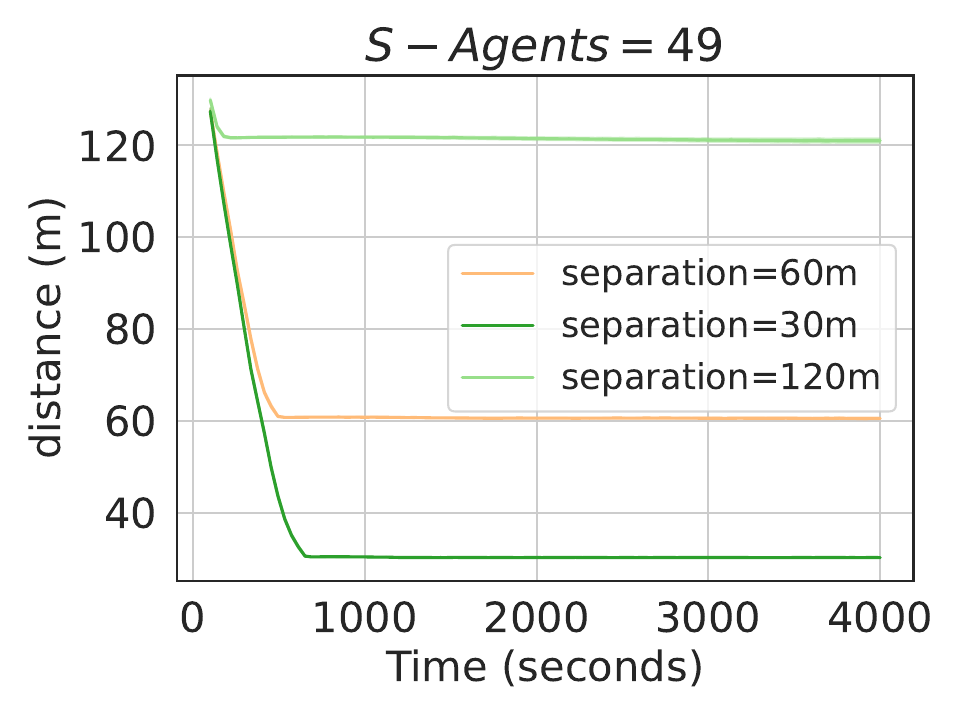}
    \caption{}\label{fig:detail-vshape-errors-0.4-10.0}
  \end{subfigure}
  \begin{subfigure}[b]{0.3\textwidth}
    \includegraphics[width=\textwidth]{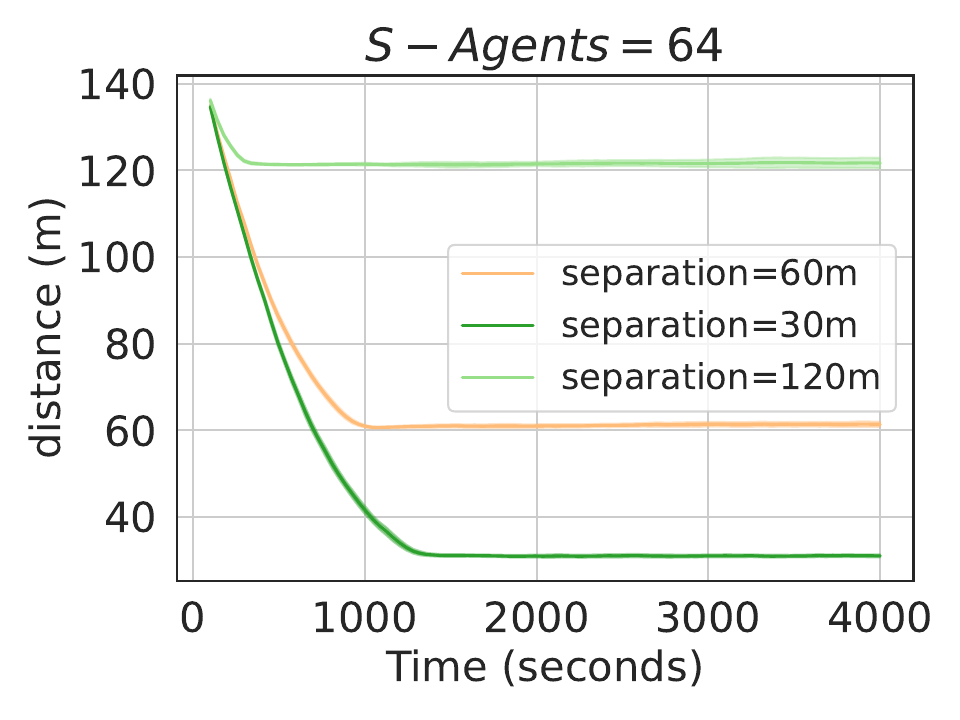}
    \caption{}\label{fig:detail-line-errors-0.4-10.0}
  \end{subfigure}
  \begin{subfigure}[b]{0.3\textwidth}
    \includegraphics[width=\textwidth]{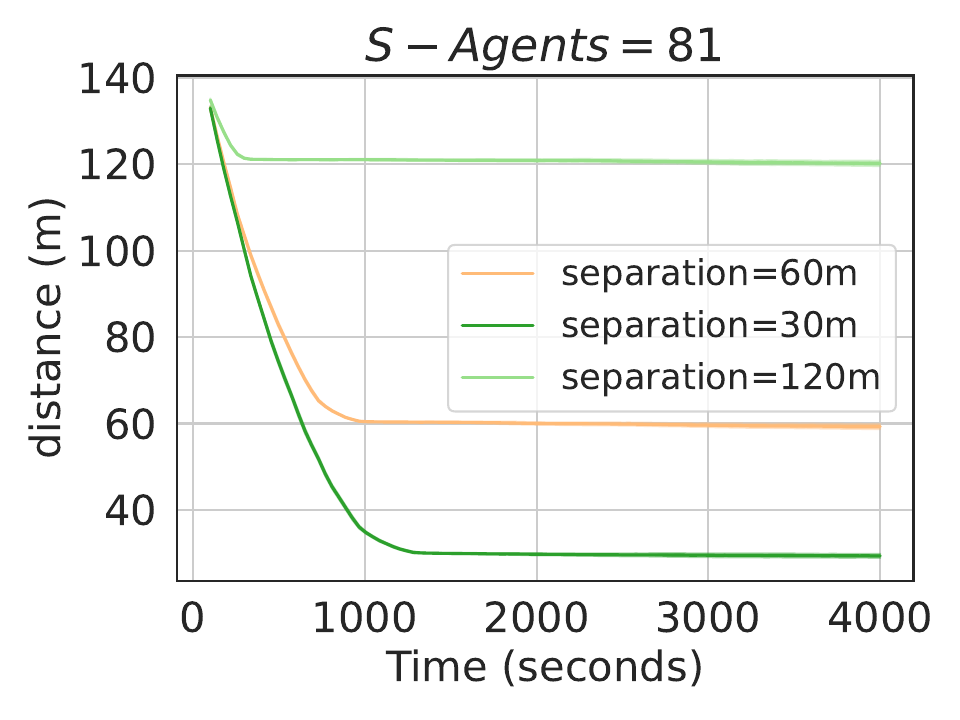}
    \caption{}\label{fig:detail-circle-errors-0.4-10.0}
  \end{subfigure}
  \begin{subfigure}[b]{0.3\textwidth}
    \includegraphics[width=\textwidth]{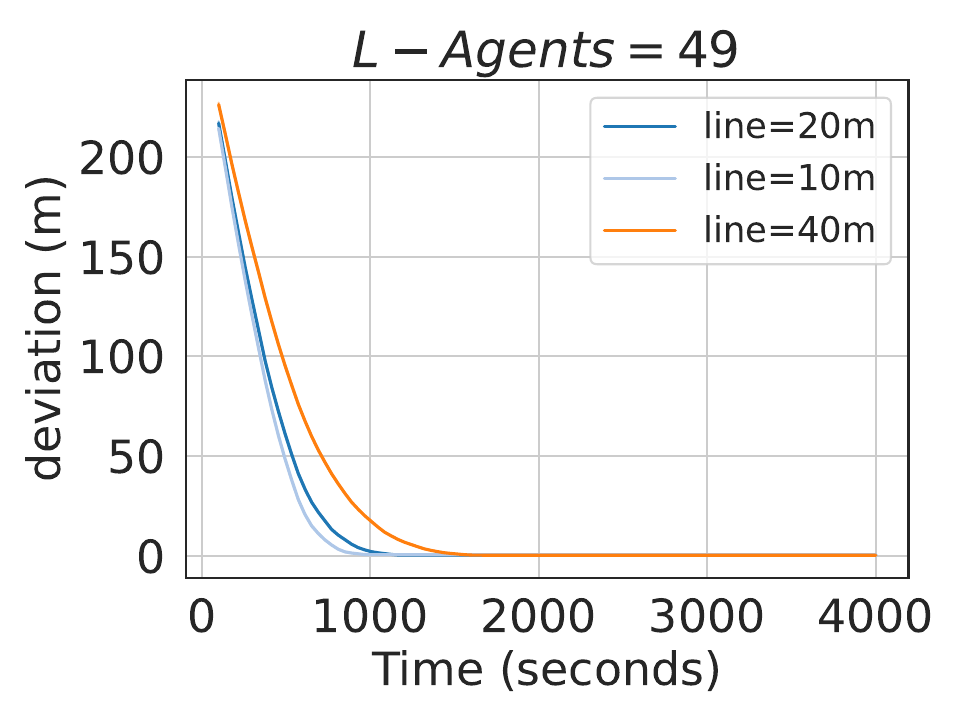}
    \caption{}\label{fig:detail-vshape-errors-0.4-10.0}
  \end{subfigure}
  \begin{subfigure}[b]{0.3\textwidth}
    \includegraphics[width=\textwidth]{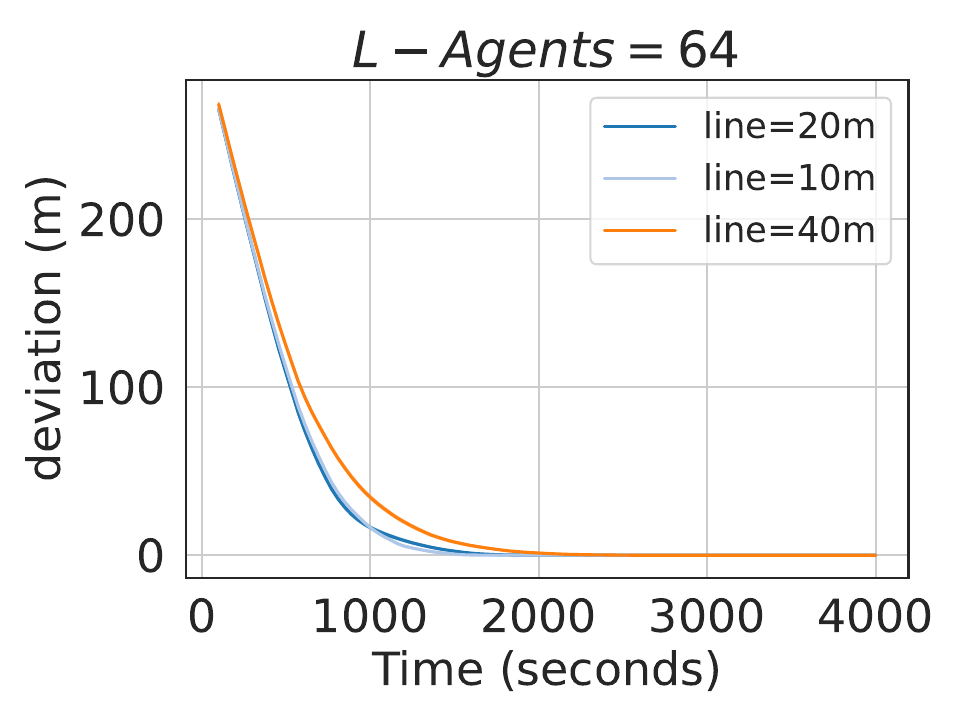}
    \caption{}\label{fig:detail-line-errors-0.4-10.0}
  \end{subfigure}
  \begin{subfigure}[b]{0.3\textwidth}
    \includegraphics[width=\textwidth]{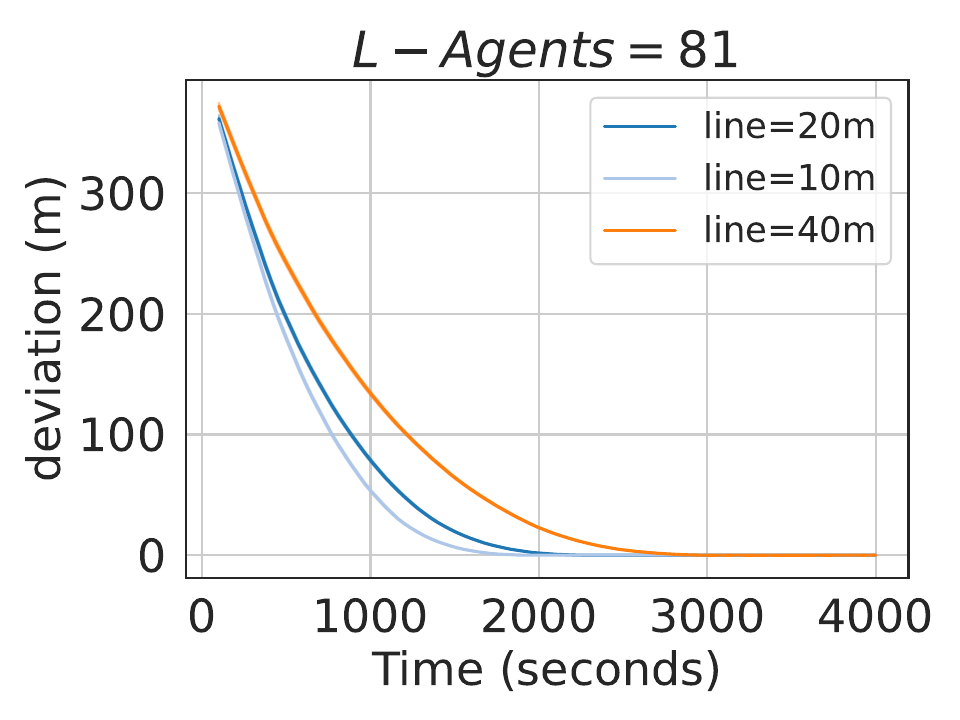}
    \caption{}\label{fig:detail-circle-errors-0.4-10.0}
  \end{subfigure}
  \begin{subfigure}[b]{0.3\textwidth}
    \includegraphics[width=\textwidth]{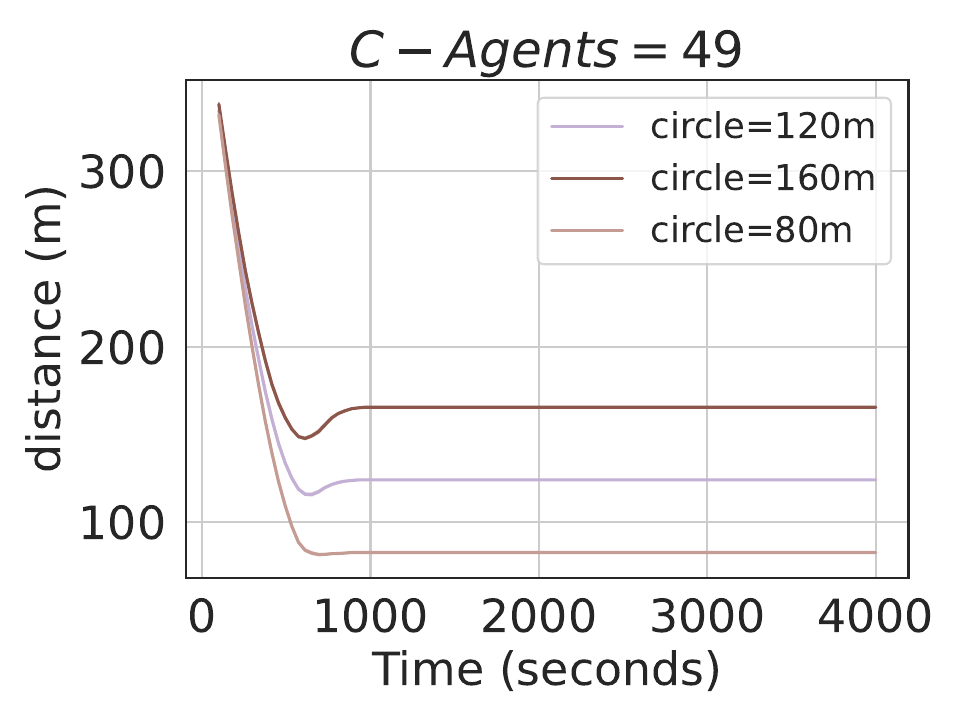}
    \caption{}\label{fig:detail-vshape-errors-0.4-10.0}
  \end{subfigure}
  \begin{subfigure}[b]{0.3\textwidth}
    \includegraphics[width=\textwidth]{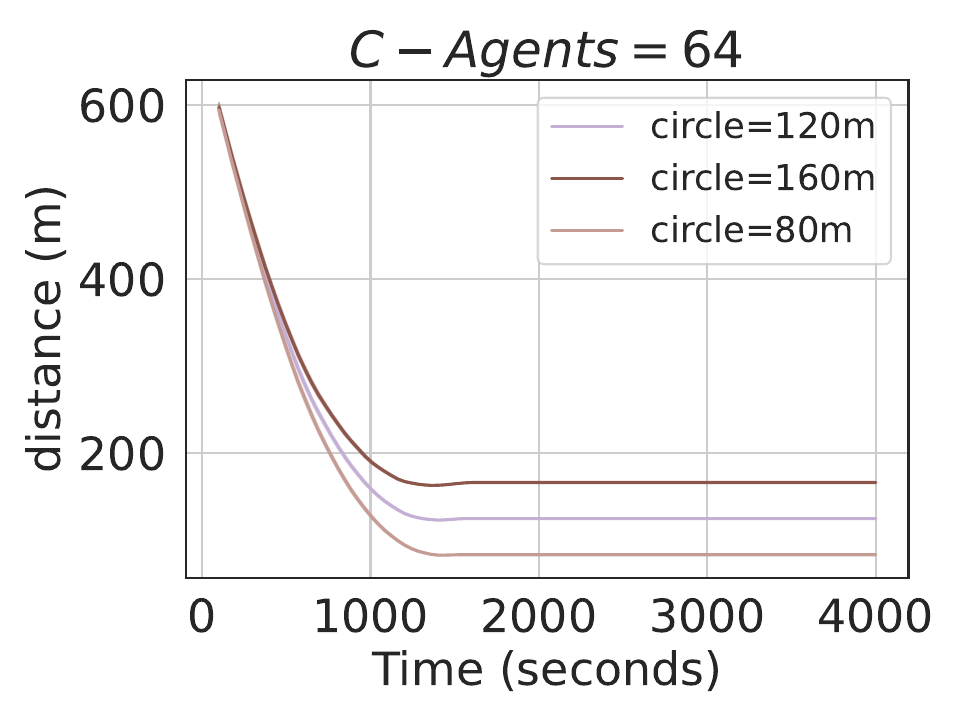}
    \caption{}\label{fig:detail-line-errors-0.4-10.0}
  \end{subfigure}
  \begin{subfigure}[b]{0.3\textwidth}
    \includegraphics[width=\textwidth]{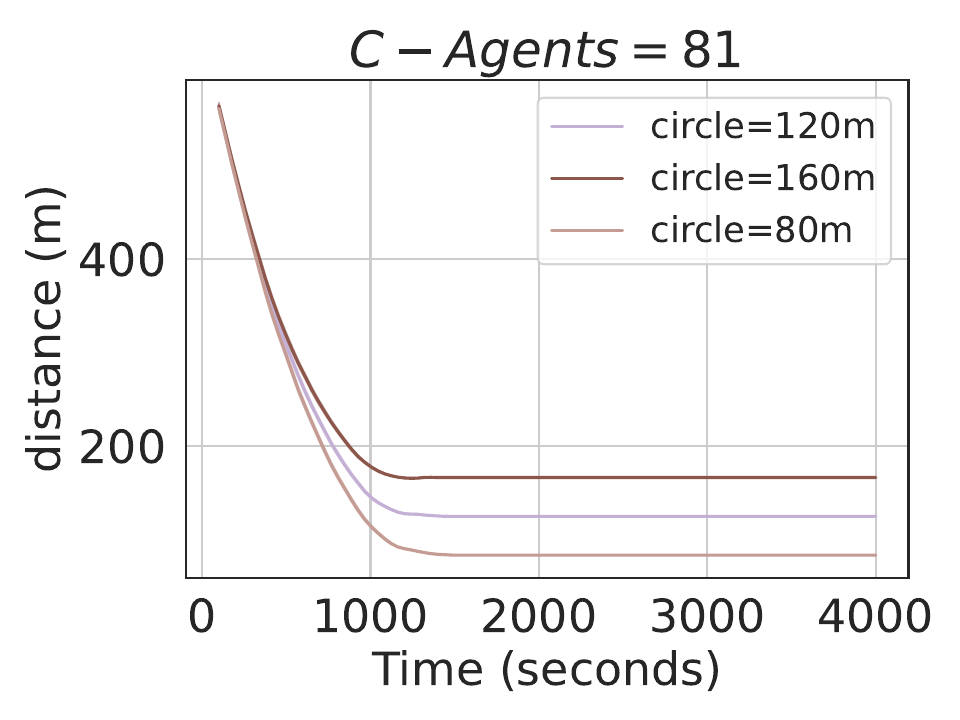}
    \caption{}\label{fig:detail-circle-errors-0.4-10.0}
  \end{subfigure}
  \caption{\revB{Effect of drones in the formation.
    The number of drones affects the formation of the structure,
    mainly in convergence time. Indeed, each of the structures converges to the desired value, but the time to reach it is different.
  }}\label{fig:pattern-eval-depth-size}
\end{figure}

\begin{figure}
  \centering
  \begin{subfigure}[b]{0.32\textwidth}
  \includegraphics[width=\textwidth]{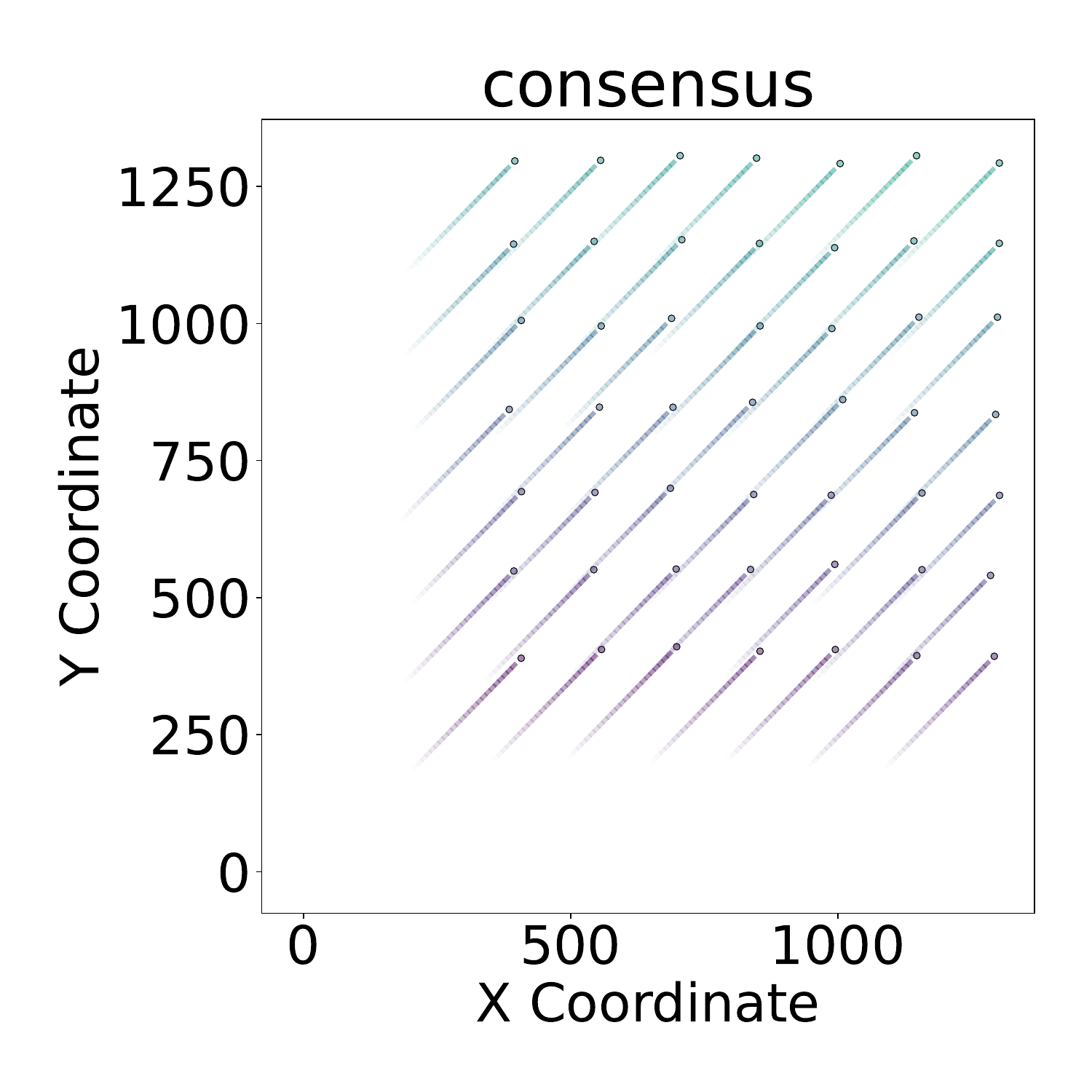}
  \caption{Traces}
  \label{fig:consensus-traces}
  \end{subfigure}

  \begin{subfigure}[b]{0.32\textwidth}
  \includegraphics[width=\textwidth]{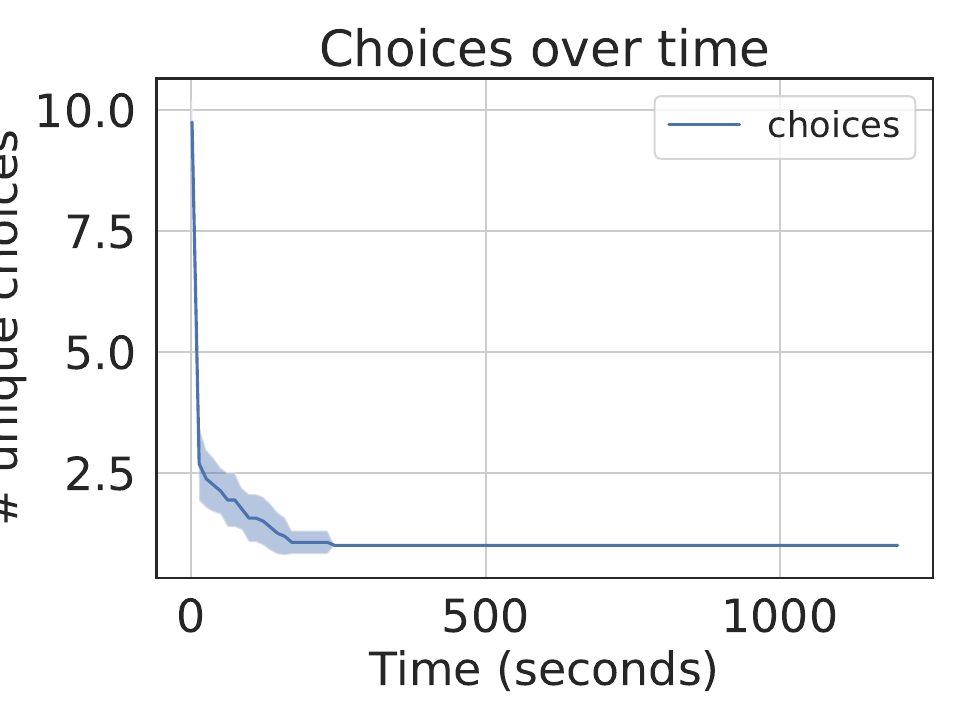}
  \caption{Selected choices (no noise)}
  \label{fig:consensus-choices}
  \end{subfigure}
  \begin{subfigure}[b]{0.32\textwidth}
    \includegraphics[width=\textwidth]{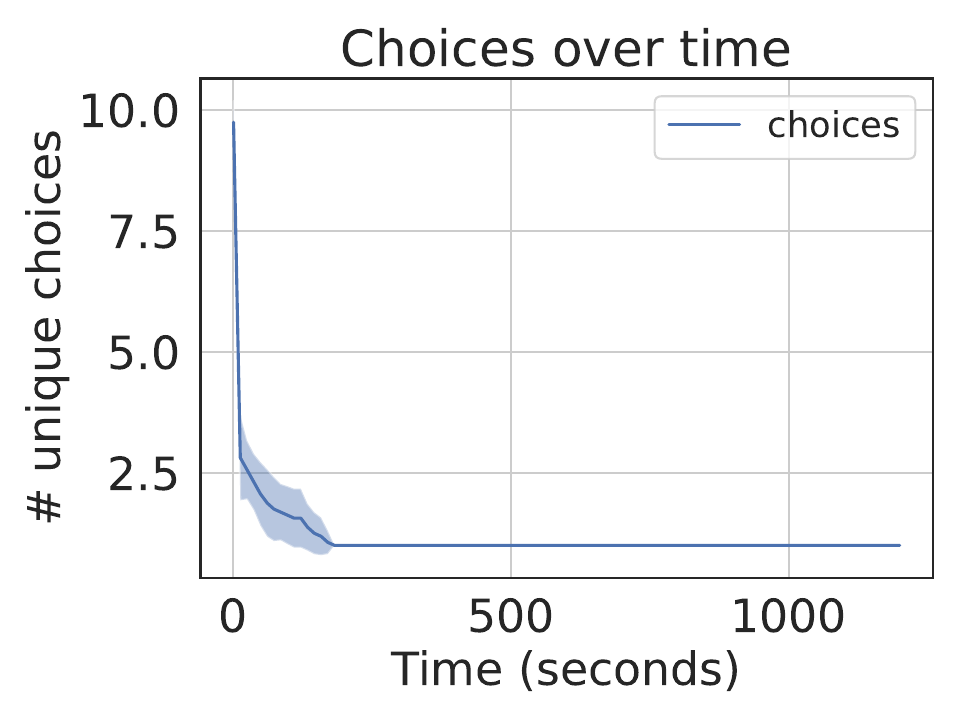}
    \caption{Selected choices ($D=0.1$)}
    \label{fig:consensus-choices}
    \end{subfigure}
  \begin{subfigure}[b]{0.32\textwidth}
    \includegraphics[width=\textwidth]{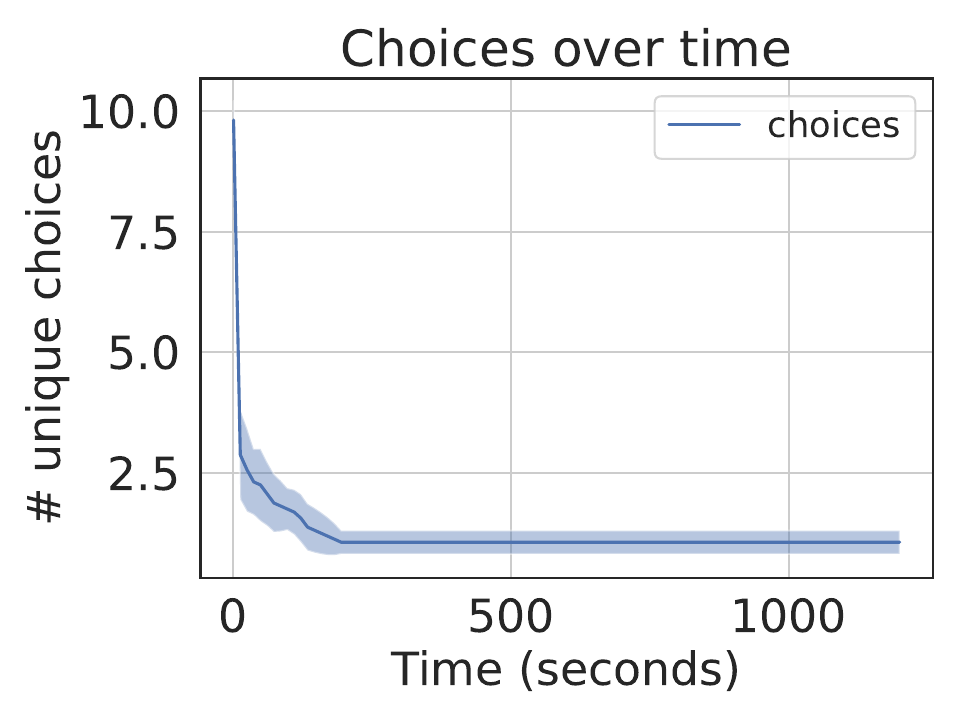}
    \caption{Selected choices ($D=0.4$)}
    \label{fig:consensus-choices}
  \end{subfigure}
  \caption{Consensus block evaluation.
   (A) shows the trace of the drones, while (B), (C), (D)
    show the selected choices over time for different message loss rates.}
  \label{fig:consensus-eval}
\end{figure}

\revB{
This section presents an evaluation of the drone formation control system, 
focusing on its performance in achieving and maintaining various geometric patterns (V-shape, separation, line, and circle) and consensus. 
The evaluation considers the impact of two key sources of error: message loss (D) and noise in position perception (P).

\Cref{fig:pattern-eval} illustrates the drone trajectories under ideal conditions (no message loss or perception noise), 
demonstrating the system's baseline effectiveness in forming the target shapes. 
Subsequent figures explore the system's robustness to error. \Cref{fig:pattern-eval-error-perception} 
shows the impact of perception noise (P=10m), 
while \Cref{fig:pattern-eval-error-both} depicts the combined effects of both message loss (D=0.7) and perception noise (P=10m). 
These figures visually demonstrate the degree to which the formations are distorted by these error sources.
A more detailed analysis of convergence and error is provided in Figures \ref{fig:pattern-eval-depth-message-loss} through \ref{fig:pattern-eval-depth-size}. 
These figures explore the effects of varying levels of message loss and perception noise on the formation shapes. 
These figures also demonstrate the system's ability to recover from transient node failures.

Finally, \Cref{fig:consensus-eval}, comprising subfigures \ref{fig:consensus-traces} and \ref{fig:consensus-choices}, evaluates the consensus algorithm. 
\Cref{fig:consensus-traces} visualises the drone trajectories as they converge to a common direction, while \Cref{fig:consensus-choices} quantifies the convergence process by showing the number of distinct directional preferences over time, 
ultimately demonstrating convergence to the leader's preference.
In the following, we elucidate the findings depicted in these charts.
}
\paragraph{V-shape formation}
\revB{
The stability and integrity of the V-shaped formation were investigated by varying the angle between the arms of the V.  
Three angles were tested: 30$^\circ$, 60$^\circ$, and 120$^\circ$.  
Under ideal conditions, the drones successfully established a V-shaped formation centred around a leader drone, as shown in \Cref{fig:pattern-eval}. 
The angle significantly influenced the effectiveness of the formation.
Indeed, 
\Cref{fig:detail-vshape-0} presents the average angular alignment of the leader drone relative to the other drones over time. 
 The data demonstrates consistent achievement of the target angle across various scenarios, with decreasing error as the simulation progresses.
Tighter angles (e.g., 60$^\circ$) resulted in faster and more stable formations, 
whereas wider angles (e.g., 30$^\circ$) led to slower formation times and reduced stability.

The formation exhibits self-healing capabilities, 
as demonstrated following the $T_k$ event in Figures~\ref{fig:pattern-eval} and \ref{fig:pattern-eval-depth-errors-message-loss}, although the transient phase can be substantial (1000s).

Message loss does not destabilise the formation until a threshold (D=0.4) is reached. 
Beyond this threshold, instability increases significantly. 
This occurs because the assumption of slower node movement relative to communication speed, 
inherent in the ideal conditions, namely the quasi-static approximation, is violated.  
Distant nodes in the V-shape require multiple steps to reach the leader, and with high message loss, 
they move without receiving the correct information.
Similarly, perception noise does not significantly impact the formation until a threshold (P=5) is exceeded.  
Beyond this threshold, while the V-shape remains recognisable (\Cref{fig:pattern-eval-error-perception}), maintaining accurate positioning becomes challenging.
Finally, the combination of message loss and perception noise has a critical impact.  Even when the formation is recognisable (\Cref{fig:pattern-eval-depth-both}), a significant proportion of nodes may be incorrectly positioned.  For example, with D=0.1 and P=5, the error rate reaches 20\%.
}

\paragraph{Separation formation}
\revB{
To analyse the impact of separation distance on achievable shapes, we varied this parameter (30m, 60m, and 120m) and observed the resulting formations. 
As shown in \Cref{fig:pattern-eval}, 
the drones successfully established formations centred around a leader in all cases. 
bigger separation distances resulted in faster, 
more stable formations, 
while smaller distances led to slower formation times and reduced stability. 
This is because smaller distances require more time to reach the desired formation in terms of movement needed to achieve the desired separation. 
Under ideal conditions, however, the drones achieved the desired formation, as illustrated by the evolution of $\rho(t)$ over time in \Cref{fig:detail-separation-0}. 
The system also demonstrated self-healing capabilities, 
recovering the desired formation within 500s after a $T_k$ event. Furthermore, 
the formation proved resilient to both message loss and perception noise, even when combined (Figures \ref{fig:pattern-eval-error-perception} and \ref{fig:pattern-eval-depth-errors-both}). 
However, with a very high message loss rate (D=0.7) and significant perception error (P=10), the formation degraded considerably introducing an error of 5 meters (on average) in the formation.}
\paragraph{Line formation}
\revB{
The line formation algorithm was evaluated by varying the inter-node distance parameter, 
which controls the target length of the drone segment.  
Three distances (10m, 20m, and 40m) were tested. 
\Cref{fig:pattern-eval} demonstrates the successful establishment of a line formation centred around a leader drone. 
This is further illustrated in \Cref{fig:detail-line-0}, which depicts the relationship between vertical deviation ($\overline{v}$) and inter-drone distance under ideal conditions, confirming that there are no misplaced nodes after the initial transient.
The formation's self-healing capabilities are evident in \Cref{fig:pattern-eval-depth-errors-message-loss}, 
where the formation recovers after node failures within approximately 1000s on average. 
Larger formations tend to be less stable, 
likely due to the inherent limitations of multi-hop communication, although this is not readily apparent from line deviation alone, 
as the algorithm initially positions nodes near their target locations, after which movement is primarily horizontal. 
Similar to the V-formation, this line formation is affected by message loss and perception noise, 
but the impact is less pronounced with smaller inter-node distances.  
A stable formation is achievable even with high message loss (0.4) and perception noise (10.0) when the inter-node distance is 10m. 
However, the combined effect of both factors significantly destabilises the formation, as shown in \Cref{fig:detail-circle-errors-0.2-5.0} and \Cref{fig:pattern-eval-error-both}
}.
\paragraph{Circle formation}
\revB{
To evaluate circle formation, we varied the distance parameter relative to the central leader (effectively the circle's radius) using values of 30m, 60m, and 120m.  As shown in \Cref{fig:pattern-eval}, 
drones successfully established a circular formation around the leader,
maintaining the desired distance with minimal error under ideal conditions (\Cref{fig:detail-circle-0,fig:detail-circle-errors-0.0}).  This centralised structure facilitates rapid recovery from node failures, demonstrated in \Cref{fig:pattern-eval-depth-errors-message-loss} for all tested radii. 
The formation also proved robust to message loss and perception noise (\Cref{fig:pattern-eval-error-perception,fig:pattern-eval-depth-errors-perception}), 
exhibiting low error and minimal deviation from the desired formation even with substantial message loss (0.4) and perception noise (10.0) (\Cref{fig:detail-circle-errors-0.4-10.0,fig:pattern-eval-error-both}).  This robustness extends to the combined presence of both disturbances (\Cref{fig:detail-circle-errors-0.4-10.0,fig:pattern-eval-error-both}). However, under extreme conditions (0.7 message loss and 10.0 perception noise), formation integrity degrades, resulting in an average error of 5 meters (\Cref{fig:detail-circle-errors-0.4-10.0}).
}
\paragraph{Consensus}
For the consensus, we create random preferences for each drone but the leader, and we let them converge to a single value.
Particularly, the leader has a preference to go diagonally to the right.
 In ideal conditions, the results 
 indicate that the drones successfully establish a consensus on a single value,
 since all the drone goes in the same direction (see \Cref{fig:consensus-traces}).
 More details can be found in \Cref{fig:consensus-choices}, in which we case see that the system always converges to a single choice (the one of the leader).
\revB{
  The noise in messages could affect the convergence to the consensus,
 indeed, in some scenario with $D=0.4$ the drones do not converge to the leader's preference (see \Cref{fig:consensus-choices}).
 This is due to the fact that, if the message loss is too high, the drones may not receive the leader's preference in time, therefore converging to another preferences.
}
\revB{
\subsection{Discussion}
\label{subsec:discussion-formationn}
This quantitative evaluation of the core resilient swarm operators (and consensus) within \MacroSwarm{} demonstrates its robustness to adversarial events, specifically message loss, perception noise, and node failures.  
Results show the system maintains the desired formation even under these conditions, exhibiting noteworthy self-healing capabilities by recovering from transient disturbances.  

However, the time required for the swarm to achieve a self-stable formation is a key consideration.  
As discussed in the properties section, the self-stabilisation of the Leader-based module's building blocks implies a form of resiliency for pattern formation. 
This resiliency holds if the evolution of the system is quasi-static, i.e.,
when node velocities are sufficiently low relative to the stabilisation speed of computations.  
High message loss rates combined with long communication paths to the leader violate this quasi-static assumption, destabilising the formation.  
For instance, in a linear formation with substantial inter-node spacing, the leftmost node may require 10 hops to reach the leader. 
Under high message loss, if this node's message is lost, re-establishing communication with the leader (and vice versa) could require an additional 10 time steps, leading to instability since the evolution is no longer quasi-static.
Nevertheless, the system's ability to achieve formation even with relatively high message loss rates (0.4) and perception noise (10.0) demonstrates the feasibility of convergence under near-equilibrium conditions. 
Future work will investigate the real-time performance of the proposed algorithms to determine the convergence threshold more precisely.

We also investigated the scalability of our blocks with respect to the number of agents. 
Indeed, we observed that even with increased time due to the larger spatial extent, 
the agents successfully reached their correct positions in various scenarios(see \Cref{fig:pattern-eval-depth-size}). 
While our aggregate approach theoretically scales to a very large number of agents, we acknowledge the limitations of our simulations. 
Specifically, crucial real-world robotic swarm considerations such as embodiment and physical spatial constraints were not addressed and are left for future work.
Nevertheless, these initial results are promising in terms of the theoretical scalability of the formations.
}




\subsection{Case Study: Find and Rescue}\label{subsec:case-study}
In our scenario, we want a fleet of drones to patrol a spatial area.
In the area, dangerous situations may arise (e.g., a fire breaks out, a person gets injured, etc.). 
 In response to these, a drone designated as a \emph{healer} 
 must approach and resolve them. 
Exploration must be carried out in groups composed of \emph{at least} one 
 healer and several \emph{explorers}, who will help the healer identify alarm situations.

\subsubsection{Goal}
The goal of the proposed case study is to demonstrate the effectiveness of the proposed \ac{api} in terms of correctness (i.e., the described behaviour collectively does what is expressed)
\revB{and to provide preliminary evidence of its expressiveness, including the ability to describe complex behaviours concisely}.
For the first point, since deploying a swarm of drones is costly, we will make use of simulations to verify that the program is functioning correctly both qualitatively (e.g., observing the graphical simulation) and quantitatively (i.e., extracting the necessary data and computing metrics that allow us to understand if the system behaves as it should).
\revB{
Regarding the second point, as it involves a qualitative assessment, 
we will outline the development process leading to the implemented code, 
highlighting its readability.  
To this end, we will examine the number of calls to the low-level, device-centric field computation functions (like \texttt{rep}, \texttt{nbr}, etc.), which are not directly related to collective behaviour.
}

\subsubsection{Setup}
Initially, 50 explorers and 5 healers are randomly positioned in an area of $1 km^2$. 
 Each drone has a maximum speed of approximately 20 km/h and a communication range of 100 meters.
The alarm situations are randomly generated at different times 
 within the spatial area in a $[0,S_k=50]$ minutes time-frame. 
Each simulation run lasts 90 simulated minutes, 
 during which we expect the number of alarm situations to reach a minimum value.
The node should form teams of at least one healer and several explorers, maintaining a distance of at least 50 meters between the node and the leader.
Also in this case, we tested the resiliency of the system to message loss and perception errors, varying the noise in the position perception and the message loss rate.

\subsubsection{Implementation details}\label{casestudy:impl}
\begin{figure}[t]
\begin{subfigure}{0.29\textwidth}
  \centering
  \fbox{\includegraphics[width=\textwidth]{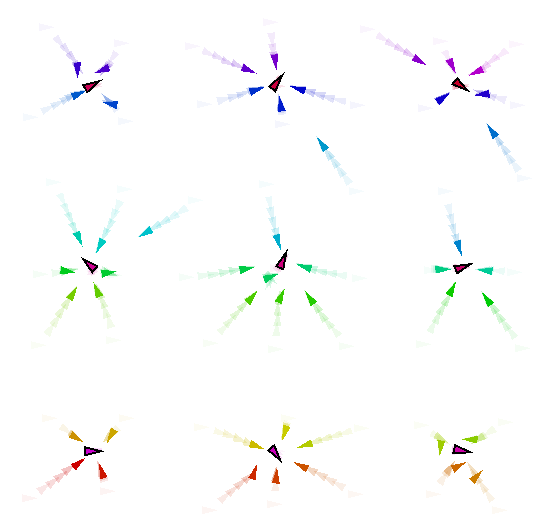}}
  \caption{Team formation}
  \label{fig:team-formation}
\end{subfigure}
\hfill
\begin{subfigure}{0.29\textwidth}
  \centering
  \fbox{\includegraphics[width=\textwidth]{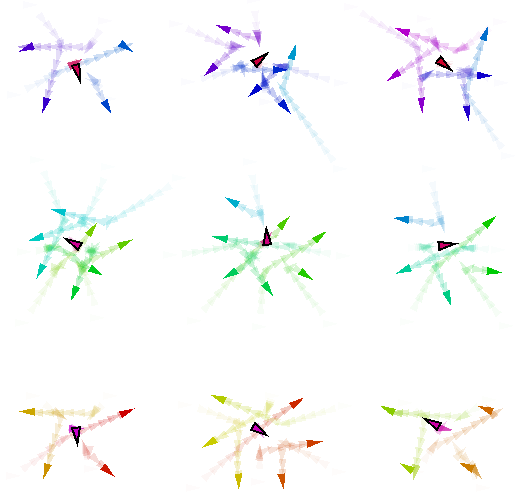}}
  \caption{Circle formation}
  \label{fig:circle-formation}
\end{subfigure}
\hfill
\begin{subfigure}{0.29\textwidth}
  \centering
  \fbox{\includegraphics[width=\textwidth]{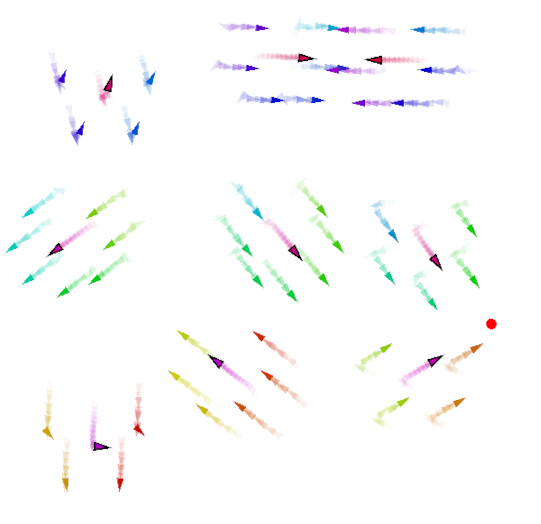}}
  \caption{Explore}
  \label{fig:explore}
\end{subfigure}
\caption{The first phases of the scenario described in \Cref{sec:eval}. 
 At the beginning, the system is split into teams; 
 afterwards, the teams assume a spatial formation (circular, in this case);
 finally, the teams start exploring the overall area.}\label{fig:scenario}
\end{figure}

To structure the desired swarm behaviour, 
 we break the problem into parts:
\begin{enumerate}
  \item the swarm must split into teams regulated by a healer, who works as a \emph{leader} 
   (\Cref{fig:team-formation});
  \item teams must assume a spatial formation promoting the efficiency of the exploration (\Cref{fig:circle-formation});
  \item the teams must explore the overall area (\Cref{fig:explore});
  \item when any node detects an alarm zone, it must point that to the healer;
  \item the healer node approaches the dangerous situation to fix it;
  \item then, the team should return to the exploration phase.
\end{enumerate}
We now describe the implementation of each part, leveraging the \MacroSwarm{} API. 
First of all, for creating teams, we can use the \textbf{Team Formation} blocks:
\begin{lstlisting}
val teamFormedLogic = 
  (leader: ID) => isTeamFormed(leader, minimumDistance + confidence)
def createTeam() = 
  teamFormation(sense("healer"), minimumDistance, teamFormedLogic)
\end{lstlisting}
where \lstinline|minimumDistance| is the minimum distance between nodes during the 
 team formation phases and \lstinline|confidence| is the confidence interval 
 used to check if the team is formed through the \lstinline|isTeamFormed| method.
Each team then should follow the aforementioned steps, 
 expressible using the \textbf{Swarm Planning} \ac{api}: 
\begin{lstlisting}[xrightmargin=-2pt]
def insideTeamPlanning(team: Team): Vector = 
 team.insideTeam {
  healerId =>
   val leading = healerId == mid() // team leader
   execute.repeat(
    plan(formation(leading)).endWhen(circleIsFormed), // shape formation
    plan(wanderInFormation(leading)).endWhen(dangerFound), // exploration
    plan(goToHealInFormation(leading, inDanger)).endWhen(dangerReached), 
    plan(heal(healerId, inDanger)).endWhen(healed(dangerFound)) // healing
   ).run() // repeat the plan
 }
\end{lstlisting}
The first step is the formation of the teams, 
 based on method \lstinline|formation| which
 internally uses \lstinline|centeredCircle| 
  to place the nodes in a circle around the leader node. 
 Function \lstinline|circleIsFormed| verifies whether the nodes are in a circle formation, i.e., that the distance between any node and the leader is less than \lstinline|radius| (set to 50 meters in this scenario).
The second step is the exploration phase, 
 implemented by method \lstinline|wanderInFormation|, 
 which uses the \lstinline|explore| function to move the nodes to a random direction
 within given bounds while keeping the circle formation.
 This leverages \lstinline|centeredCircle|,  passing 
 the movement logic of the healer (leader) to the block.
Exploration will go on until someone finds a danger node, 
 denoted by predicate \lstinline|dangerFound|.
This internally uses \lstinline|C| and \lstinline|G| to collect the danger nodes' positions
 and share them within the team:
\noindent\begin{minipage}{\textwidth}
\begin{lstlisting}
def dangerFound(healer: Boolean): Boolean = {
  val dangerNodes = 
    C(sense("healer"), combinePosition, List(sense("danger")), List.empty)
  broadcast(healer, dangerNodes.nonEmpty)
}
\end{lstlisting}
\end{minipage}
The third step is the movement towards the danger node, 
 which is implemented by the \lstinline|goToHealInFormation| method, 
 which uses again the \lstinline|centeredCircle| function
 with a delta vector that moves the leader node towards the danger node.
\lstinline|inDanger| is computed similarly to \lstinline|dangerFound|, 
 but, in this case, the position will be shared instead. 
 \lstinline|dangerReached| is a Boolean field indicating if the healer node is close enough to the danger node.
The last step is the healing of the danger node, which is modelled as an actuation of the healer. 
 The rescue ends when the danger node is healed.
As a final note, we also want the nodes to be able to avoid 
 each other when they are too close, even if they are not in the same team.
 For this, we leverage the \textbf{Flocking} \ac{api} the \texttt{separation} block outside the team logic.
\revB{
Note that the steps are continuosly verified---if some condition is not met, the system will go back to the previous step and re-evaluate the situation.}
Then, the main program is as follows:
\noindent\begin{minipage}{\textwidth}
\begin{lstlisting}
val team = createTeam()
rep(Vector.Zero) { v =>
  insideTeamPlanning(team) + 
  separation(v, OneHopNeighbourhoodWithinRange(avoidDistance))
}.normalize
\end{lstlisting}
\end{minipage}
This program shows that the \ac{api} 
 is flexible enough to create complex behaviours handling various coordination aspects.

\subsubsection{Results}
\begin{figure}[t]
\centering
\begin{subfigure}{0.32\textwidth}
  \centering
  \includegraphics[width=\textwidth]{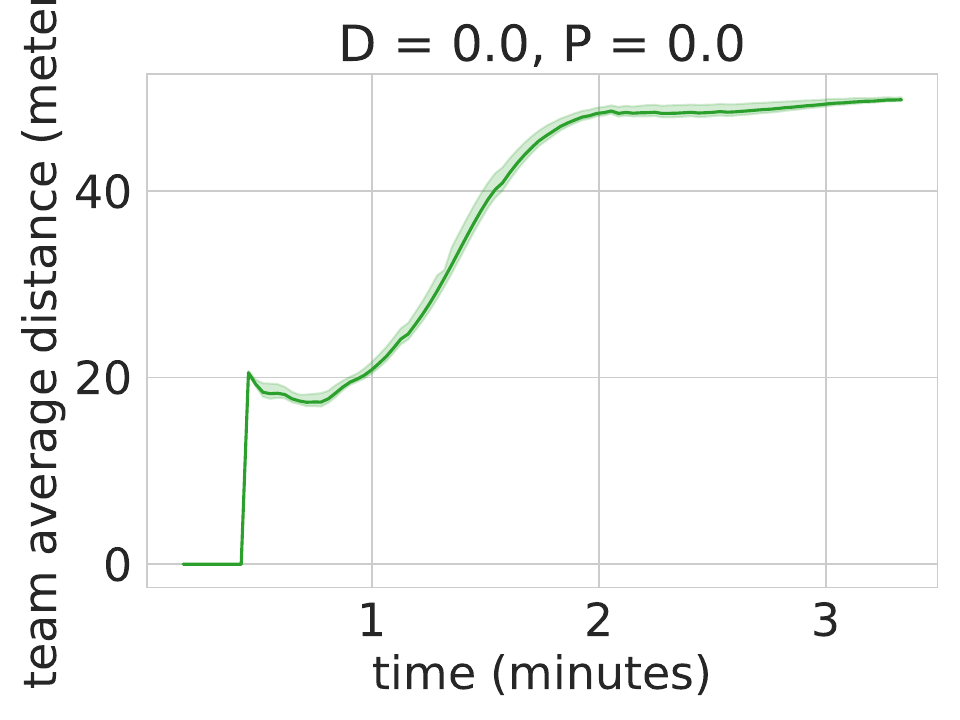}
  \caption{}\label{fig:average_intra_team_distance_0.0_0.0}
\end{subfigure}
\hfill
\begin{subfigure}{0.32\textwidth}
  \centering
  \includegraphics[width=\textwidth]{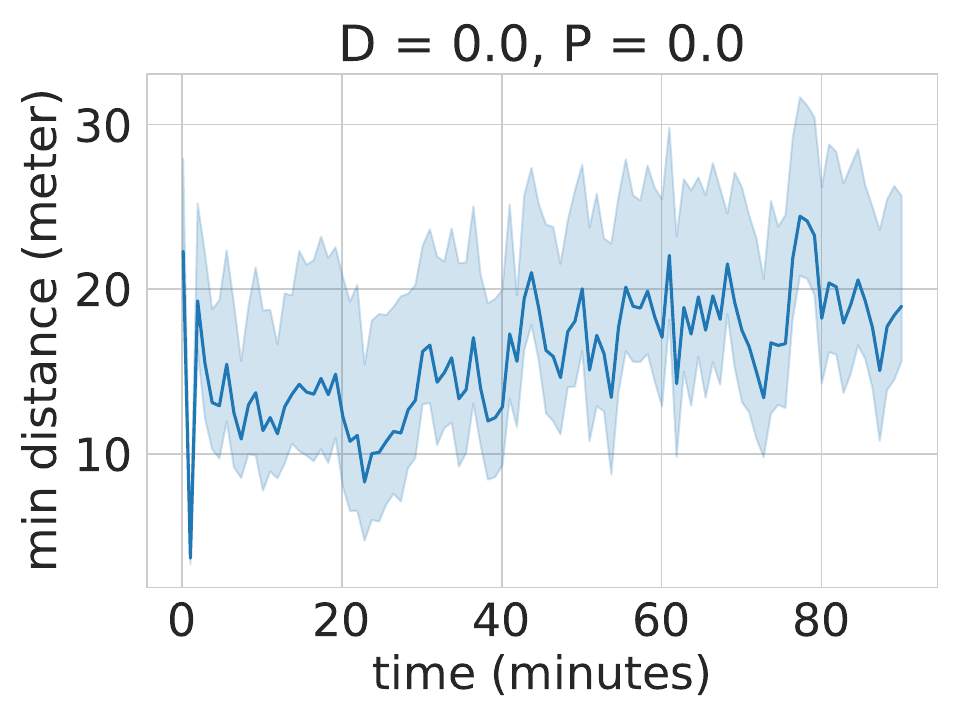}
  \caption{}\label{fig:min_distance_0.0_0.0}
\end{subfigure}
\hfill
\begin{subfigure}{0.32\textwidth}
  \centering
  \includegraphics[width=\textwidth]{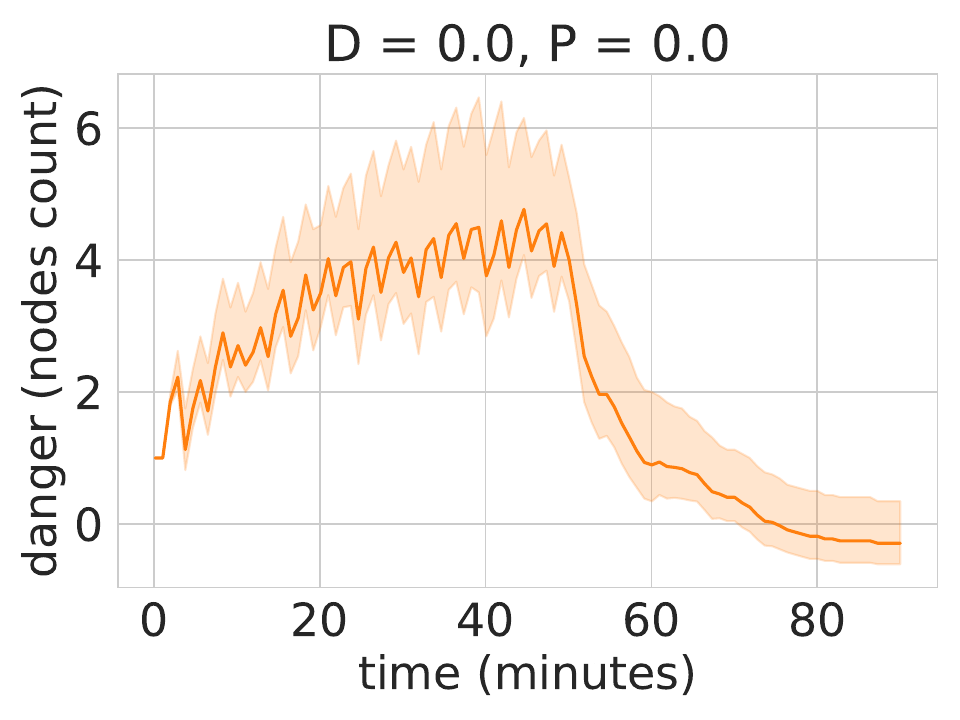}
  \caption{}\label{fig:in_danger_0.0_0.0}
\end{subfigure}

\begin{subfigure}{0.32\textwidth}
  \centering
  \includegraphics[width=\textwidth]{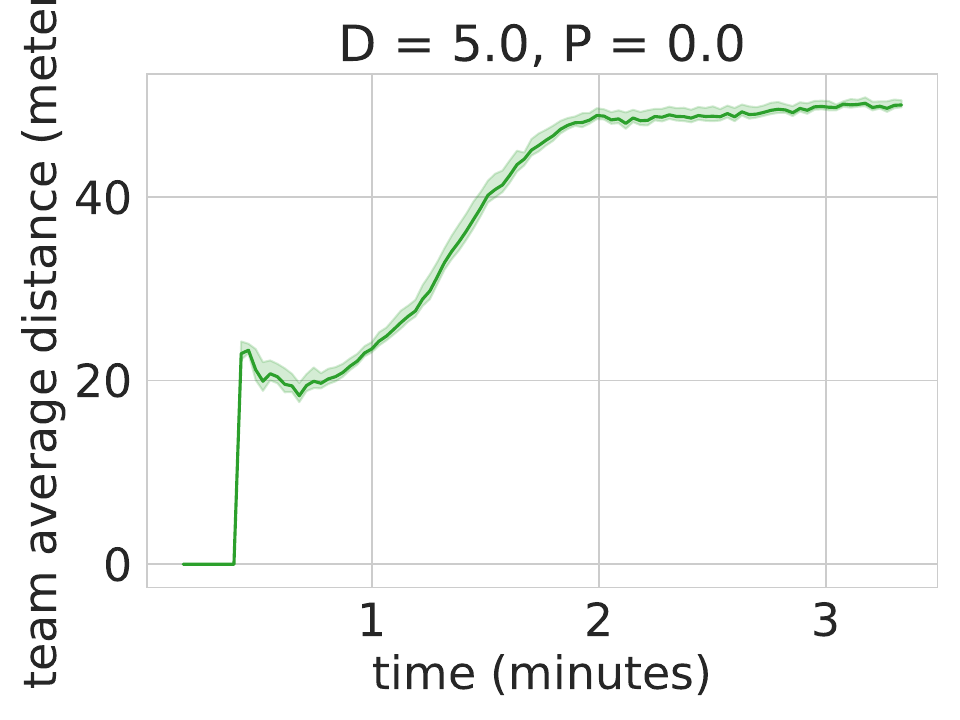}
  \caption{}
\end{subfigure}
\hfill
\begin{subfigure}{0.32\textwidth}
  \centering
  \includegraphics[width=\textwidth]{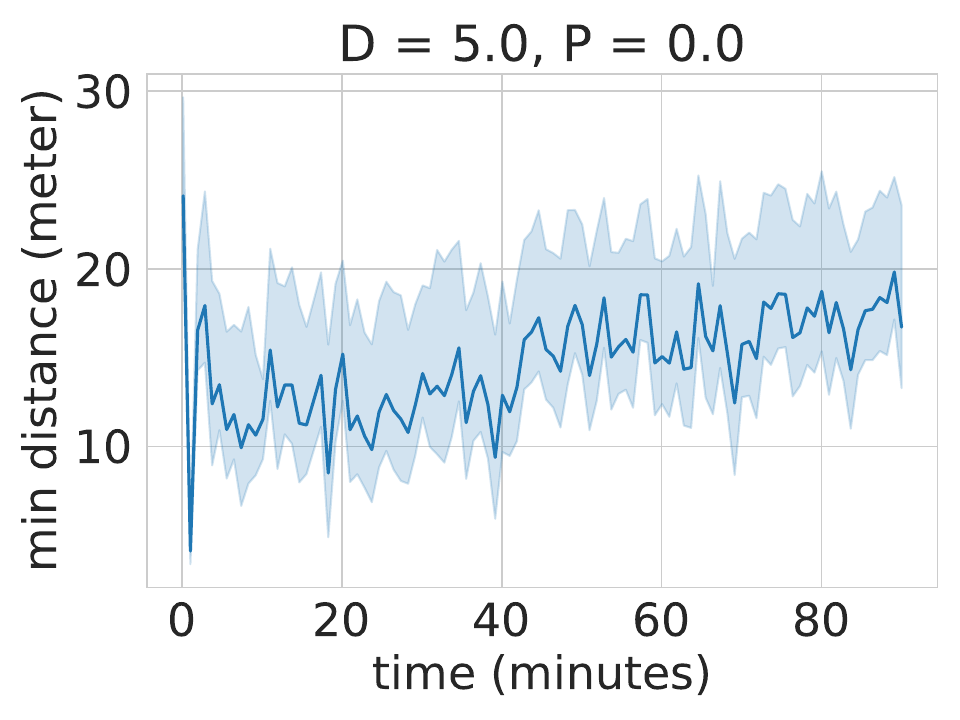}
  \caption{}
\end{subfigure}
\hfill
\begin{subfigure}{0.32\textwidth}
  \centering
  \includegraphics[width=\textwidth]{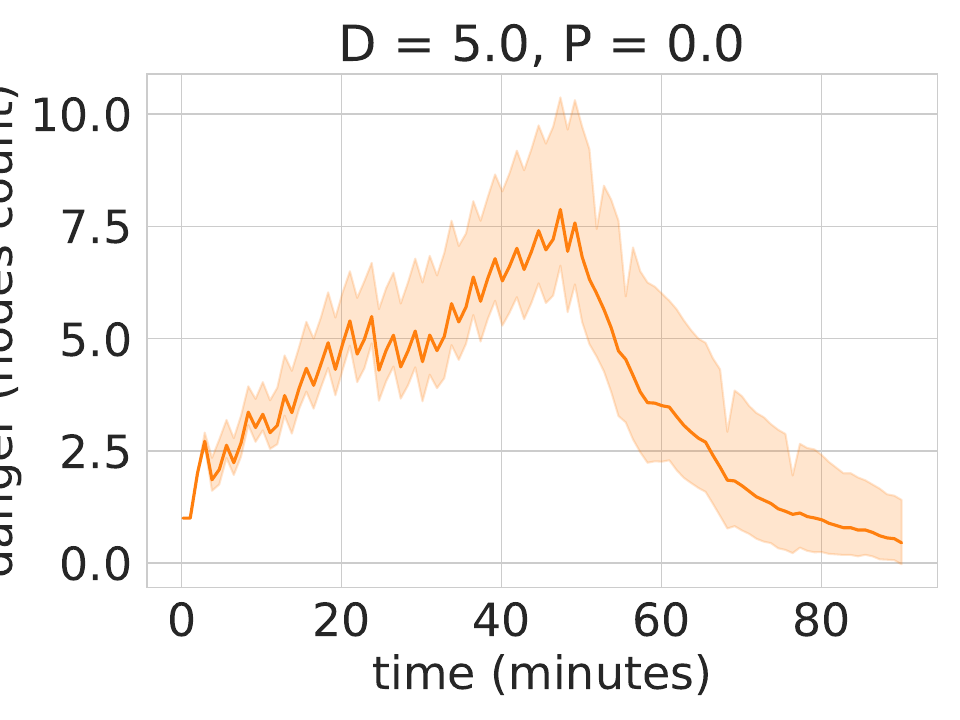}
  \caption{}
\end{subfigure}
\begin{subfigure}{0.32\textwidth}
  \centering
  \includegraphics[width=\textwidth]{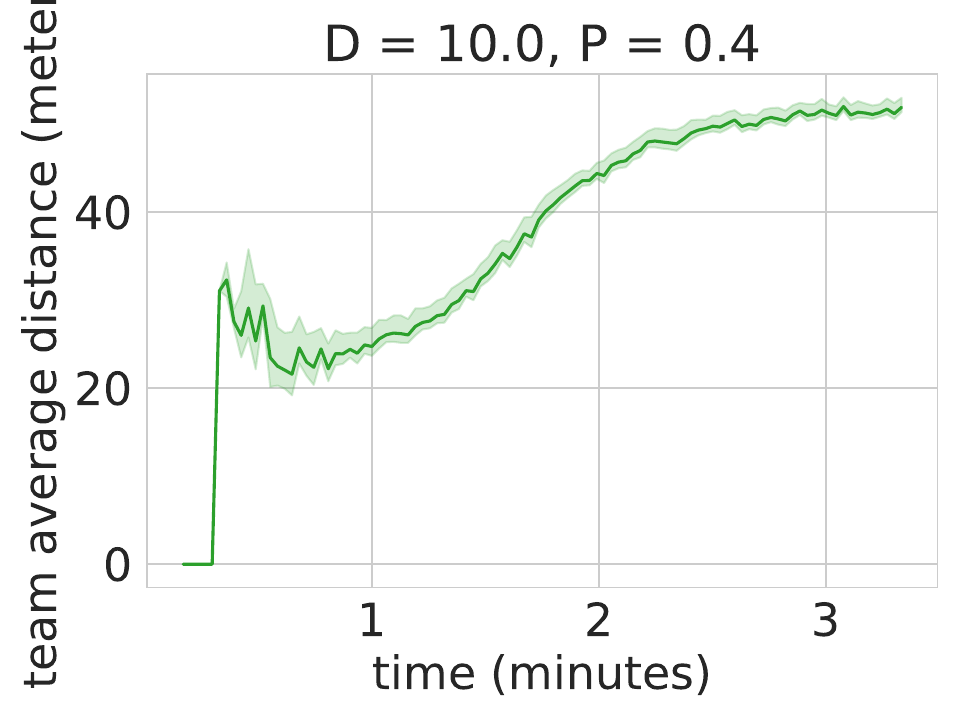}
  \caption{}
\end{subfigure}
\hfill
\begin{subfigure}{0.32\textwidth}
  \centering
  \includegraphics[width=\textwidth]{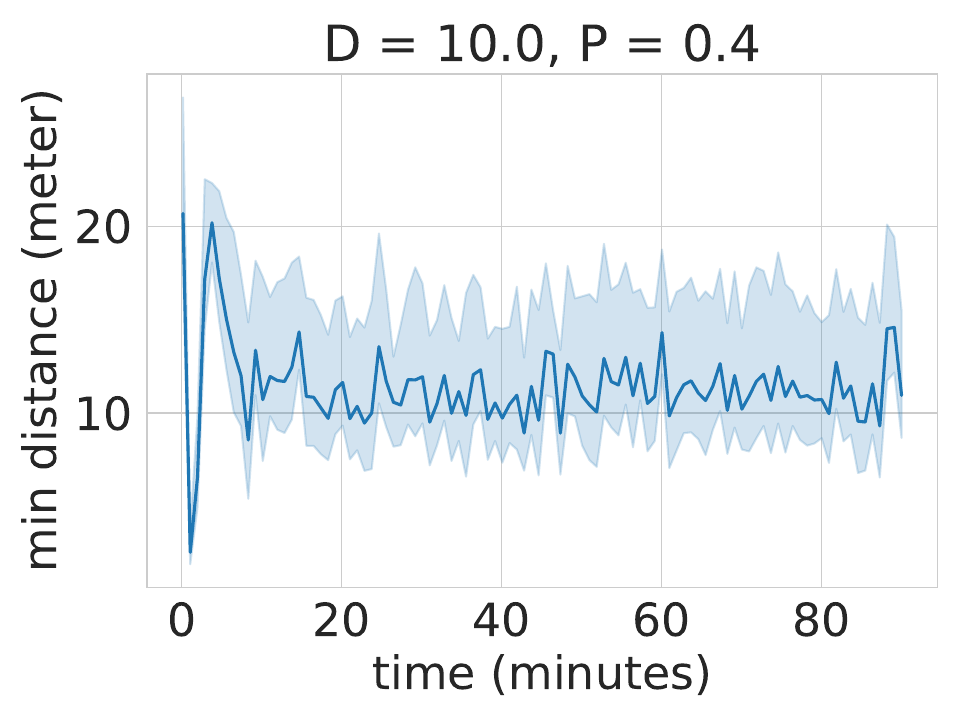}
  \caption{}
\end{subfigure}
\hfill
\begin{subfigure}{0.32\textwidth}
  \centering
  \includegraphics[width=\textwidth]{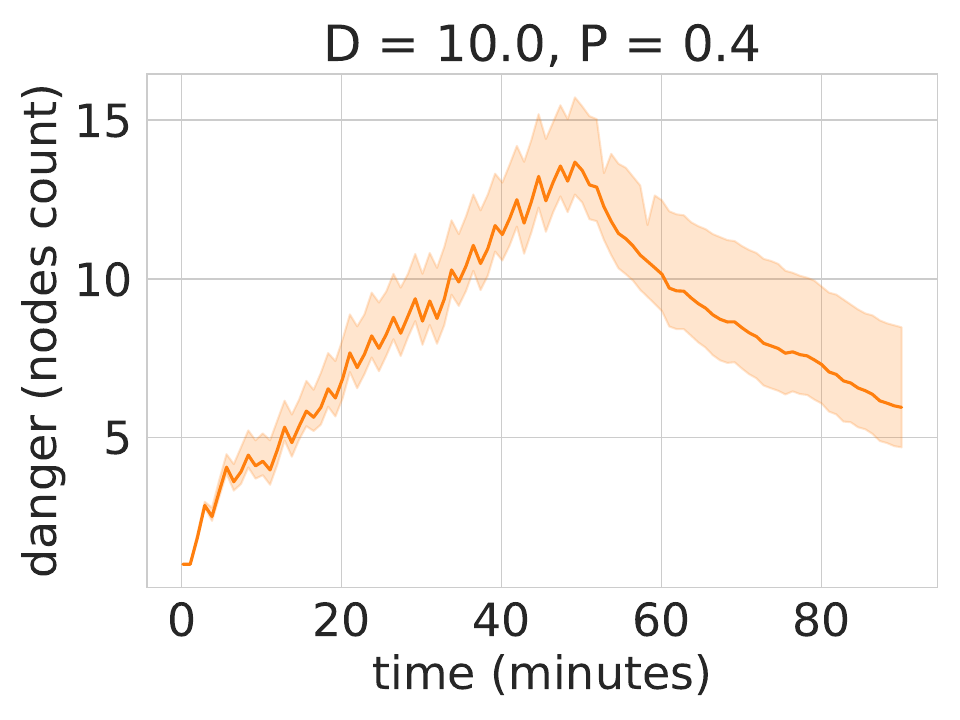}
  \caption{}
\end{subfigure}

\begin{subfigure}{0.32\textwidth}
  \centering
  \includegraphics[width=\textwidth]{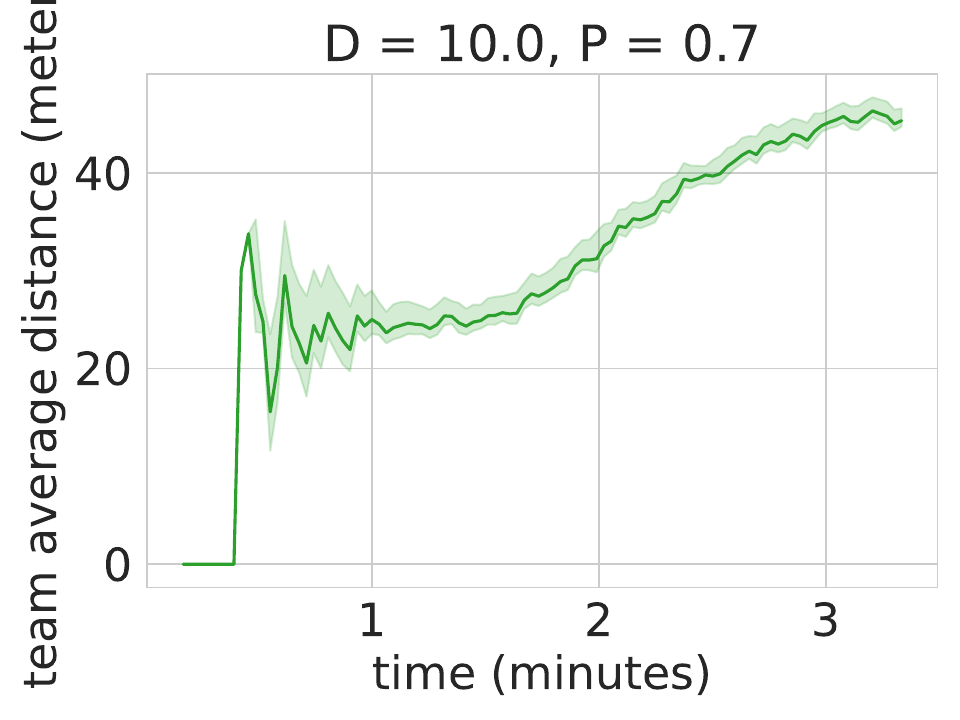}
  \caption{}
\end{subfigure}
\hfill
\begin{subfigure}{0.32\textwidth}
  \centering
  \includegraphics[width=\textwidth]{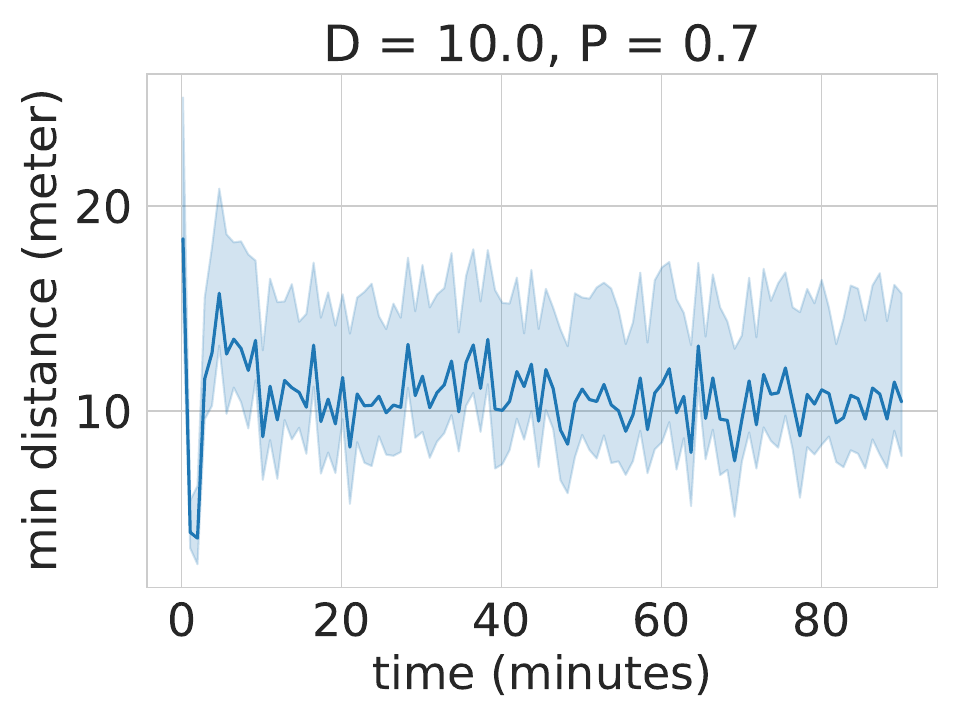}
  \caption{}
\end{subfigure}
\hfill
\begin{subfigure}{0.32\textwidth}
  \centering
  \includegraphics[width=\textwidth]{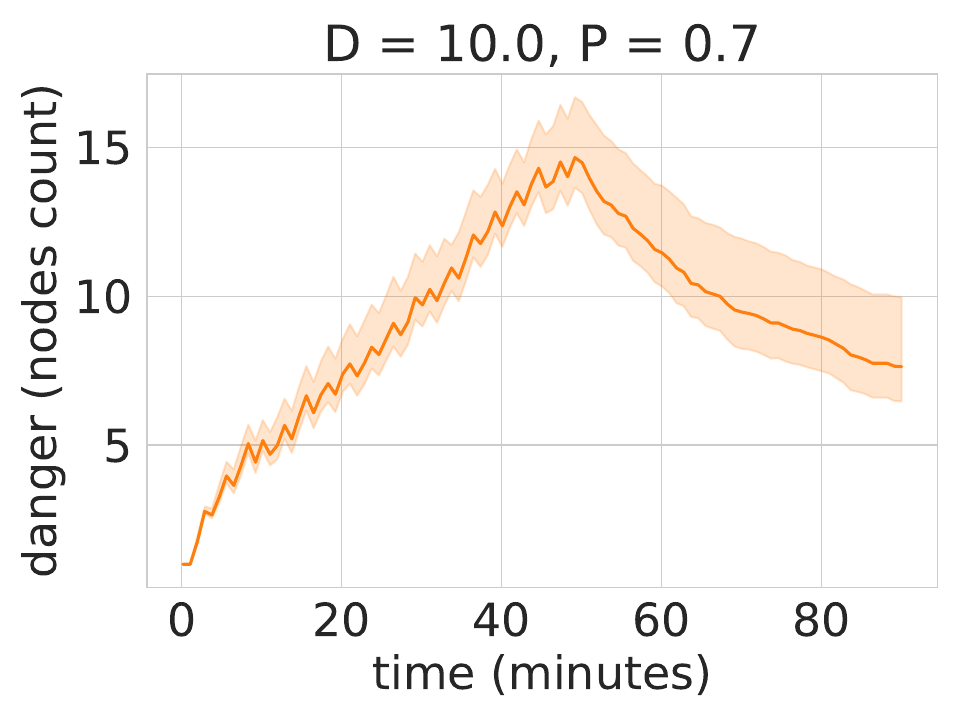}
  \caption{}
\end{subfigure}

\caption{
  Quantitative plots of the simulated scenario. 
  The first column shows the average team distance in the first two minutes. 
  The second column shows the minimum distance between nodes. 
  The third column shows the nodes in danger through time. 
}\label{fig:results}
\end{figure}

We validated the results by effectively running \revA{Alchemist} simulations. 
We launched 64 simulation runs with different random seeds: 
 \Cref{fig:results} shows the average results obtained. 
We extracted the following data (see \Cref{fig:results}):
\begin{itemize}
  \item \emph{intra-team distance}:
   after an initial adjustment phase, 
   the system should converge to an average distance of 50 meters~\Cref{fig:average_intra_team_distance_0.0_0.0};
  \item \emph{minimum distance between each node}: 
    as we want to avoid collisions, 
    the minimum distance between 
    two nodes should always be greater than zero~\Cref{fig:min_distance_0.0_0.0};
  \item \emph{number of nodes in danger}: 
   we expect the nodes in danger to increase 
   up to $S_k$ minutes and then decrease 
   tending towards zero~\Cref{fig:in_danger_0.0_0.0}.
   This happens since danger nodes stop to to be spawn after $S_k$ minutes and therefore
   the healer nodes can heal all the remaining nodes.
   
\end{itemize}
\revB{
The results (\Cref{fig:results}) show that the system can achieve the expected outcomes in ideal condition.
\Cref{fig:results} demonstrates the system's performance under various conditions.  
In ideal conditions (first row), 
convergence to the correct inter-peer distance is achieved in under three minutes.
The middle chart confirms collision avoidance, as the minimum inter-node distance remains positive.  
Finally, the last chart illustrates successful detection and recovery of nodes in danger, with their count converging to zero as expected.
Introducing perception noise (second row) still allows the system to form the desired shape, 
but with slight misalignment and increased convergence time.  Crucially, node in danger detection and recovery remain functional. 
This robustness, even with combined layers, highlights the composability of the API, maintaining coherent structure.

However, under combined high message loss and high perception error (bottom two rows), 
system performance degrades significantly.  
Convergence time doubles and endangered node recovery fails.  
This degradation is anticipated, 
mirroring the performance limitations of the underlying system layer observed in the previous section.}
\subsection{Discussion}\label{subsec:discussion}
Despite its simplicity, 
this use case allowed us to demonstrate the capability of \MacroSwarm{}, 
both in qualitative terms (i.e., \revB{the produced code appears concise and potentially easier to understand due to the higher level collective abstractions since it uses rep only once}) 
and quantitative terms (i.e., the data show that the swarm follows the given instructions correctly).

That being said, there are several things to consider when using the library in real-world contexts. Ours is a top-down approach, in which we have defined an evaluation and implementation system that is general enough to be executed in various multi-robot systems. Specifically, we require that at least: 
\emph{i)} nodes can perceive and interact with neighbours and approximate a direction vector to each of them; 
\emph{ii)} they can move in a specific direction with a certain velocity; and 
\emph{iii)} they can perceive distance and direction for certain obstacles; and 
\emph{iv)} they communication velocity is high enough to allow the system to converge in a reasonable time w.r.t. node speed.
As for point \emph{i)}, this can be developed using specific local sensors (e.g., range and bearing systems~\cite{DBLP:conf/antsw/BilalogluSAST22}), by using GPS, by approximating distances using cameras mounted on each drone, or by using Bluetooth direction finding~\cite{DBLP:conf/wcnc/SambuW22}.
Concerning the point \emph{ii)} the velocity vector can be mapped to the motors of the UAVs, or the motor's wheels of the ground robots~\cite{DBLP:conf/icra/KorenB91}, \revB{suggesting a potential pathway for real-world implementation.}
Finally, concerning \emph{iii)}, there are several solutions for perceiving the direction of obstacles by leveraging various sensors, like \emph{\ac{lidar}} systems~\cite{DBLP:conf/icinfa/PengQZXLG15}.
For \emph{iv)}, the communication velocity can be increased by using more powerful communication systems, like 5G or 6G networks.

That being said, we know that the reality gap for real-world scenarios could introduce divergences from the behaviours shown, as the used simulator, although general, does not simulate many aspects of reality, such as friction. 
We aim to test the \ac{api} in more realistic simulators (like Gazebo~\cite{DBLP:conf/iros/KoenigH04}) or real systems as a future work.

\begin{table}[htbp]
    \centering\footnotesize
    \caption{\label{tab:rw}\revB{Comparison of \MacroSwarm{} with other decentralised swarm programming \acp{dsl} covered in \Cref{rw:dec}.
The dimensions of analysis include:
(i) 
the \textbf{generality/focus} of the proposal, which may affect some of its assumptions;
(ii) 
the \textbf{execution} style, if centralised or decentralised;
(iii) 
the \textbf{abstractions} provided by the programming model;
(iv) 
the \textbf{compositionality} of the programming model \emph{at the level of collective behaviours};
(v) 
the \textbf{scalability} of the approach with the number of agents;
(vi)
the \textbf{formality} of the approach, i.e., the extent to which the language/platform is formally described and enables forms of formal analysis;
(vii)
the \textbf{pragmatism} of the approach, namely the support of tools and reusable components for easy application development;
  and
\textbf{operational flexibility}, namely the support for flexible deployment and execution.
Bullets are used to denote full support (\cmark{}),
 partial support (\pmark{}), 
 or no support (\xmark{}).
The details about the choice of bullets can be found in \Cref{ssec:ac-features-swarm} and \Cref{sec:contrib} for \MacroSwarm{},
 and in \Cref{rw:dec} for the other related works.
}}
    \begin{tabularx}{\textwidth}{|X|c|c|c|c|c|c|c|}
        \hline

\textbf{Language} 
    & \makecell{\textbf{Generality}\\/\textbf{Focus}}
    & \makecell{\textbf{Abstractions}} 
    & \textbf{Compos.} 
    & \textbf{Scalab.} 
    & \makecell{\textbf{Formal.}} 
    & \textbf{Pragm.} 
    & \makecell{\textbf{Op. } \\ \textbf{Flex.}} 
    \\
        \hline
\textbf{\MacroSwarm{}} 
	& \makecell{\pmark{}\\Swarms} 
	& \makecell{Comput. fields, \\field functions} 
	& \cmark 
	& \cmark 
	& \cmark 
	& \pmark 
	& \cmark 
\\\hline\hline
\textbf{Buzz} \cite{DBLP:conf/iros/PinciroliB16} 
	& \makecell{\pmark{}\\Heterog.\\Swarms} 
	& \makecell{First-class swarm, \\virtual stigmergy} 
	& \pmark 
	& \cmark 
	& \xmark 
	& \cmark 
	& \pmark 
\\\hline
\textbf{Voltron} \cite{Mottola2014voltron} 
	& \makecell{\xmark{}\\Drone fleets,\\ mobile\\sensing} 
	&  \makecell{Abstract drone,\\spatial\\variables and \\iteration} 
	& \pmark 
	& \pmark 
	& \xmark  
	& \cmark  
	& \cmark  
\\\hline
\textbf{Meld} \cite{Meld2007} 
	& \makecell{\xmark{}\\Modular\\robotics} 
	& \makecell{Facts/rules,\\side-effectful\\facts,\\aggregate rules}
	& \pmark 
	& \cmark 
	& \pmark 
	& \pmark 
	& \xmark 
\\\hline
    \end{tabularx}
\end{table}

\section{Related Work}
\label{sec:rw}
\revA{In this section, 
 we review related work on swarm engineering.
First, to properly position this work, we cover related swarm engineering methods (\Cref{rw:methods});
 then, we consider languages and \acp{dsl} for swarm programming, first on those resulting in \emph{decentralised} behaviour \revB{(\Cref{rw:dec})},
 which are the most related to \MacroSwarm{},
 and finally on task orchestration languages \revB{(\Cref{rw:orch})}.
}

\revA{
\subsection{Swarm engineering methods}\label{rw:methods}

Brambilla et al.~\cite{DBLP:journals/swarm/BrambillaFBD13} provide a comprehensive review on engineering swarm robotic systems.
They distinguish between  
 (i) \emph{automatic} design methods, which do not require explicit intervention by the developer, such as those based on learning, 
 and (ii) \emph{behaviour-based} design methods, where swarm behaviours are iteratively developed using languages, often taking inspiration by social animals~\cite{bonabeau1999swarmintelligence-book}.
Classes of automatic methods include
 \emph{multi-agent reinforcement learning}~\cite{DBLP:journals/tsmc/BusoniuBS08} 
 and \emph{evolutionary robotics}~\cite{DBLP:series/sci/2008-108}.
Classes of behaviour-based methods 
 include 
 \emph{probabilistic finite state machines}~\cite{DBLP:conf/swis/SoysalS05},
 virtual physics-based techniques, such as those based on \emph{artificial potential fields}~\cite{DBLP:books/sp/90/Khatib90}, 
 and other design methods like 
 \emph{aggregate computing}~\cite{DBLP:journals/jlap/ViroliBDACP19} (reviewed in \Cref{sec:background}).
In particular, the term \emph{macroprogramming}~\cite{Casadei2023,regiment}
 refers to languages and approaches
 aiming to simplify the design of collective or macroscopic behaviours, often leveraging macro-level abstractions such as
 computational fields~\cite{DBLP:journals/pervasive/MameiZL04,DBLP:journals/jlap/ViroliBDACP19}, ensembles~\cite{Meld2007,scel2014taas}, collective communication interfaces~\cite{scel2014taas},
 and spatial abstractions~\cite{Ni2005spatialviews}.
Behaviour-based design is also supported by a literature of \emph{patterns} of collective and self-organising behaviour~\cite{DBLP:journals/nc/Fernandez-MarquezSMVA13,DBLP:journals/ras/OhSSJ17,DBLP:journals/tomacs/ViroliABDP18,DBLP:journals/swarm/BrambillaFBD13,DBLP:journals/fgcs/PianiniCVN21}.
}

\subsection{\revA{Decentralised swarm programming approaches}}\label{rw:dec}

Buzz~\cite{DBLP:conf/iros/PinciroliB16}
 is a mixed imperative-functional language for programming swarms.
In Buzz, swarms are first-class abstractions:
 they can be explicitly created,
 manipulated,
 joined (e.g., based on local conditions),
 and used as a way to address individual members (e.g., for tasking them).
For individual robots, 
 the language provides access to local features
 and the local set of neighbours, for interaction.
\revB{
The support for neighbouring operations (iteration, transformation, reduction)
 makes Buzz similar to aggregate programming languages like \MacroSwarm{}.
}
For swarm-wide consensus, 
 a notion of \emph{virtual stigmergy} is leveraged,
 based on distributed tuple spaces.
\revB{
The paper~\cite{DBLP:conf/iros/PinciroliB16} reports simulations in ARGoS with up to 1000 robots.
}
Buzz is designed to be an extensible language, since new primitives can be added.
Indeed, Buzz is based on a set of quite effective but ad-hoc mechanisms.
By contrast, \MacroSwarm{} uses few general and expressive primitives, 
  and supports swarm programming
  through a library of reusable, composable blocks.
Additionally, \MacroSwarm{} can leverage theoretical results from field calculi~\cite{DBLP:journals/jlap/ViroliBDACP19,DBLP:journals/tomacs/ViroliABDP18}, making programs amenable for formal analysis.
\revB{
Also, we consider Buzz to provide a partial support to compositionality of behaviours
  since the responsibility of disciplined programming
  is left to the programmer (cf. arbitrary mixes of mechanisms involving publish-subscribe, global variables, etc.), whereas in aggregate computing (and hence \MacroSwarm{})
  the functional paradigm and the notion of \emph{``field alignment''}~\cite{DBLP:conf/ecoop/AudritoCDSV22} ensures
  multiple instances of behaviours are properly scoped 
  to the right set of devices.
}

Voltron~\cite{Mottola2014voltron} is a programming model \revB{(an ``extension language'' meant to be applied to host languages)} for team-level design of drone systems\revB{, focussing on mobile sensing}. It represents a group of individual drones through a \emph{team abstraction}, which is responsible for the overall task \revB{and abstracts from low-level details, hence providing operational flexibility (such as transparently changing the number of drones)}. The details of individual drone actions and their timing are delegated to the platform system during runtime. The programmer issues \emph{action commands} to the drone team, along with \emph{spatiotemporal constraints}. The tasks in Voltron are associated with spatial locations, \revB{which can be iterated through a \texttt{foreachlocation} construct; then,} the team self-organises to populate \emph{multisets of future values} that represent the task's eventual result at a specific location. 
\revB{Java and C++ prototypes of the runtime system have been developed, supporting both centralised and state-replicated executions. The paper~\cite{Mottola2014voltron} reports experimental results for up to 100 drones in simulation, as well as a real-world experiment with 7 drones.}
\revB{Regarding the programming model, Voltron 
  provides
  an abstract \ac{api} for starting/stopping arbitrary actions at specific locations,
  and to read/write data in a global registry.
Specifically, the \ac{api} 
  mixes declarative and imperative constructs  
  tailored to tasking groups of drones
  for sensing activities at specific locations and specifying spatiotemporal constraints.
As such, the generality is partial,
 the support for behaviour composition
 is also partial (concurrent sensing tasks have to be handled via futures);
 we are also not aware of formalisations and specific formal guarantees. 
}

Meld~\cite{Meld2007} is a logic-based language for programming modular ensembles,
for systems where communication is limited to immediate neighbours. 
It leverages 
\emph{facts with side-effects} to handle actuation, 
\emph{production rules} to generate new facts from existing facts, 
and \emph{aggregate rules} to combine multiple facts into one fact by folding (e.g., maximisation or summation).
The runtime deals with communication of facts and removal of invalidated facts.
\revB{Decentralised execution is supported by fact localisation and (neighbour-based) fact sharing.
The paper~\cite{Meld2007} reports experiments in the Claytronics simulator with hundreds or thousands of robots (depending on the use case), showing scalability.
A semantics for Meld is provided in \cite{pfenning-meld}, though we are unaware of specific formal guarantees on programs.
}
The declarativity and logical foundation 
 make Meld an interesting macroprogramming system;
 however, it is not clear how it can scale with the complexity of general swarm behaviour.
Indeed, it is mainly adopted for shape formation and self-reconfiguring ensembles.

\revA{
Finally, we mention that there exist several calculi and languages that may support swarm programming (e.g. attribute-based communication languages~\cite{scel2014taas,abd2020programming-cas-attribute-based}), but we do not relate them in detail since
 they may not come with full-fledged implementations, libraries of reusable behaviours, or support top-down design through macroscopic abstractions.
}
 
\revA{
\subsection{Centralised orchestration approaches}\label{rw:orch}

Finally, we mention another category of related works, which are \emph{task orchestration languages} for swarms (e.g.,  TeCoLa~\cite{Koutsoubelias2016tecola},
 Dolphin~\cite{lima2018dolphin},
 Maple-Swarm~\cite{DBLP:conf/isola/KosakHBWHR20}, 
 PARoS~\cite{paros},
 Resh~\cite{DBLP:conf/icra/CarrollNS21}, 
 and \cite{DBLP:conf/iros/YiDLD0WY20}): they adopt quite a different approach that leverages centralised entities to control the activity of the swarm members based on the provided task descriptions.

In \MacroSwarm{}, 
 a program provides a description of the collective tasks,
 but also acts as a control program for the individual agents,
 and is hence executed in a decentralised fashion.
In the following, we review task orchestration languages for swarms, which adopt a quite different approach that leverages centralised entities to control the activity of the swarm members based on the provided task descriptions.

TeCoLa~\cite{Koutsoubelias2016tecola} is a programming framework that is designed to coordinate robotic teams and is implemented in Python. 
It provides abstractions for controlling individual robots and teams of robots, with most of the team management activities being handled automatically in the background. 
The framework uses the concept of nodes, which possess resources and services consisting of methods and properties that can be remotely invoked using proxies. 
TeCoLa leverages the notion of a \emph{mission group}, i.e., a dynamic set of nodes that participate in a mission, controlled and monitored by a \emph{coordinator entity} like a leader node or command control center. A mission group can also be split into teams whose shape is managed automatically, based on a \emph{membership rule} that specifies the set of services that team members should support. 
When all team members provide a particular service, that service is said to be promoted at the \emph{team-level}, allowing it to be requested on the team itself, triggering a corresponding service request on all team members and returning a vector of results.

PARoS~\cite{paros} is a Java framework for programming swarms that, similarly to the other reviewed approaches, provides an \emph{abstract swarm} abstraction, to support orchestration and spatial organisation of multiple robots. 
The \ac{api} provides various functions that include path planning, declaration of points of interest  (e.g., spatial locations that need to be inspected), enumeration swarm members, and event handling (e.g., to respond to robot failure).
The PARoS language combines elements from imperative, declarative, and event-driven programming.
At the execution level, PARoS uses a centralised coordinator, limiting the scalability of the approach.

Maple-Swarm~\cite{DBLP:conf/isola/KosakHBWHR20} (``Multi-Agent script Programming
Language for multipotent Ensembles'')
  is an approach to swarm programming
  that is based on concepts like
  agent groups (for addressing subsets of agents), 
  virtual swarm capabilities, 
  and hierarchical task plans.
Maple-Swarm supports the orchestration of individual agents
 through tasks that are derived from a composition of context-oriented partial plans.
In Maple, compositionality is obtained by connecting partial plans, which are defined in terms of swarm capabilities--the analogue of collective behaviours in \MacroSwarm{}. 
In \MacroSwarm{}, we directly support programming swarm capabilities by leveraging the field-based framework.

Dolphin~\cite{lima2018dolphin} is a Groovy \ac{dsl} designed specifically for task-oriented programming for autonomous vehicle networks. 
In Dolphin, the main abstraction is the \emph{vehicle set}, which is a dynamic group of vehicles that can be manipulated using set operations and pick/release operators---similarly to swarms in Buzz.
In Dolphin, the macro-level program defines how groups of vehicles are formed and tasked, essentially supporting the orchestration of individual behaviours, which are specified separately.

In~\cite{DBLP:conf/iros/YiDLD0WY20},
 a mixed decentralised-centralised actor-based framework
 is proposed for programming swarms.
The authors propose a \emph{collective actor} mechanism
 to centrally manage a swarm
 and provide primitives for high-level coordination
 (e.g., barrier synchronisation, branching and aggregation of swarms).
Though interesting, the approach -- unlike \MacroSwarm{} -- does not provide a well-defined programming model: instead, it is based on XML dialects to define actor configurations and task scripts.

Resh~\cite{DBLP:conf/icra/CarrollNS21} is 
 a \ac{dsl} for orchestration of multi-robot systems.
It leverages 
 the notion of a \emph{task} as a compositional block,
 robot capabilities (which are to be advertised by the robots),
 and spatiotemporal primitives (e.g., for waiting for events, specifying the location where a task is to be executed, etc.).
However, Resh is not Turing-complete, to simplify synthesis of task orchestration programs.
}

\revB{
\subsection{Properties of swarm behaviour}\label{rw:properties}
Luckcuck et al.~\cite{DBLP:journals/csur/LuckcuckFDDF19} 
 provide a comprehensive survey on the formal specification and verification of multi-robot systems.
Adopted formalisms include 
  set-theoretic approaches,
  transition systems, 
  logics (e.g., temporal or spatial logics),
  process algebras,
  ontologies,
whereas tools include model checkers, theorem provers, runtime monitors, and integrated toolchains.
Relevant swarm properties 
 that are addressed in the literature include
 safe navigation (e.g., collision avoidance),
 liveness properties,
 robustness against faulty sensors and failures,
 and convergence results (such as self-stabilisation---cf.  \Cref{sec:acblocks}). 

Specifically, the community of collective adaptive systems~\cite{DBLP:journals/sttt/NicolaJW20}
 also emphasises formal modelling and verification
 of large ensembles of devices, hence finding application in swarm robotics.
Recent work on this line
 includes GLoTL~\cite{DelGiudice2024global-prop-cas}, a linear temporal logic enabling to express both local and global properties of collective computations, the latter expressing that a certain property is satisfied by a defined fraction of agents.
Spatial logics such as SLCS~\cite{DBLP:journals/corr/CianciaLLM16}
 have been proposed to express spatial/topological properties,
 and for which run-time monitors can be synthesised, e.g., with field calculus-based encodings~\cite{DBLP:journals/jss/AudritoCDSV21}.
Properties of dynamically evolving ensembles can be
expressed in \emph{dynamic logics}~\cite{Hennicker2018dynamic-logic}.

}

\section{Conclusion and Future Work}
\label{sec:conc}

\revA{
In this article, 
 we presented \MacroSwarm{},
 a framework 
 for top-down behaviour-based swarm programming
 that offers a library of composable blocks 
 capturing common patterns of 
 decentralised swarm behaviours.
\MacroSwarm{} has been designed 
 in aggregate computing,
 a paradigm formally founded on field-based coordination,
 and implemented as an extension of the \scafi{} toolkit/\ac{dsl}.
We describe \MacroSwarm{} through examples 
 and case studies,
 evaluating by simulation 
 that the proposed approach is expressive, 
 compositional, and practical.
}

\revA{
In the future,
 we aim to more comprehensively extend the \MacroSwarm{} library,
 by implementing more algorithms 
 while drawing inspiration from surveys and classifications of swarm and collective behaviour~\cite{DBLP:journals/swarm/BrambillaFBD13}.
Also, interesting directions to be explored include the synthesis of \MacroSwarm{} programs, e.g., by following the approach in~\cite{DBLP:conf/coordination/AguzziCV22} based on reinforcement learning and sketching,
\revB{ 
the extension of the automatic verification framework presented in previous work~\cite{audrito23-coq} to self-stabilising properties under near-equilibrium conditions~\cite{DBLP:journals/jlap/ViroliBDACP19}},
and the support for \emph{low-code}~\cite{DBLP:journals/cacm/Hirzel23} aggregate programming, e.g., building on previous work on {\sc{}ScaFi-Web}~\cite{DBLP:conf/coordination/AguzziCMPV21} (and its blockly variant~\cite{DBLP:conf/coordination/AguzziCCV24}).
\revB{
Furthermore, 
we aim to investigate the cognitive load imposed by our proposed platform in comparison to the underlying ScaFi framework, 
to assess the simplicity and understandability of our approach. 
Qualitatively,
 the provided \ac{api}
 provides high-level constructs 
 that emphasise swarm programming at the macro-level;
 this is in contrast with requiring programmers to think at the micro-or node-level 
(which, in aggregate computing, often reduces to calls to low-level constructs such as \texttt{rep} and \texttt{nbr}). 
In the future, we intend to complement this qualitative evaluation with a quantitative study to provide an objective comparison. This will allow us to better understand how the library can be improved to reduce the complexity of the programming model in the context of swarm robotics.
}
Finally, we plan to deploy and test \MacroSwarm{} on physical robotic systems,
 for real-world scenario assessment, by leveraging on-going work on aggregate computing middlewares~\cite{DBLP:conf/acsos/AguzziCPSV21}.
}

\section*{Acknowledgements}

This work was supported by the Italian PRIN project ``CommonWears'' (2020HCWWLP). 

\bibliographystyle{alphaurl}
\bibliography{bibliography}

\end{document}

%% file: scala-def.tex
\lstdefinelanguage{scala}{
  keywords={abstract,case,catch,class,def,%
    do,else,extends,false,final,finally,%
    for,if,implicit,import,match,mixin,%
    new,null,object,override,package,%
    private,protected,requires,return,sealed,%
    super,this,throw,trait,true,try,lazy,%
    type,val,var,while,with,yield,forSome},
  otherkeywords={=>,<-,<\%,<:,>:,\#},
  sensitive=true,
  columns=fullflexible,
  morecomment=[l]{//},
  morecomment=[n]{/*}{*/},
  morestring=[b]",
  stringstyle=\ttfamily\color{red!50!brown},
  showstringspaces=false,
  morestring=[b]',
  morestring=[b]""",
  basicstyle=\sffamily\lst@ifdisplaystyle\footnotesize\fi\ttfamily,
  emphstyle=\sffamily\bfseries\ttfamily
}
\definecolor{ddarkgreen}{rgb}{0,0.5,0}
\lstdefinelanguage{scafi}{
  frame=single,
  basewidth=0.5em,
  language={scala},
  keywordstyle=\color{blue}\textbf,
  commentstyle=\color{ddarkgreen},
  keywordstyle=[2]\color{red}\textbf,
  keywords=[2]{rep,nbr,foldhood,foldhoodPlus,aggregate,branch,spawn,mux,mid},
  keywordstyle=[3]\color{gray},
  keywords=[3]{Me,AroundMe,Everywhere,Forever}, 
  keywordstyle=[4]\color{red}\textbf,
  keywords=[4]{in,out,rd},
  keywordstyle=[5]\color{violet},
  keywords=[5]{evolve,when,andNext,workflow,G,C,broadcast,gossip, S},
  keywordstyle=[6]\color{orange},
  keywords=[6]{Available,Serving,Done,Waiting,Removing,None,Set}
}